\documentclass[final,5p,times,twocolumn,authoryear]{elsarticle}

\usepackage{amssymb}
\usepackage{amsmath}
\usepackage{booktabs}
\usepackage{multirow}
\usepackage{amsmath}
\usepackage{graphicx}
\usepackage{subcaption}
\usepackage{cleveref}

\usepackage{caption}
\crefname{figure}{Fig.}{Figs.}
\Crefname{figure}{Fig.}{Figs.}
\crefname{equation}{Eq.}{Eqs.}
\Crefname{equation}{Eq.}{Eqs.}
\crefname{table}{Table}{Tables}
\Crefname{table}{Table}{Tables}
\usepackage[figuresright]{rotating}
\usepackage{soul}
\usepackage{xcolor}
\usepackage{float}
\usepackage{stfloats}
\usepackage{url}
\usepackage[most]{tcolorbox}

\usepackage{lineno}

\newtcolorbox{promptbox}[3][]{
    colback=gray!5, % Light gray background
    colframe=gray!50, % Muted border
    fonttitle=\bfseries\sffamily,
    coltitle=black,
    title=Diffusion Model Prompt,
    sharp corners,
    boxrule=0.5pt,
    left=10pt,
    right=10pt,
    top=5pt,
    bottom=5pt,
    enhanced,
    #1
}

\journal{ISPRS Journal of Photogrammetry and Remote Sensing}

\begin{document}
%%\begin{linenumbers}
\begin{frontmatter}



\title{SatSplatDiff: Geometry-preserving generative refinement for high-fidelity satellite Gaussian Splatting}


\author[1]{Jiyong Kim\fnref{equal}}
\ead{kim.9854@osu.edu}

\author[1]{Shuang Song\fnref{equal}}
\ead{song.1634@osu.edu}

\author[1,2]{Rongjun Qin\corref{cor1}}
\ead{qin.324@osu.edu}

\fntext[equal]{These authors contributed equally to this work.}
\cortext[cor1]{Corresponding author}

%% Author affiliation (OSU Department 1)
\affiliation[1]{organization={Department of Civil, Environmental and Geodetic Engineering, The Ohio State University},
            % addressline={2070 Neil Avenue}, 
            city={Columbus},
            postcode={43210}, 
            state={OH},
            country={USA}}

%% Author affiliation (OSU Department 2)
\affiliation[2]{organization={Department of Electrical and Computer Engineering, The Ohio State University},
            % addressline={2015 Neil Avenue}, 
            city={Columbus},
            postcode={43210}, 
            state={OH},
            country={USA}}

%% Abstract
\begin{abstract}
Gaussian Splatting has been recently explored for satellite 3D reconstruction, demonstrating flexibility and efficiency in representing radiometrically diverse satellite scenes. However, the limited top viewpoint of satellite imagery results in insufficient supervision on building facades, leaving surface holes and degraded visual fidelity. Generative refinement, which leverages pretrained generative priors to iteratively refine and update the rendered images used as supervision targets, has recently been investigated to improve the visual fidelity of Gaussian-rendered images. However, since these models refine each view independently, the resulting images can generate hallucinations and break photo-consistency, leading to geometric degradation. To address these limitations, we propose SatSplatDiff, which aims to minimize geometric degradation prevalent in generative refinement. Building on photogrammetric digital surface model (DSM) initialization and 2DGS-based shadow casting established in our prior work SatSplat, we first introduce monocular depth supervision and multi-scale geometric refinement to establish a geometrically accurate and well-regularized surface representation. We then apply shadow-guided generative refinement, where calculated shadow maps guide the Gaussians to maintain consistency with the underlying geometry, improving visual fidelity while reducing geometric degradation. Extensive evaluations on the IARPA2016 and DFC2019 datasets demonstrate state-of-the-art performance, reducing geometric $\mathrm{MAE}_{reg}$ by up to 18\% and improving visual fidelity (FID-CLIP) by 28–45\% over existing baselines. Our method delivers up to 5 $\times$ resolution enhancement with minimal hallucination and sensor-consistent appearance. By preserving geometric accuracy throughout the pipeline, it further demonstrates seamless cross-tile consistency and strong scalability for large-scale reconstruction. Source code is available at \url{https://github.com/GDAOSU/SatSplatDiff} %\RQ{the link is not avilable, it could be an empty link saying that the code will be released upon acceptance}. 
\end{abstract}

\end{frontmatter}

% \linenumbers


\section{Introduction}
\label{sec:intro}

The 3D reconstruction of large-scale urban scenes from satellite imagery is a critical task that has emerged at the intersection of photogrammetry, remote sensing, and computer vision \citep{li2023whu}. For example, satellite sensors running 24/7 allow capturing and reconstructing scenes at a consistent temporal rate, it also provides the advantage of enabling extensive coverage of inaccessible regions without physical presence. Classical multi-view stereo (MVS) methods process these images with heuristic pairwise selection, feature matching, and dense stereo matching to generate Digital Surface Model (DSM) \citep{de2014automatic, beyer2018ames, qin2016rpc, zhao2023review}. These pipelines establish a robust technical foundation by producing tractable surface representations through rigorous geometric modeling of the sensors and ray-based triangulation. However, despite their explicit rigor and geometric accuracy, traditional photogrammetric methods are often limited in handling complex radiometric variations and multi-temporal inconsistencies inherent in satellite imagery including shadow effects.

These limitations have motivated the adoption of learnable scene representations such as NeRF~\citep{mildenhall2021nerf} and 3D Gaussian Splatting~\citep{kerbl20233d}, which have been integrated into satellite 3D reconstruction pipelines due to their ability to model the scene with complex light environments. While much of the recent literature explores various aspects of 3D Gaussian Splatting in representing large-scale urban scenes from ground or aerial views \citep{liu2024citygaussian, lin2024vastgaussian, liu2024citygaussianv2, li2025ulsr} for visualization, some studies have focused on reconstructing accurate 3D surfaces from satellite imagery \citep{mari2023multi, aira2025gaussian, luo2026shadowgs}. 

A key strategy to improve geometric accuracy in these learnable representations is to decouple scene geometry from appearance by separating shadow and albedo components. Shadow visibility, in particular, can be explicitly rendered from scene geometry under known solar illumination, offering a direct geometric signal while albedo components address appearance reconstruction. Most notably, our previous work SatSplat~\citep{satsplat} extended this approach by incorporating 2D Gaussian Splatting (2DGS)~\citep{huang20242d} and shadow casting, which has shown that 2DGS, with oriented disk primitives, is more effective in representing scene sourced from high-altitude satellite images, leading to more convergent optimization and more accurate 3D reconstruction. SatSplat also integrates differentiable shadow casting for geometric supervision, which has achieved state-of-the-art accuracy in DSM generation. Despite the demonstrable results showing the superior quality of the GS models and the resulting DSM accuracy, these works consistently suffer from numerical instability associated with inaccurate GS initializations and pose errors from very small intersection angles among these high-nadir satellite images. Existing works partially investigated these problems by exploring different GS initializations and pose optimization through GS reconstruction, while they were not systematically integrated into a coherent approach to increase the solution stability. 

Another limitation stems from the nature of satellite observations: although existing approaches are able to arguably produce highly accurate DSM from the nadir-view, they suffer from insufficient supervision on building facades due to the limited top viewpoints of satellite imagery and occlusions, resulting in surface holes and degraded visual quality of the models. Diffusion-based generative refinement, which uses pretrained diffusion models to iteratively refine and update the rendered images used as supervision targets, has recently emerged as a promising direction to recover this missing appearance information by injecting appearance details from large and pretrained generative priors (i.e., diffusion models). However, a direct re-application of diffusion model for view generation will produce plausible but hallucinated textures drastically deviating from the reality (the actual, weak textures from the original images), which then leads to geometry artifacts when these generated views are used in 3DGS optimization \citep{lee2025SkyfallGS}. This calls for the need to place strong control on the diffusion model to yield enhanced views that are not only plausible, but also respect the geometry and appearance reality sourced from the actual satellite images.

%\RQ{you are supposed to talk about why 2DGS outperform the 3DGS in this introduction of the SatSplat}

%Most notably, SatSplat extended this paradigm by incorporating 2D Gaussian Splatting (2DGS) \citep{huang20242d} and shadow casting. By leveraging the ability of 2DGS to represent 'geometrically accurate' surfaces and shadow casting to provide geometric constraints, SatSplat achieved state-of-the-art accuracy in DSM generation.

%\RQ{check the sentence, I've added some more adjectives}
In this context, we present SatSplatDiff, a unified, accuracy preserving, "splats \& diffusion" driven pipeline that reconstructs highly accurate Gaussians from multi-date images. The proposed work resolves the fragmented gaps in the existing literature with a single pipeline consisting of three main novel strategies: 1) photogrammetric initialization to improve the poor convergence of Gaussians for satellite images; 2) geometric optimization via in GS pose estimation to improve the high-nadir ray convergence; 3) strong control for diffusion models to facilitate generative refinement of the GS model via shading guidance. Specifically, starting from a DSM generated by traditional photogrammetry, the geometric optimization stage extends SatSplat with monocular depth supervision, multi-scale geometric refinement, and densification to establish geometrically well-regularized surfaces with fully opaque Gaussians. This foundation is then followed by shadow-guided generative refinement, where geometrically calculated shadow maps from location and solar metadata are cast onto rendered images to condition the Gaussians on the underlying scene geometry. Furthermore, since shadow structure remains a stable geometric cue throughout refinement, the diffusion-refined image helps preserve geometry while improving visual fidelity. %\RQ{based on the first sentence, i think you can state at the end the diffusion did not degrade the accuracy but actually variably increased the accuracy in some cases (that's why we call it accuracy-preserving)}

Together, these components enable SatSplatDiff to reduce geometric $\textrm{MAE}_{reg}$ by up to 18\% while delivering up to 5× resolution enhancement with high-fidelity texture and minimal hallucination. Furthermore, by preserving geometric accuracy and sensor-consistent appearance throughout the pipeline, our method demonstrates seamless cross-tile consistency, demonstrating scalability for larger area. Our major contributions can be listed as follows:
\begin{itemize}
\item \textbf{Integrated Multi-date Satellite 3D Reconstruction Framework:} SatSplatDiff is a unified framework for multi-date satellite 3D reconstruction. It leverages the differentiable nature of Gaussian Splatting to separate lighting from appearance while incorporating generative priors for enhanced geometric precision and texture richness. SatSplatDiff establishes state-of-the-art results across the DFC2019 and IARPA2016 datasets.%\RQ{I do not think we did anything special to make it scalable, i can hardly see that this is a contribution, so i changed}.% \RQ{i edited this sentence a bit, please check} 
\item \textbf{Enhanced Geometric Optimization:} We extend SatSplat with monocular depth supervision, multi-scale geometric refinement and densification to address insufficient facade supervision and establish a solid foundation for generative refinement.
\item \textbf{Geometry-preserving Generative Refinement:} We introduce a shadow-guided generative refinement stage that improves visual fidelity by up to 5$\times$ effective resolution, while improving rather than degrading the mean $\mathrm{MAE}_{reg}$ from 1.27m to 1.23m by using shadow as a geometric cue.
\end{itemize}
%\RQ{you need a paragraph talking about: the remainder of the paper is organized as follows: }

We first review related work in \Cref{sec:related}, positioning our approach within the ongoing research on satellite 3D reconstruction and generative refinement. \Cref{sec:method:preliminary} and \Cref{sec:method} then introduce the preliminaries and detail our proposed method. \Cref{sec:experiments} evaluates our method against state-of-the-art baselines and presents ablation studies on each component. Finally, \Cref{Discussion} discusses sensitivity and practical considerations, and \Cref{sec:conclusion} concludes the paper.
%\RQ{so the condition is accurately both shadow mask and the albedo, not the rendered image (shading on the albedo) as the condition?}. 

\section{Related work}
\label{sec:related}

\subsection{Satellite 3D reconstruction}
\label{sec:related:nvs}

%Classical satellite 3D reconstruction generally refers to photogrammetric pipeline, which consist of stereo rectification, stereo matching, and triangulation~\citep{zhao2023review}. Among them, stereo matching has been widely investigated due to the ill-posed nature of stereo correspondence, with algorithms such as SGM~\citep{hirschmuller2008stereo} and MGM~\citep{facciolo2015mgm}, establishing strong baselines. 

\begin{figure}[h]
    \centering
    \includegraphics[width=\linewidth]{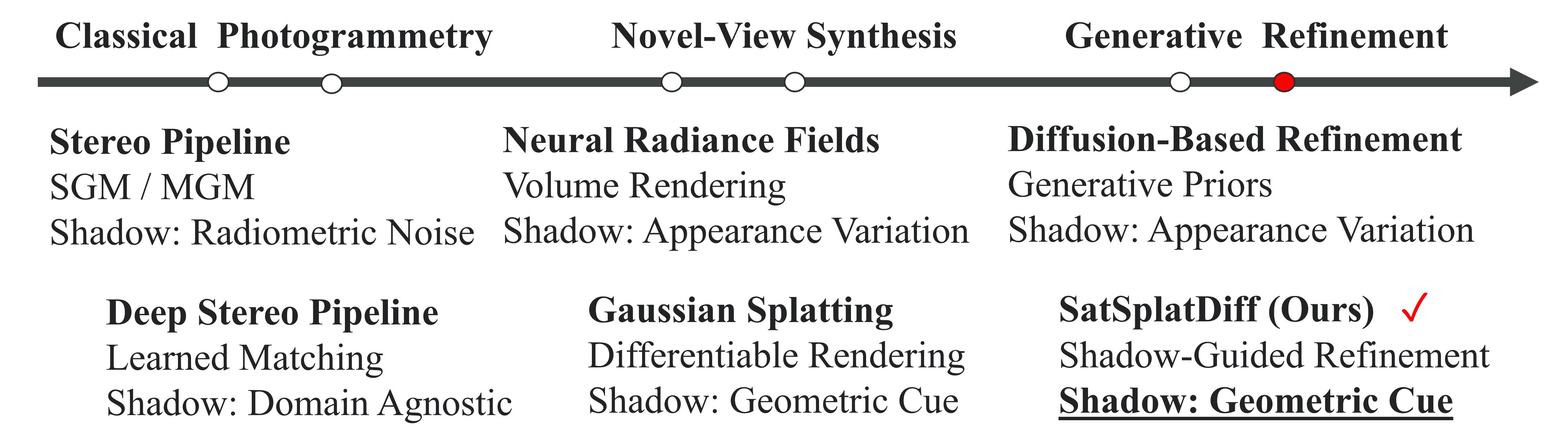}
    \caption{Evolution of satellite 3D reconstruction and the role of shadow modeling: from a source of radiometric noise in classical photogrammetry, to a geometric cue for Gaussian Splatting optimization. SatSplatDiff extends shadow as a geometric cue to diffusion-based generative refinement stage.}
    \label{fig:related}
\end{figure}

\Cref{fig:related} summarizes the evolution of satellite 3D reconstruction across three distinct stages, marking a paradigm shift from treating shading as image noise to leveraging it as a valuable geometric observation. Classical satellite 3D reconstruction is primarily based on photogrammetric pipelines consisting of stereo rectification, stereo matching, and triangulation~\citep{zhao2023review}. Stereo matching algorithms such as semi global matching (SGM)~\citep{hirschmuller2008stereo} and more global matching (MGM)~\citep{facciolo2015mgm} established strong geometric baselines by aggregating matching costs along multiple paths while enforcing smoothness constraints, with robust cost measures such as census transform reducing sensitivity to illumination variations including cast shadows. Subsequent deep learning approaches further improved robustness through learned feature representations that are domain-agnostic to radiometric variations~\citep{he2022hmsm, kim2025improving}. In both cases, reconstruction requires photo-consistency across pairs, with shadows treated as radiometric variation to be suppressed rather than as geometry-aware supervision cues. Domain-agnostic methods demonstrate robustness, but only within the distribution represented in their training data, which covers only a small fraction of the available satellite image archive.

%\RQ{we also have the shadow as the geoemtric cues, we additionally have this for the diffusion guidance, what is the difference between geomeric cues under gaussian splatting and the geometric guidance under the satsplatdiff}

Recently, novel-view synthesis methods including Neural Radiance Fields (NeRF)~\citep{mildenhall2021nerf} and Gaussian Splatting ~\citep{kerbl20233d} have been adopted in satellite scene reconstruction due to its flexibility in representing diverse radiometric variations. The differentiable rendering structure of NeRF and GS allows shading models to be injected directly into the optimization, naturally separating per-image solar illumination from albedo. S-NeRF~\citep{derksen2021shadow} modeled illumination using two components, direct sunlight with a local light source visibility field, and indirect sky light with a color field conditioned on the sun position, learning to predict a shading component and an albedo for each image. Sat-NeRF~\citep{mari2022sat} extended this shadow-aware formulation with a transient embedding to additionally absorb appearance changes from small transient objects, such as cars and trees. EO-NeRF~\citep{mari2023multi} took a different approach, rather than learning shadows as an appearance factor, it casts shadows directly from the estimated geometry and Sun direction, so that geometric inconsistencies surface as shadow artifacts that in turn provide a supervision signal to refine the geometry. However, despite achieving strong reconstruction quality with shadow-albedo decomposition, NeRF-based methods remain computationally expensive to train, which motivated SatelliteRF~\citep{zhou2024satelliterf} and Sat-NGP~\citep{billouard2024sat} to adopt multi-resolution hash encoding for faster convergence.

%\RQ{i understand whatever you mention before this point is about the nerf, but you are oragnizing the logic by addressing challenges not the categorization of representations, I do not think GS addresses these challenges better, it was just faster, and (perhaps) easier to converge, your presents the challenge and then cheerlead the GS does not sound to me correct. I would say that both nerf and GS, due to its differeiable structure, allows one to inject shading modeling, that naturally seperate the solar illumination per image and the albedo (proxy of the shaodw-free appearance). early works (prior to GS) address this problem progressively, and GS was later approached in addressing this challenge because it is faster and converge better.}
Gaussian Splatting~\citep{kerbl20233d} has emerged as an efficient alternative, offering comparable representational flexibility with substantially faster training and rendering, motivating recent efforts to bring the similar shadow casting explored in NeRF into the Gaussian Splatting framework. Motivated by the faster convergence of Gaussian Splatting, EOGS~\citep{aira2025gaussian} brought the geometry-driven shadow casting introduced by EO-NeRF into 3DGS optimization, demonstrating that shadow-based geometric cues can achieve DSM accuracy on par with classical photogrammetric methods when validated against LiDAR ground truth. The effectiveness of this shadow casting formulation within 3DGS established a foundation for subsequent works. EOGS++~\citep{bournez2025eogs++} expanded EOGS with optical-flow-based alignment between rendered and training images, while ShadowGS~\citep{luo2026shadowgs} adopted a similar shadow casting mechanism to sharpen surface reconstruction. Most closely aligned with our work, SatSplat \citep{satsplat} adopted a 2D Gaussian Splatting representation~\citep{huang20242d} and photogrammetric DSM initialization, achieving state-of-the-art geometric accuracy. The flattened, surface-aligned primitives of 2DGS, designed for accurate surface reconstruction, are particularly well suited to geometry-driven shadow casting, since shadow boundaries are rendered directly from the reconstructed surface rather than from a volumetric approximation.

While these shadow-aware Gaussian Splatting methods achieve strong near-nadir view synthesis quality and DSM generation, their performance degrades at oblique viewpoints, where top viewpoint satellite imagery provides sparse coverage. This results in insufficient supervision on building facades, leaving surface holes and degraded visual fidelity, motivating the need for additional geometric regularization and appearance refinement on these under-observed regions.

%\RQ{why? this is a too simplistic statement.}

\subsection{Pretrained priors for external supervision}
\label{sec:related:generative}

Within the satellite domain, pretrained priors have been increasingly used to provide explicit guidance, compensating for the sparse supervision inherent to satellite acquisition. Sat-DN~\citep{liu2025sat} and SkySplat~\citep{huang2025skysplat} both incorporated depth priors estimated by a pretrained monocular network, Depth Anything V2~\citep{depth_anything_v2}, but applied them differently. Sat-DN registered the relative depth predictions to sparse point clouds to recover absolute altitude, while SkySplat instead supervised geometry through a Pearson correlation loss~\citep{lee1988thirteen} between rendered and predicted depth. GU-GS~\citep{ding2026gu} instead leveraged DINOv2~\citep{oquab2023dinov2} feature embeddings to model uncertainty in the discrepancy between rendered and ground-truth features, automatically downweighting regions, such as shadows and transient objects, where photo-consistency is not ensured.

Meanwhile, the generative capacity of diffusion models~\citep{esser2024scaling, rombach2022high, peebles2023scalable, saharia2022photorealistic} in the computer vision domain has inspired a broader line of work that leverages these models as generative priors to enhance 3D scene representations. Early Score Distillation Sampling (SDS)-based methods such as DreamFusion~\citep{poole2022dreamfusion} and Magic3D~\citep{lin2023magic3d} demonstrated that 3D synthesis can be guided by 2D diffusion priors. However, these methods incur substantial computational overhead.

%\RQ{additionally, you should write a bit more here why it is possible nowadays to use geneartive priors, i want to make sure we build the rationale why it is possible to use that in the aerial case, i do not want people to think about diffusion still for movies, cups and tables: "Foundation generative/diffusion models such as .. have been trained with a large volume high-resolution aerial and facade images, which holds the distribution of plausible high-resolution facade images of its kind. Taking these diffusion models as the generative prior, with proper condition, it is possible to augument weak satellite-based textures or make up missing facade information with accuracy"}.

Owing to its computational efficiency, more recent works concentrate on what we refer to as generative refinement. A 3D representation is first optimized from the input views, and a diffusion model is then used to improve its rendered outputs, such as novel views, depth, or appearance. These refined outputs are fed back as additional supervision for the 3D representation, rather than distilling it directly from the diffusion model. This reduces memory consumption because unlike SDS-based methods, which keep both the diffusion model and the 3D representation in memory simultaneously throughout optimization, generative refinement alternates between generating a batch of refined views with the diffusion model and distilling them into the 3D representation. Following this pipeline, Deceptive-NeRF/3DGS~\citep{liu2024deceptive} directly injected diffusion-generated images as training views, while FlowR~\citep{fischer2025flowr} and CAT3D~\citep{gao2024cat3d} adopted multi-view consistent diffusion models for more coherent priors. 3DGS Unbounding~\citep{liu20243dgs} leveraged video diffusion to maintain temporal consistency, and DIFIX3D+~\citep{wu2025difix3d+} and ArtiFixer~\citep{de2026artifixer} focused on accelerating inference through single-step and auto-regressive distillation, respectively. Across these methods, generative refinement effectively recovers fine-grained appearance details, while reducing the computational overhead of SDS-based distillation. However, several recent works~\citep{liu2026had} have reported that diffusion-generated content can introduce hallucinations inconsistent with the actual scene. %\RQ{people have done this, what is the conclusion, was it sucessful, any gaps, regardless of whether it was applied in the satellite domain}

%\RQ{actually, what is the hurdle of the current mv consistent diffusion models for aerial cases, the datasets they are trained from?}

%Direct application of these refinement priors to satellite imagery was limited until recently, as the earlier diffusion models were trained predominantly on ground-level natural images with little exposure to aerial viewpoints. This has changed with recent foundation diffusion models~\citep{khanna2024diffusionsat, wu2025qwen, flux-2-2025} trained on large volumes of aerial and facade imagery, making it possible to take them as a generative prior to augment weak satellite textures or recover missing facade appearance with reasonable accuracy.

Building on this trend in computer vision, generative refinement has recently been applied to satellite imagery, enabled by foundation diffusion models~\citep{khanna2024diffusionsat, wu2025qwen, flux-2-2025} trained on large volumes of aerial and facade imagery, which can be taken as generative priors to augment weak satellite textures or recover missing facade appearance with reasonable accuracy. Skyfall-GS~\citep{lee2025SkyfallGS} generates high-fidelity synthetic views with a diffusion model to supervise Gaussian optimization, demonstrating the potential of generative refinement to substantially elevate visual fidelity in satellite 3D reconstruction. However, Skyfall-GS also exhibits the same hallucination issue observed in general-domain generative refinement. Since these hallucinations break the photo-consistency assumption underlying geometric reconstruction, Skyfall-GS exhibits degraded geometric accuracy after generative refinement. Orbit2Ground~\citep{yu2025orbit} mitigates this by anchoring generative priors to a 2.5D mesh, but the structural rigidity of a 2.5D representation limits its ability to capture complex urban surfaces.

Hallucination thus remains a key challenge in generative refinement, as it can break photo-consistency and consequently degrade geometric accuracy. While hallucination itself may be unavoidable, its impact on geometry can be mitigated by constraining the refinement process with external geometric cues. However, neither Skyfall-GS nor Orbit2Ground extends the shadow-based geometric cues already established in EO-NeRF~\citep{mari2023multi}, EOGS~\citep{aira2025gaussian}, and SatSplat~\citep{satsplat} into the generative refinement stage.

\section{Preliminaries}
\label{sec:method:preliminary}

This section introduces the preliminaries adopted in our pipeline: the 2D Gaussian Splatting representation, the affine camera model, and the differentiable shadow casting mechanism. These components, previously validated in our prior work SatSplat~\citep{satsplat}, form the foundation upon which our proposed extensions, detailed in \Cref{sec:method}, are built. Following subsections describe the details of each component.

%\RQ{why this name, sounds like the basics for the whole methodology, but this corresponds mostly to the initialization? as i said, the correspondences of your sections and the three -step process should be made more explicit, if you truely want to include a subsection about the preliminaries, it should an independent subseciton not piggy-bagging  one of the steps }

%\RQ{you need to say something here to introduce what you will introduce in this section}
\subsection{2D Gaussian Splatting}
\label{sec:method:affine_2dgs}

The key difference in 2DGS and 3DGS is that 2DGS computes the exact intersection of each ray with an oriented disk primitive, rather than projecting the disk's center onto the image plane~\citep{huang20242d}. This ray-disk intersection allows 2DGS to represent geometrically accurate surfaces, in contrast to the volumetric primitives used in 3DGS. The alpha blending procedure follows 2DGS~\citep{huang20242d}:
\begin{equation}
\mathcal{R}(\mathbf{x}) = \sum_{i=1}^{N} \mathbf{a}_i \alpha_i \hat{\mathcal{G}}_i(\mathbf{u}(\mathbf{x})) \prod_{j=1}^{i-1} (1 - \alpha_j\hat{\mathcal{G}}_j(\mathbf{u}(\mathbf{x}))),
\label{eq:render}
\end{equation}
where $\mathcal{R}(\mathbf{x})$ denotes the rasterized features at pixel $\mathbf{x}$, and $\mathbf{a}$ includes color, normal, and depth. Index $i$ orders Gaussian primitives by depth, and $\alpha$ denotes opacity.

To ensure the Gaussians represent a consistent and continuous surface rather than disconnected primitives, 2DGS \citep{huang20242d} introduces two critical loss functions: distortion loss and normal consistency loss. The distortion loss $\mathcal{L}_{dist}$ is designed to minimize the depth ambiguity along the ray by penalizing the spread of Gaussians in the depth direction, encouraging them to collapse onto a thin surface layer:
\begin{equation}
\mathcal{L}_{dist} = \sum\limits_{i,j} \omega_i \omega_j |z_i - z_j| ,
\end{equation}
where $\omega_i$ is the blending weight and $z_i$ is the depth of the $i$-th Gaussian. 

The 2D Gaussians define a surface normal $\mathbf{n}$ in world space, which is the last column of the rotation matrix for each Gaussian. The rendered normal map $\textbf{N}$ and depth map $\textbf{D}$ are computed via alpha blending following \Cref{eq:render}. The normal consistency loss $\mathcal{L}_{norm}$ reduces the discrepancy between the per-Gaussian normals $\mathbf{n}_{i}$ and the normals estimated from the gradient of the rendered depth map $\mathbf{N}$:
\begin{equation}
\mathcal{L}_{norm} = \sum\limits_{i} \omega_i (1 - \mathbf{n}_i^T \mathbf{N}) ,
\end{equation}

However, since these geometric losses are computed only from the available satellite viewpoints, Gaussians in occluded regions such as building facades receive weak gradient signals due to limited pixel coverage, motivating the multi-scale geometric refinement introduced in \Cref{sec:method:multiscale}.

\subsection{Affine camera model fitting}
\label{sec:camera_sampling}

Satellite images employ rational polynomial coefficients (RPCs) as their camera model, which provide high geometric accuracy but involve complex nonlinear transformations. Several prior works have demonstrated the effectiveness of approximating RPCs with affine camera models for satellite-based Gaussian splatting~\citep{aira2025gaussian, bournez2025eogs++}. These methods approximate the RPC models with an affine camera as \Cref{eq:affine_camera}.

\begin{equation}
\mathbf{x} = \mathbf{F} \cdot \begin{bmatrix}\mathbf{p}\\1\end{bmatrix} .
\label{eq:affine_camera}
\end{equation}

This simplification allows efficient mapping of 3D points to the image plane. Specifically, an affine camera maps a 3D world point $\mathbf{p} \in \mathbb{R}^3$ to a 2D image coordinate $\mathbf{x} \in \mathbb{R}^2$ via a $2 \times 4$ projection matrix $\mathbf{F}$, as shown in \Cref{eq:affine_camera}. We note that we also describe the camera model $\mathbf{F}$ as combination of $2 \times 3$ matrix $\mathbf{W}$ and translation matrix $\mathbf{b}$ with $[\mathbf{W} | \mathbf{b}]$.

\subsection{Shadow casting and rendering}
\label{subsubsec:shadow_casting}
An effective strategy adopted in several works~\citep{aira2025gaussian} to optimize geometry is to model the sun as an inverse camera, which makes shadow casting differentiable and enables direct geometric supervision on Gaussians. Specifically, for a sun camera with azimuth and elevation angles $(\phi_{sun}, \theta_{sun})$, world points are flattened onto the XOY plane along the sun direction and then project onto the main camera. To preserve depth information along the sun direction for shadow testing, the original elevation is retained as the sun depth.

\begin{equation}
\mathbf{W}_{sun} 
%= \mathbf{B}_{xy} (\mathbf{I} - \left[ \mathbf{0} \quad \mathbf{0} \quad \mathbf{t} \right] ) = \mathbf{B}_{xy} \mathbf{T}
= \mathbf{W} (\mathbf{I} - \begin{bmatrix}
    0 & 0 & \frac{\cos\phi_{sun}}{\tan\theta_{sun}} \\
    0 & 0 & \frac{\sin\phi_{sun}}{\tan\theta_{sun}} \\
    0 & 0 & 1
\end{bmatrix}).
\label{eq:sun_camera_Bxy}
\end{equation}

After generation, $\mathbf{W}_{sun}$ and \Cref{eq:affine_merge} are used to build the sun camera $\mathbf{F}_{sun}$. Using the sun camera $\mathbf{F}_{sun}$, we render the depth map and compute the sun visibility mask $\mathbf{V}$, determined by the discrepancy between the projected depth $\mathbf{D}_{proj}$ and the sun-space depth reprojected onto image space $\hat{\mathbf{D}}_{sun}$:
\begin{equation}
\mathbf{V} = \exp\left( -\rho \cdot \max(0, \mathbf{D}_{proj} - \hat{\mathbf{D}}_{sun}) \right),
\end{equation}
where $\rho$ equals to shadow density. 

Finally, the image $\mathbf{R}_{final} \in \mathbb{R}^{C\times H \times W }$ can be modeled as a combination of the surface albedo $\mathbf{R}_{albedo} \in \mathbb{R}^{C \times H \times W}$, the sun visibility mask $\mathbf{V} \in \mathbb{R}^{1\times H \times W}$, and a global ambient light component $\mathbf{L}_a \in \mathbb{R}^{C\times 1 \times 1}$. The sun visibility equals 1 if the point is lit by the sun and 0 if it is in shadow. To model radiometric variation and sensor characteristics across different acquisitions, $\mathbf{C} \in \mathbb{R}^{C\times C}$ and $\mathbf{c} \in \mathbb{R}^{C}$ could be used to represent per-image camera gain and bias. The final rendered color is given by \Cref{eq:image_formation}.

\begin{equation} 
\mathbf{R}_{final} =  \underbrace{(\mathbf{C} \cdot \mathbf{R}_{albedo}  + \mathbf{c} )}_{\text{Corrected albedo}} \odot \underbrace{(\mathbf{V} + (1 - \mathbf{V}) \mathbf{L}_a)}_{\text{Shadow map}},
\label{eq:image_formation}
\end{equation}
where $\odot$ denotes element-wise multiplication, and $C$ is the number of color channels (i.e., 3 for RGB), $H$ is the image height, and $W$ is the image width. Aside from the sun visibility mask $\mathbf{V}$, which is determined at render time, all other components are learned during training. 

%\RQ{ditto, but i am not exactly sure why there is a transparency and why it needs to be reduced, perhaps I forgot this detail in the original paper, but by just reading it blantantly, this triggers a question mark} 

As noted in EOGS~\citep{aira2025gaussian}, much of the scene texture becomes embedded in the shadow map unless shadow transparency is penalized, since semi-transparent shadows effectively bake appearance information into the visibility mask. EOGS addresses this by applying an entropy penalty on the visibility mask $\mathbf{V}$ for each pixel grid $(u,v)$, using the entropy function $\mathcal{H}(x) = -\left[x\log(x) + (1-x)\log(1-x)\right]$:
\begin{equation}
\mathcal{L}_{shadow} = \sum\limits_{u, v} \mathcal{H}(\mathbf{V}(u, v)).
\end{equation}\label{eq:loss_shadow}

\section{Methodology}
\label{sec:method}
%\RQ{I d like to have your figure present early on, figure 1. generally, there seems not to be a good match between the text structure and the general pipeline, the title of subsections seems to be disjointly if you view the entire title/sub-title structure. Even reading the intro section and the intro paragraph, when looking at the subtitles still no idea why they are needed how these are connected.}

%\RQ{calling a photogrammetric initiation as photogrammetric rigor is not appropriate. initialization place influence but no constraint in this process. you can say we found that this reconstruction process is sensitive to initialization, therefore we started with a lower-accuracy results from photogrammetric reconstruction, i think you shbould also in your discuss, study that}

% add section numbers to enhance readability

\begin{figure*}[tb]
\centering
\includegraphics[width=0.97\textwidth]{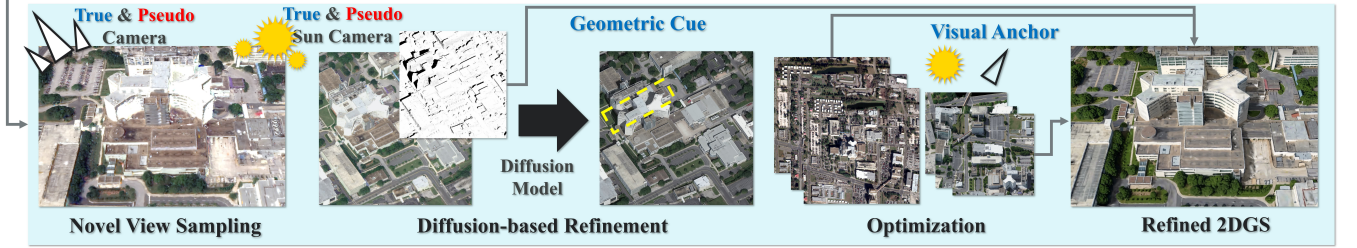}
\caption{Overall architecture of the SatSplatDiff framework. The pipeline achieves geometric accuracy and visual fidelity through three key stages: (1) photogrammetric initialization (\Cref{sec:method:preliminary}) establishes the initial 3D representation through bundle adjustment and DSM generation; (2) geometric optimization (\Cref{sec:geometric_optimization}) transforms the baseline into a 2DGS representation and refines the geometry, utilizing multi-scale geometric refinement, monocular depth supervision and Gaussian densification; and (3) shadow-guided generative refinement (\Cref{sec:shadow-guided}) incorporates diffusion-based refinement to improve visual fidelity, with geometry guided by shadow casting to preserve geometric accuracy.}%\RQ{different blocks should mention where it was introduced. place the relevant section number in these subtitles of the process, readers can then easily relay to the corresponding sections if they are particularly interested in one of the modules/process}
%\RQ{sub-figure are not all high res. with high quality pdf people typically zoom in to see details }
\label{fig:overall_flowchart}
\end{figure*}

As established in \Cref{sec:related}, satellite 3D reconstruction faces two major challenges: insufficient supervision for occluded structures such as building facades, and the difficulty of improving visual fidelity without geometric degradation during generative refinement. As illustrated in \Cref{fig:overall_flowchart}, SatSplatDiff addresses these challenges through a three-stage pipeline consisting of photogrammetric initialization, geometric optimization, and shadow-guided generative refinement. Starting from a DSM generated by traditional photogrammetry, the geometric optimization stage extends SatSplat with monocular depth supervision, multi-scale geometric refinement, and Gaussian densification. The generative refinement stage further improves appearance quality using diffusion-refined images as supervision targets while preserving geometry through shadow-based structural cues. In the following subsections, we describe each stage in detail.

%\RQ{you spend effort in generating fig 1 but the above text barely touch it. I did not meant to say that you should talk more about that but less only if the matching of the mainbody text/structure and the figure is so obvious. Also you need some layout adjustment to keep figure close to where it was mentioned}

% \subsection{Preliminary}
% \label{sec:method:preliminary}

% \subsubsection{Bundle Adjustment and DSM Initialization}

% Following the SatSplat approach, we first perform global refinement of the camera parameters via Bundle Adjustment (BA) using the Sat-Bundle Adjust \citep{ipol.2021.352} to mitigate absolute localization errors in the initial RPCs. This process provides critical elevation priors and scene boundaries for casting the coordinate system into NDC space. We also generate an initial Digital Surface Model (DSM) to produce a structural DSM that maintains global consistency. This DSM is subsequently converted into 2D Gaussian primitives for initialization.

\subsection{Photogrammetric initialization}

Most Gaussian Splatting methods utilize sparse point clouds for initialization~\citep{kerbl20233d, huang20242d, lee2025SkyfallGS, luo2026shadowgs}, while some satellite-specific approaches for DSM generation spread Gaussians volumetrically and prune during optimization~\citep{aira2025gaussian, bournez2025eogs++, ding2026gu}. However, as demonstrated in SatSplat~\citep{satsplat}, initializing from a photogrammetric DSM provides more accurate geometric initialization, which is essential in our pipeline because both shadow casting and subsequent generative refinement rely on geometry-consistent scene representations.

Following our previous work SatSplat, we initialize 2D Gaussians from a photogrammetric DSM generated through classical stereo matching with ASP~\citep{beyer2018ames} or s2p~\citep{de2014automatic}, after refining camera parameters through bundle adjustment. The DSM is converted into 2D Gaussian primitives, where each Gaussian center is sampled from DSM pixels. To better represent high-frequency structures such as building facades, we adopt an area-weighted sampling strategy that assigns higher sampling probability to regions with larger local surface variation. Specifically, for each DSM pixel, we estimate the local surface area as
\begin{equation}
a = \| dx \times dy \|,
\end{equation}
where $dx$ and $dy$ denote local tangent vectors computed from neighboring points. The estimated area serves as a sampling weight for Gaussian initialization, allowing geometrically complex regions to receive denser Gaussian representation.

\subsection{Geometric optimization}
\label{sec:geometric_optimization}

In the geometric optimization stage (\Cref{fig:overall_flowchart}), we extend the SatSplat optimization framework to improve surface completeness and geometric consistency in under-observed regions. We introduce multi-scale geometric refinement, monocular depth supervision, and adaptive Gaussian densification to enhance the reconstruction of regions with limited geometric constraints. Multi-scale geometric refinement optimizes Gaussians from diverse viewpoints to improve surface continuity, while monocular depth supervision provides a coarse geometric prior that stabilizes early-stage optimization. Finally, Gaussian densification improves surface coverage and enables fine-grained structure recovery.

%In the geometric optimization stage (\Cref{fig:overall_flowchart}), shadow casting serves as the primary geometric supervision. While major components of SatSplat are described in \Cref{sec:method:preliminary}, we extend this foundation with additional components to improve geometric completeness of facade surfaces. Multi-scale geometric refinement improves facade surfaces, monocular depth supervision stabilizes early-stage optimization, and densification ensures efficiency of Gaussian optimization. Following subsections describe the details of our extensions.

%\RQ{this section is started with talking about the shaodws and your next section is "shadow-guided generative refinement", this is confusing, it sounds like your 3.2 are 3.3. are shadow related, why you bother to have two subsections. The introduction should focus on the geometry, shading is a cue. so it should read "In the geometric optimization stage, we utilize a series of cues to reflect the mulit-date and sub-optimal geometric configuration nature for multi-view satellite images, this includes: 1) explicitly model the shading process using the per-image solar information, to scalfolding the time-varying radiometric inconsistency to achieve superior geometry recovery..... "}

\subsubsection{Multi-scale geometric refinement} %\RQ{the "multi-scale" is new to me, you never mention the need for multi-scale earlier, you need to mention that somewhere earlier why multi-scale is the need in your solution}
\label{sec:method:multiscale}

As discussed in \Cref{sec:geometric_optimization}, satellite imagery provides limited observations of vertical structures, leaving facade regions insufficiently supervised. Since facade regions are typically rendered at a much lower effective resolution than the original satellite imagery, optimizing Gaussians at a single fixed scale is insufficient to recover consistent surface geometry. We therefore render facade regions at multiple scales, applying geometric optimization at each resolution to better capture fine-grained structural detail. To address this issue, we introduce multi-scale geometric refinement by generating affine camera models with varying zoom scales, elevation, and rotation angles:
\begin{equation}
\mathbf{F} = f(t, \phi, \theta, s) , \label{eq:generate_camera}
\end{equation}
where $t$ refers to target interest vector, $\theta$ and $\phi$ refers to the angle of camera position and $s$ refers to zoom scale. The viewing direction is determined by the azimuth angle $\phi$ and elevation angle $\theta$, which directly define the unit viewing vector $\mathbf{f} \in \mathbb{R}^3$:
\begin{equation}
\mathbf{f} = \begin{bmatrix} -\cos\theta \sin\phi , -\cos\theta \cos\phi , \sin\theta 
\end{bmatrix}^\top.\label{eq:direct_f}
\end{equation}

We then construct an orthonormal basis $\{\mathbf{u}, \mathbf{v}, \mathbf{f}\}$ for the camera coordinate system, where $\mathbf{u} = \frac{\mathbf{f} \times \mathbf{up}}{\|\mathbf{f} \times \mathbf{up}\|}$ with the world-up vector $\mathbf{up} = [0, 0, -1]^\top$, and $\mathbf{v} = \mathbf{f} \times \mathbf{u}$ is computed following the right-handed rule. The refined affine parameters $\mathbf{W}$ and $\mathbf{b}$ are formulated by scaling the image-plane basis with a zoom factor $s$ and the original scene scale $\sigma$:

\begin{equation}\mathbf{W} = s \sigma \cdot \begin{bmatrix} \mathbf{u}^\top \ \mathbf{v}^\top \end{bmatrix}, \quad \mathbf{b} = -\mathbf{W} \mathbf{t}\end{equation}
\begin{equation}
\mathbf{F} = [\mathbf{W} | \mathbf{b}]. \label{eq:affine_merge}
\end{equation}

Based on this camera sampling function, we define an affine camera model $\mathbf{F}$ with \Cref{eq:generate_camera}. Using $\mathbf{F}$, we render the albedo image $\mathbf{R}_{albedo}$ and depth map $\mathbf{D}$ for geometric supervision. We then perform an additional optimization step using geometric loss terms: $\mathcal{L}_{norm}$ and $\mathcal{L}_{dist}$. 

Furthermore, to encourage solid and stable surface structures while promoting the removal of transparent Gaussians, we introduce an opacity entropy term:
\begin{equation}
\mathcal{L}_{ent} = \sum\limits_{i} \mathcal{H}(\alpha_i) ,
\end{equation}
where $\alpha_i$ denotes the opacity of the $i$-th Gaussian. The final refinement loss is defined as:
\begin{equation}
\mathcal{L}_{refine} = \lambda_{dist} \mathcal{L}_{dist} + \lambda_{norm} \mathcal{L}_{norm} + \lambda_{ent}\mathcal{L}_{ent}.
\end{equation}

\begin{figure}[!htbp]
    \centering
    \begin{subfigure}[t]{0.42\columnwidth}
        \centering
        \includegraphics[width=\textwidth]{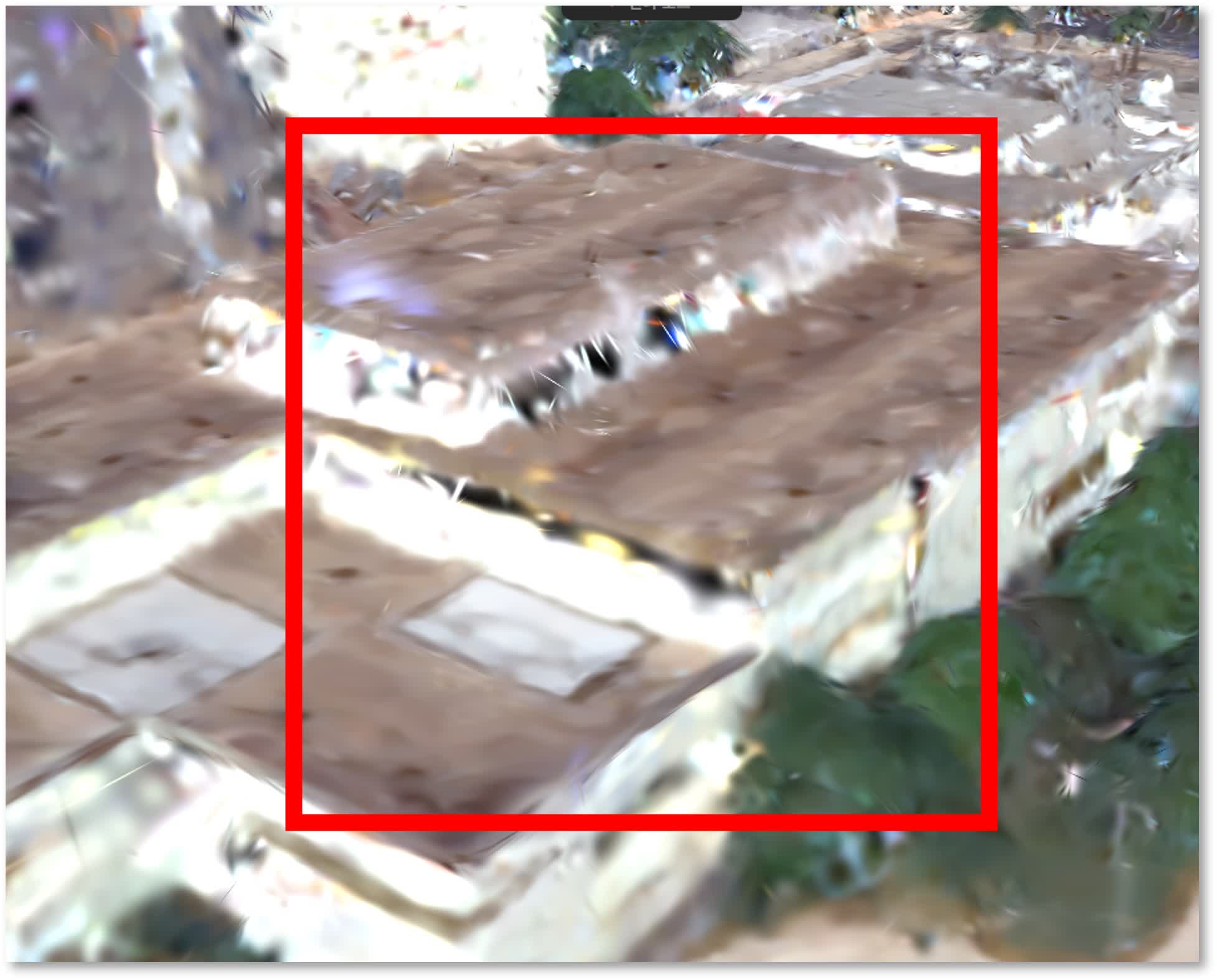}
        \caption{w/o multi-scale geometric refinement}
        \label{fig:before_zoom}
    \end{subfigure}
    \hspace{0.02\columnwidth}
    \begin{subfigure}[t]{0.42\columnwidth}
        \centering
        \includegraphics[width=\textwidth]{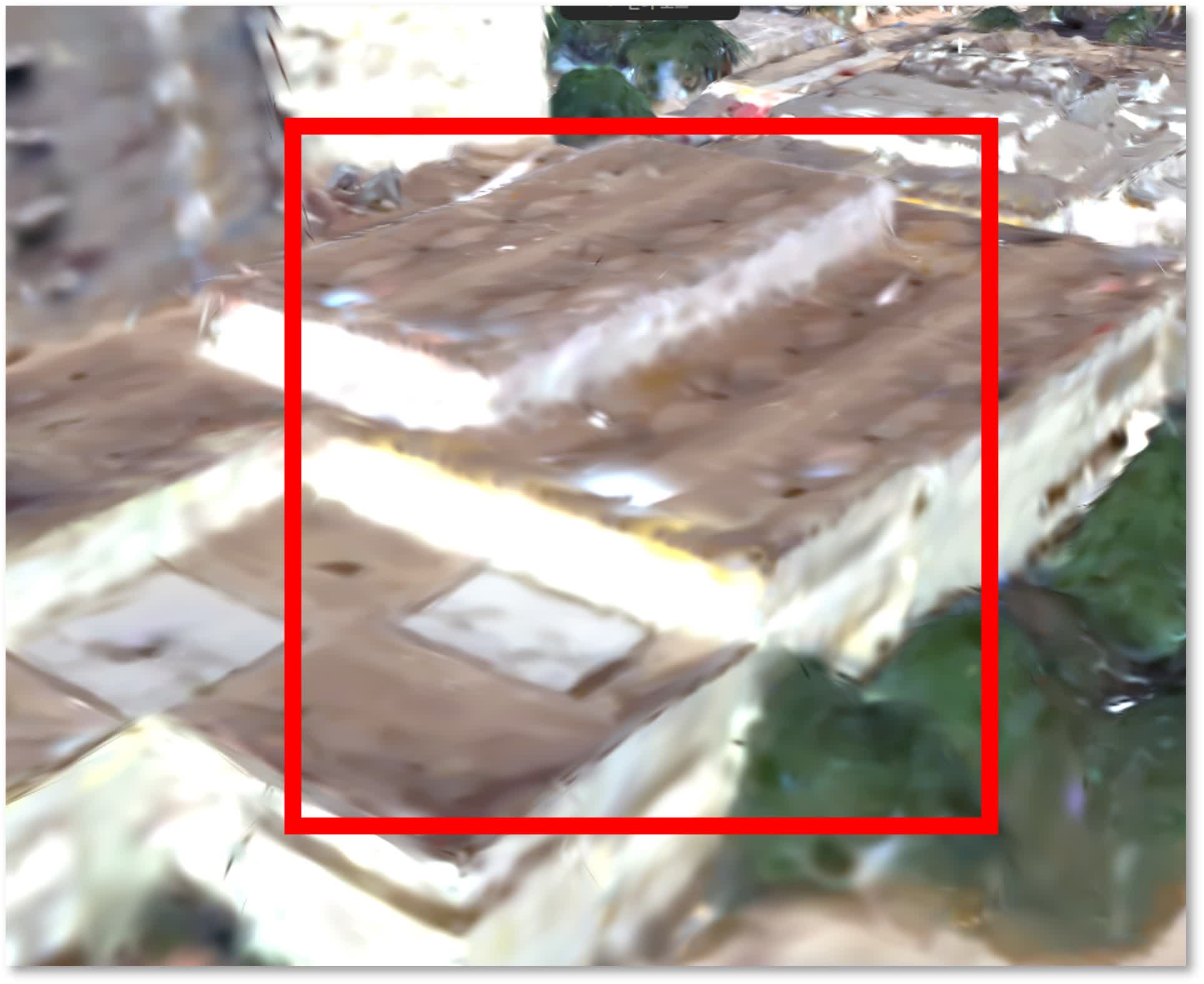}
        \caption{w/ multi-scale geometric refinement}
        \label{fig:after_zoom}
    \end{subfigure}
    \caption{Impact of multi-scale geometric refinement on surface solidity, effectively eliminating holes in building facades.}
    \label{fig:zoom_comparison}
\end{figure}

At each training iteration, an additional optimization step is performed using $\mathcal{L}_{refine}$, providing geometric gradients for occluded regions and eliminating surface holes (\Cref{fig:zoom_comparison}).

\subsubsection{Monocular depth supervision}
\label{subsubsec:monocular}

Although photogrammetric initialization provides a reliable geometric starting point, early-stage Gaussian optimization can still suffer from local minima due to sparse observations. To provide additional structural guidance during this stage, we incorporate monocular depth supervision following previous works~\citep{lee2025SkyfallGS, li2024dngaussian}. Monocular depth supervision is applied only during the first 3,000 iterations to provide a coarse geometric prior and stabilize the initial optimization. Since monocular depth supervision provides only relative geometric information, it is removed after the initial stabilization stage to prevent accumulated depth biases.

Specifically, we use Depth Anything V2~\citep{depth_anything_v2} to predict a monocular depth map and supervise the rendered depth using the Pearson Correlation Coefficient~\citep{lee1988thirteen} as a scale-invariant loss function. Let $\mathbf{D}_{rend}$ be the rendered depth map from the Gaussians and $\mathbf{D}_{mono}$ be the depth map predicted by the monocular foundation model. The depth supervision loss $\mathcal{L}_{depth}$ is defined as:%\RQ{it does not sound like as significant difference if we start with 3000 iterations, you need to provide a strong justification on this, why it matters to start with 3,000 intead of 1,000}.
\begin{equation}
\mathcal{L}_{depth} = 1 - \frac{\sum\limits_{u,v} (\mathbf{D}_{rend}(u,v) - \bar{\mathbf{D}}_{rend})(\mathbf{D}_{mono}(u,v) - \bar{\mathbf{D}}_{mono})}{\sqrt{\sum\limits_{u,v} (\mathbf{D}_{rend}(u,v) - \bar{\mathbf{D}}_{rend})^2 \sum\limits_{u,v} (\mathbf{D}_{mono}(u,v) - \bar{\mathbf{D}}_{mono})^2}},
\label{eq:pearson_depth}
\end{equation}
where $\bar{\mathbf{D}}$ denotes the mean depth value of the respective map. By maximizing the Pearson correlation (minimizing $\mathcal{L}_{depth}$), we enforce the Gaussians to align with the relative geometric structure and relief of the scene without being constrained by absolute depth values at initial stage. This initialization phase provides a coarse starting point for shadow casting, preventing the optimization from collapsing into local minima.

\subsubsection{Gaussian densification}
%\RQ{densification of what?}
\label{sec:densification}

Recovering facade details requires sufficient Gaussian density in occluded regions that are sparsely covered by satellite viewpoints. We therefore adopt Adaptive Density Control (ADC)~\citep{kerbl20233d} to dynamically adjust Gaussian density during optimization. However, standard ADC preserves opacity values during cloning, leading to accumulation in densified regions. To address this, we apply the revised opacity formulation of \citet{rota2024revising}, such that the combined transmittance of two split Gaussians along the viewing ray equal to that of the original Gaussian in the alpha blending: %\RQ{in the depth direction? or planemetrically you enforce the splitted GS opacity summed to the quanity prior to split, the use of alpha make me think the alpha blending in the depth direction}
\begin{equation}
\hat{\alpha} = 1 - \sqrt{1 - \alpha}.
\end{equation}

Given the vast geographical extent and varying GSD inherent to satellite data, we employ scale-based densification via gsplat~\citep{ye2025gsplat} to handle the wide range of surface detail levels across the scene, preventing over-expansion in flat terrains while preserving fine-grained structural detail. %The learning schedule of our method is defined in \Cref{fig:mae_per_iter}.

%\begin{figure}[!htbp]
%    \centering
%    \includegraphics[width=\linewidth]{Figure_others/MAE_per_iter.jpg}
%    \caption{Learning schedule of the geometric optimization stage. The plot illustrates the geometric Mean Absolute Error (MAE) relative to iterations, highlighting the transition from monocular depth supervision (applied during the initial 3,000 iterations) to shadow casting and multi-scale refinement.\RQ{we do not like repetitions in paper, you have this figure in your main figure (figure 1) and then presented here. The main purpose of figure 1 is to show the pipeline, a roadmap }}
%    \label{fig:mae_per_iter}
%\end{figure}

\subsection{Shadow-guided generative refinement}
\label{sec:shadow-guided}

%As described in \Cref{sec:intro}, the limited top-down viewpoint coverage of satellite imagery leaves building facades insufficiently supervised, motivating the integration of generative refinement \RQ{This does not sound like a core motivation, it was demostrated that the generative priors were shown to be effective in improving the texture quality, especially in low-resolution and occluded regions}. To prevent geometric degradation during this process, we extend the shadow casting pipeline established in the geometric optimization stage into the generative refinement stage, conditioning the diffusion model on geometrically calculated shadow maps to preserve geometric accuracy while improving visual fidelity. Details of each component are described in the following subsections.

Generative refinement has been demonstrated to be effective in improving texture quality, particularly in building facades~\citep{lee2025SkyfallGS}. However, diffusion-based refinement tends to degrade geometric accuracy without explicit geometric cues. To address this, we extend the shadow casting pipeline into the generative refinement stage, where shadow structures are preserved during diffusion-based appearance refinement. This enables diffusion models to improve visual fidelity while maintaining the geometry established during the Gaussian optimization process. The following subsections describe each component of this process.

%\begin{figure*}[tb]
%    \centering
%    \includegraphics[width=\linewidth]{Figure_others/Diffusion_outline.jpg}
%    \caption{Pipeline of our shadow-cast diffusion refinement. A generative diffusion model is employed to improve visual fidelity, while introducing stochastic shadow sampling to impose geometric constraints. To prevent distributional drift and maintain radiometric consistency, the resulting pseudo-dataset incorporates original satellite observations, keeping the distribution aligned.\RQ{is it a sub-figure of your figure 1? if so why not directly reference to the figure 1}}
%    \label{fig:diffusion_outline}
%\end{figure*}

\subsubsection{Novel View Sampling}
\label{sec:novel-view}
% what is the criteria for novel-view synthesis?
% why is shadow casting random? 

%Satellite images provide highly constrained observations, where most building surfaces are observed from near-nadir viewpoints under limited illumination conditions. Such observation bias limits the geometric constraints available during generative refinement, particularly for weakly observed regions such as building facades. 
To provide more diverse supervision, we synthesize novel-view images with varying viewpoints and scales using the affine camera generation function from \Cref{eq:generate_camera}. The camera parameters $(t, \phi, \theta, s)$ are randomly sampled from predefined ranges and grids, where $t$ is drawn from a predefined $3 \times 3$ grid in NDC space, $\phi \sim \mathcal{U}(0^\circ, 360^\circ)$, $\theta \sim \mathcal{U}(\theta_{\min}, \theta_{\max})$, and $s \sim \mathcal{U}(s_{\min}, s_{\max})$. This sampling reduces the dependency on the original satellite acquisition geometry and provides diverse geometric cues for optimizing the Gaussian representation.

Once the novel-view camera $\mathbf{F}$ is generated, we render an image from the Gaussians $\mathcal{G}$ optimized in the previous stage. Since novel-view cameras do not contain radiometric parameters $\mathbf{C}$ and $\mathbf{c}$, we transform the Gaussian colors $\mathbf{a}$ into the corresponding camera color space using $\mathbf{C}\mathbf{a}+\mathbf{c}$ and generate modified Gaussians $\mathcal{G}_{modified}$. The rendered albedo image $\mathbf{R}_{albedo}$ is then obtained using $\mathbf{F}$ and $\mathcal{G}_{modified}$ with \Cref{eq:generate_camera}. Because radiometric correction is directly embedded into $\mathcal{G}_{modified}$, the image formation model in \Cref{eq:image_formation} uses $\mathbf{C}=\mathbf{I}$ and $\mathbf{c}=0$ for novel-view rendering.

%\subsubsection{Random shadow casting}

%Original satellite images are acquired under specific solar illumination conditions, providing only a limited set of shadow patterns for geometric supervision. Optimizing Gaussians under a fixed illumination condition risks biasing the geometry toward specific shadow patterns, rather than learning a representation that is consistent across diverse illumination conditions.

After rendering the novel-view image, we compute the shadow visibility mask $\mathbf{V}$ from the Gaussian geometry of $\mathcal{G}_{modified}$. Unlike the original satellite observations that provide only a single illumination configuration, random solar sampling generates diverse shadow patterns while preserving the same underlying geometry. The illumination direction is diversified by randomly sampling a timestamp $\tau \sim \mathcal{U}(\tau_{\min}, \tau_{\max})$ from ten-year window centered on the acquisition date, and compute the corresponding solar elevation $\theta_{sun}$ and solar azimuth $\phi_{sun}$ using a solar position model, given the geographic latitude $\phi_{lat}$ and longitude $\lambda_{lon}$:
\begin{equation}
\phi_{sun}, \theta_{sun} = \mathcal{S}_{solar}(\tau, \phi_{lat}, \lambda_{lon}).
\label{eq:solar_direction}
\end{equation}

To avoid extreme low-angle illumination and unstable shadow formation, we constrain the solar elevation angle to $\theta_{sun} \in [45^\circ, 90^\circ]$. If the sampled timestamp violates this constraint, we resample $\tau$ until a valid configuration is obtained. The resulting solar direction is used to generate a shadow visibility mask $\mathbf{V}$ from Gaussian positions with \Cref{subsubsec:shadow_casting}. This shadow visibility mask is then cast onto the rendered albedo image following \Cref{eq:image_formation} to produce the shadow-guided input for the diffusion model.

\subsubsection{Diffusion-based refinement}
\label{sec:diffusion_refinement}

%We employ generative refinement to improve visual fidelity while preserving underlying scene geometry. To achieve this, we apply a diffusion model to shadow-cast rendered images rather than plain albedo images. 

%After random view sampling\RQ{I thought this is not a random view sampling, but follows a circular pattern?}, we employ a diffusion model on shadow-cast images to improve visual fidelity while preserving the underlying scene geometry. Since our shadow maps are geometrically calculated from the underlying scene structure (\Cref{subsubsec:shadow_casting}), conditioning the diffusion model on shadow-cast images prevents the generation of illumination patterns inconsistent with the scene geometry (\Cref{fig:refined_image}). 

After novel view sampling and shadow casting, we employ a diffusion model on the shadow-cast rendered image (as in \Cref{sec:novel-view}) to improve visual fidelity while preserving the underlying scene geometry. The shadow-cast image is passed as a conditioning input alongside a text prompt, producing a diffusion-refined image $\mathbf{R}_{\text{diff}}$. During generative refinement, we adopt FLUX.2~\citep{flux-2-2025} as our diffusion backbone. The prompts are designed to emphasize surface detail enhancement rather than introducing structural modifications:
\begin{promptbox}
\small\texttt{
"Restored satellite photo, 8k ultra-sharp resolution. Strictly rectilinear architectural lines and cleaner textures. Zero vehicles, empty roads, and vacant parking lots. Razor-sharp facades. Crisp building boundaries. Google Earth style. Pitch black background for no-data areas, zero atmospheric haze, no sky, no clouds. Clean and smooth but crisp surfaces. Shadows are perfectly linear and semi-transparent. Maintain consistent radiometric exposure and original lighting levels from the input. Zero sensor noise, anti-aliased urban concrete. Strictly aligned with input geometry."
}
\end{promptbox}

\begin{figure}[!h]
    \centering
    \begin{subfigure}[b]{0.4\linewidth}
        \centering
        \includegraphics[width=\linewidth]{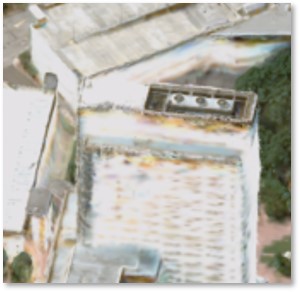}
        \caption{Rendered albedo}
    \end{subfigure}
    \begin{subfigure}[b]{0.4\linewidth}
        \centering
        \includegraphics[width=\linewidth]{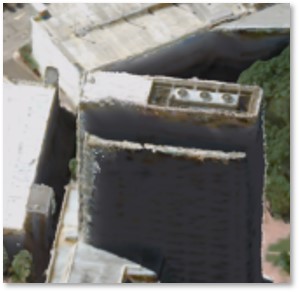}
        \caption{Shadow-cast image}
    \end{subfigure}

    \begin{subfigure}[b]{0.4\linewidth}
        \centering
        \includegraphics[width=\linewidth]{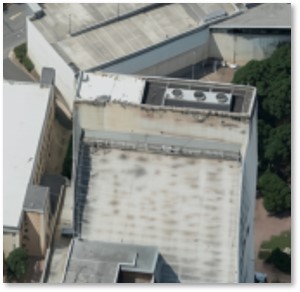}
        \caption{Refined albedo}
    \end{subfigure}
    \begin{subfigure}[b]{0.4\linewidth}
        \centering
        \includegraphics[width=\linewidth]{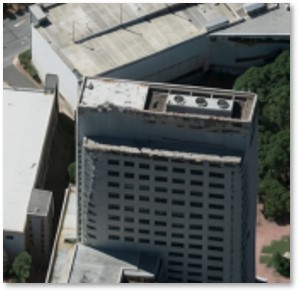}
        \caption{Refined shadow-cast image}
    \end{subfigure}

    \caption{Visualization of diffusion refinement combined with shadow casting. The diffusion model enhances texture details while preserving the shape of cast shadows.}
    \label{fig:refined_image}
\end{figure}

The diffusion model is not explicitly optimized to preserve shadow boundaries. However, instead of refining a shadow-free appearance representation, we provide the diffusion model with an input image that already incorporates geometrically calculated shadows derived from the sun visibility mask $\mathbf{V}$. Therefore, the refinement process enhances surface appearance while maintaining the shadow present in the input representation. As shown in \Cref{fig:refined_image}, the refined images improve local texture details while preserving consistent shadow boundaries.

%\RQ{i think you need to state clearer about this one, does the diffusion model know what are shadows in the input image? say, did not seperately feed in the shadow and albeod as two conditions for the diffusion model, or you rendered the shadows already and then give this single image as the condition to the diffusion model. If the latter, how it is guaranteed that the shadows remain the same if the diffusion does not specifically understand where the region of the shadow is in that image, a justification of why this can be the case, should be noted.}

Since $\mathbf{R}_{\text{diff}}$ remains fixed until the diffusion-refined supervision targets are regenerated in the next refinement cycle, the geometry of $\mathcal{G}_{modified}$ is optimized using image supervision aligned with the shadow structure rendered from the current Gaussian geometry. Gradients through the shadow visibility mask $\mathbf{V}$ continue to provide a valid geometric cue for updating $\mathcal{G}_{modified}$. The remaining discrepancy is primarily attributed to texture and color variations, which are handled by the albedo component $\mathbf{R}_{\text{albedo}}$ instead.

%\RQ{in general when you describe technical terms, I generally need you to be amguity free}

Furthermore, diffusion-refined images may exhibit color distributional shifts from the original satellite observations. To mitigate this effect, we mix $N_{pseudo}$ diffusion-refined images with $N_{original}$ original satellite images during each optimization step, aligning the refinement process to real sensor observations. Each sample is paired with its corresponding camera model $\mathbf{F}$ and sun camera model $\mathbf{F}_{sun}$.

%\RQ{the title of this section is not very self-explanatory, there must be a need for you to do this, the title is formed as a named of a solution without a problem to solve}

\subsection{Loss functions and optimization}

The proposed pipeline is trained through two sequential stages. In the geometric optimization stage, all loss terms are active, with $\mathcal{L}_{depth}$ applied only during the first 3,000 iterations. In the shadow-guided generative refinement stage, $\mathcal{L}_{depth}$ is deactivated and $\mathcal{L}_{img}$ is applied to the mixed pseudo-dataset $\mathcal{D}_{pseudo}$ consisting of diffusion-refined and original satellite images. The total objective function is:
\begin{equation}
\mathcal{L}_{total} = \lambda_{img}\mathcal{L}_{img} + \lambda_{depth}\mathcal{L}_{depth} + \mathcal{L}_{reg}
\end{equation}

Following the standard 3DGS and 2DGS pipelines  \citep{kerbl20233d, huang20242d}, the image reconstruction loss $\mathcal{L}_{img}$ is formulated as a combination of the $\mathcal{L}_1$ loss and the Structural Similarity (SSIM) term:
\begin{equation}
\mathcal{L}_{img} = (1 - \lambda_{ssim})\mathcal{L}_1 + \lambda_{ssim}\mathcal{L}_{ssim} \label{eq:loss_img} ,
\end{equation} 
where $\lambda_{ssim}$ balances the pixel-wise intensity difference and the structural consistency. The regularization term $\mathcal{L}_{reg}$ combines geometric supervision terms:
\begin{equation}
\mathcal{L}_{reg} = \lambda_{ent}\mathcal{L}_{ent} + \lambda_{shadow}\mathcal{L}_{shadow} + \lambda_{norm}\mathcal{L}_{norm} + \lambda_{dist}\mathcal{L}_{dist} \label{eq:loss_reg}.
\end{equation}

At each iteration of the geometric optimization stage, an additional backpropagation step is performed using $\mathcal{L}_{refine}
$ defined in \Cref{sec:method:multiscale}.

\section{Experiments and analysis}
\label{sec:experiments}

\subsection{Implementation}
\label{sec:implementation}

% Bundle Adjustment & DSM generation
For bundle adjustment, we utilize the Sat Bundle Adjust library  \citep{ipol.2021.352}. Features are extracted from the SIFT \citep{lowe1999object} descriptors with epipolar-based matching, constrained by RANSAC \citep{fischler1981random} threshold of 0.3 to eliminate outliers. After bundle adjustment, DSM generation was conducted with the ASP \citep{beyer2018ames} pipeline. Based on the refined RPC parameters from the bundle adjustment stage, we select optimal stereo pairs by filtering the intersection angles between $5^\circ$ and $30^\circ$. For each selected pair, we perform dense stereo matching using the MGM (More Global Matching) \citep{facciolo2015mgm} algorithm.

% Depth Anything V2
%\RQ{how do you respond to a reviewer's question why not use depthanything v3?}
After DSM generation, the DSM is converted into 2DGS and geometrically optimized. For generating the monocular depth priors, we utilized the Depth Anything V2 \citep{depth_anything_v2} ViT-Large variant model instead of recent V3 version. We empirically observe that V3 produces less stable depth estimates for satellite imagery, likely due to the domain gap between its training distribution and near-nadir satellite views. Since the model output represents relative disparity, we invert the estimated values to obtain a relative depth map, which is then utilized as a monocular depth prior $\mathcal{L}_{depth}$ during the initial $3,000$ iterations of the geometric optimization stage.

For generating shadow map, we utilized the shadow density $\rho$ value of 0.4 and initial ambient value $\mathbf{L}_{a}$ of 0.5 for optimization. Since our pipeline is initialized with the solid DSM-based pipeline, we utilized the relatively high value of ambient to reduce the instability of shadows on initial training iterations. 

% Pseudo Camera Generation
Based on \Cref{eq:generate_camera}, we sample $N_{pseudo}=180$ views on shadow-guided generative refinement. We set the target vector $t$ from a $3 \times 3$ grid $\{-0.25, 0.0, 0.25\}^2$ to ensure that the generative refinement could be held evenly on all scenes. Since our camera models are affine, the distance between camera and the result image doesn't affect the output. However, to ensure that all Gaussians are placed in front of the affine camera plane, we set the distance value $d$ as 5 in NDC space. The camera elevation value $\theta$ was sampled from $[5.0^\circ, 86.0^\circ]$ and azimuth $\phi$ was sampled from $[-180.0^\circ, 180.0^\circ]$ with fixed increments to cover the entire visibility hemisphere. The zoom scale $s \sim \mathcal{U}(4.0, 6.0)$, allowing the diffusion model to refine multi-scale high-frequency details. For sun camera random date and time $\tau$ was simulated within a ten-year window (2014--2024).

% Diffusion refinement
After generating the rendered image and shadow-cast image pairs with a resolution of $1024 \times 1024$, we apply the diffusion model to refine their appearance. Following the refinement strategy described in \Cref{sec:diffusion_refinement}, we designed prompts to encourage texture enhancement while minimizing structural modifications and preserving the shadow layout.
%\RQ{it there a reference or rationale you want to talk about why this prompt? Also, i think what prompt used should be in your methodology section where you talk about why  this prompt and in the implementation section you can always relay to that in the methodology}

% Densification & learning schedule
The geometric optimization stage was run for a total of 12,000 iterations. Monocular depth supervision was applied only for the first 3,000 iterations, while shadow casting was enabled from the start, as the pipeline is initialized from a solid photogrammetric base. Loss weights were set as follows: $\lambda_\text{img} = 1.0$, $\lambda_\text{depth} = 0.05$, $\lambda_\text{ssim} = 0.2$, $\lambda_\text{ent} = 0.05$, $\lambda_\text{shadow} = 0.01$, and $\lambda_\text{norm} = \lambda_\text{dist} = 0.05$. Similarly, learning rates were tuned to prioritize geometry over image refinement: 1e-4 for color correction, 2.5e-3 for colors, 1e-2 for ambient $\mathbf{L}_a$, 2.5e-2 for Gaussian means (positions), 7.5e-2 for opacities, and 1e-3 for rotations and scales. Densification was applied every 100 iterations according to the 2DGS scale gradient, and Gaussians with opacity below 5e-3 were pruned.

The Gaussians on shadow-guided generative refinement stage were optimized for a total of 10,000 iterations, repeated for 5 refinement steps. Monocular depth supervision and shadow entropy were disabled in this stage. Learning rates were reduced for Gaussian means to 1e-2 to minimize fluctuations. Densification started at iteration 2,000 and was applied every 2,000 iterations. A single NVIDIA RTX 6000 ADA Generation GPU was used to accelerate the optimization.

\subsection{Datasets}

%\RQ{the source is google earth, studio is a software to facilitate the access, i think you should call it google earth image, unless this becomes a convention by other papers}
 
We utilized two primary datasets in our experiments: DFC2019 and IARPA2016. DFC2019~\citep{le2019data, bosch2019semantic} covers Jacksonville, Florida, and Omaha, Nebraska. The dataset contains 2048 $\times$ 2048 pixel images at a 35 cm/pixel resolution. Ground truth LiDAR data is available for a 512 $\times$ 512 pixel area at 0.5 m/pixel resolution. Metadata such as Rational Polynomial Coefficients (RPCs), acquisition dates, and solar angles are also provided. For visual assessment, we follow the evaluation protocol of Skyfall-GS~\citep{lee2025SkyfallGS}, incorporating lower-view Google Earth images using 300 images for each of four JAX sites (JAX-004, JAX-068, JAX-214, and JAX-260) and three Omaha sites (OMA-203, OMA-212, OMA-315). These Google Earth images are strictly used for evaluation and not for training. %\RQ{ditto problems identified earlier, i suggest you can state that: The experimental protocol is consistent with Skyfall-GS and other works (ofc if any)}.

IARPA2016 \citep{bosch2016multiple} provides approximately 50 commercial satellite images covering 100 $\text{km}^2$ along with corresponding 3D point clouds. Digital Surface Models (DSMs) were derived from the LiDAR data to serve as ground truth for geometric evaluation. To ensure evaluation consistency, we generated corresponding lower-view Google Earth images similar to the DFC2019 setup. As with DFC2019, these Google Earth images were used only for evaluation purposes and excluded from any training procedure. %\RQ{you should provide relevant citations when talking about the dataset}

\subsection{Evaluation metrics}

We categorize our evaluation criteria into three distinct groups to assess the quality of our reconstruction from multiple perspectives. To evaluate visual fidelity, we used PSNR \citep{huynh2008scope} and SSIM \citep{wang2004image} compared to Google Earth images. In particular, we employ histogram-matched PSNR to compensate for global illumination differences, focusing on structural accuracy rather than radiometric variations.

However, since these metrics are often sensitive to minor pixel-level misalignment common in satellite imagery, we primarily incorporate CW-SSIM \citep{sampat2009complex}, which leverages complex wavelet transform coefficients to ensure invariance against small geometric distortions. 
\begin{equation}
\tilde{S}(\mathbf{c}_x, \mathbf{c}_y) = \frac{2 \left| \sum_{i=1}^{N} c_{x,i} c_{y,i}^* \right| + K}{\sum_{i=1}^{N} |c_{x,i}|^2 + \sum_{i=1}^{N} |c_{y,i}|^2 + K} ,
\label{eq:cw_ssim}
\end{equation}
where $c^*$ denotes the complex conjugate and $K$ is a small positive constant to ensure numerical stability. Furthermore, LPIPS \citep{zhang2018unreasonable} is included to capture high-level perceptual similarity by comparing deep feature activations, providing a more human-centric assessment of visual fidelity. All criteria are calculated using lower-view Google Earth images as ground truth. 

To assess generative quality and distribution alignment, we utilize metrics based on the CLIP \citep{radford2021learning} backbone, which aligns more closely with human perception than traditional InceptionV3-based methods. We report FID-CLIP \citep{kynkaanniemi2022role} to measure the Fréchet Distance between feature distributions and CMMD \citep{jayasumana2024rethinking}, a CLIP Maximum Mean Discrepancy-based metric, to provide a robust distance measure between sets of image embeddings.

%\RQ{this wording is not accuracy, you can call geometric assessment}

%For geometric assessment, we compare our reconstructed Digital Surface Model (DSM) against LiDAR ground truth. The altitude map is rendered from a nadir perspective ($512 \times 512$ pixels, 0.5m resolution for DFC2019 and $863 \times 863$, 0.3m resolution) to calculate the registration-based Mean Absolute Error ($\text{MAE}_{reg}$) :
%\begin{equation}
%\text{MAE}_{reg} = \min_{\mathbf{o}} \frac{1}{N} \sum_{i=1}^{N} \left| \hat{M}(x_i + o_x, y_i + o_y) + o_z - M(x_i, y_i) \right|,\label{eq:maereg_fixed}
%\end{equation}
%\RQ{this is problematic, because this will be strongly affect by outliers and floaters, for example, if you have a very good result but just one a couple of points flying really high, the overall MAE will be poor, so your method should consider eliminate very large outliers when calculating the offset}
%where $M$ is the ground truth DSM, $\hat{M}$ is the reconstructed DSM, and $\mathbf{o}=(o_x, o_y, o_z)$ represents the 3D translation offset where minimize $MAE_{reg}$. This metric effectively compensates for both horizontal and vertical registration shifts, ensuring that geometric accuracy is evaluated independently of global alignment errors.

For geometric assessment, we compare our reconstructed Digital Surface Model (DSM) against LiDAR ground truth. The altitude map is rendered from a nadir perspective ($512 \times 512$ pixels, 0.5m resolution for DFC2019 and $863 \times 863$, 0.3m resolution for IARPA2016) to calculate the registration-based Mean Absolute Error ($\text{MAE}_{reg}$). To mitigate the influence of outliers and floating artifacts, the error values are clipped to the range of $[-10, 10]$ meters prior to evaluation:
\begin{equation}
\text{MAE}_{reg} = \min_{\mathbf{o}} \frac{1}{N} \sum_{i=1}^{N} \left| \hat{M}(x_i + o_x, y_i + o_y) + o_z - M(x_i, y_i) \right|,\label{eq:maereg_fixed}
\end{equation}
where $M$ is the ground truth DSM, $\hat{M}$ is the reconstructed DSM, and $\mathbf{o}=(o_x, o_y, o_z)$ represents the 3D translation offset that minimizes $\text{MAE}_{reg}$. This metric effectively compensates for both horizontal and vertical registration shifts, ensuring that geometric accuracy is evaluated independently of global alignment errors.

\subsection{Experimental results}
Our method is compared against several state-of-the-art baselines. For geometric comparison, we included Sat-NGP \citep{billouard2024sat}, EOGS \citep{aira2025gaussian}, Skyfall-GS \citep{lee2025SkyfallGS}, GU-GS \citep{ding2026gu} and ASP \citep{beyer2018ames}. Since ASP, Sat-NGP, EOGS, and GU-GS provide their own DSM generation procedures, we follow the original implementation of each method to obtain reconstructed DSMs. For Skyfall-GS, we generate the DSM using the same nadir-view altitude rendering procedure adopted in our evaluation pipeline. For visual comparison, we include EOGS, Mip-Splatting \citep{yu2024mip}, and Skyfall-GS. These baselines allow us to evaluate the effectiveness of the proposed refinement strategy from both geometric and visual perspectives. \Cref{sec:results:visual} presents the assessment of visual fidelity, followed by the assessment of geometric accuracy in \Cref{sec:results:geometric}.

\subsubsection{Assessment of visual fidelity}
\label{sec:results:visual}

\begin{figure*}[p]
    \centering
    \small
    
    \newcommand{\myimg}[1]{%
        \includegraphics[width=\linewidth, keepaspectratio, trim=0 10 0 10, clip]{#1}%
    }
    \def\colw{0.18\linewidth} 

    % Header Row
    \makebox[\colw]{Google Earth}
    \makebox[\colw]{EOGS}
    \makebox[\colw]{Mip-Splatting}
    \makebox[\colw]{Skyfall-GS}
    \makebox[\colw]{Ours} \\
    \vspace{1mm}

    % --- JAX Section ---
    \rotatebox{90}{\makebox[0.08\linewidth][c]{\footnotesize JAX-068}}
    \begin{subfigure}{\colw}\myimg{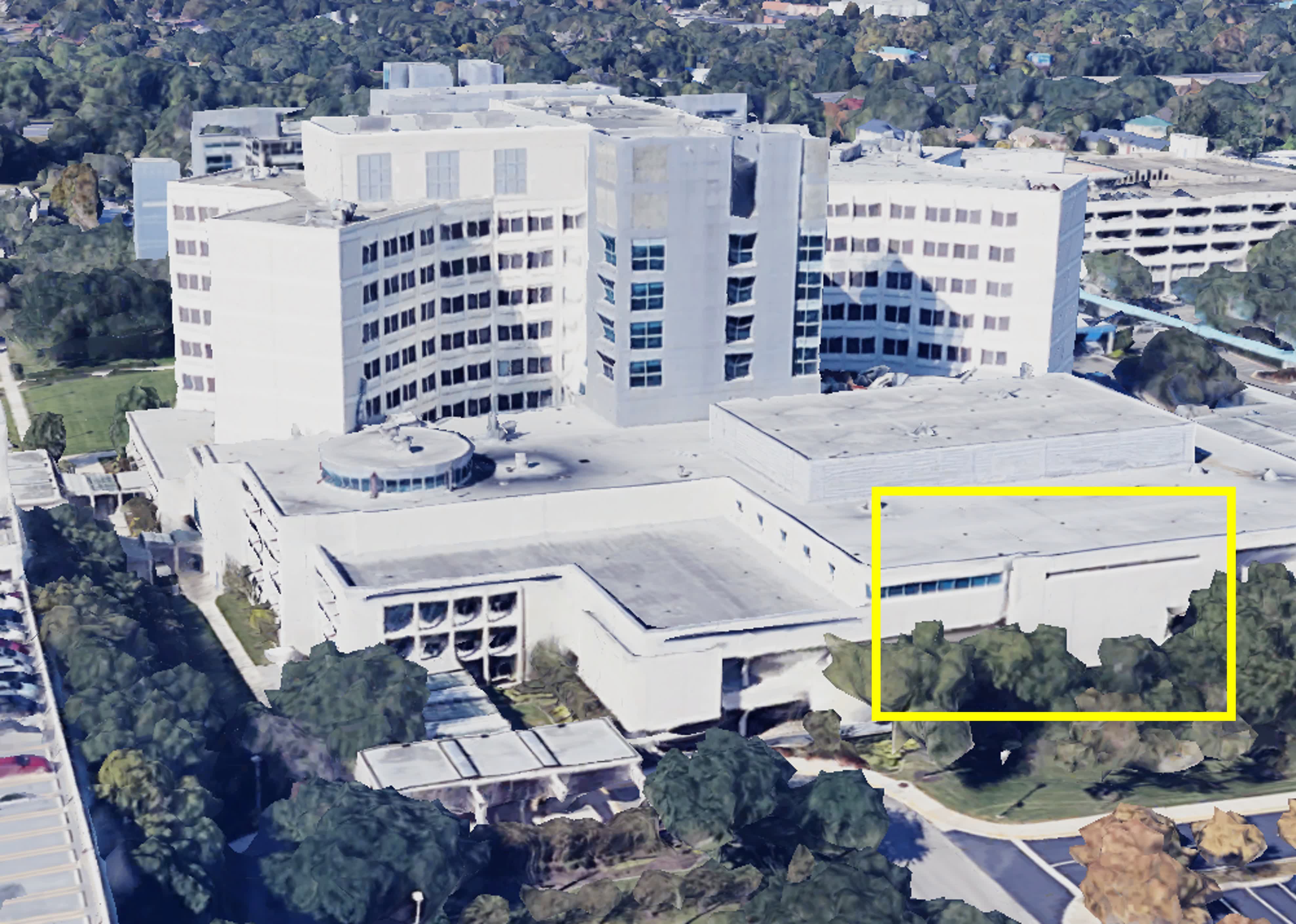}\end{subfigure}
    \begin{subfigure}{\colw}\myimg{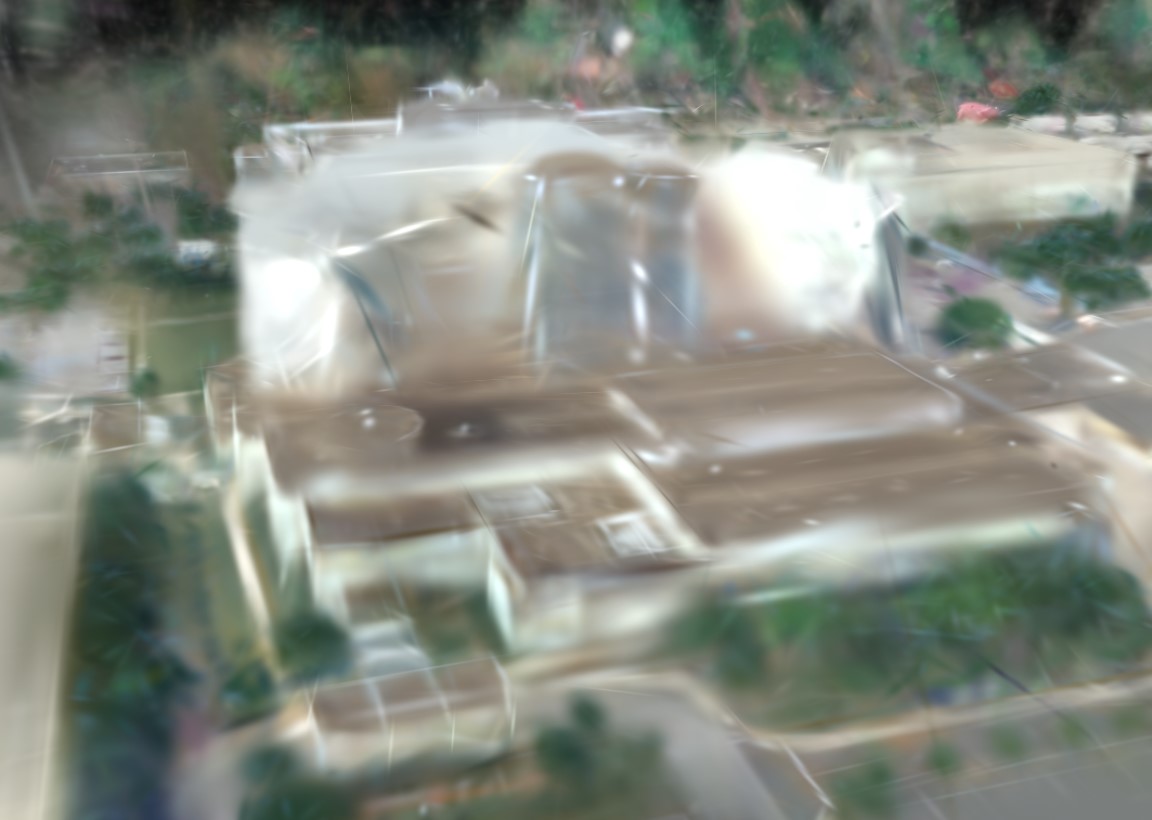}\end{subfigure}
    \begin{subfigure}{\colw}\myimg{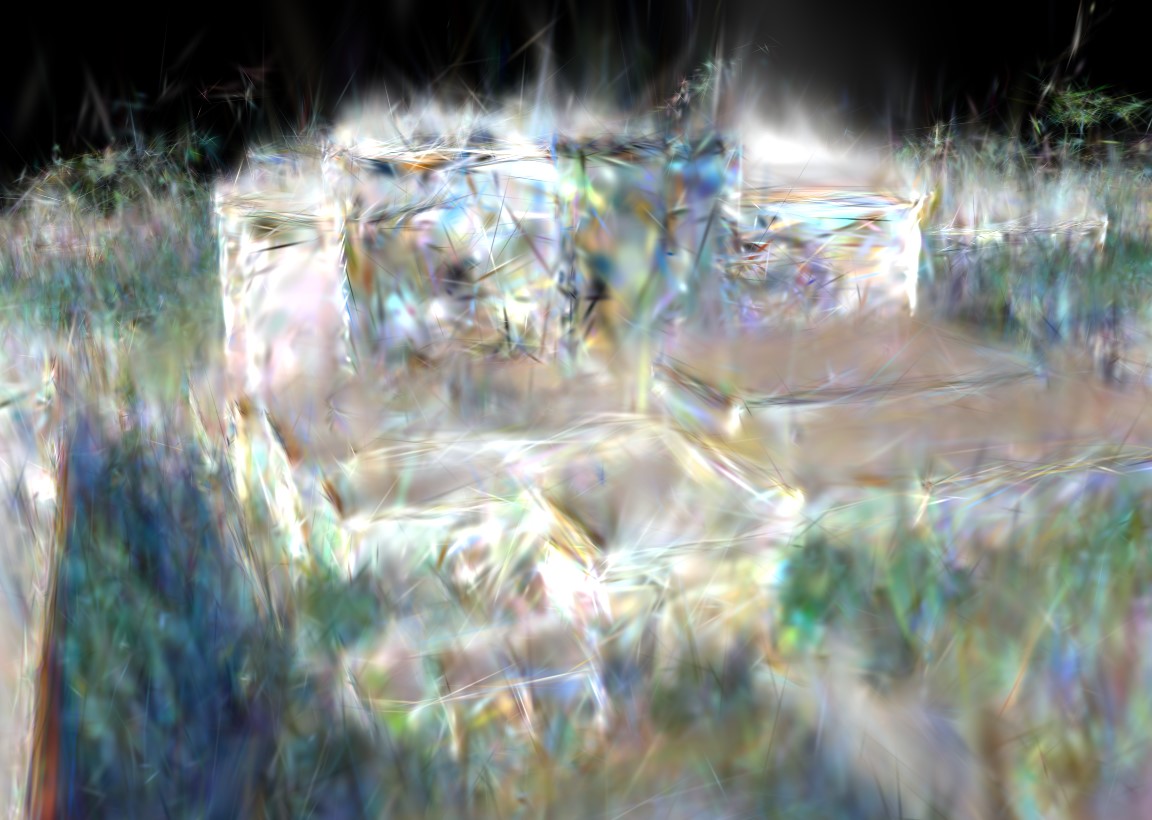}\end{subfigure}
    \begin{subfigure}{\colw}\myimg{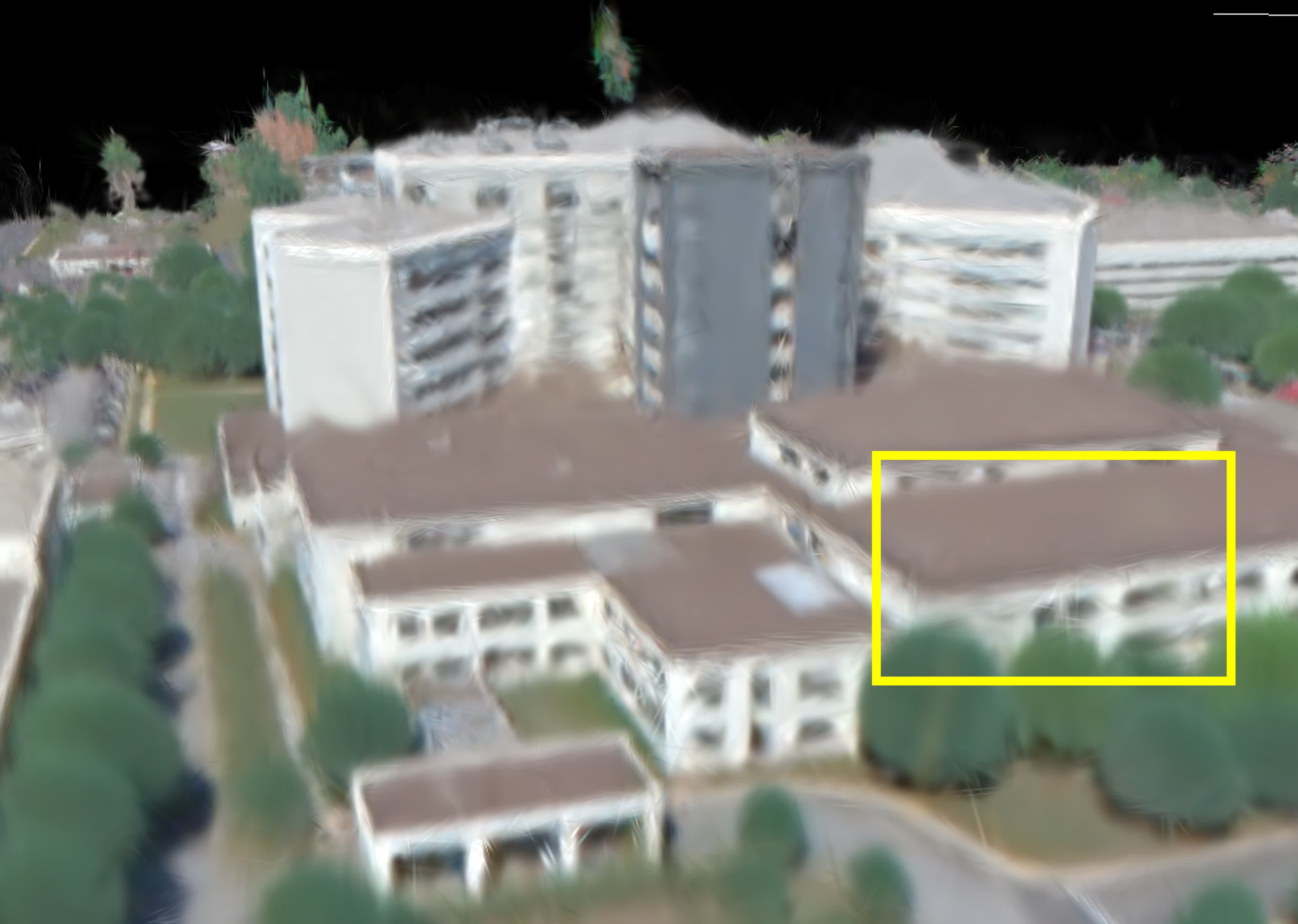}\end{subfigure}
    \begin{subfigure}{\colw}\myimg{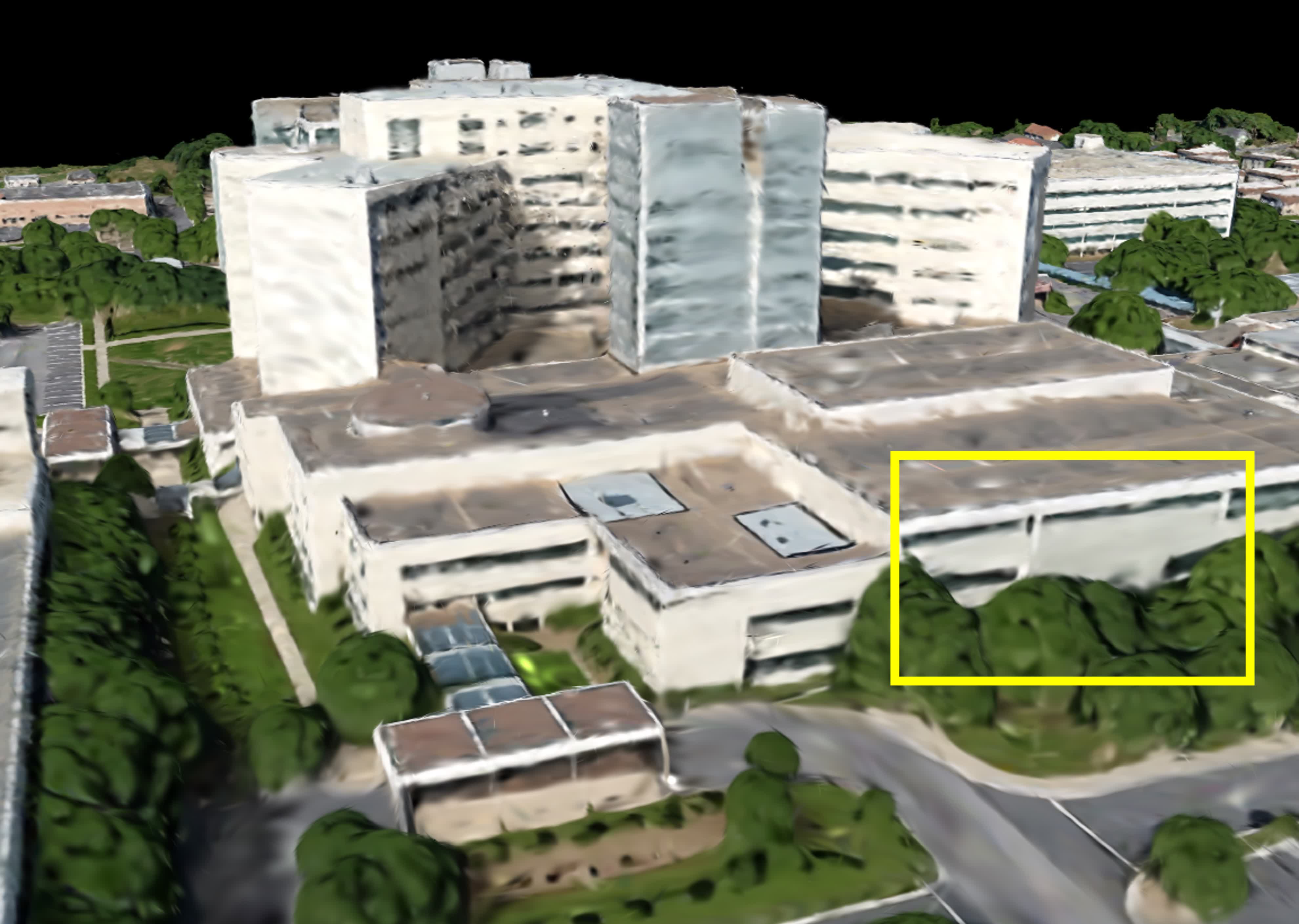}\end{subfigure} \\
    \vspace{0.5mm}

    \rotatebox{90}{\makebox[0.08\linewidth][c]{\footnotesize JAX-214}}
    \begin{subfigure}{\colw}\myimg{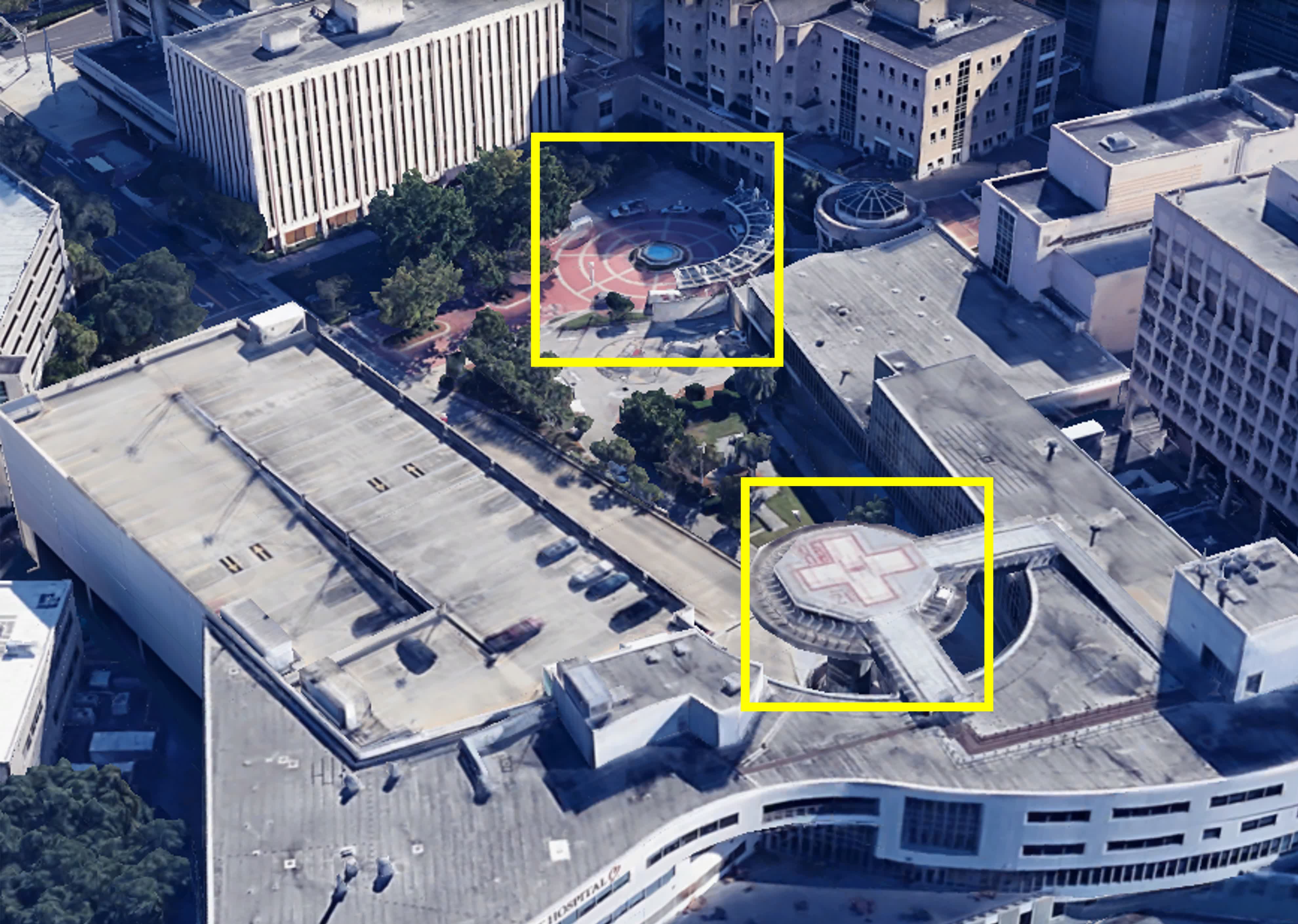}\end{subfigure}
    \begin{subfigure}{\colw}\myimg{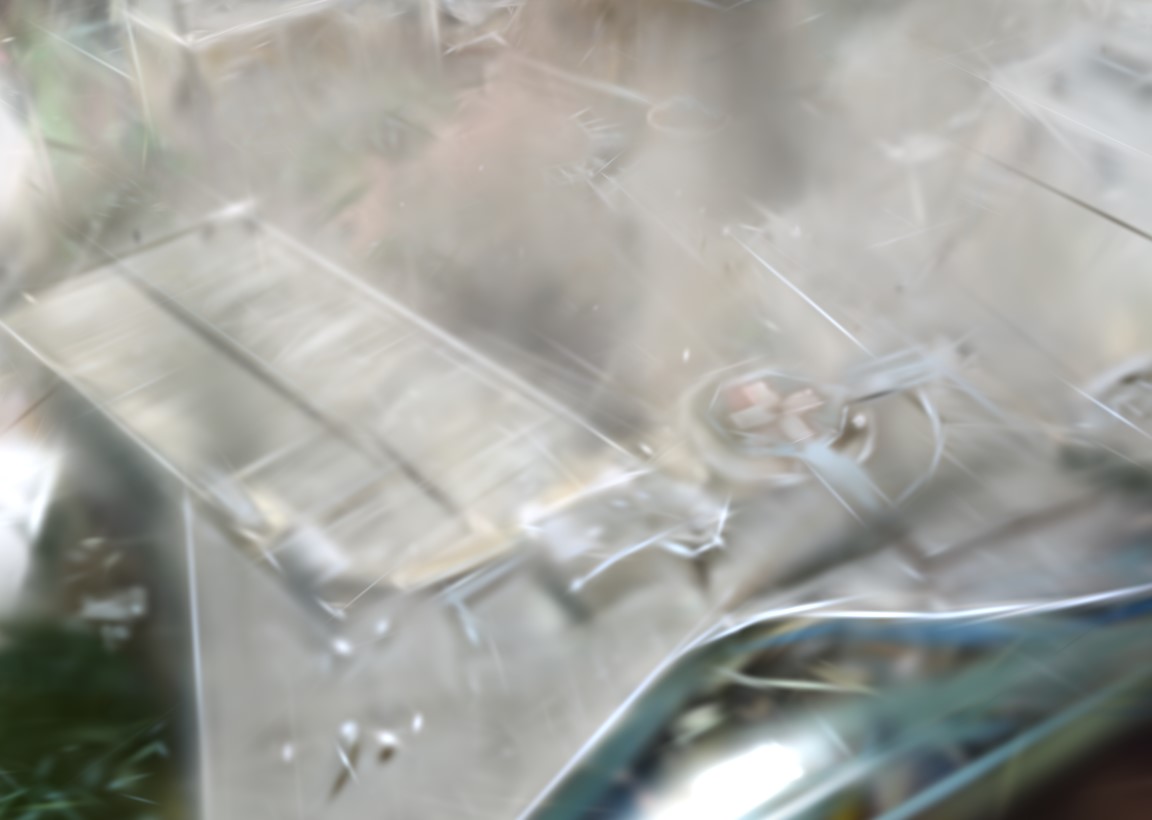}\end{subfigure}
    \begin{subfigure}{\colw}\myimg{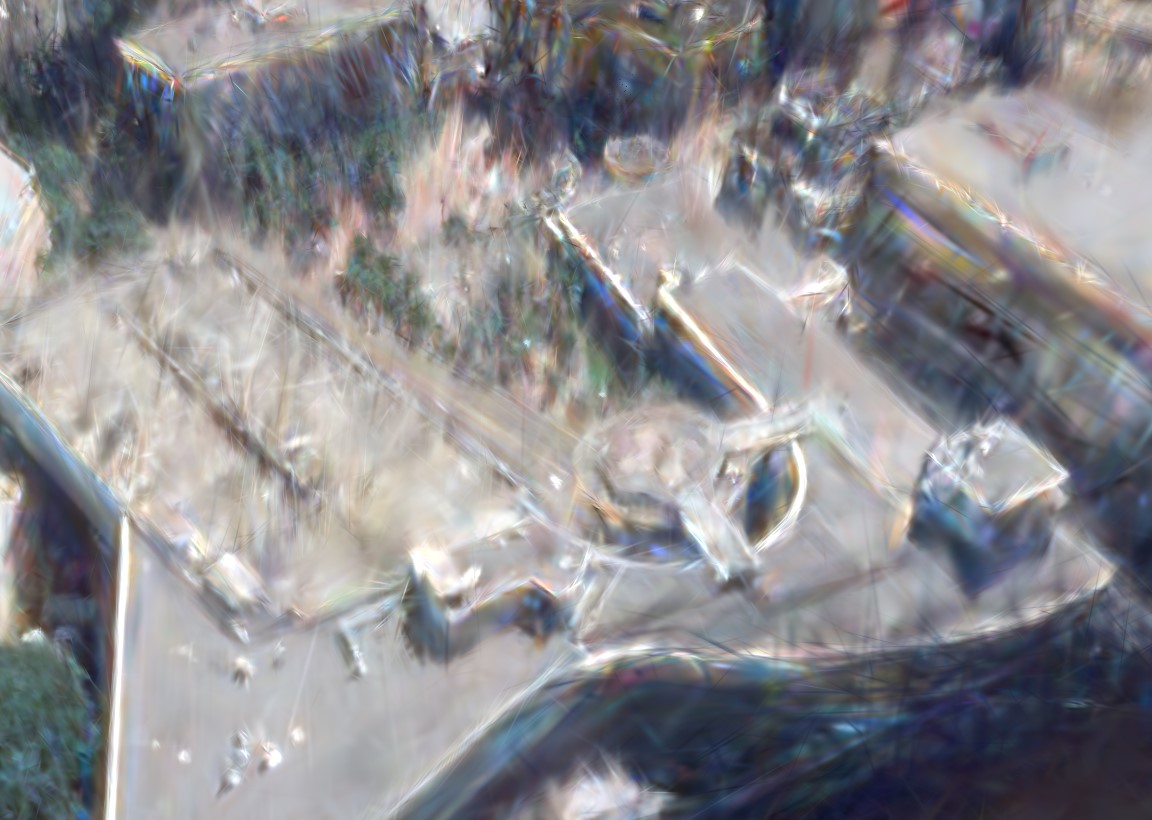}\end{subfigure}
    \begin{subfigure}{\colw}\myimg{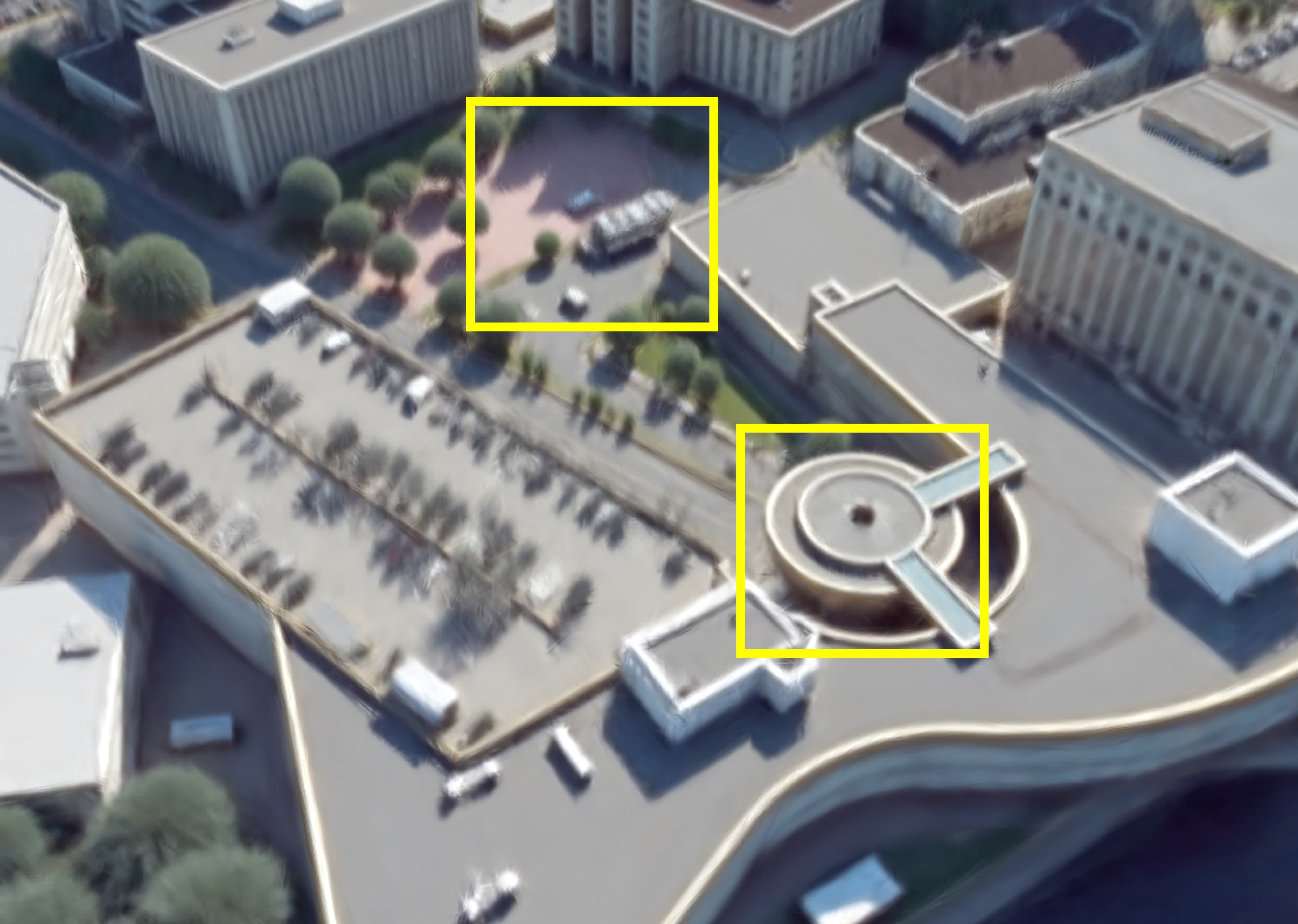}\end{subfigure}
    \begin{subfigure}{\colw}\myimg{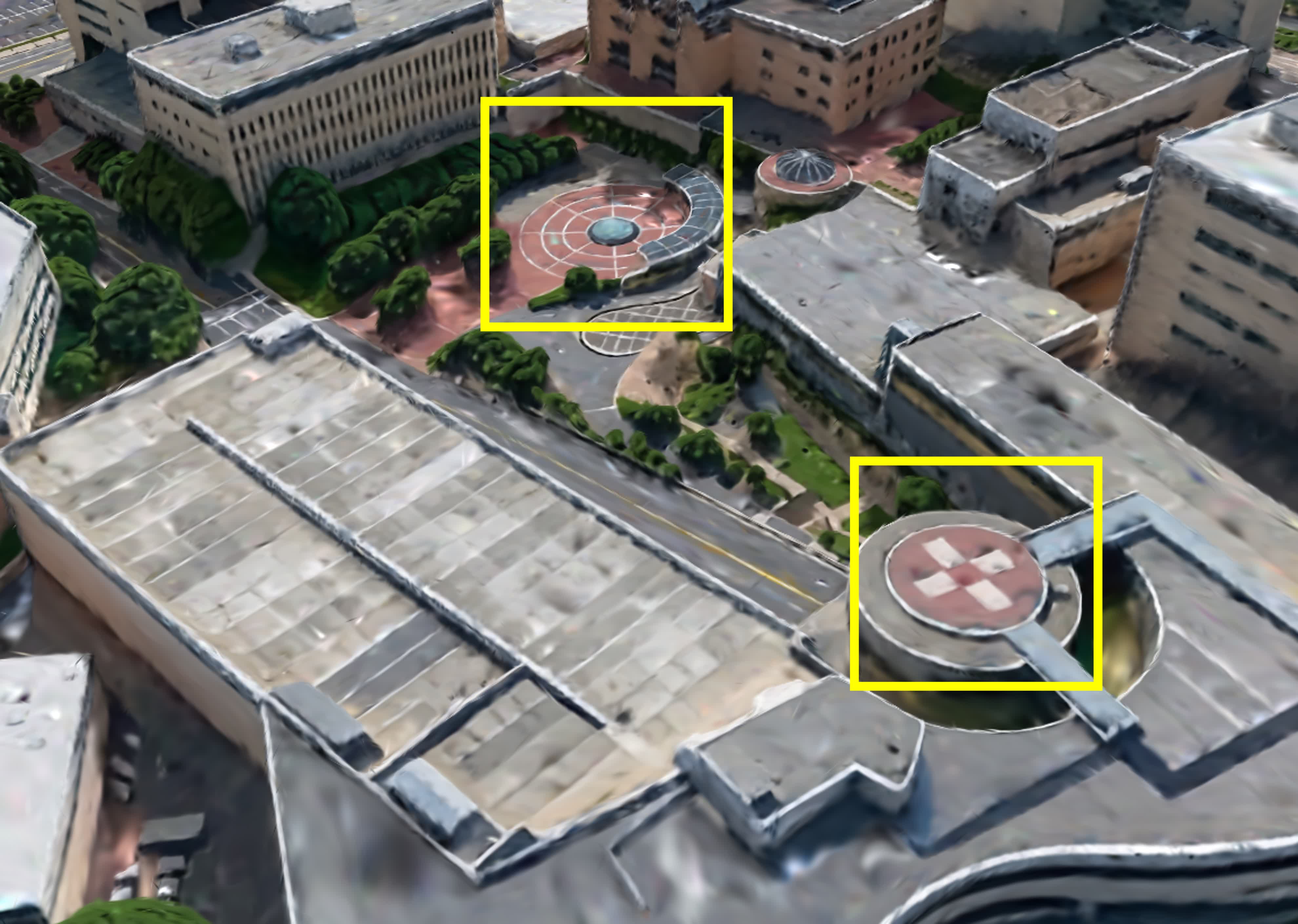}\end{subfigure} \\
    \vspace{0.5mm}

    \rotatebox{90}{\makebox[0.08\linewidth][c]{\footnotesize JAX-260}}
    \begin{subfigure}{\colw}\myimg{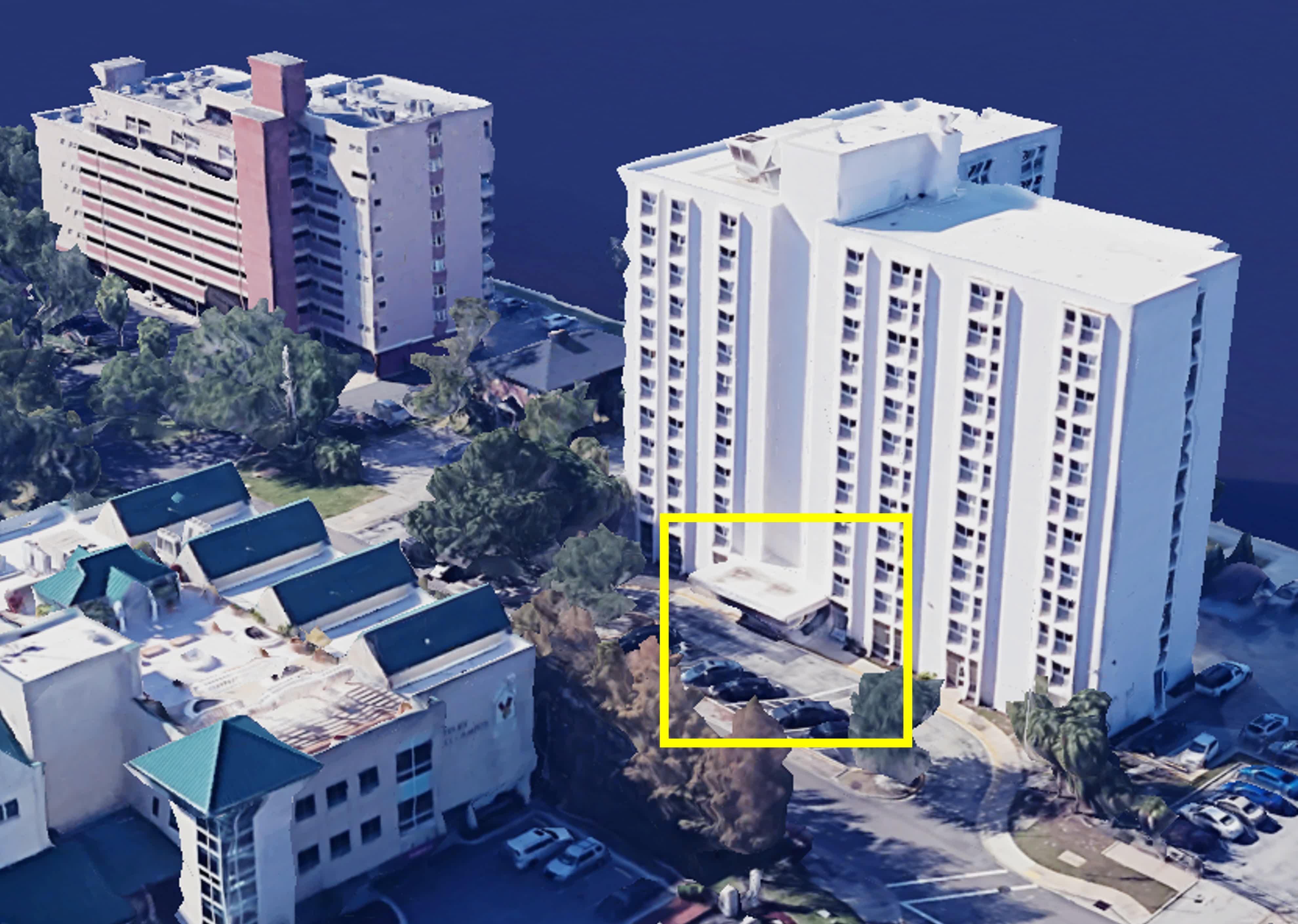}\end{subfigure}
    \begin{subfigure}{\colw}\myimg{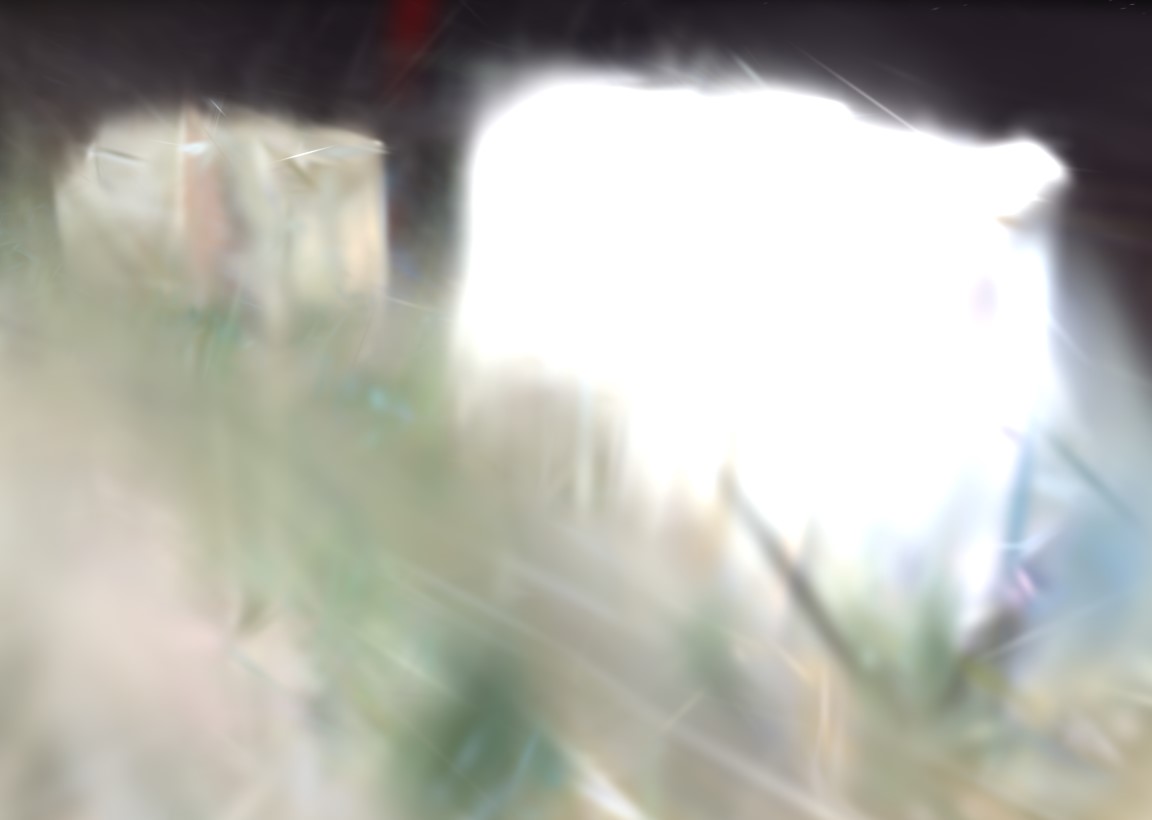}\end{subfigure}
    \begin{subfigure}{\colw}\myimg{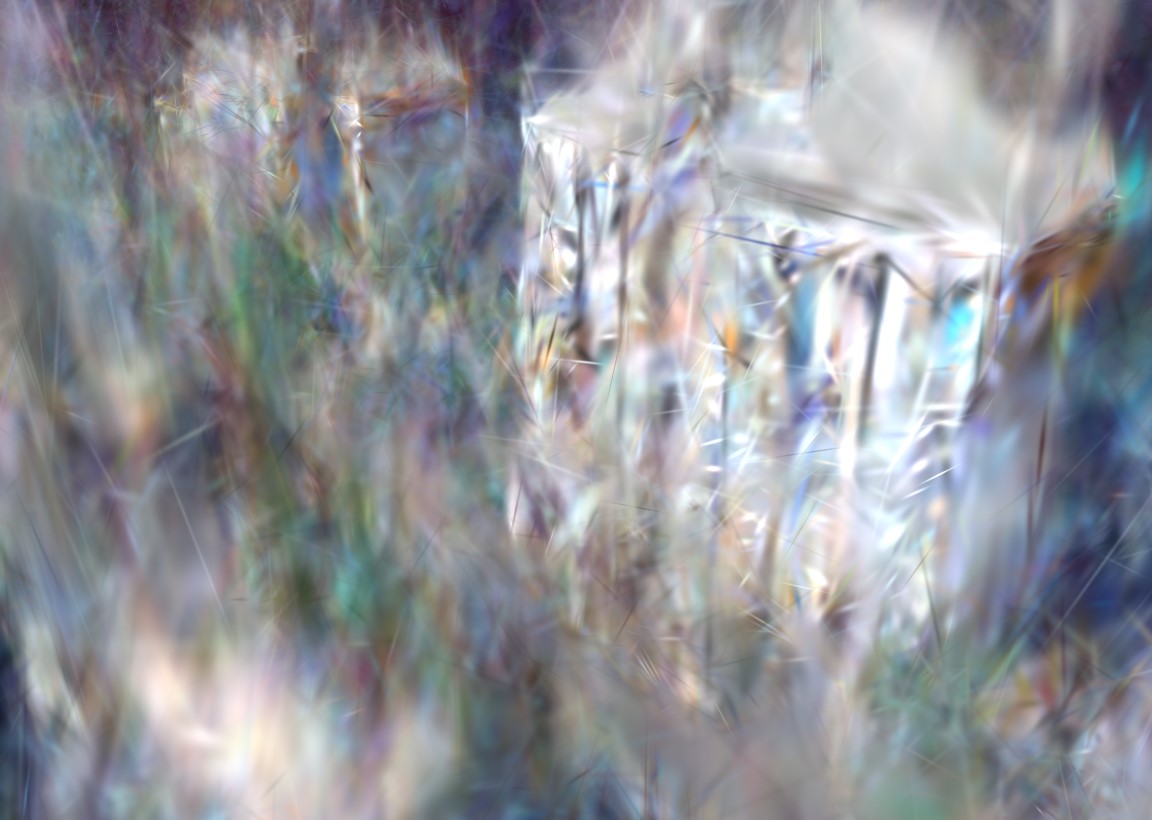}\end{subfigure}
    \begin{subfigure}{\colw}\myimg{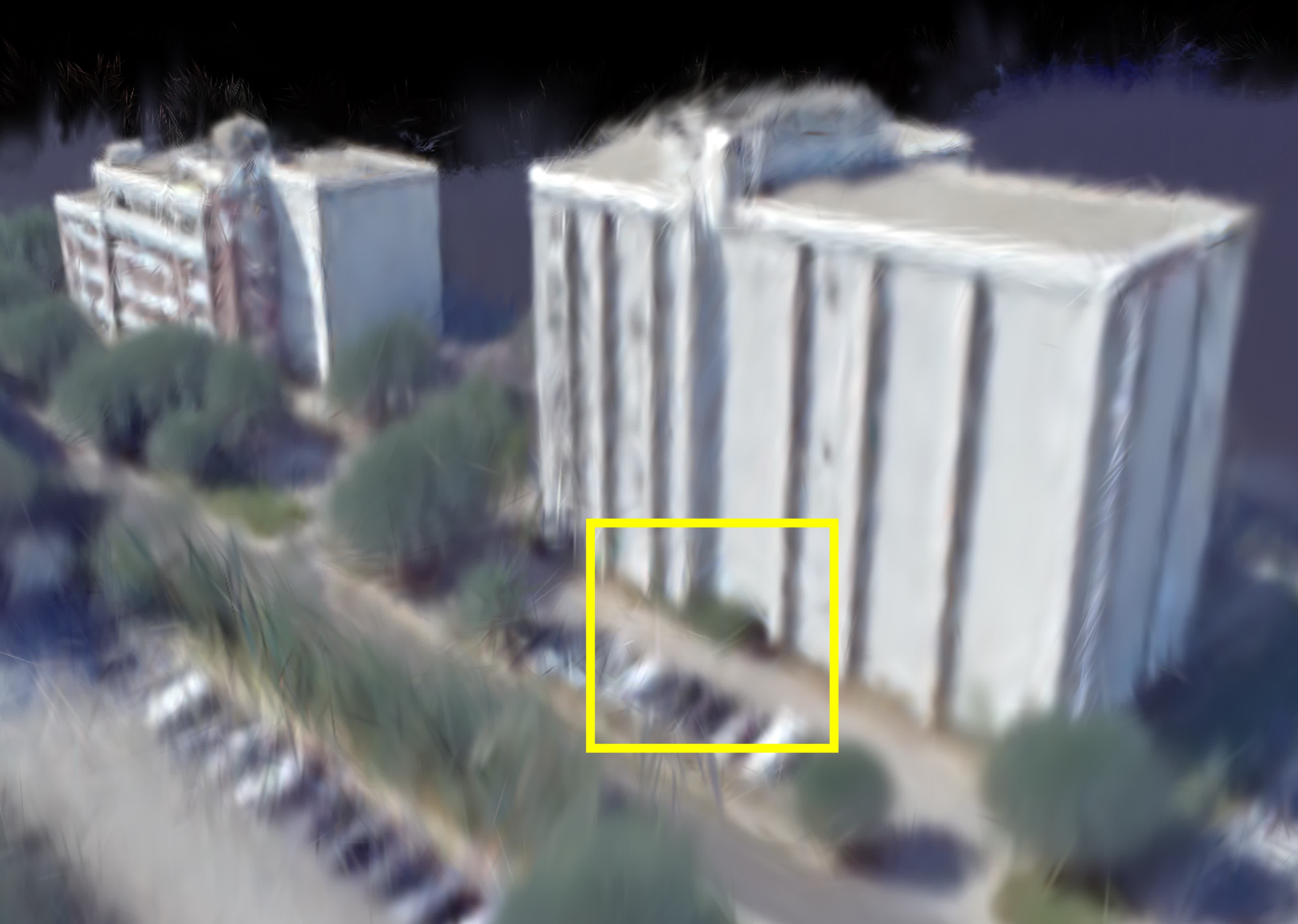}\end{subfigure}
    \begin{subfigure}{\colw}\myimg{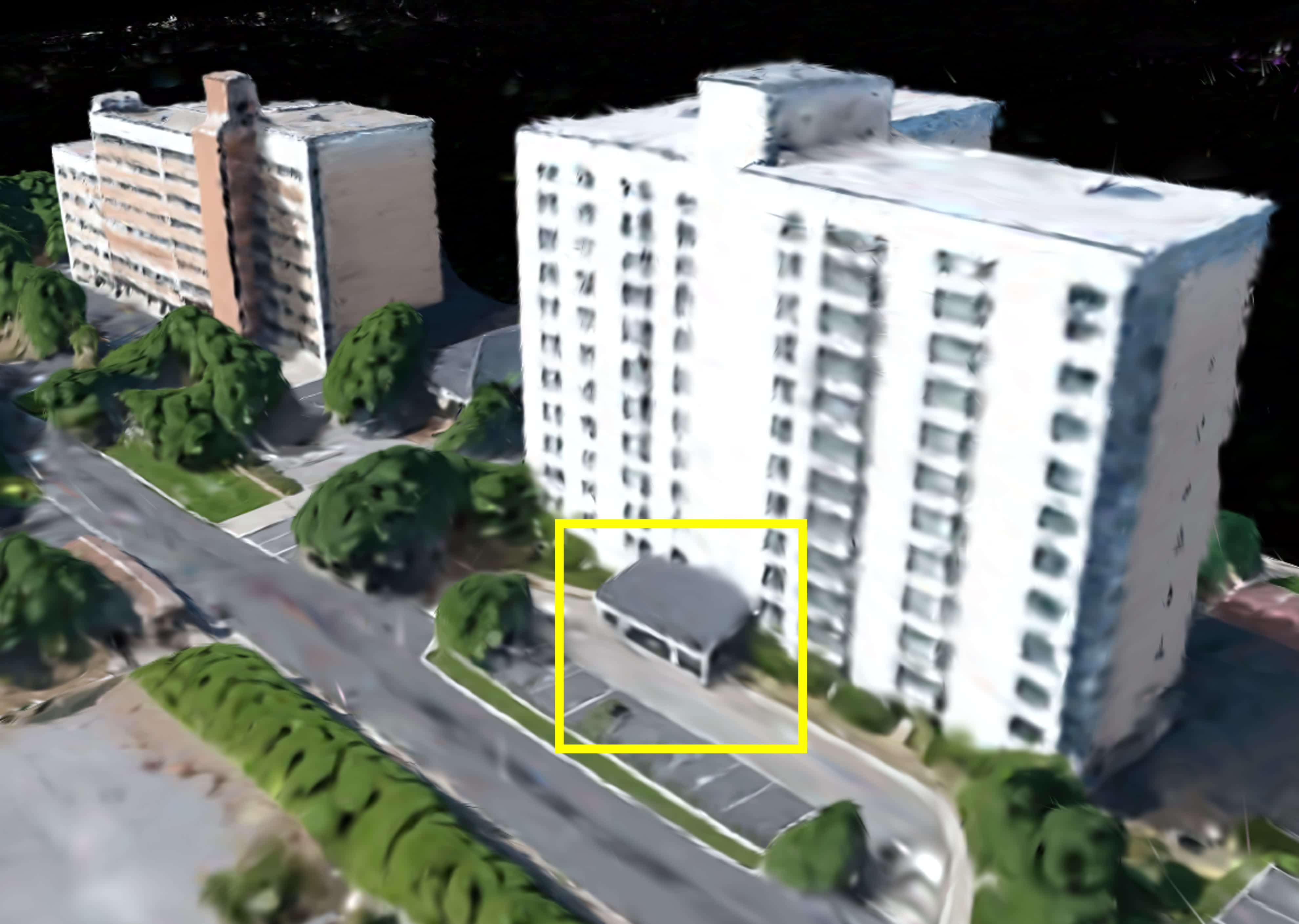}\end{subfigure} \\
    \vspace{0.5mm}
    
    \vspace{2mm} \hrule \vspace{2mm}
        
    % --- OMA Section ---
    \rotatebox{90}{\makebox[0.08\linewidth][c]{\footnotesize OMA-203}}
    \begin{subfigure}{\colw}\myimg{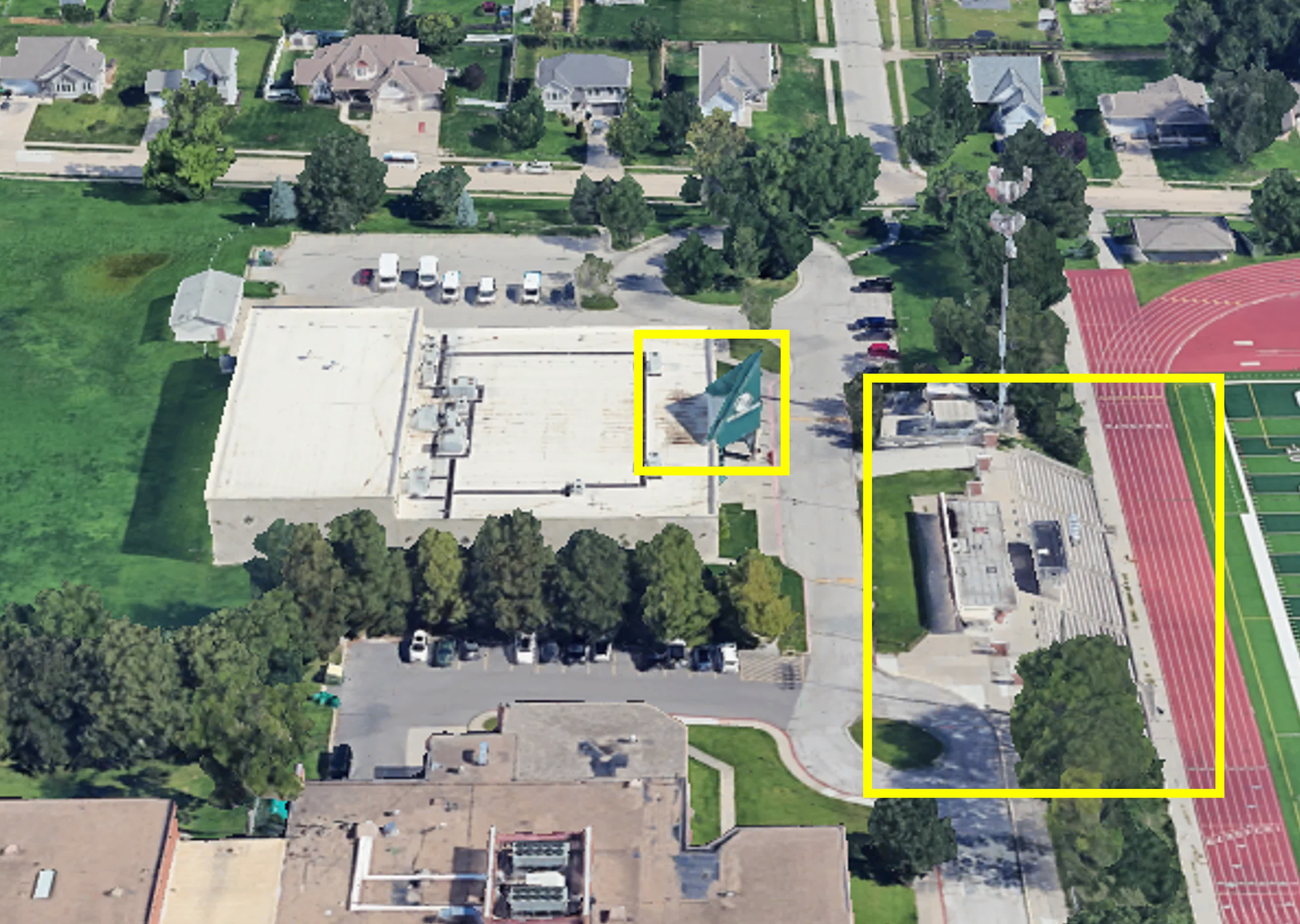}\end{subfigure}
    \begin{subfigure}{\colw}\myimg{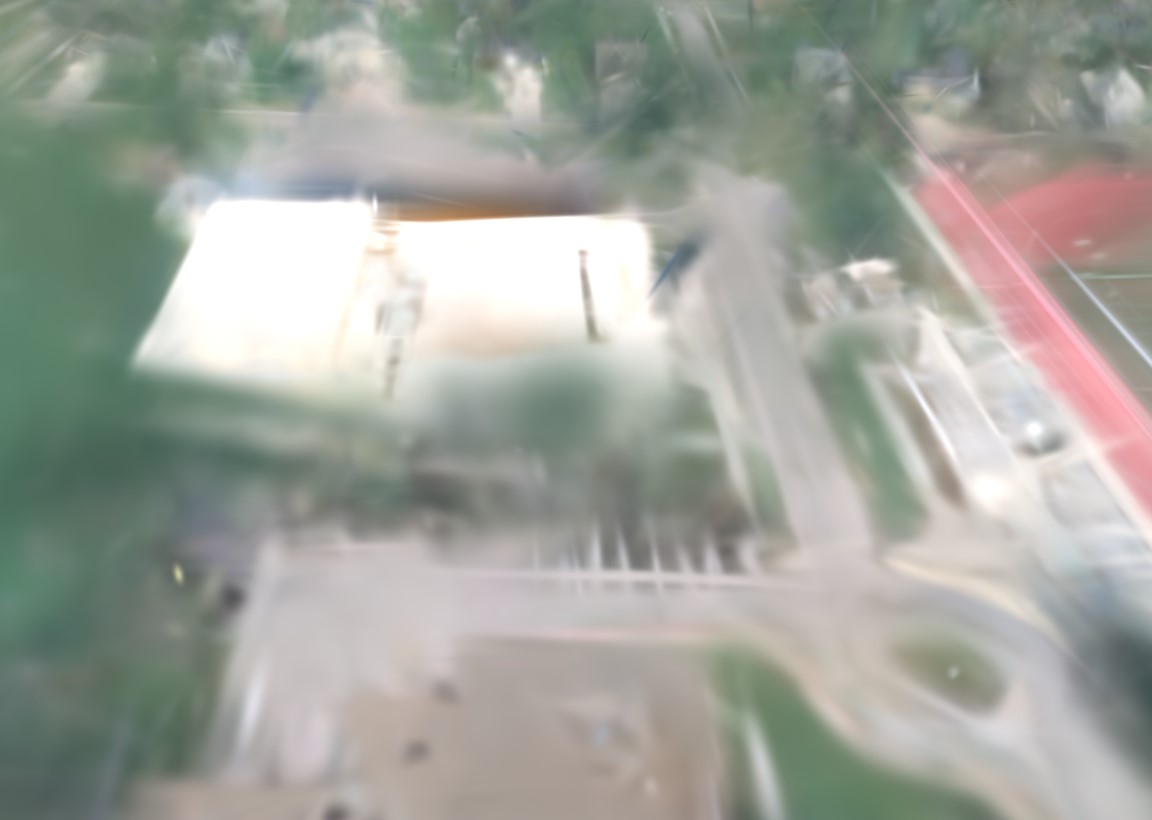}\end{subfigure}
    \begin{subfigure}{\colw}\myimg{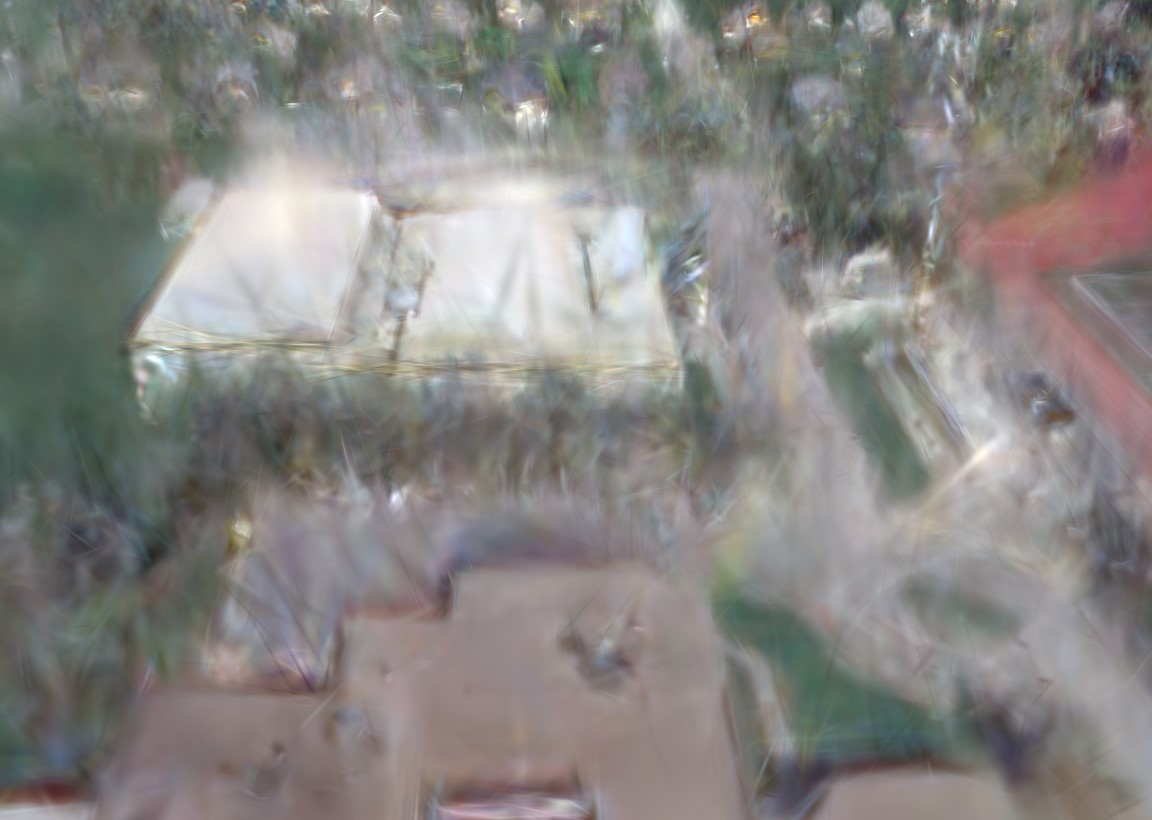}\end{subfigure}
    \begin{subfigure}{\colw}\myimg{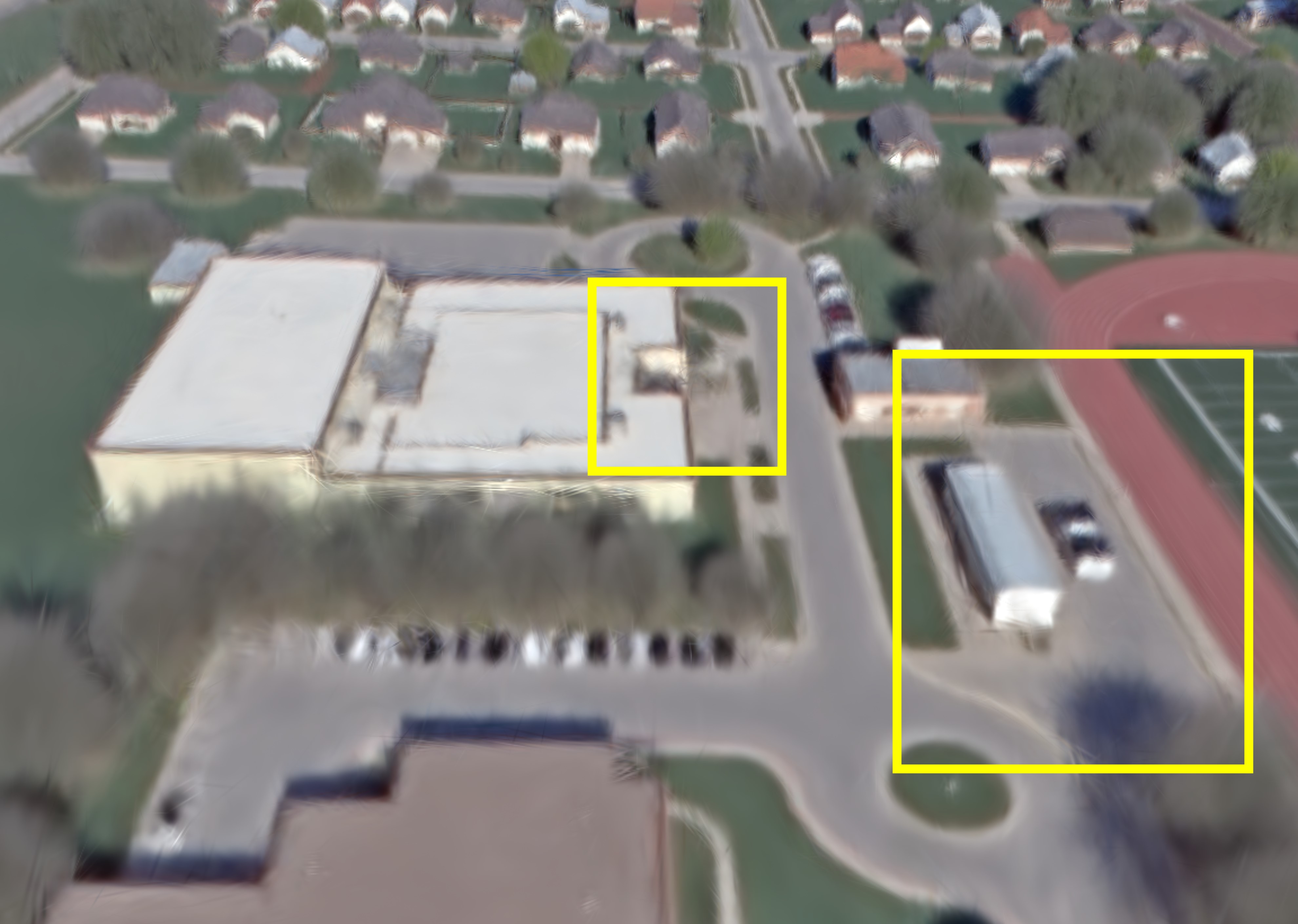}\end{subfigure}
    \begin{subfigure}{\colw}\myimg{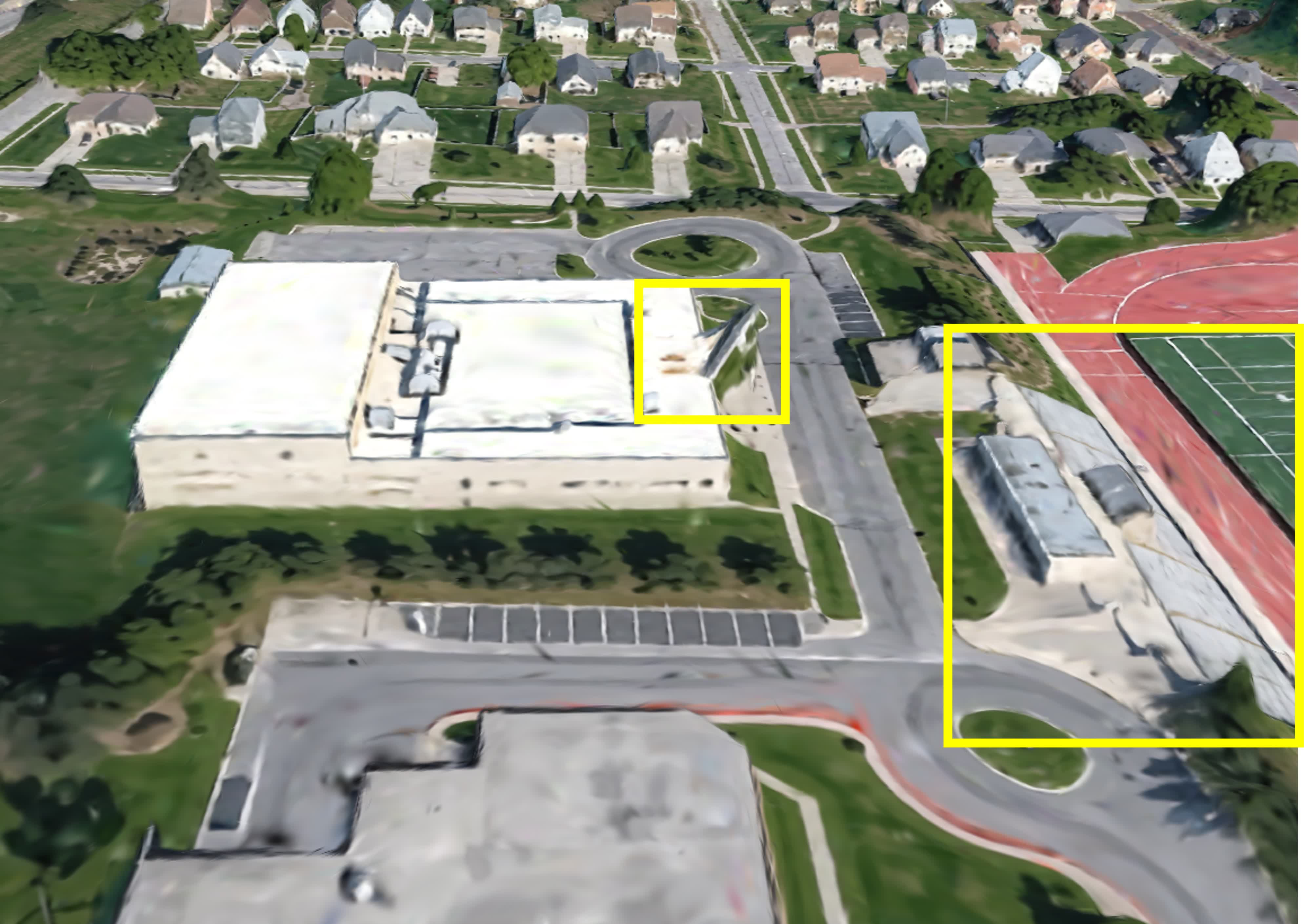}\end{subfigure} \\
    \vspace{0.5mm}

    \rotatebox{90}{\makebox[0.08\linewidth][c]{\footnotesize OMA-212}}
    \begin{subfigure}{\colw}\myimg{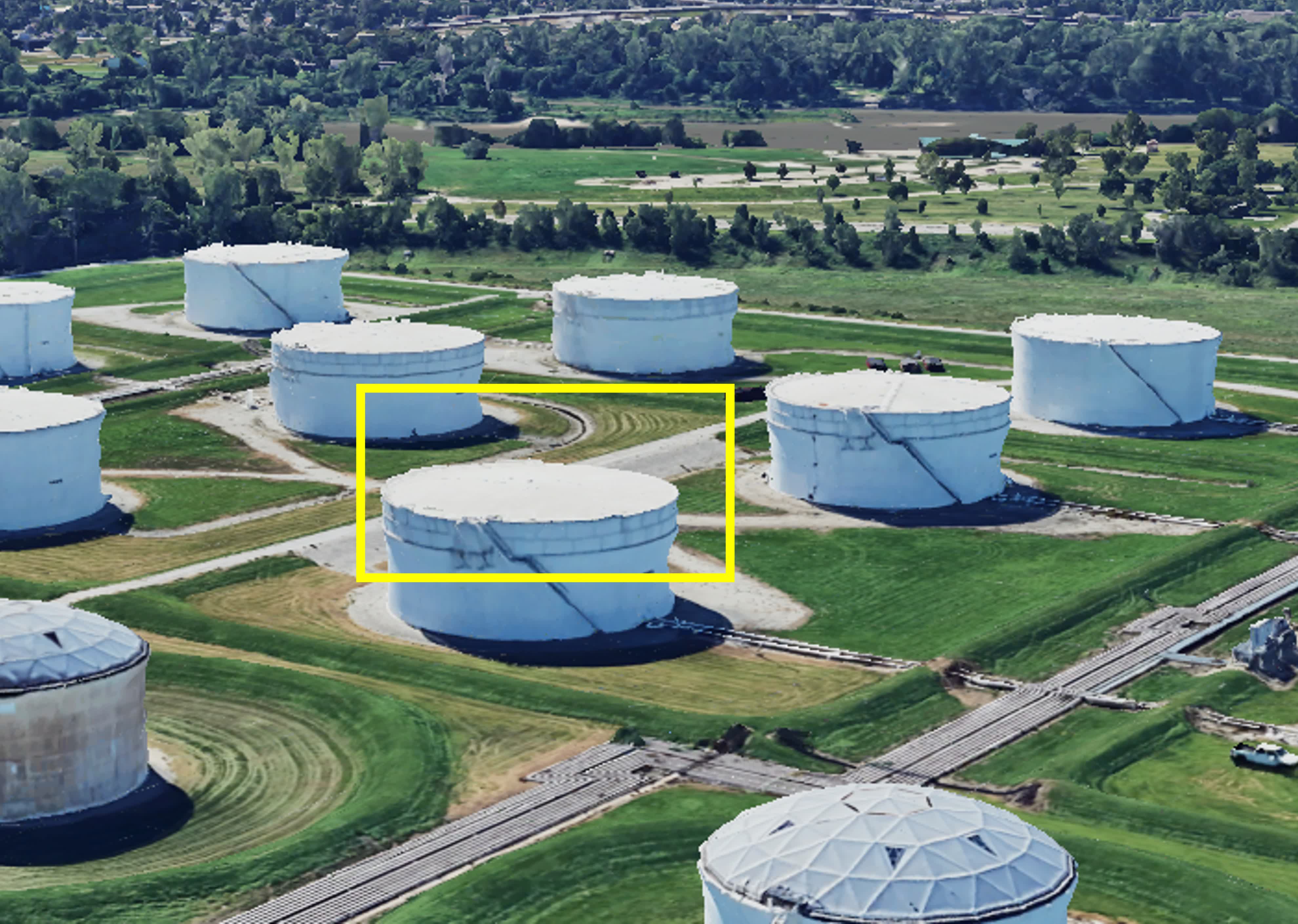}\end{subfigure}
    \begin{subfigure}{\colw}\myimg{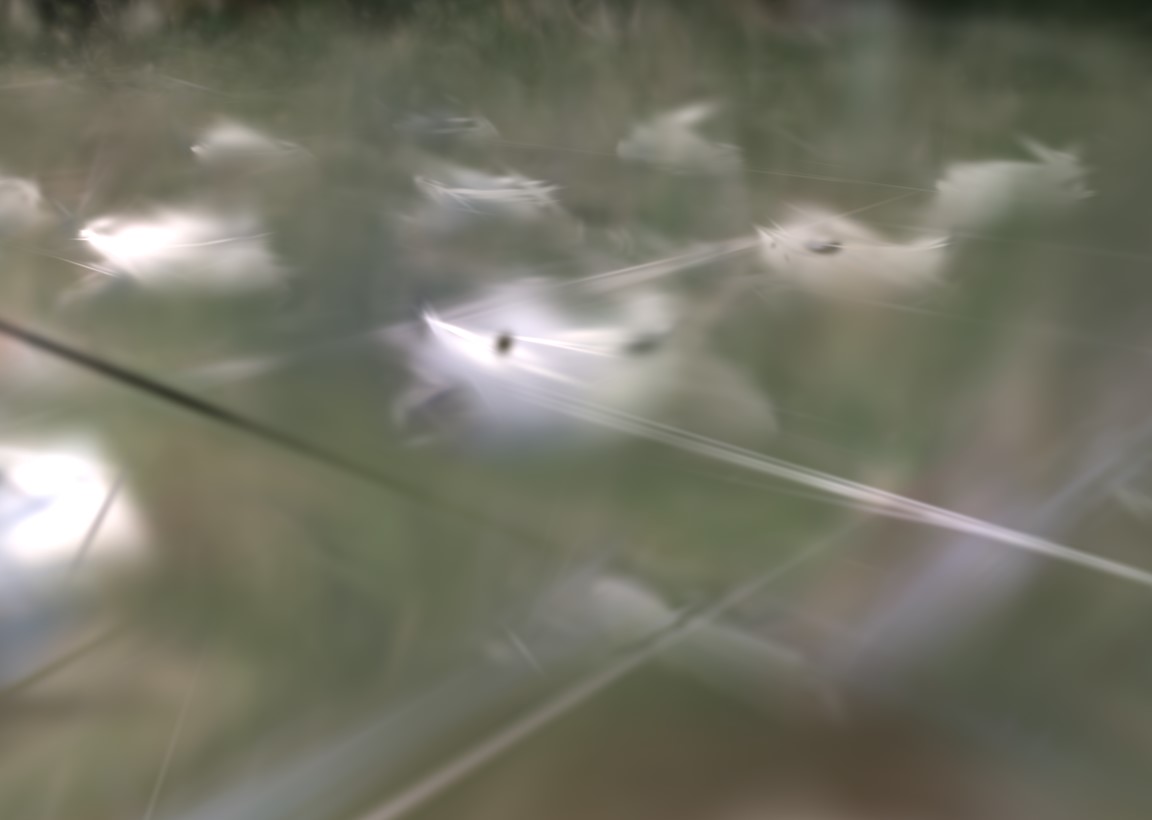}\end{subfigure}
    \begin{subfigure}{\colw}\myimg{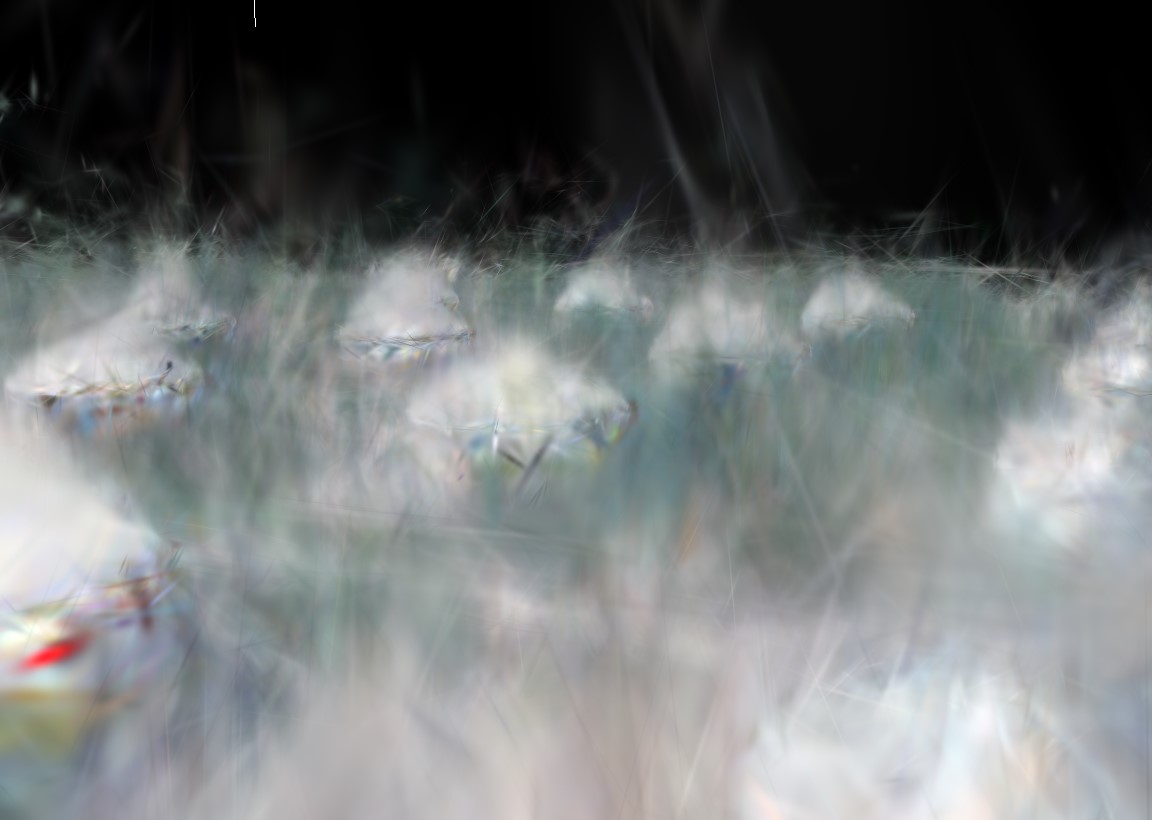}\end{subfigure}
    \begin{subfigure}{\colw}\myimg{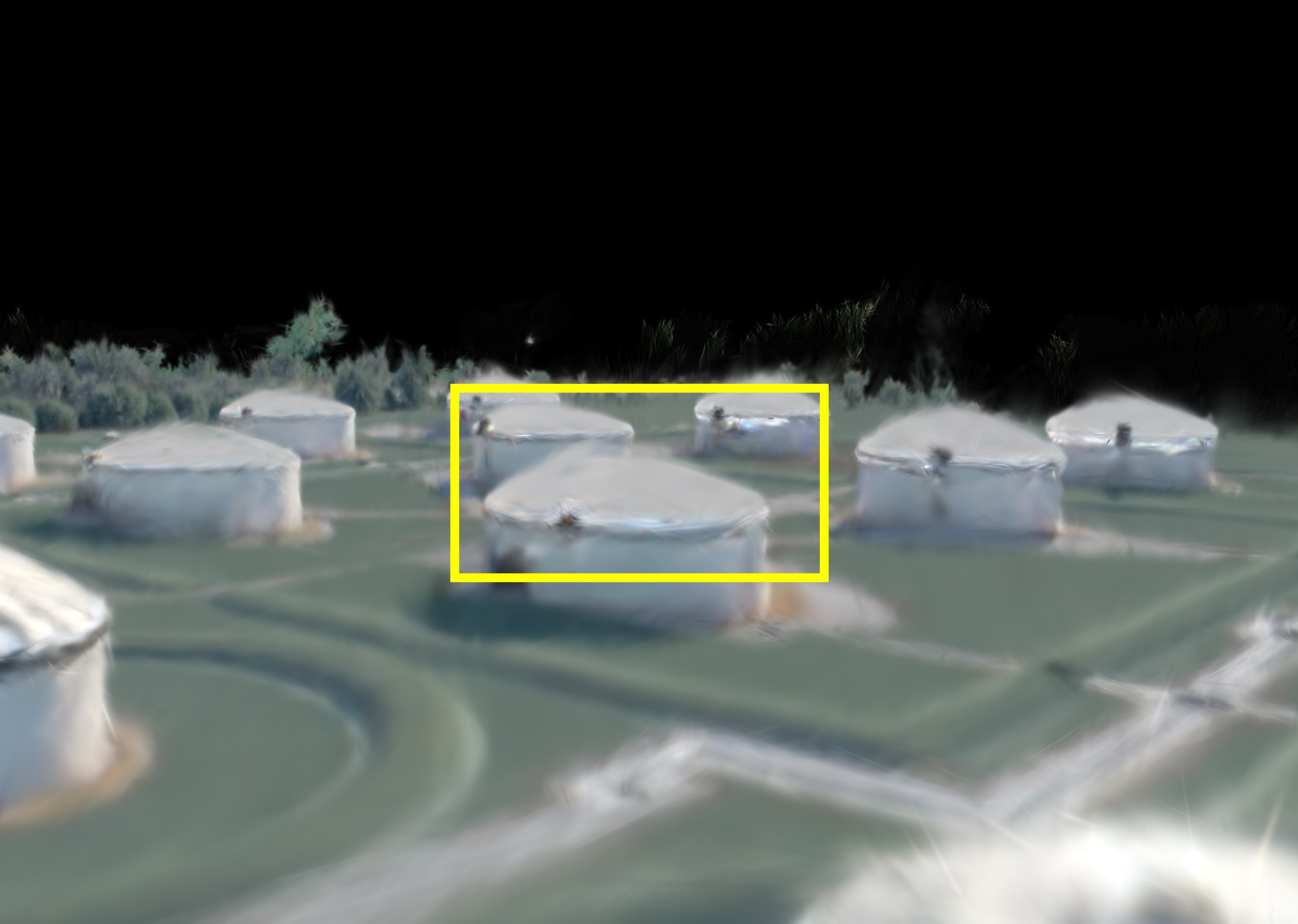}\end{subfigure}
    \begin{subfigure}{\colw}\myimg{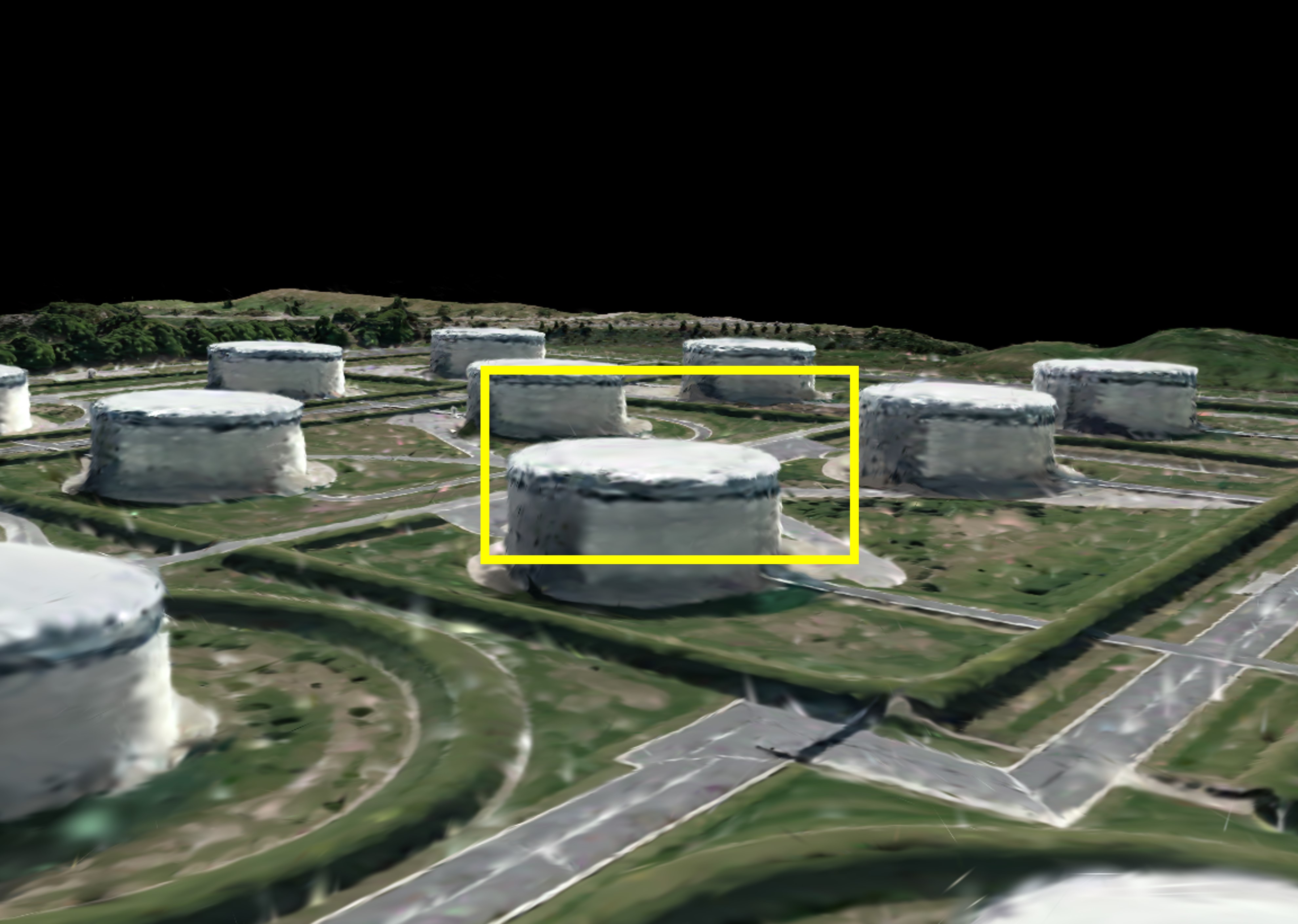}\end{subfigure} \\
    \vspace{0.5mm}

    \rotatebox{90}{\makebox[0.08\linewidth][c]{\footnotesize OMA-315}}
    \begin{subfigure}{\colw}\myimg{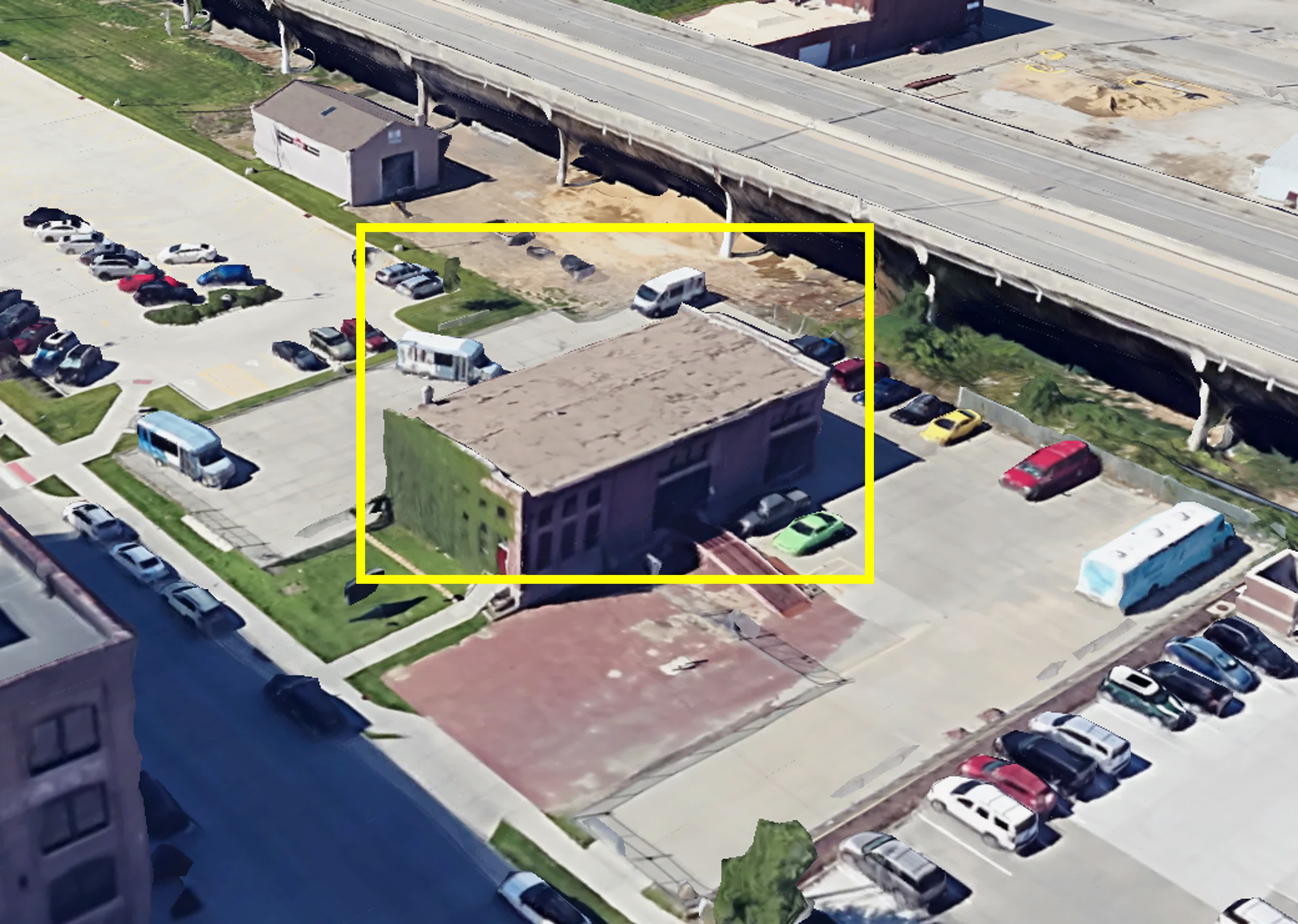}\end{subfigure}
    \begin{subfigure}{\colw}\myimg{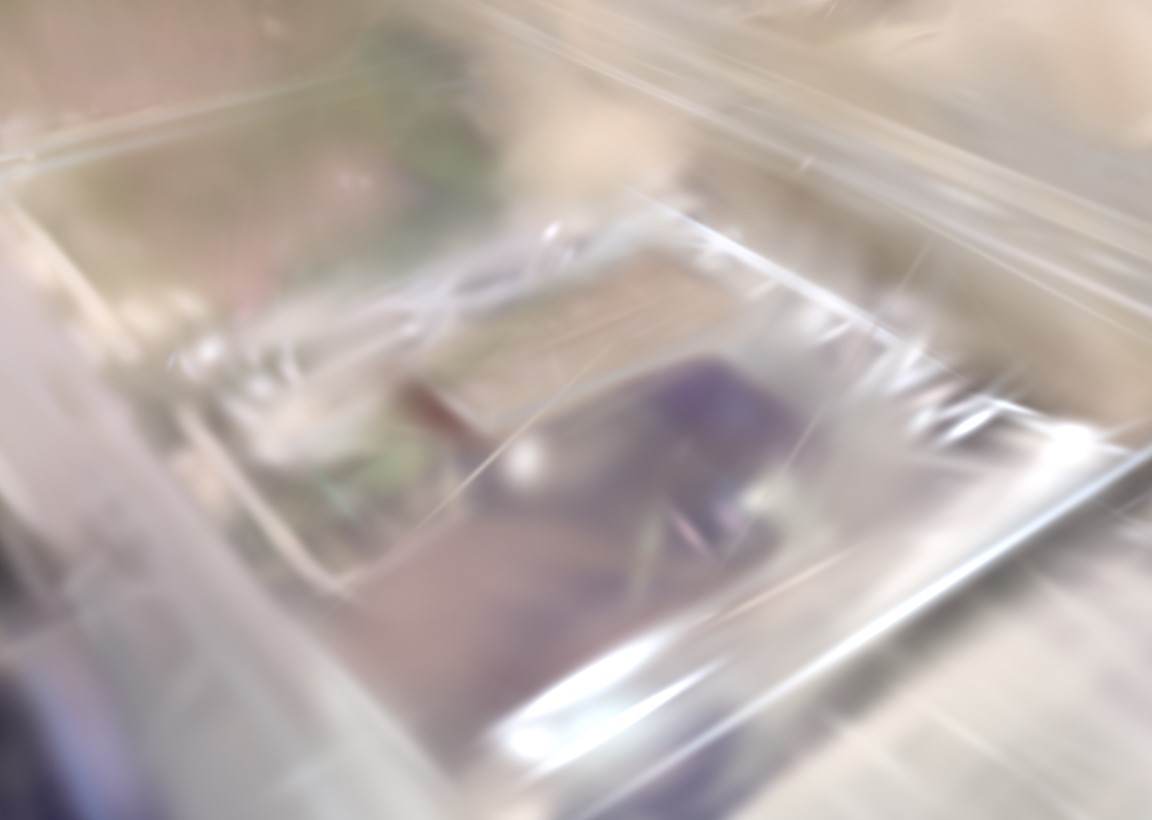}\end{subfigure}
    \begin{subfigure}{\colw}\myimg{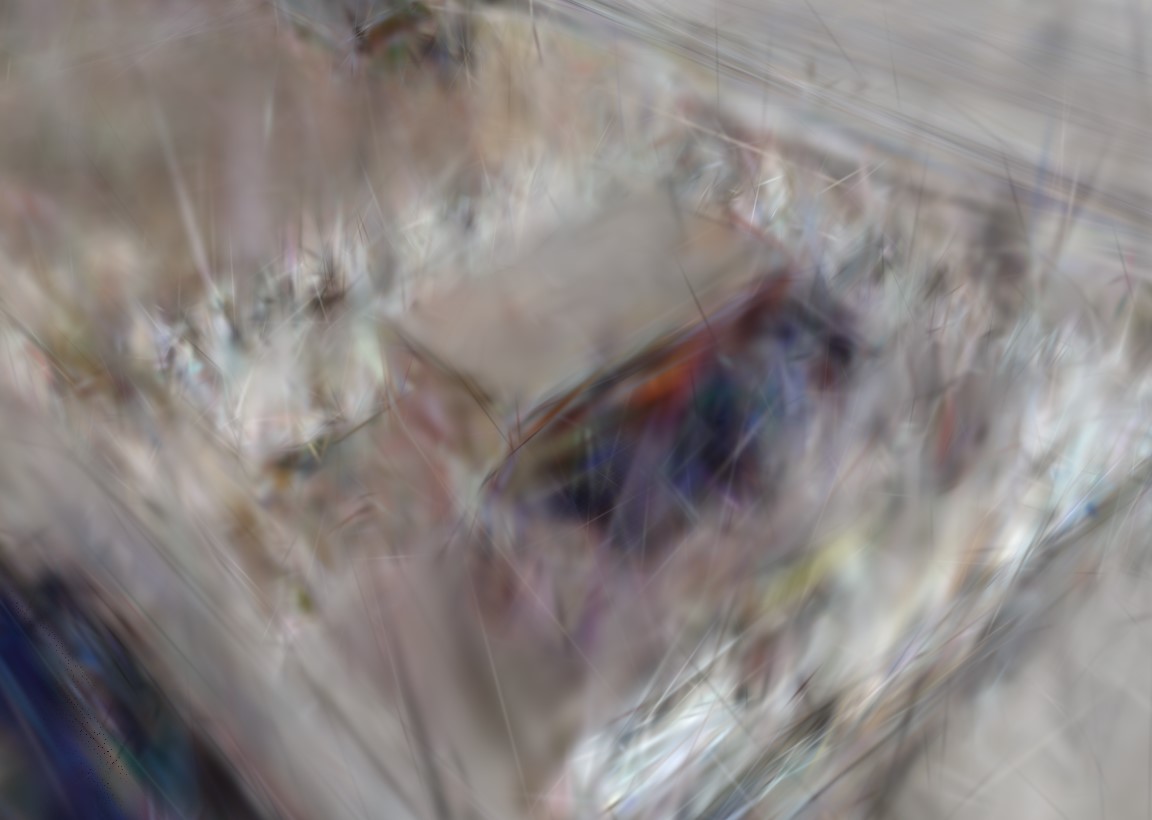}\end{subfigure}
    \begin{subfigure}{\colw}\myimg{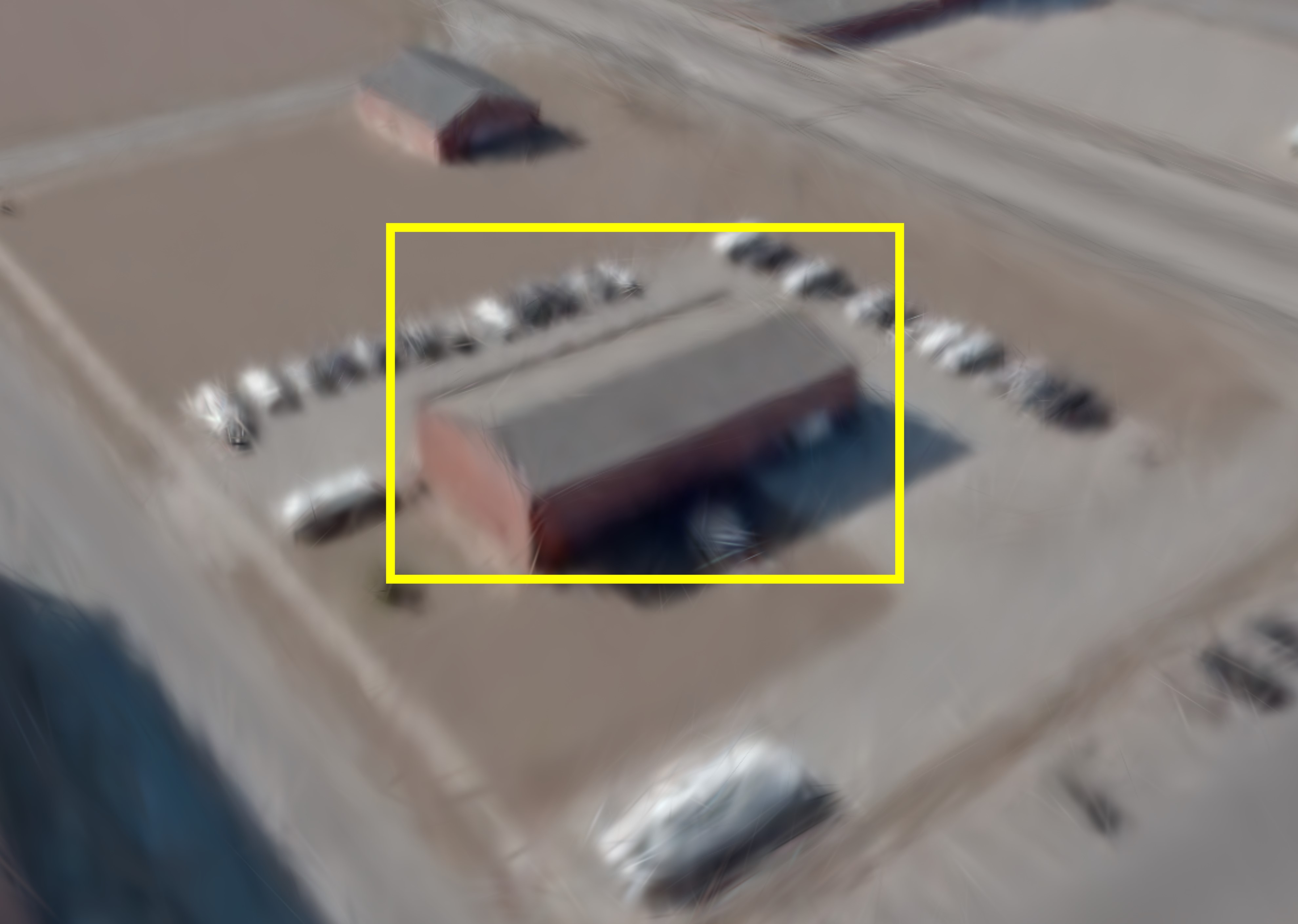}\end{subfigure}
    \begin{subfigure}{\colw}\myimg{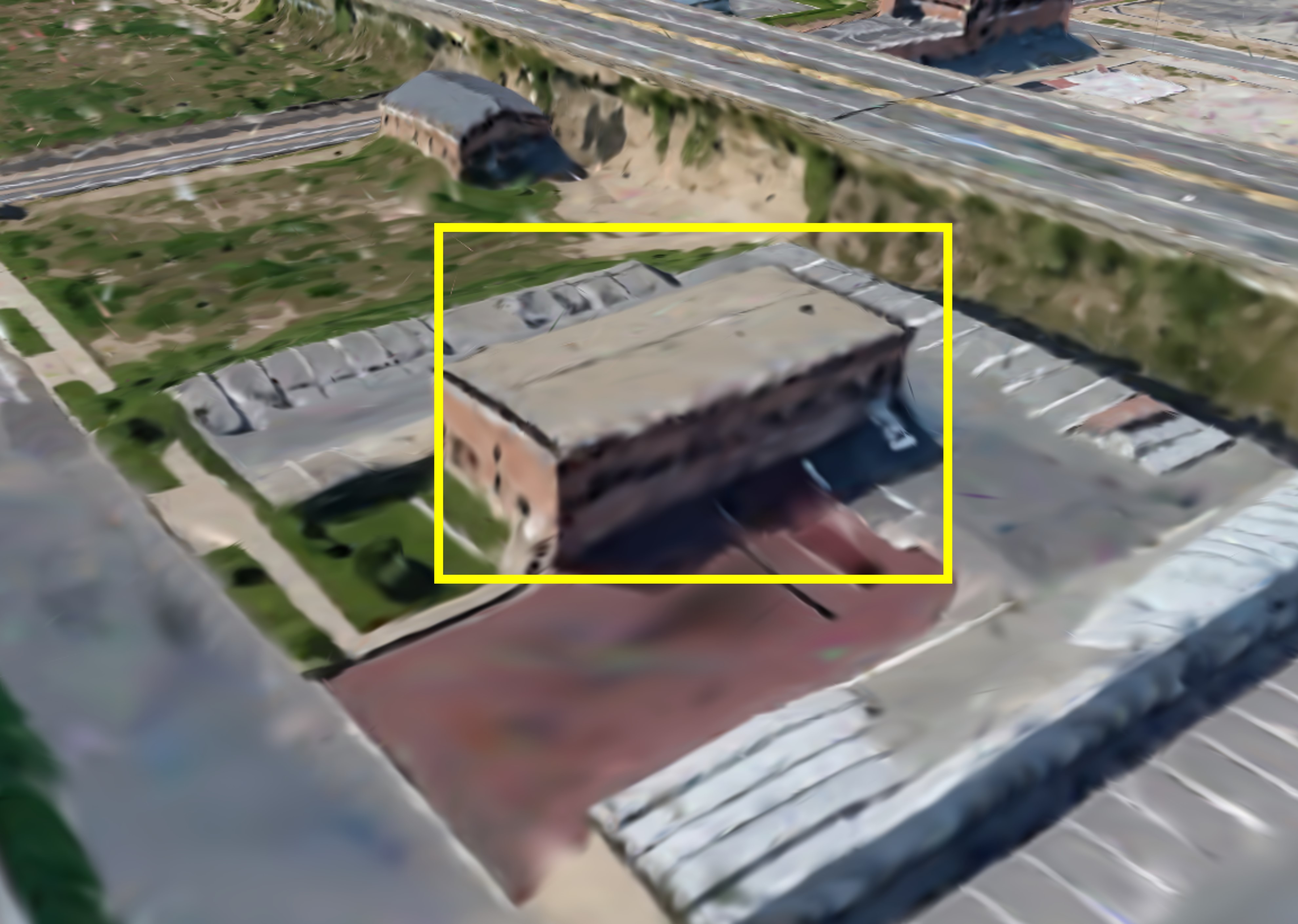}\end{subfigure} \\
    \vspace{0.5mm}
    
    \vspace{2mm} \hrule \vspace{2mm}
    
    % IARPA-001
    \rotatebox{90}{\makebox[0.08\linewidth][c]{\footnotesize IARPA-001}}
    \begin{subfigure}{\colw}\myimg{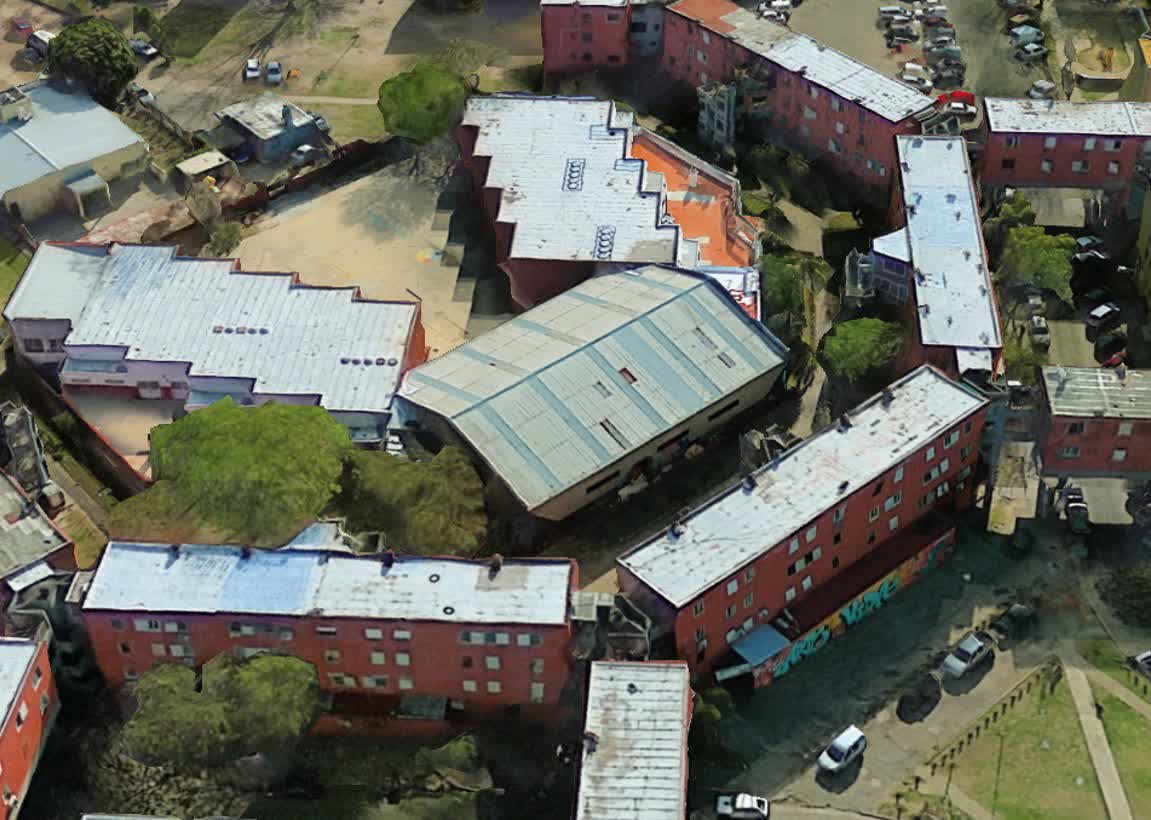}\end{subfigure}
    \begin{subfigure}{\colw}\myimg{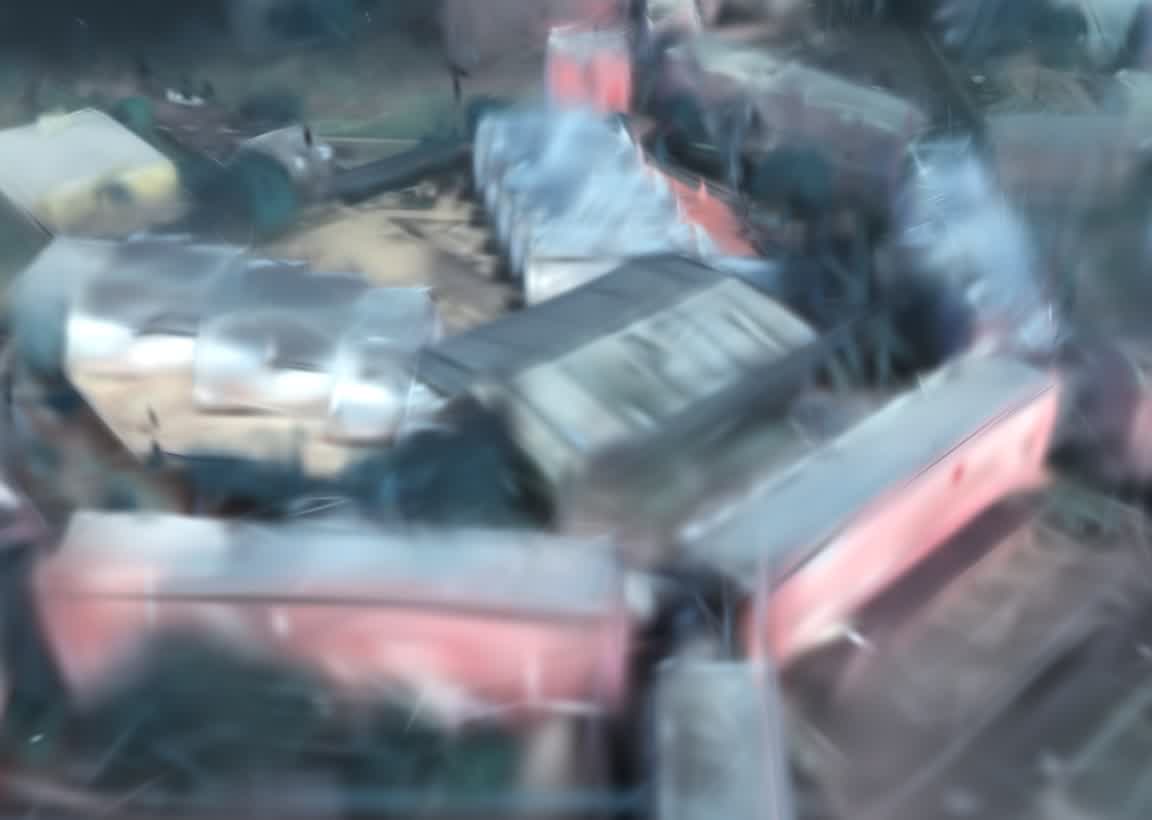}\end{subfigure}
    \begin{subfigure}{\colw}\myimg{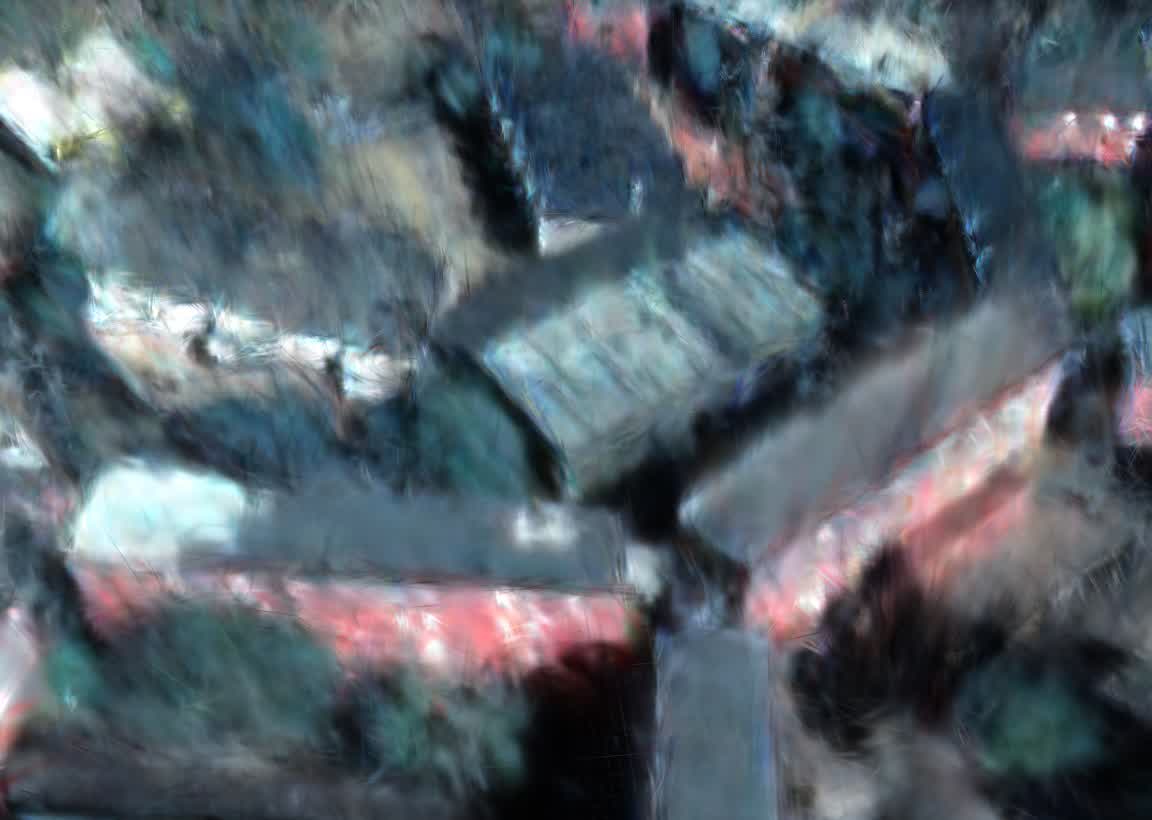}\end{subfigure}
    \begin{subfigure}{\colw}\myimg{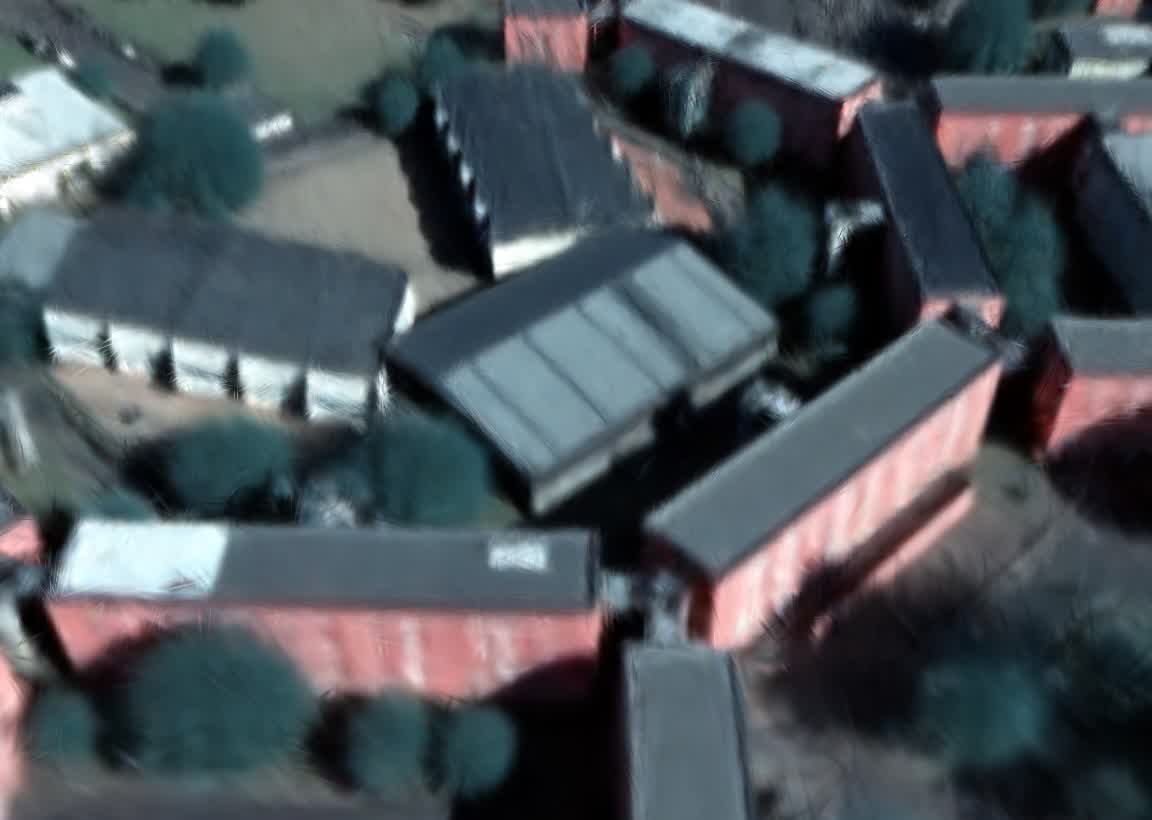}\end{subfigure}
    \begin{subfigure}{\colw}\myimg{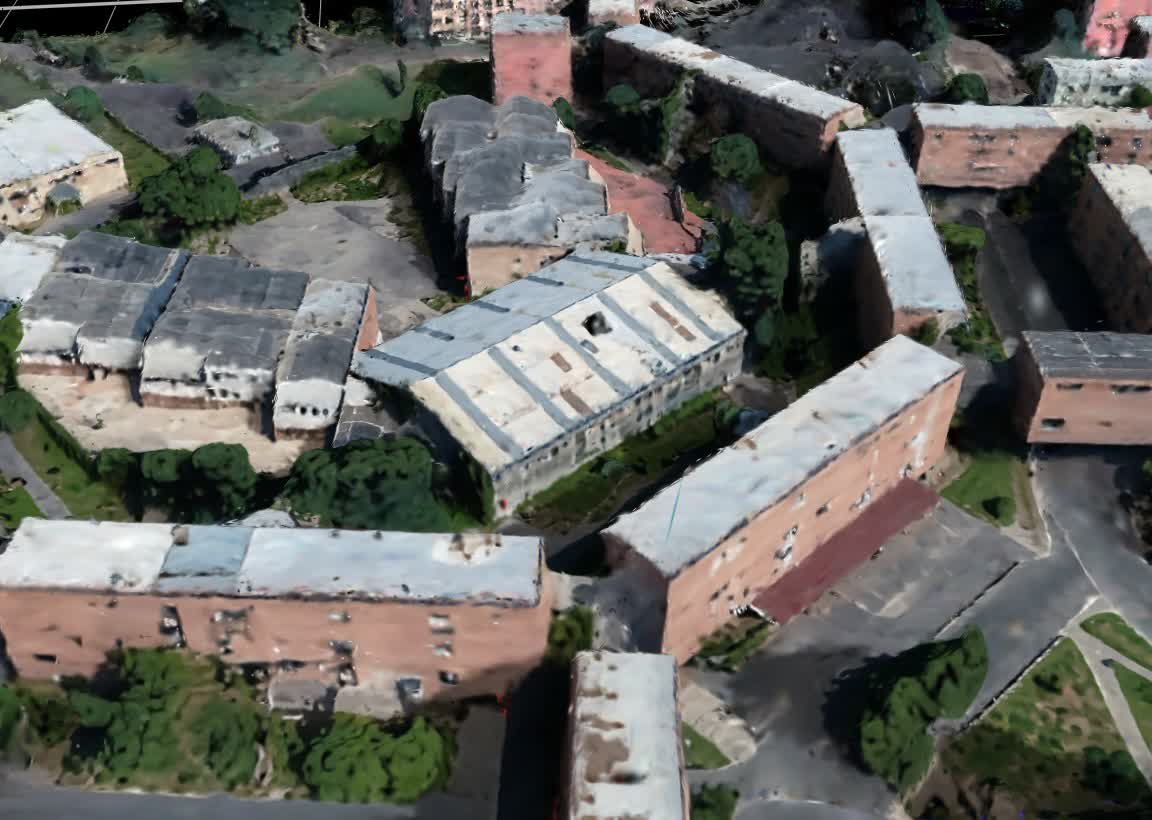}\end{subfigure} \\
    \vspace{1mm}

    % IARPA-002
    \rotatebox{90}{\makebox[0.08\linewidth][c]{\footnotesize IARPA-002}}
    \begin{subfigure}{\colw}\myimg{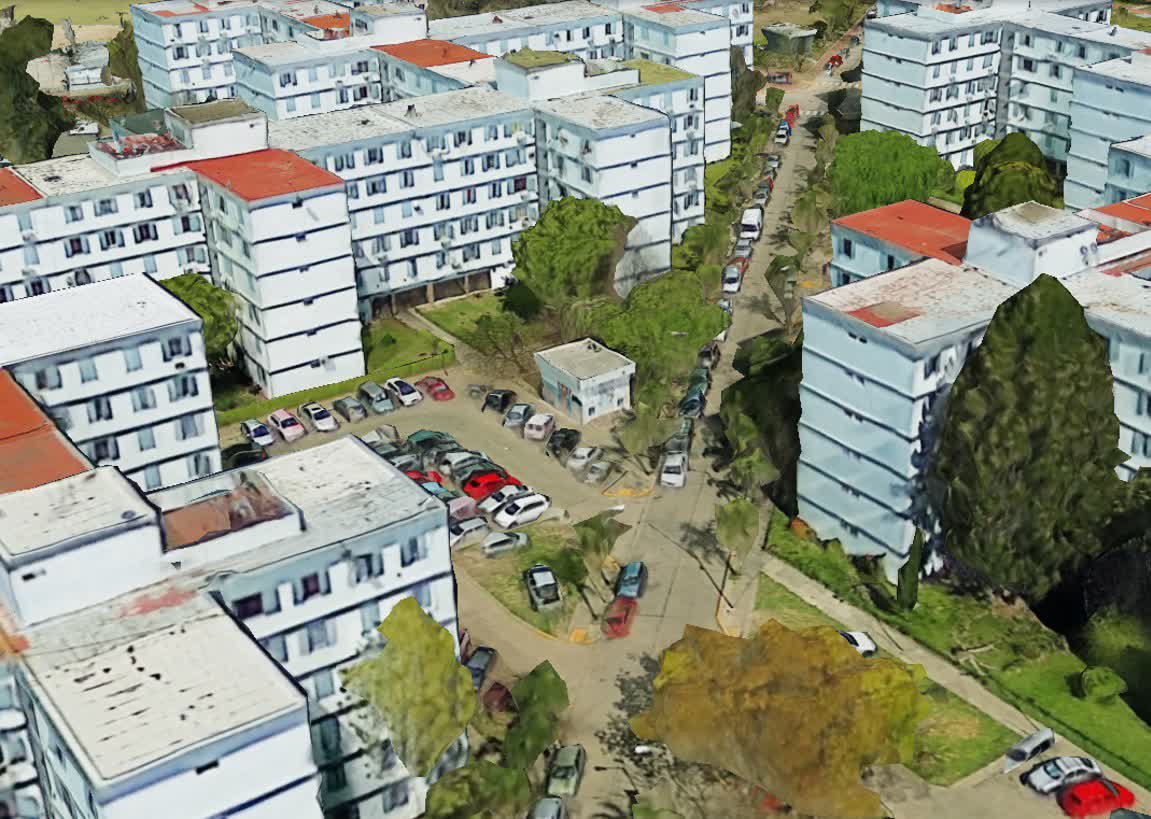}\end{subfigure}
    \begin{subfigure}{\colw}\myimg{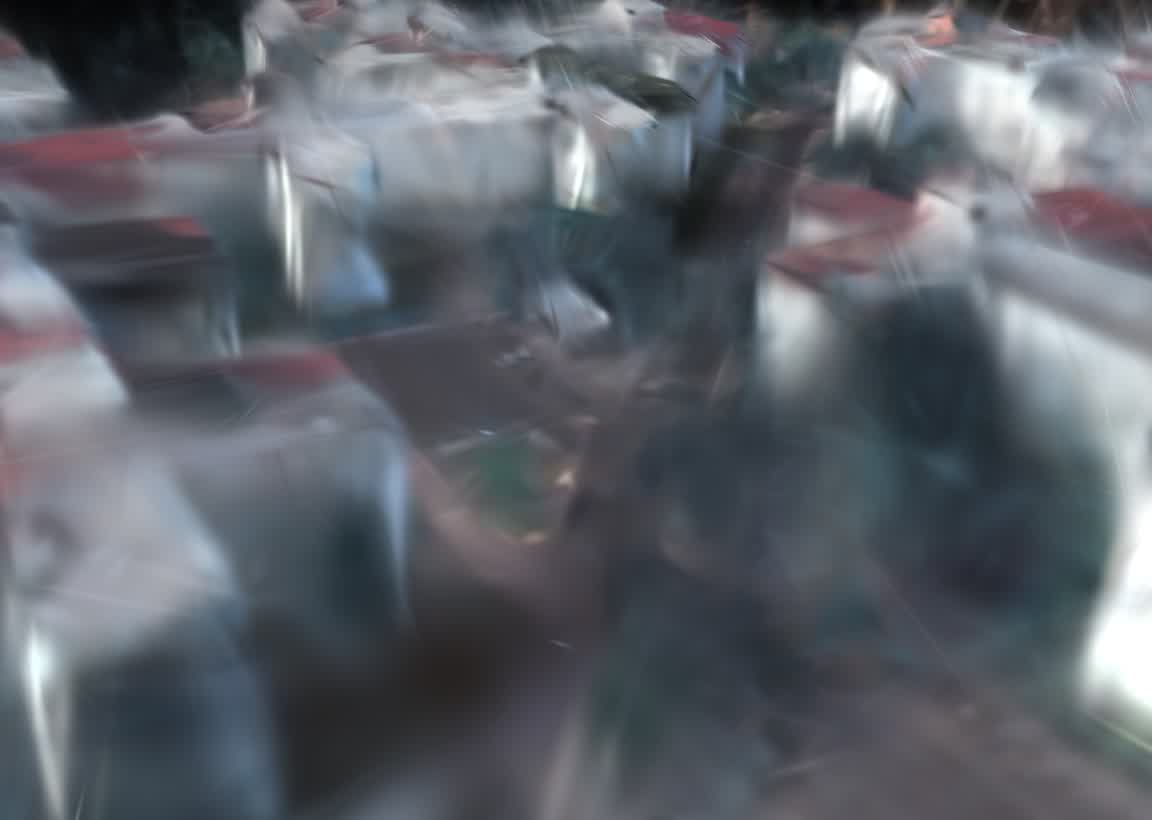}\end{subfigure}
    \begin{subfigure}{\colw}\myimg{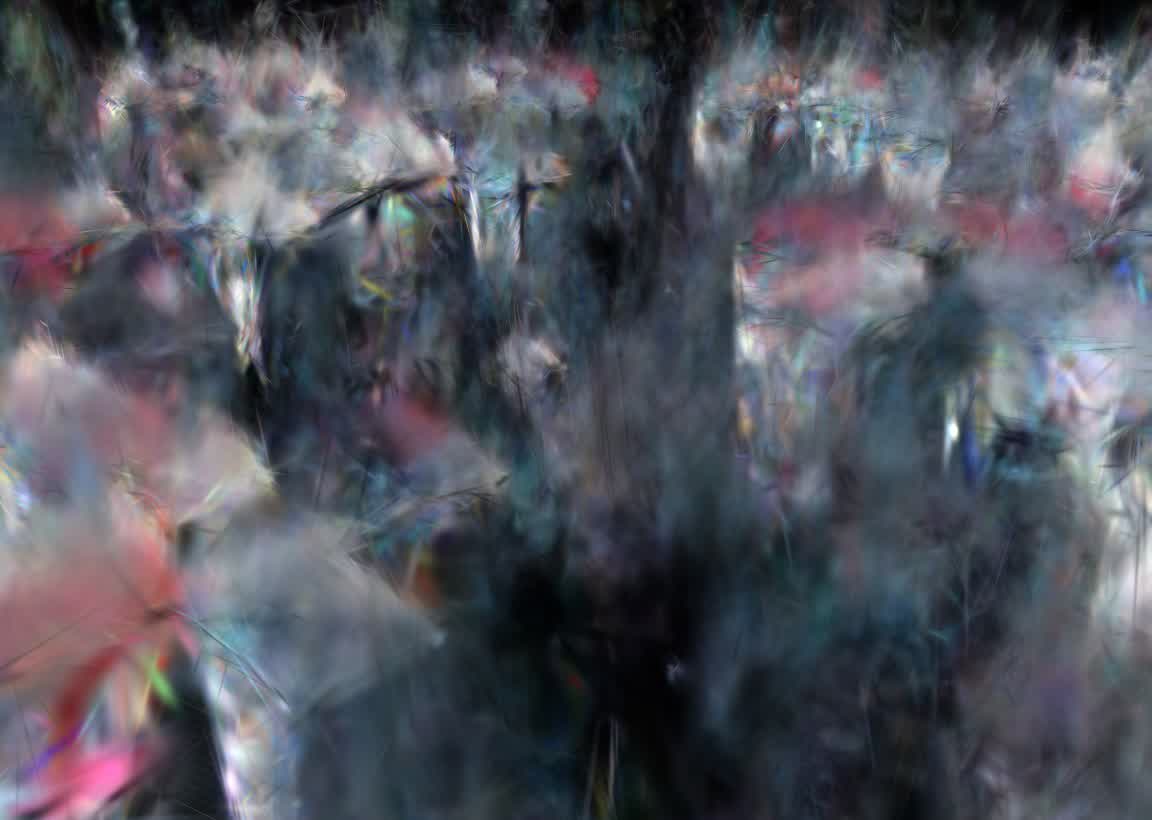}\end{subfigure}
    \begin{subfigure}{\colw}\myimg{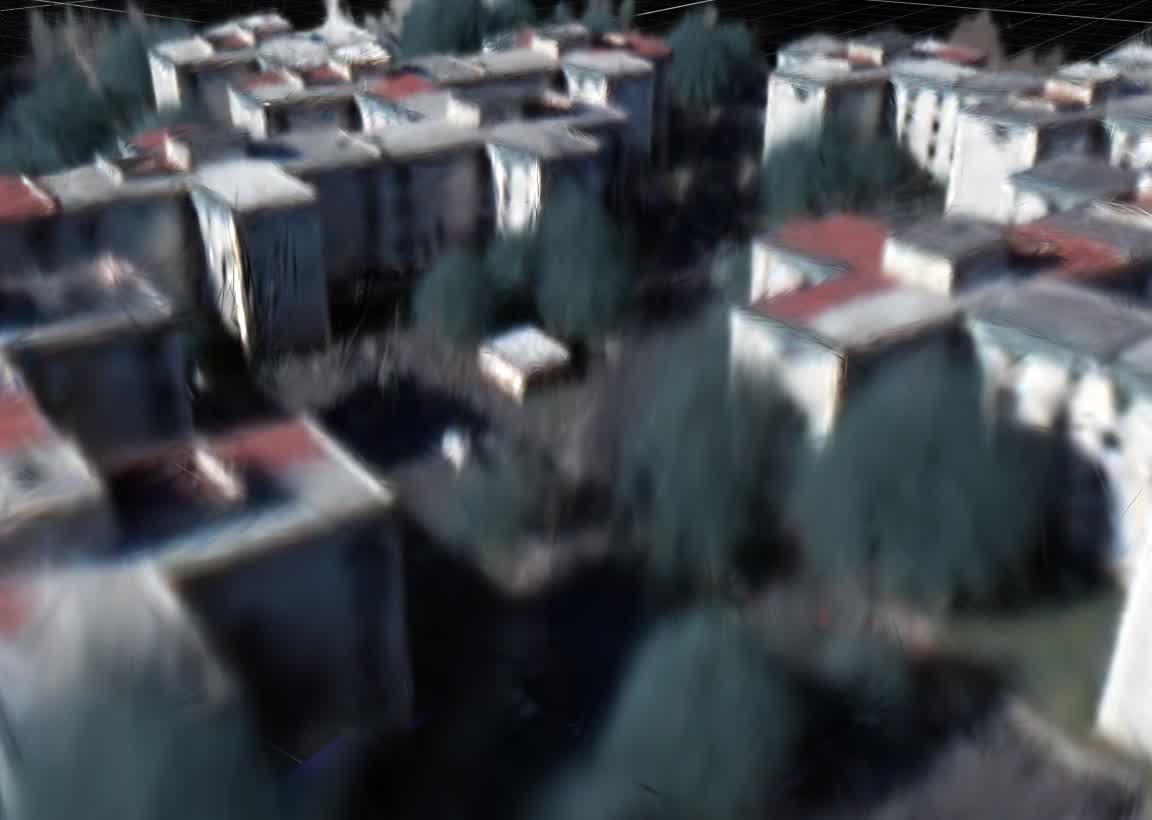}\end{subfigure}
    \begin{subfigure}{\colw}\myimg{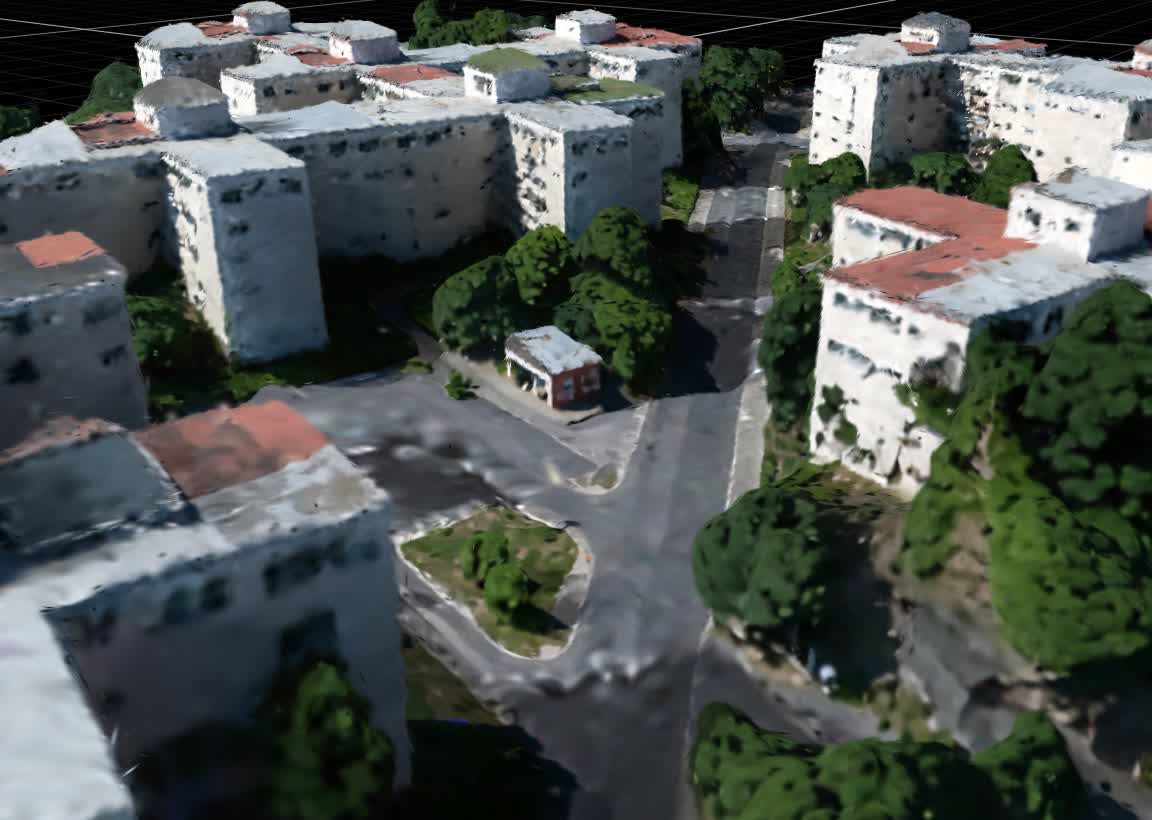}\end{subfigure} \\
    \vspace{1mm}

    % IARPA-003
    \rotatebox{90}{\makebox[0.08\linewidth][c]{\footnotesize IARPA-003}}
    \begin{subfigure}{\colw}\myimg{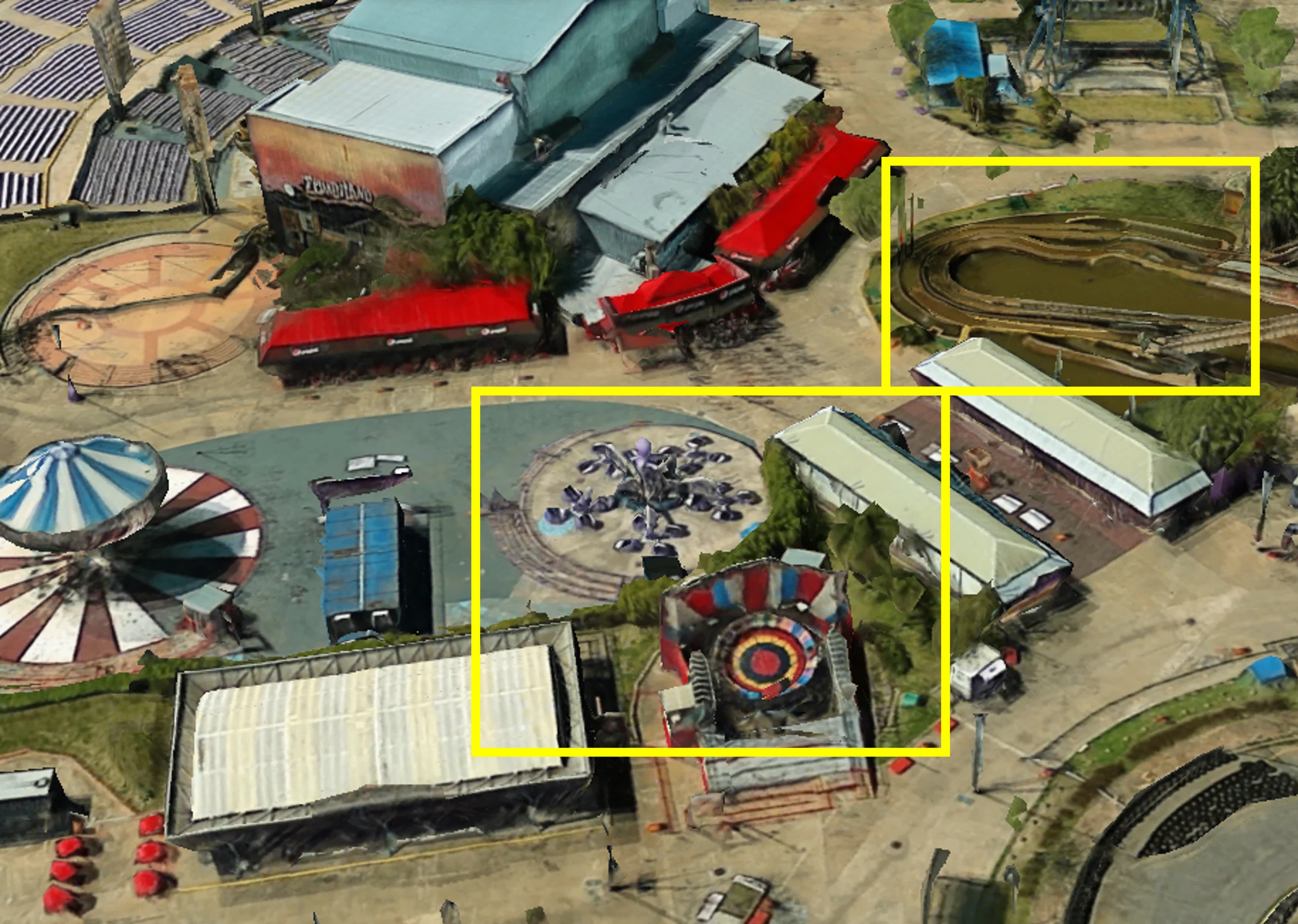}\end{subfigure}
    \begin{subfigure}{\colw}\myimg{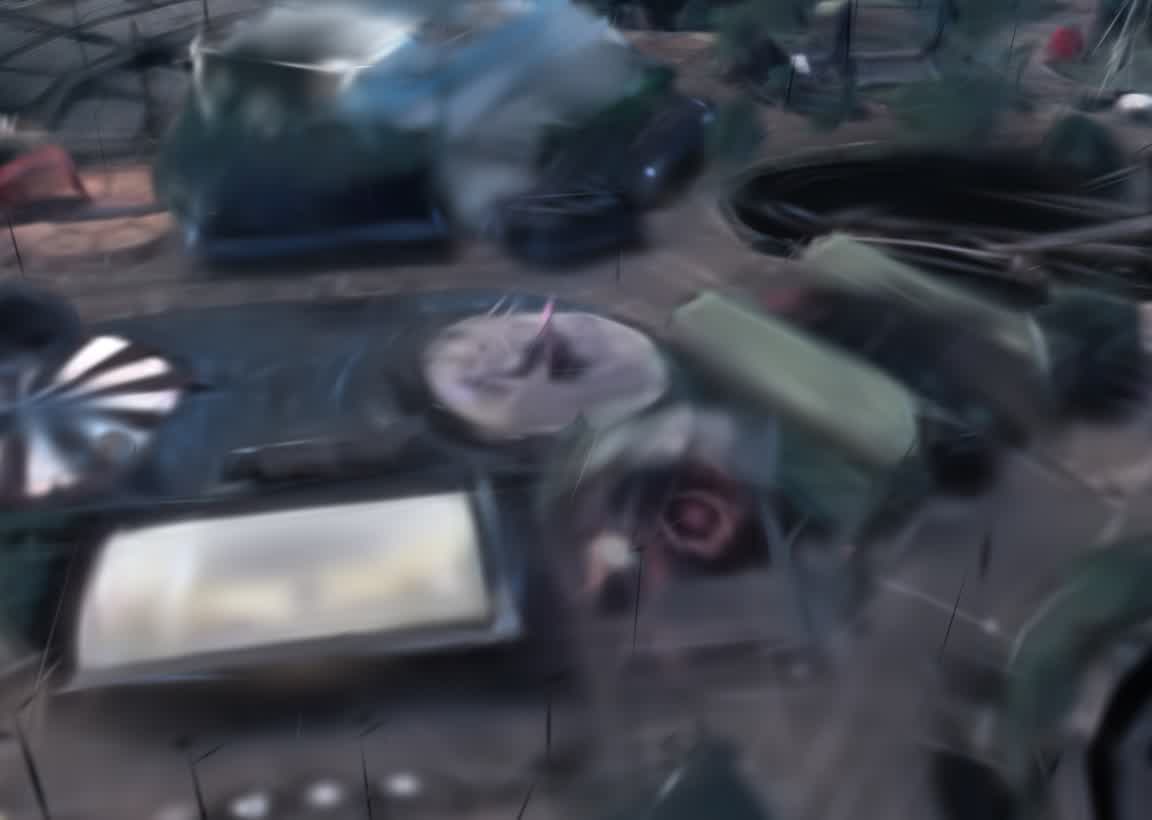}\end{subfigure}
    \begin{subfigure}{\colw}\myimg{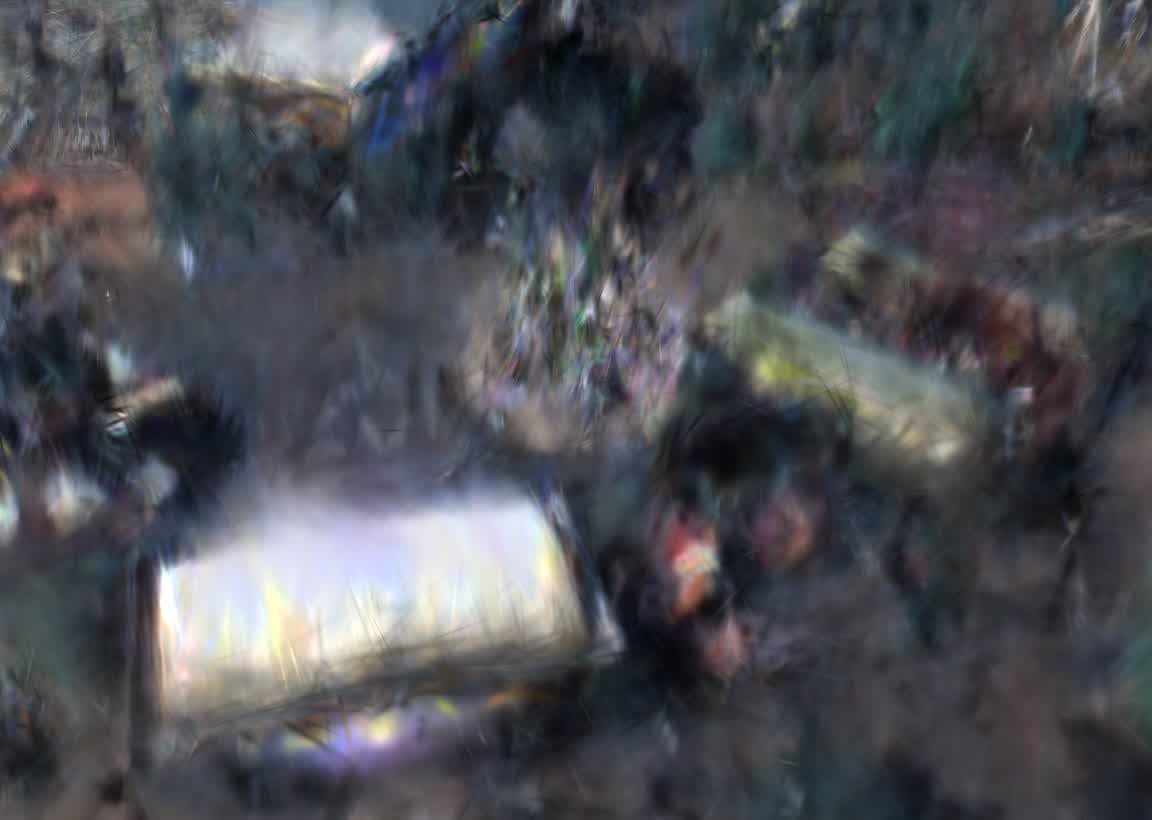}\end{subfigure}
    \begin{subfigure}{\colw}\myimg{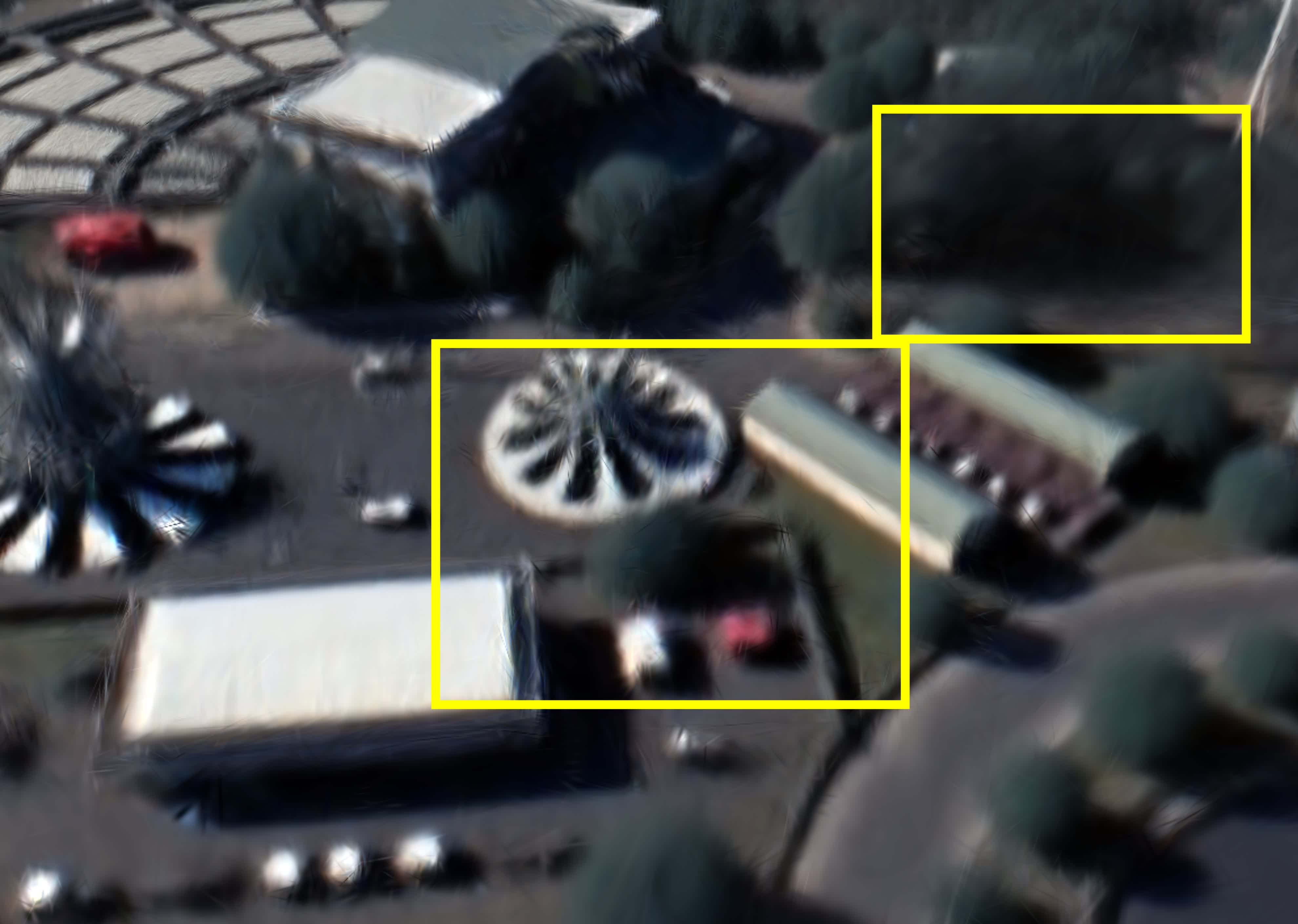}\end{subfigure}
    \begin{subfigure}{\colw}\myimg{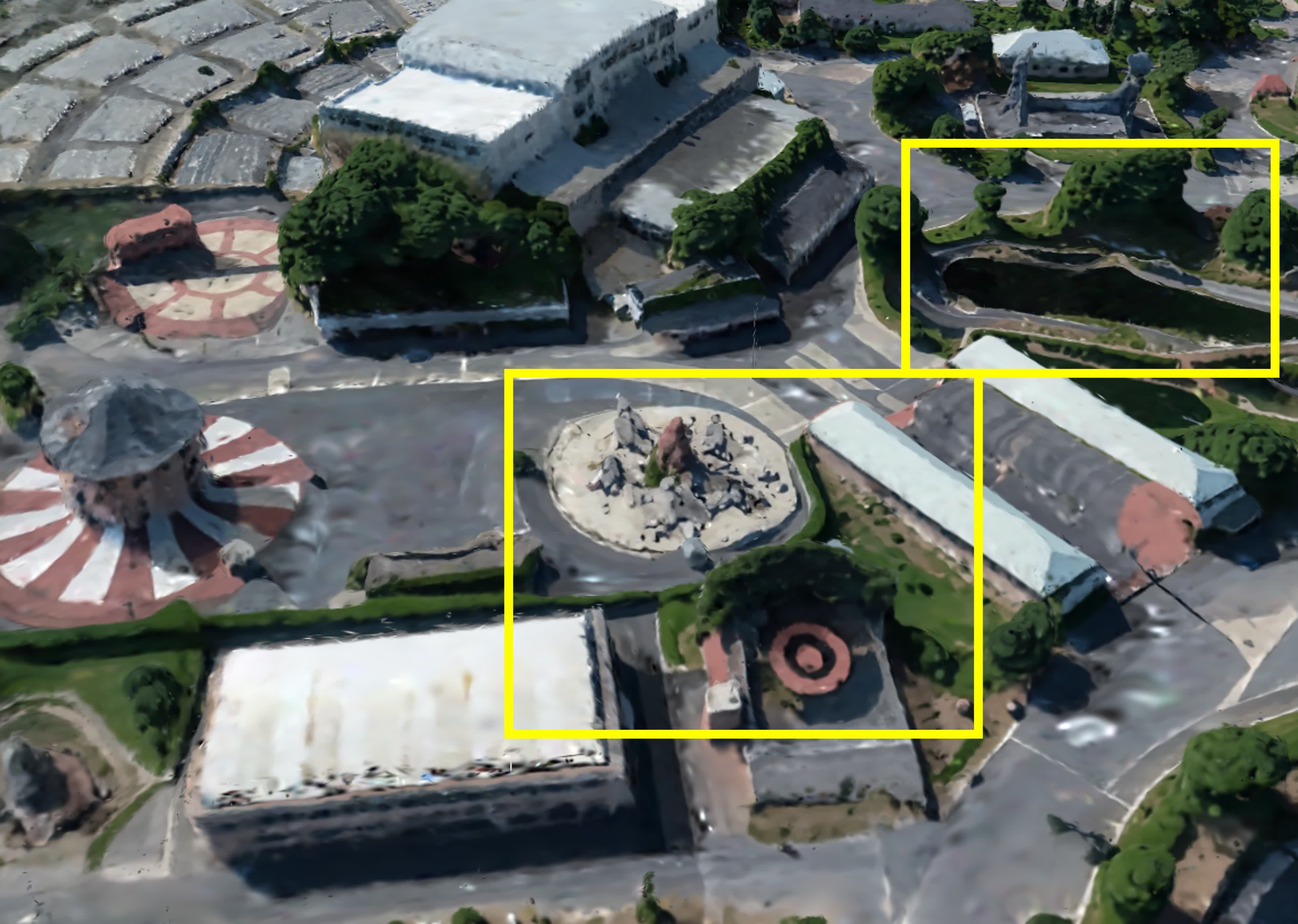}\end{subfigure} \\

    \caption{Qualitative visual comparison of synthesized views across JAX, OMA, and IARPA sites. Compared to existing 3DGS-based baselines that often produce floating artifacts around building roofs, our method generates sharper synthesized views with well-defined structural boundaries. With our shadow-guided generative refinement, our pipeline effectively filters out transient features such as vehicles and maintains strict edge consistency.}
    \label{fig:dfc_comparison}
\end{figure*}

Both qualitative and quantitative comparisons are presented in \Cref{fig:dfc_comparison} and \Cref{tab:metrics_comparison_combined}. Our method produces sharper textures and more clearly defined boundaries than existing baselines, whereas other 3DGS-based methods tend to generate transparent Gaussian clouds with ambiguous surface boundaries. This distinction is particularly evident at the JAX-214 site, where Skyfall-GS introduces floating artifacts above building roofs, primarily due to the influence of dynamic objects such as cars, while our method effectively filters out these transient features to maintain a clean roof surface.

Our method also demonstrated results with reduced artifacts on \Cref{fig:dfc_comparison}. In the JAX-214 and JAX-260 sites, Skyfall-GS removes or modifies original structures, whereas the proposed method reconstructs the geometry without artifacts. Similar results are observed at OMA sites. While Skyfall-GS generates artifacts and causes both visual and geometric misalignments at OMA-203 and OMA-315, our method produces realistic surfaces with accurate geometric alignment. IARPA site evaluations further confirm the pipeline’s consistency across locations. Specifically, at IARPA-003, Skyfall-GS fails to capture the intricate details of circular amusement structures, often simplifying them into generic conical forms, whereas our method preserves the authentic geometric characteristics of the objects.

\begin{figure*}[t] 
    \centering
    \small
    
    \newcommand{\myfigcrop}[1]{%
        \includegraphics[width=\linewidth, keepaspectratio, trim=0 5 0 5, clip]{#1}%
    }
    
    \def\colwthree{0.31\linewidth} 

    \makebox[\colwthree]{Google Earth}
    \makebox[\colwthree]{Skyfall-GS}
    \makebox[\colwthree]{Ours} \\
    \vspace{1mm}

    % --- JAX-168 Section ---
    \rotatebox{90}{\makebox[0.12\linewidth][c]{\footnotesize JAX-168}}
    \begin{subfigure}{\colwthree}\myfigcrop{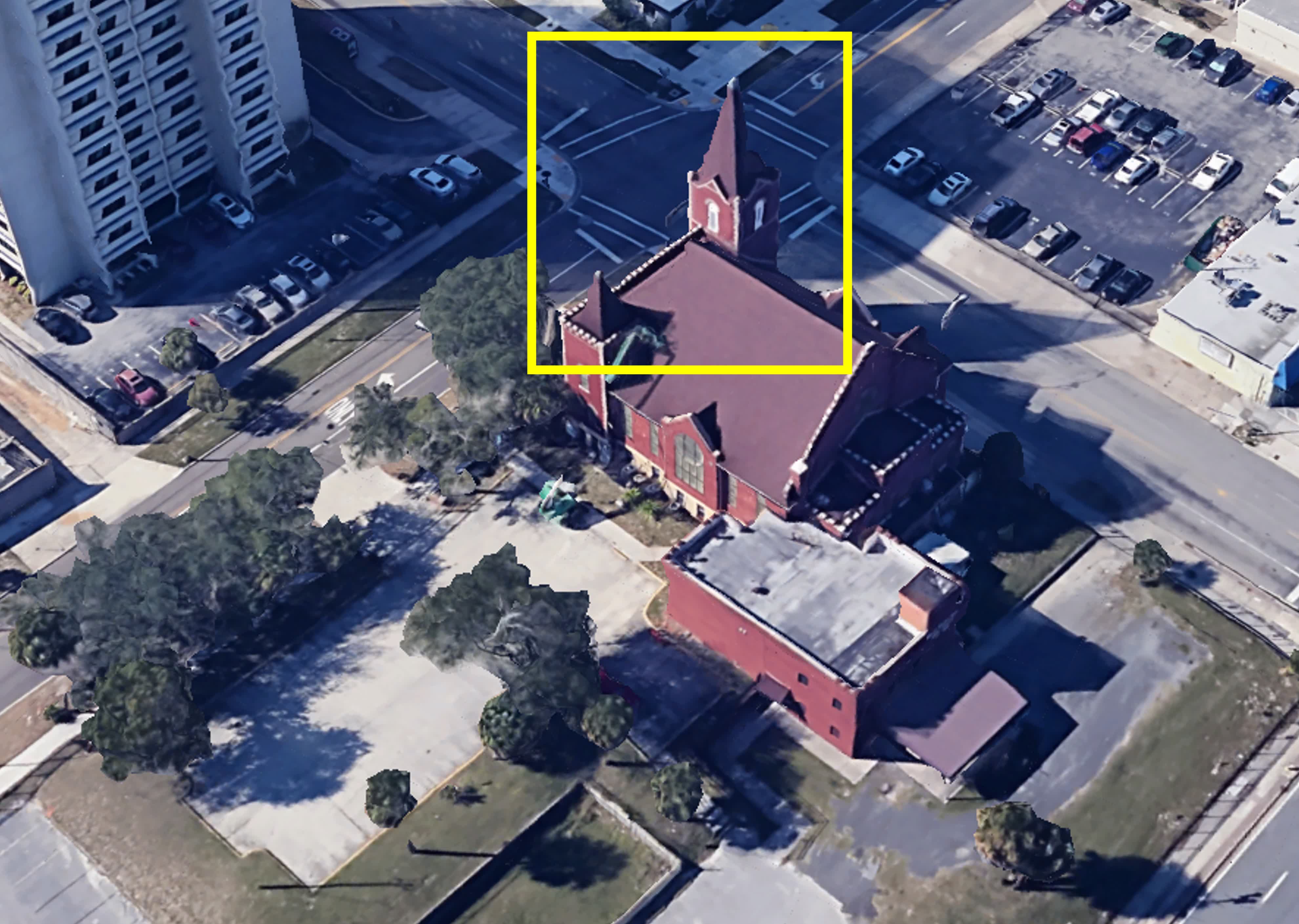}\end{subfigure}
    \begin{subfigure}{\colwthree}\myfigcrop{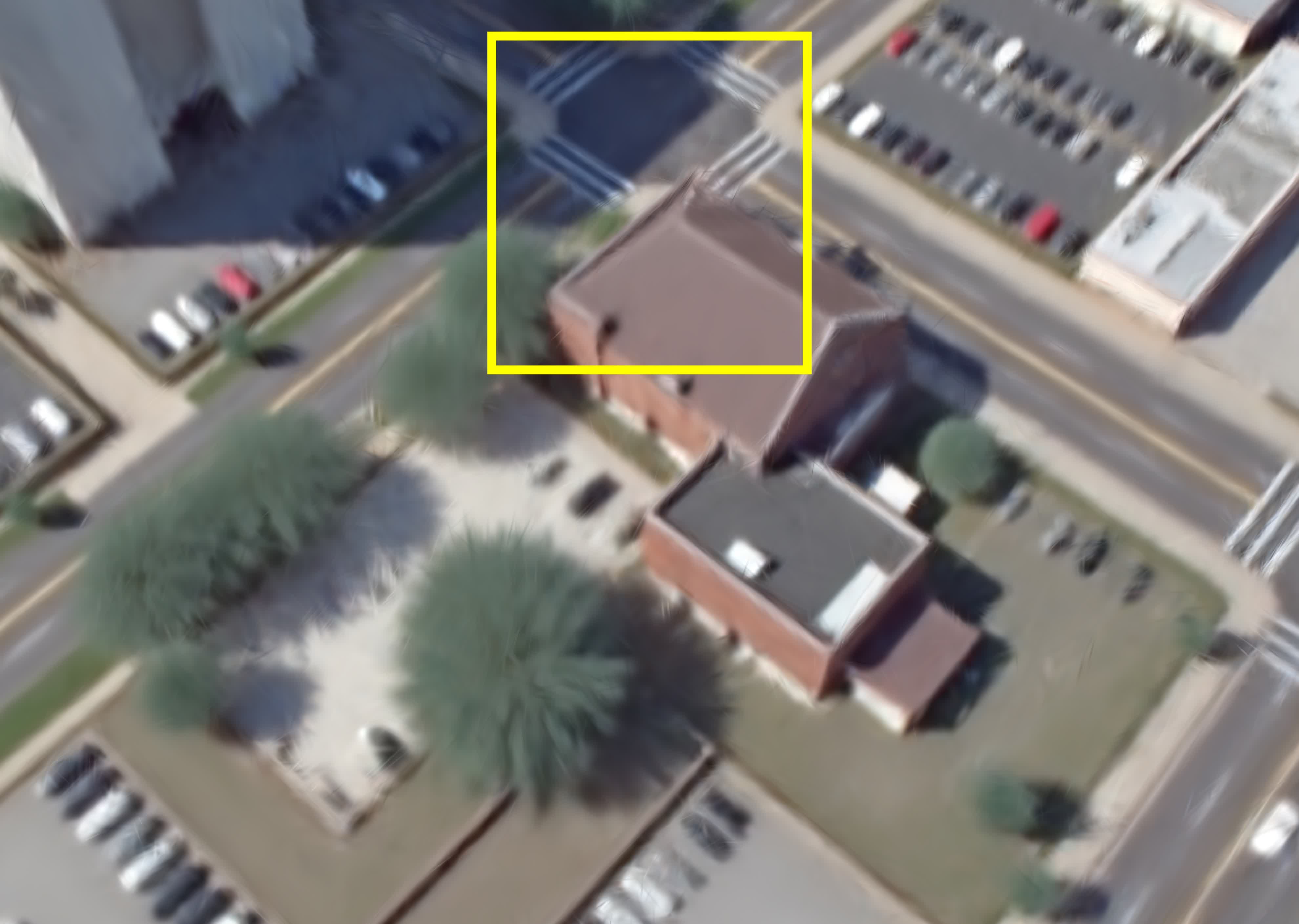}\end{subfigure}
    \begin{subfigure}{\colwthree}\myfigcrop{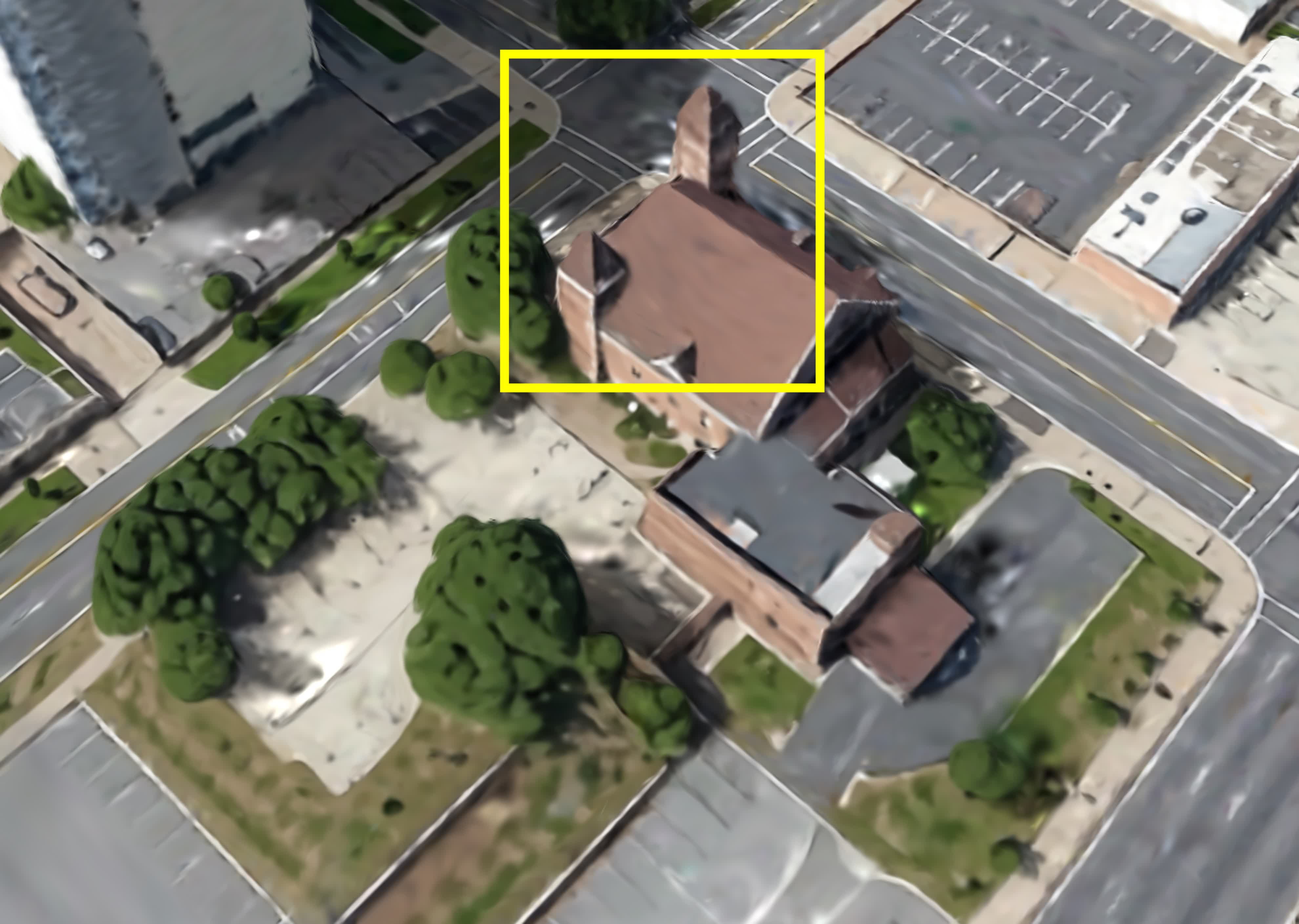}\end{subfigure} \\
    \vspace{0.5mm}

    % --- JAX-214 Section ---
    \rotatebox{90}{\makebox[0.12\linewidth][c]{\footnotesize JAX-214}}
    \begin{subfigure}{\colwthree}\myfigcrop{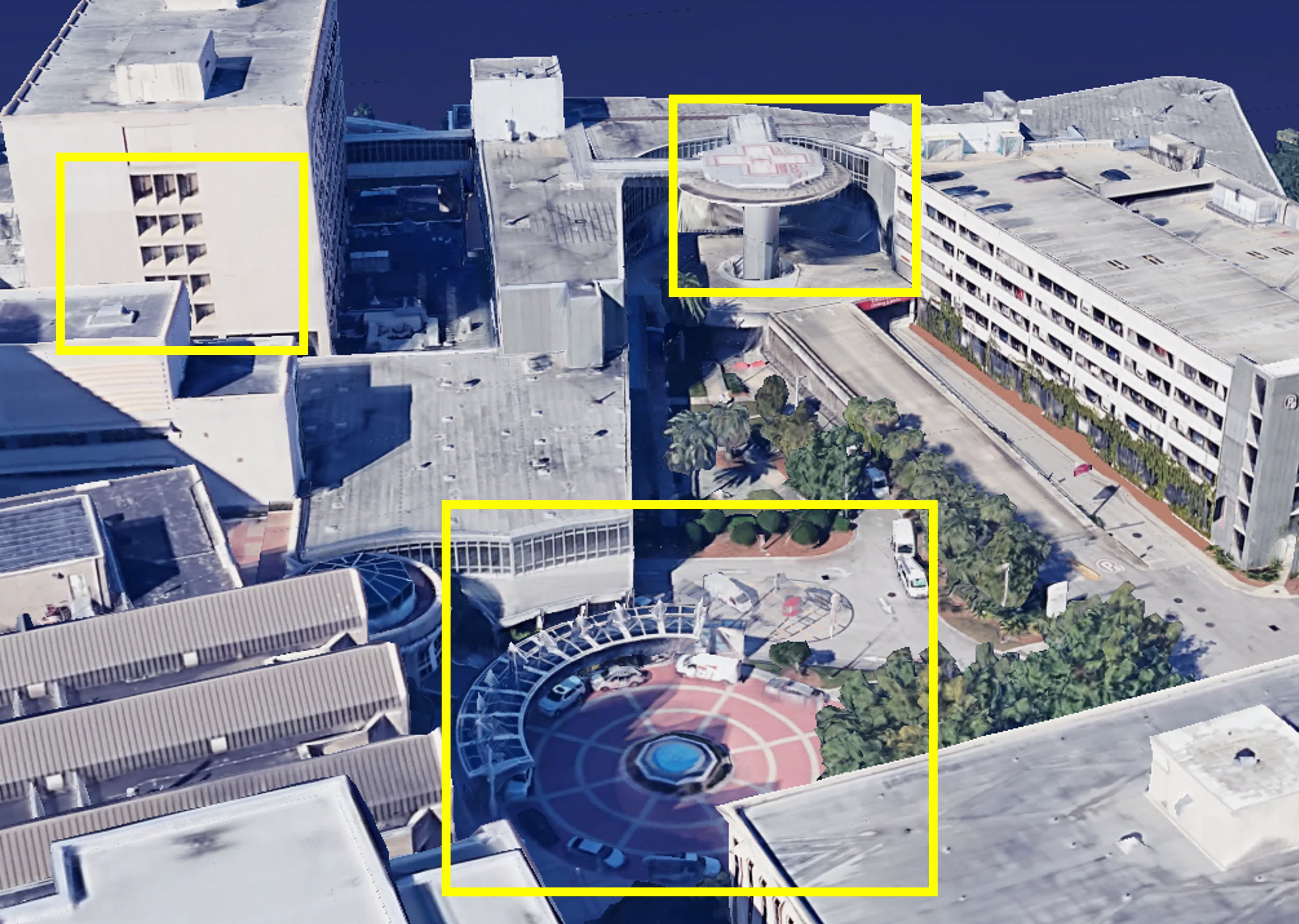}\end{subfigure}
    \begin{subfigure}{\colwthree}\myfigcrop{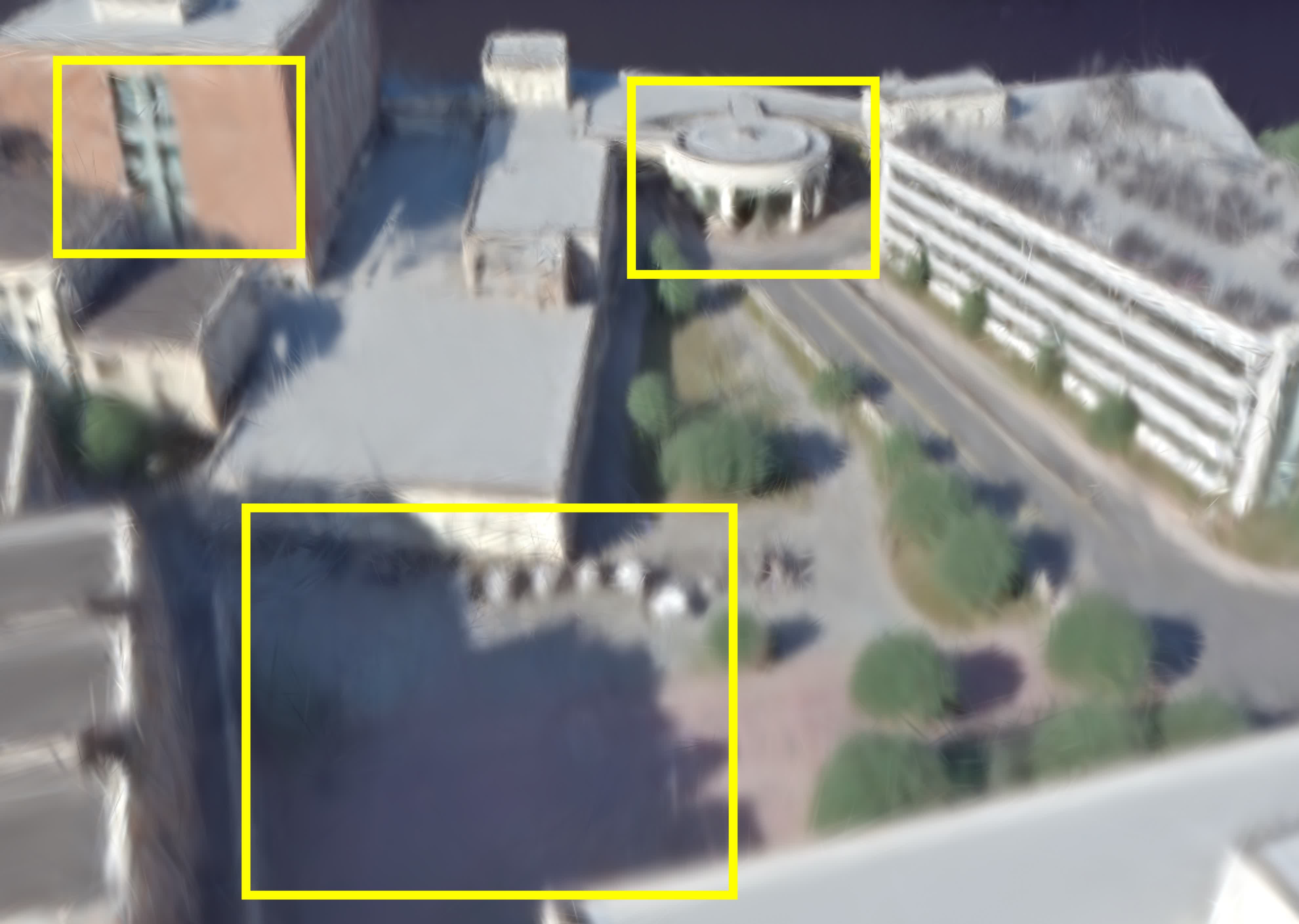}\end{subfigure}
    \begin{subfigure}{\colwthree}\myfigcrop{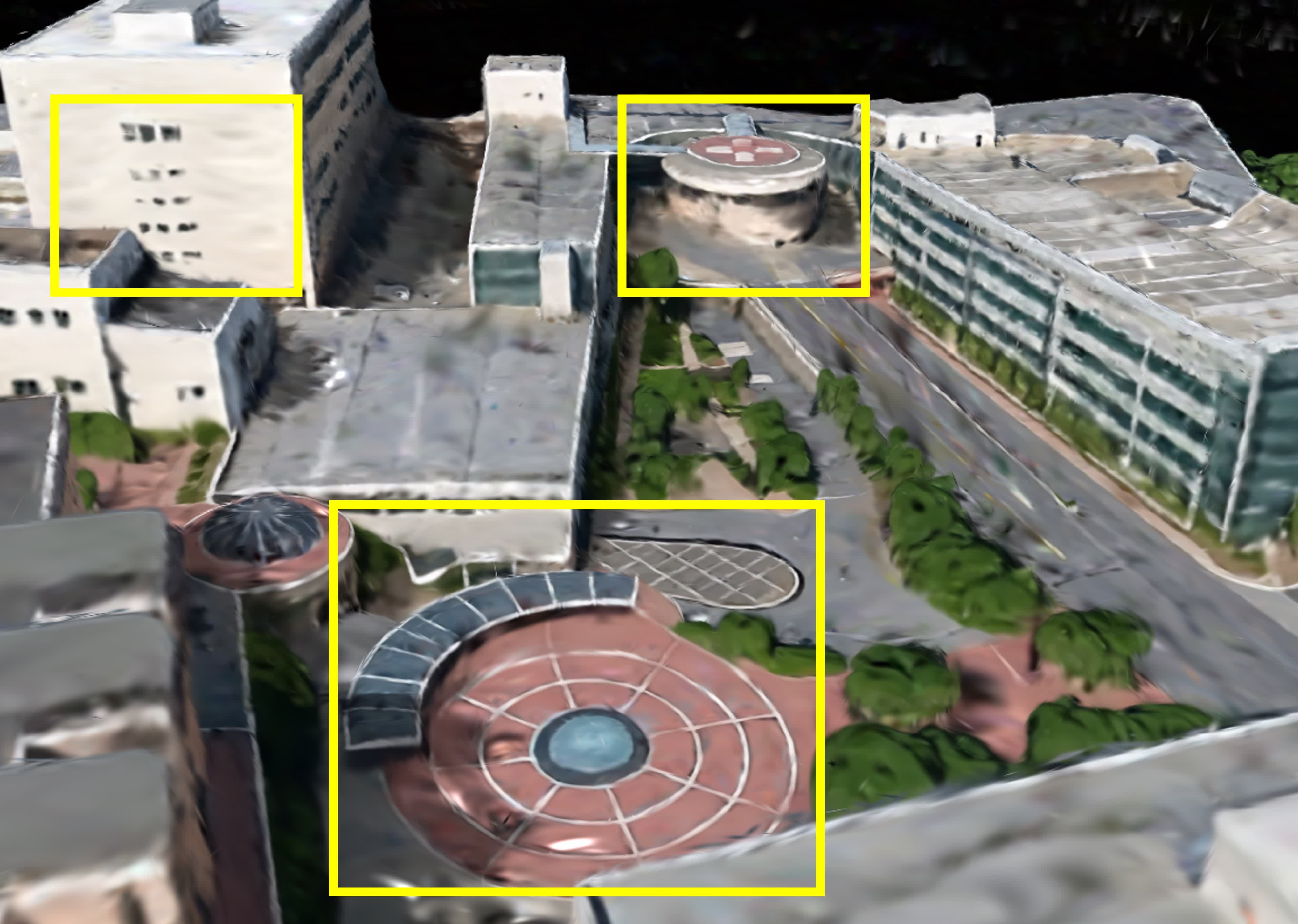}\end{subfigure} \\
    \vspace{0.5mm}

    % --- JAX-251 Section ---
    \rotatebox{90}{\makebox[0.12\linewidth][c]{\footnotesize JAX-251}}
    \begin{subfigure}{\colwthree}\myfigcrop{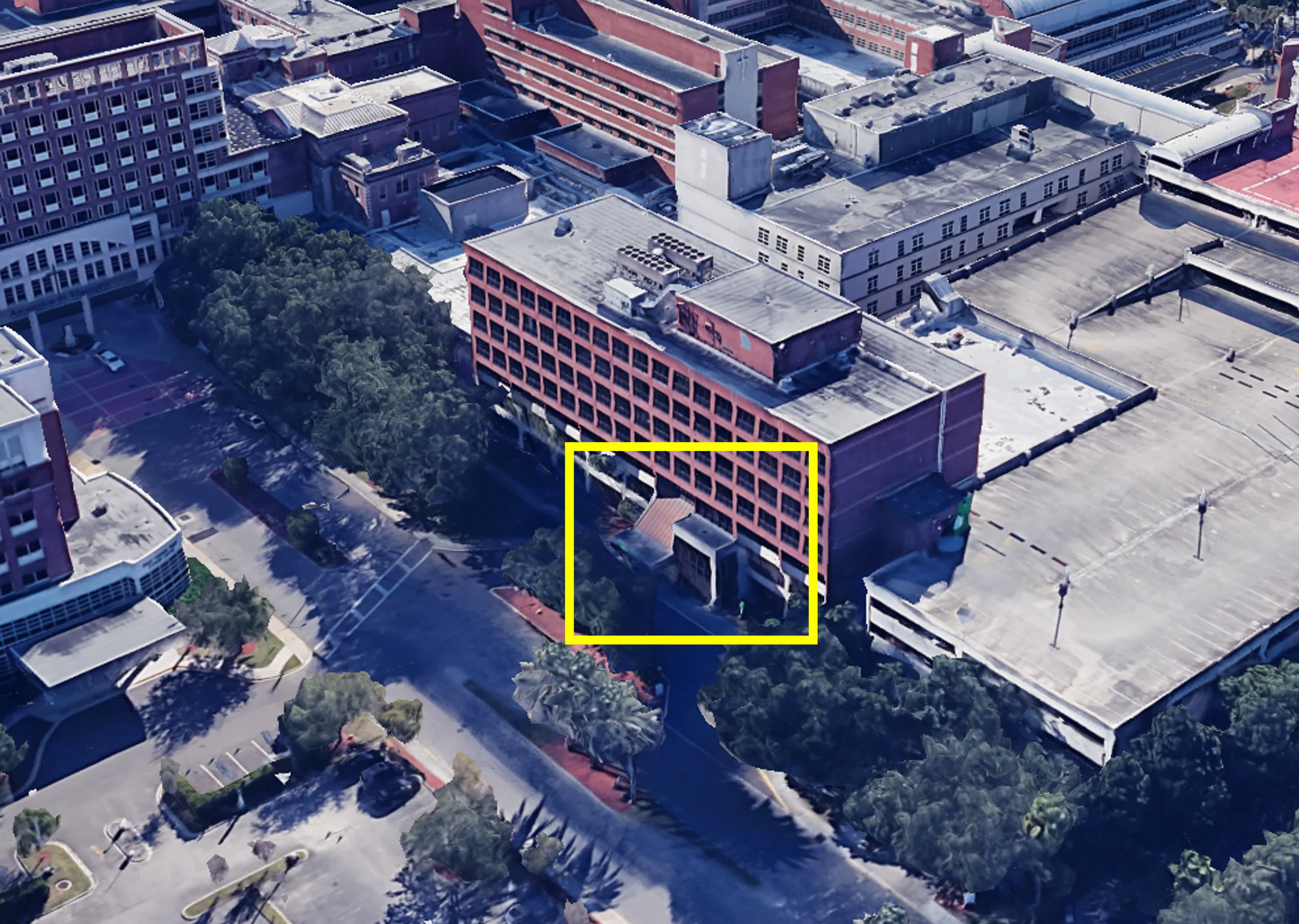}\end{subfigure}
    \begin{subfigure}{\colwthree}\myfigcrop{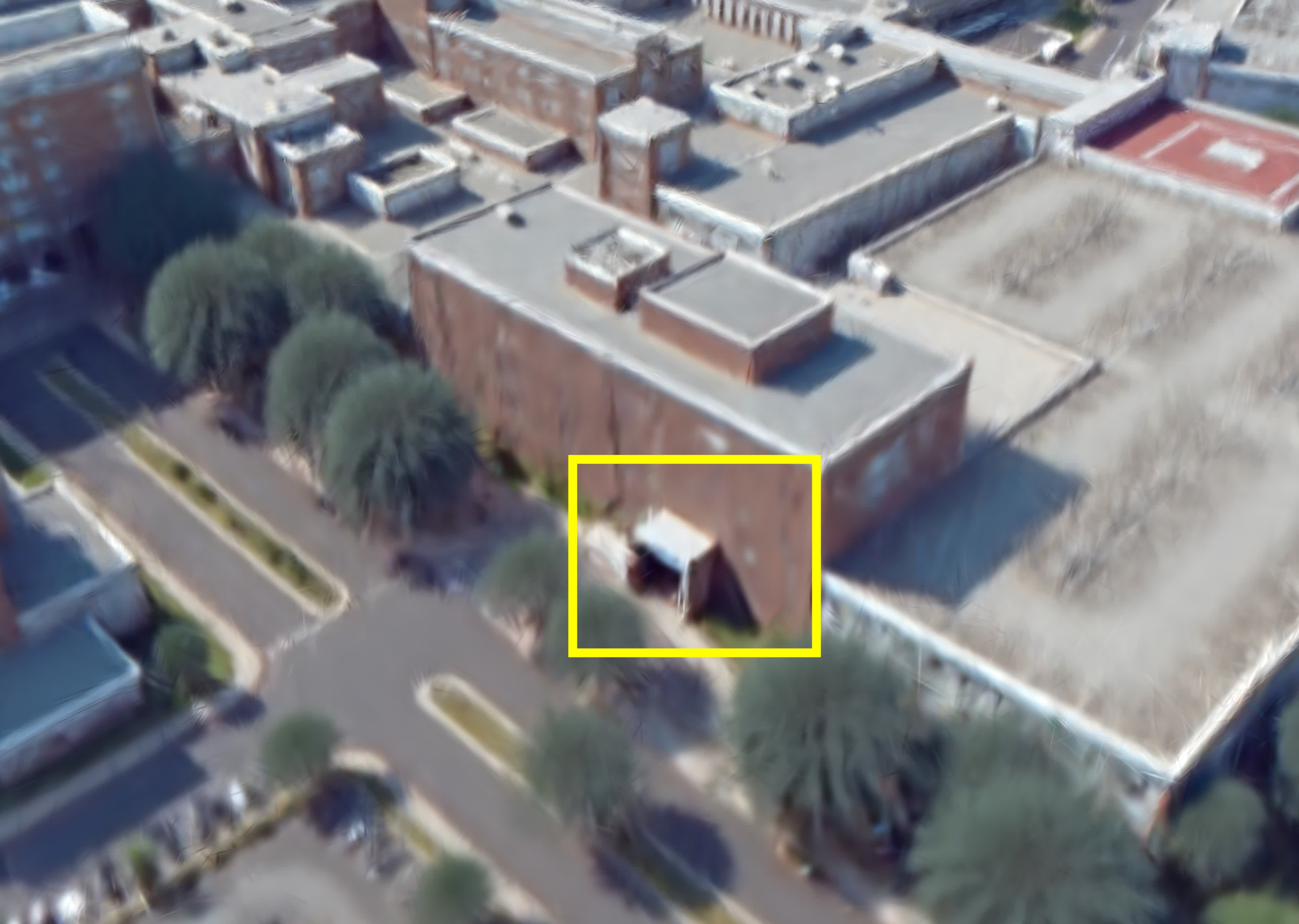}\end{subfigure}
    \begin{subfigure}{\colwthree}\myfigcrop{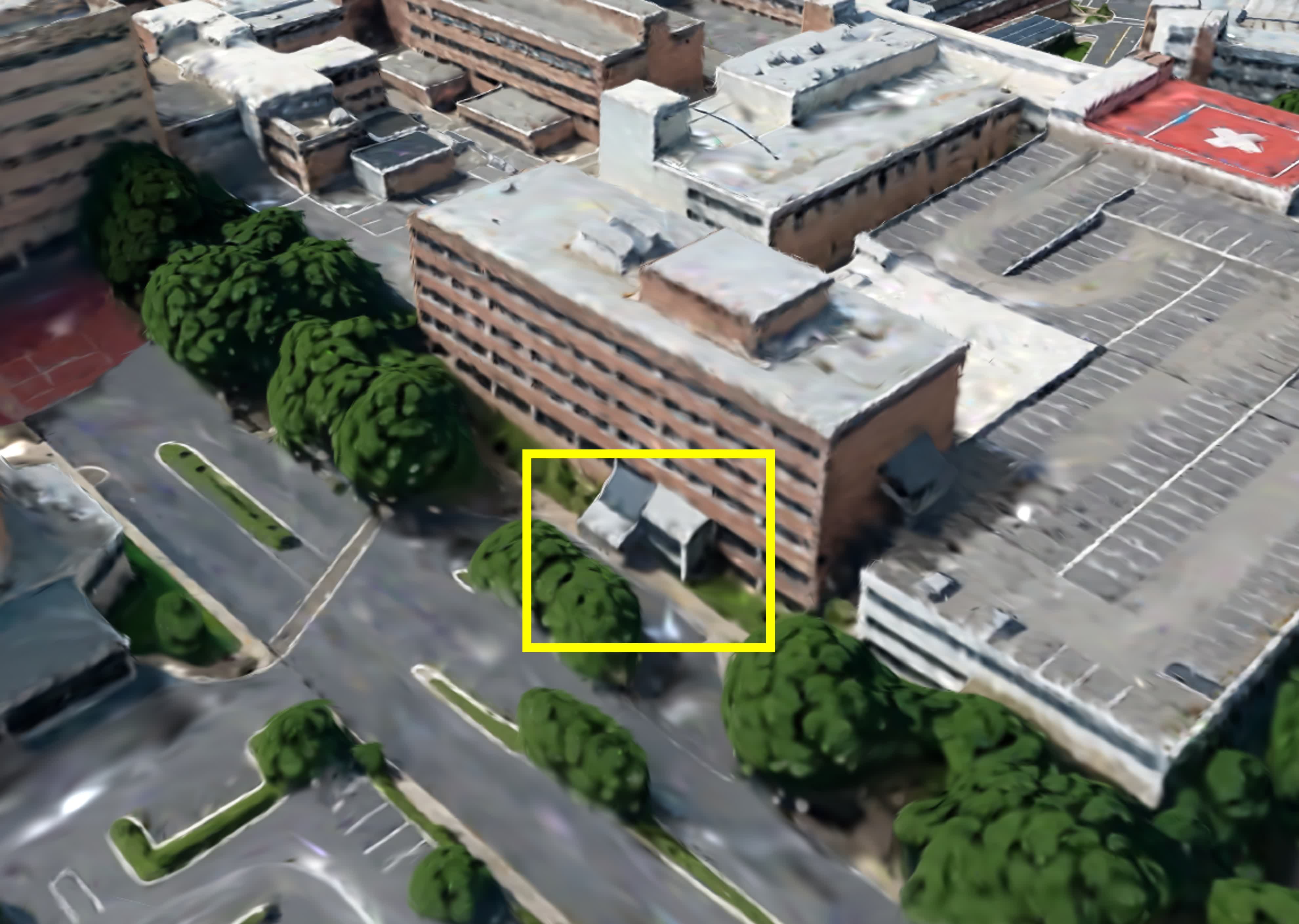}\end{subfigure} \\
    \vspace{1mm} 

    \caption{Comparative analysis of structural stability and artifacts. This figure contrasts our framework with Skyfall-GS and Google Earth images. Yellow boxes highlight regions where the baselines introduce structural hallucinations. In contrast, our method remains geometrically aligned with actual observations.}
    \label{fig:artifact_comparison}
\end{figure*}

A closer comparison with other diffusion-integrated frameworks further highlights the robustness of our approach (\Cref{fig:artifact_comparison}). Skyfall-GS often introduces artifacts that do not correspond to the actual scene when compared against Google Earth imagery. For example, it removes the roof of a church (JAX-168) and alters the structure of nearby buildings (JAX-214). In contrast, our pipeline exhibits substantially fewer such artifacts, with reconstructed geometry that remains visually consistent with the underlying structures observed in Google Earth.

Quantitative results on JAX, OMA, and IARPA sites align with the qualitative findings. Our method outperforms all baselines across all metrics (\Cref{tab:metrics_comparison_combined}). The distribution metrics (FID-CLIP, CMMD) indicate that our method aligns with the overall data distribution of Google Earth images. While baseline models show higher discrepancy in these metrics, our method achieves a FID-CLIP of 19.50 on JAX, 19.06 on OMA, and 29.50 on IARPA, which are 28\%, 28\%, and 45\% lower than the second-best results, respectively. These scores confirm that our method generates outputs that are structurally similar to Google Earth.

\begin{table}[!htb]
\centering
\caption{Quantitative evaluation of visual fidelity on JAX, OMA and IARPA sites. Metrics are validated against Google Earth reference imagery. $^\dagger$JAX rendered outputs for baseline methods are reproduced from \citet{lee2025SkyfallGS}; metrics are computed by us on the same evaluation set. Sat-NeRF results are only available for JAX.}
\label{tab:metrics_comparison_combined}
\resizebox{\linewidth}{!}{%
\begin{tabular}{llccccc}
\toprule
\multirow{2}{*}{\textbf{Dataset}} & \multirow{2}{*}{\textbf{Methods}} & \multicolumn{2}{c}{\textbf{Distributional}} & \multicolumn{3}{c}{\textbf{Pixel-level}} \\
\cmidrule(lr){3-4} \cmidrule(lr){5-7}
& & FID-CLIP $\downarrow$ & CMMD $\downarrow$ & PSNR $\uparrow$ & CW-SSIM $\uparrow$ & LPIPS $\downarrow$ \\
\midrule
\multirow{5}{*}{\textbf{JAX$^\dagger$}} 
& Sat-NeRF       & 86.90 & 4.786 & 9.77  & 0.302 & 0.773 \\
& EOGS           & 87.47 & 5.294 & 11.33 & 0.236 & 0.749 \\
& Mip-Splatting  & 86.94 & 5.414 & 11.21 & 0.322 & 0.753 \\
& Skyfall-GS     & 27.06 & 2.125 & 11.83 & 0.386 & 0.716 \\
& \textbf{Ours}  & \textbf{19.50} & \textbf{1.681} & \textbf{12.26} & \textbf{0.414} & \textbf{0.611} \\

\midrule
\multirow{4}{*}{\textbf{OMA}} 
& EOGS            & 70.56 & 4.760 & 10.47 & 0.322 & 0.778 \\
& Mip-Splatting  & 90.10 & 5.583 & 10.64 & 0.313 & 0.798 \\
& Skyfall-GS     & 26.31 & 1.725 & 10.61 & 0.352 & 0.759 \\
& \textbf{Ours}  & \textbf{19.06} & \textbf{1.520} & \textbf{10.68} & \textbf{0.358} & \textbf{0.677} \\

\midrule
\multirow{4}{*}{\textbf{IARPA}} 
& EOGS            & 86.26 & 4.887 & 11.17 & 0.258 & 0.806 \\
& Mip-Splatting  & 88.20 & 5.031 & 11.11 & 0.294 & 0.816 \\
& Skyfall-GS     & 53.72 & 2.588 & 11.08 & 0.300 & 0.824 \\
& \textbf{Ours}  & \textbf{29.50} & \textbf{1.790} & \textbf{11.65} & \textbf{0.300} & \textbf{0.769} \\
\bottomrule
\end{tabular}%
}
\end{table}

In terms of pixel-level metrics (PSNR, CW-SSIM, LPIPS), our model demonstrates higher correspondence in direct pixel values and structural details. The improved PSNR and LPIPS scores across both datasets show that the reconstructed geometry and textures are closer to the target pixels. In particular, the CW-SSIM score (0.414) reflects that our approach maintains edge consistency and building boundaries even under varying satellite viewpoints.

\subsubsection{Geometric accuracy}
%\RQ{just geometric accuracy}
\label{sec:results:geometric}

\begin{figure*}[p]
    \centering
    \small
    
    \newcommand{\myimg}[1]{%
        \includegraphics[width=\linewidth, height=\linewidth, keepaspectratio=false, trim=0 30 0 30, clip]{#1}%
    }
    \def\colw{0.14\linewidth}

    % Header Row
    \makebox[0.02\linewidth]{}
    \makebox[\colw]{GT}
    \makebox[\colw]{ASP}
    \makebox[\colw]{SAT-NGP}
    \makebox[\colw]{EOGS}
    \makebox[\colw]{Skyfall-GS}
    %\makebox[\colw]{GU-GS}
    \makebox[\colw]{Ours} \\
    \vspace{1mm}

    % --- JAX Section ---

    \rotatebox{90}{\makebox[\colw]{\footnotesize JAX-004}}
    \begin{subfigure}{\colw}\myimg{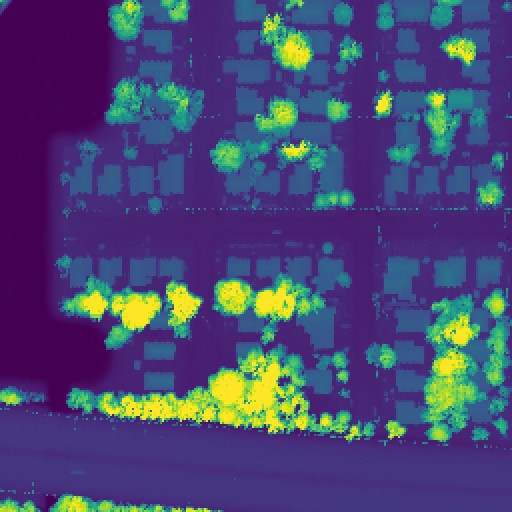}\end{subfigure}
    \begin{subfigure}{\colw}\myimg{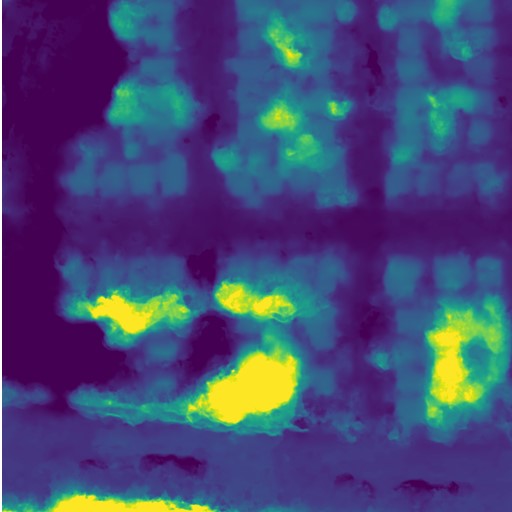}\end{subfigure}
    \begin{subfigure}{\colw}\myimg{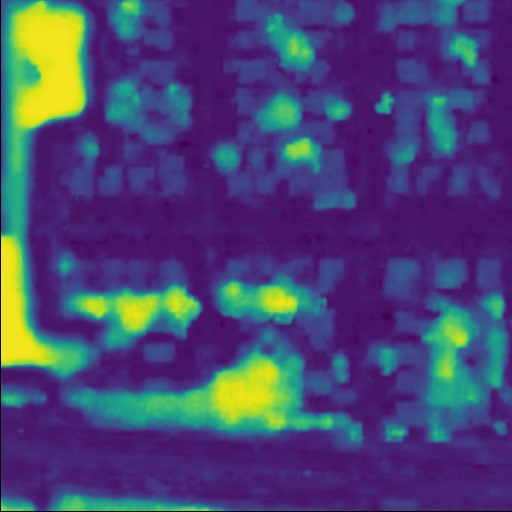}\end{subfigure}
    \begin{subfigure}{\colw}\myimg{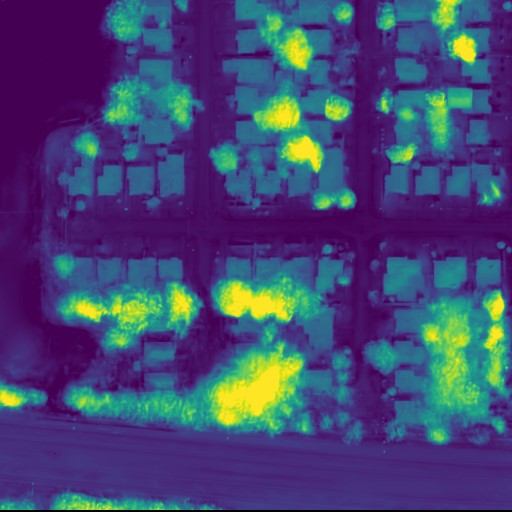}\end{subfigure}
    \begin{subfigure}{\colw}\myimg{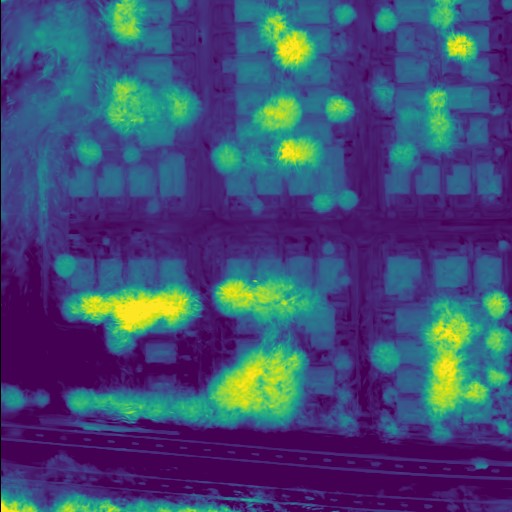}\end{subfigure}
    %\begin{subfigure}{\colw}\myimg{Figures/Figure_mae/JAX_004_GU-GS_vis.png}\end{subfigure}    
    \begin{subfigure}{\colw}\myimg{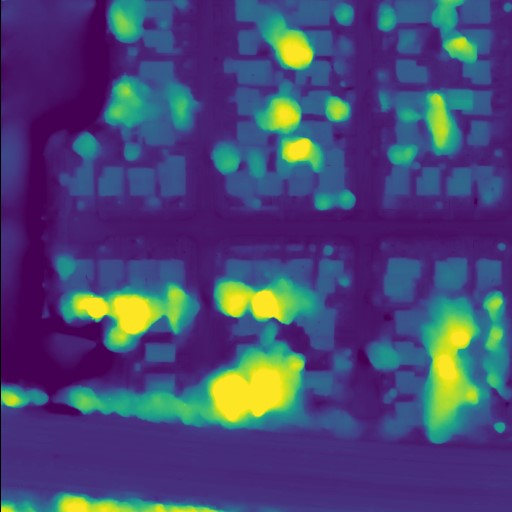}\end{subfigure} \\
    \vspace{0.5mm}
    
    \rotatebox{90}{\makebox[\colw]{\footnotesize JAX-068}}
    \begin{subfigure}{\colw}\myimg{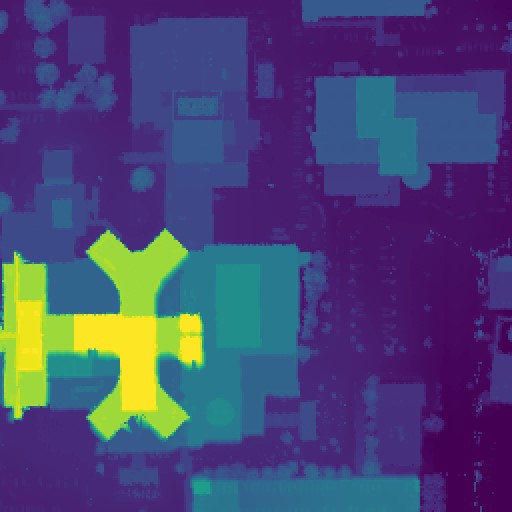}\end{subfigure}
    \begin{subfigure}{\colw}\myimg{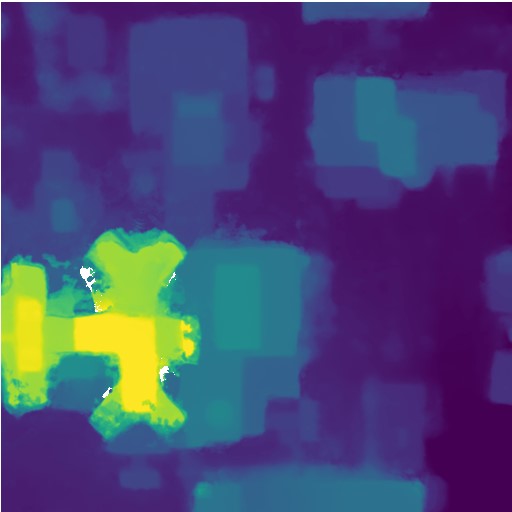}\end{subfigure}
    \begin{subfigure}{\colw}\myimg{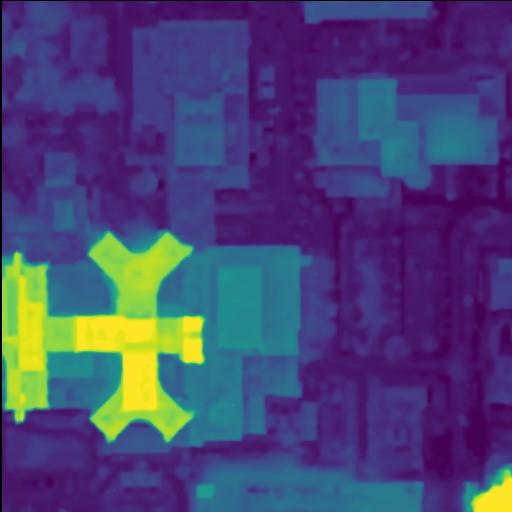}\end{subfigure}
    \begin{subfigure}{\colw}\myimg{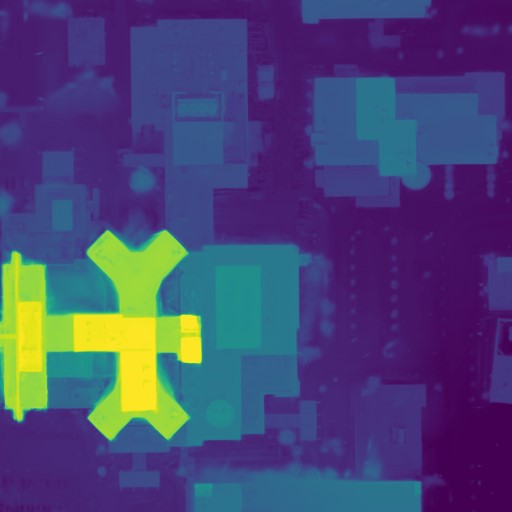}\end{subfigure}
    \begin{subfigure}{\colw}\myimg{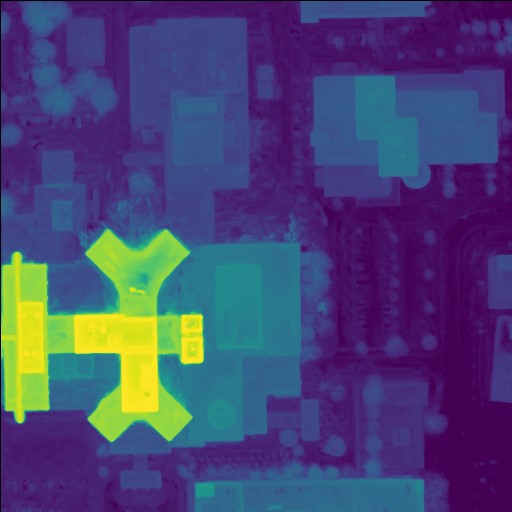}\end{subfigure}
    %\begin{subfigure}{\colw}\myimg{Figures/Figure_mae/JAX_068_GU-GS_vis.png}\end{subfigure}    
    \begin{subfigure}{\colw}\myimg{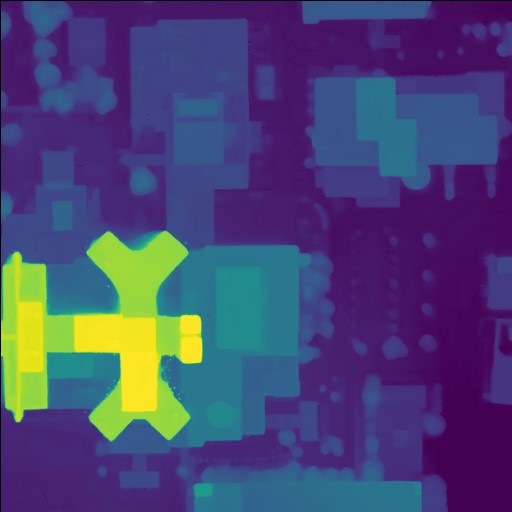}\end{subfigure} \\
    \vspace{0.5mm}

    \rotatebox{90}{\makebox[\colw]{\footnotesize JAX-214}}
    \begin{subfigure}{\colw}\myimg{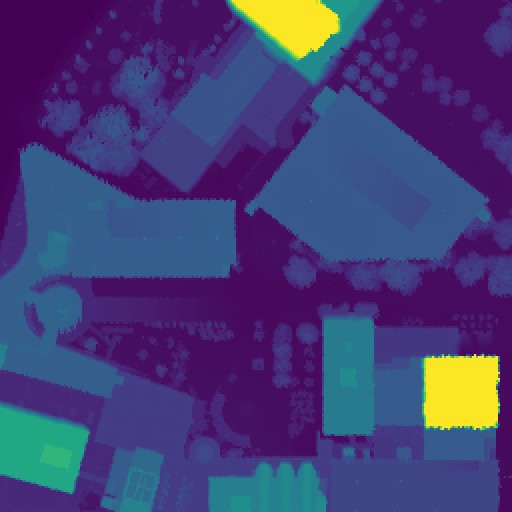}\end{subfigure}
    \begin{subfigure}{\colw}\myimg{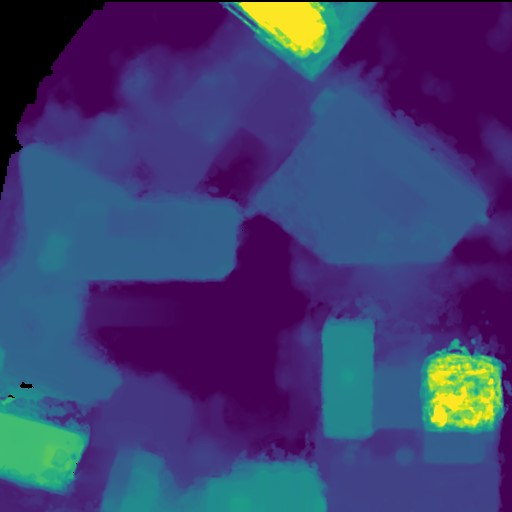}\end{subfigure}
    \begin{subfigure}{\colw}\myimg{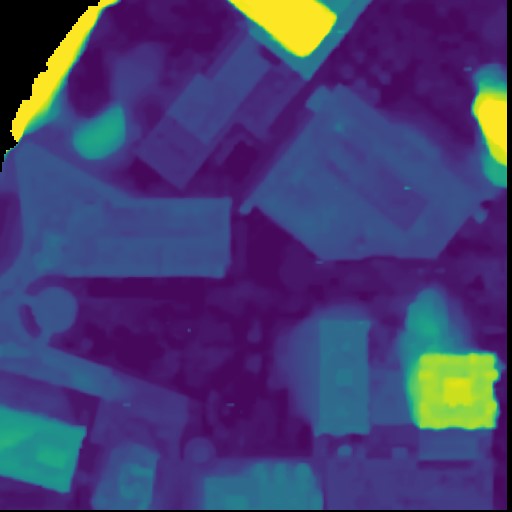}\end{subfigure}
    \begin{subfigure}{\colw}\myimg{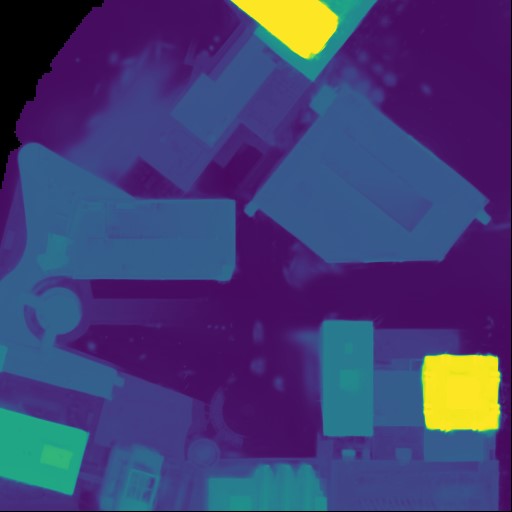}\end{subfigure}
    \begin{subfigure}{\colw}\myimg{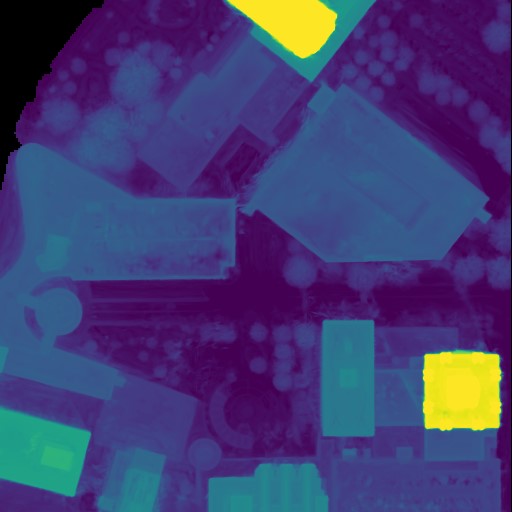}\end{subfigure}
    %\begin{subfigure}{\colw}\myimg{Figures/Figure_mae/JAX_214_GU-GS_vis.png}\end{subfigure}   
    \begin{subfigure}{\colw}\myimg{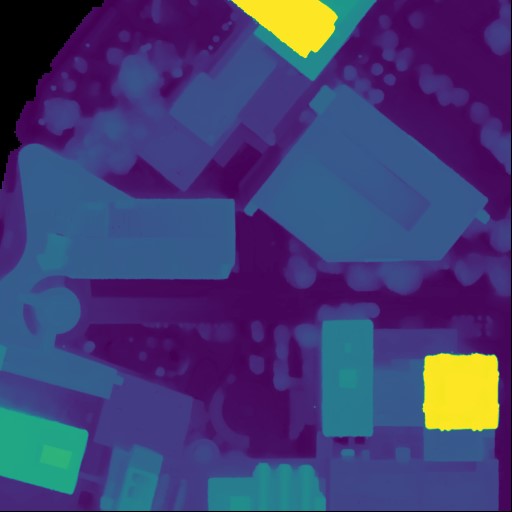}\end{subfigure} \\
    \vspace{0.5mm}

    \rotatebox{90}{\makebox[\colw]{\footnotesize JAX-168}}
    \begin{subfigure}{\colw}\myimg{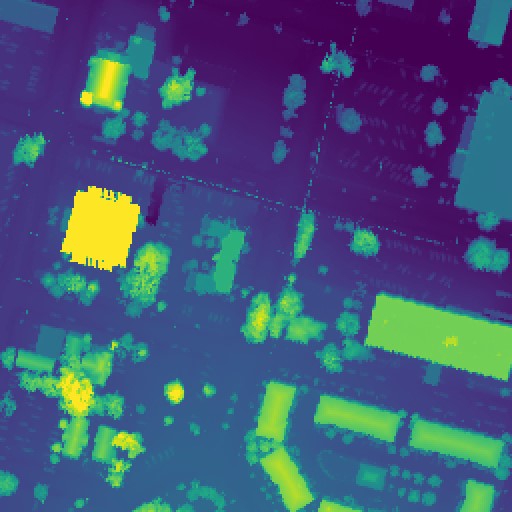}\end{subfigure}
    \begin{subfigure}{\colw}\myimg{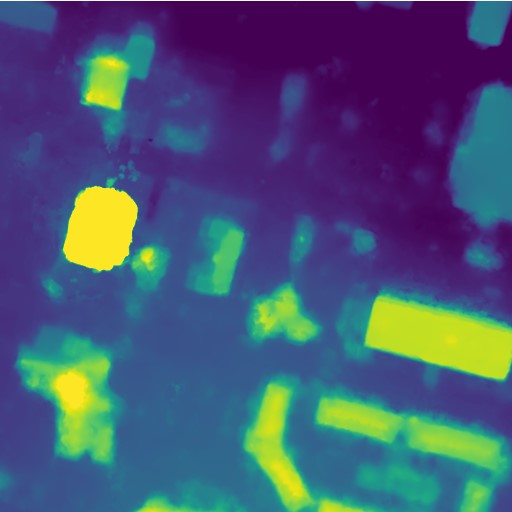}\end{subfigure}
    \begin{subfigure}{\colw}\myimg{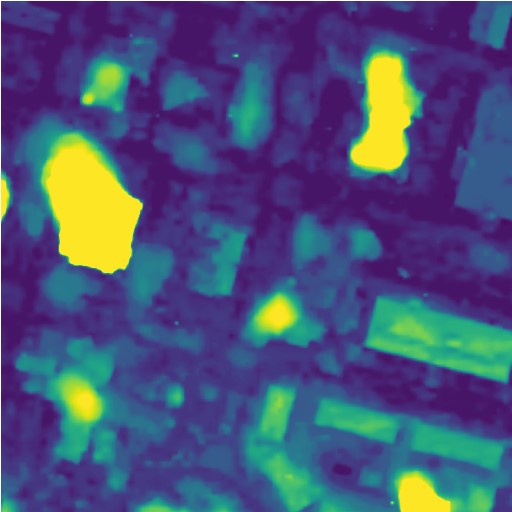}\end{subfigure}
    \begin{subfigure}{\colw}\myimg{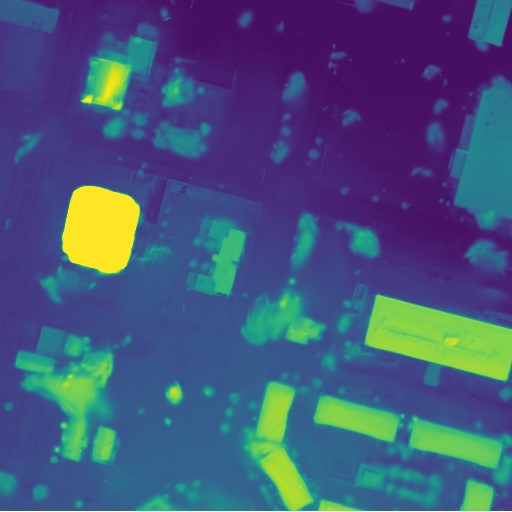}\end{subfigure}
    \begin{subfigure}{\colw}\myimg{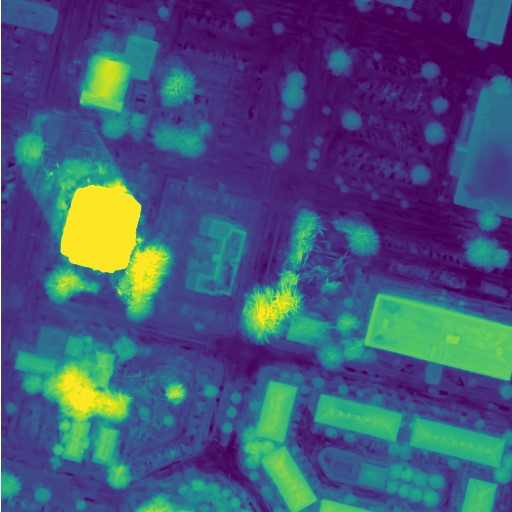}\end{subfigure}
    %\begin{subfigure}{\colw}\myimg{Figures/Figure_mae/JAX_168_GU-GS_vis.png}\end{subfigure}  
    \begin{subfigure}{\colw}\myimg{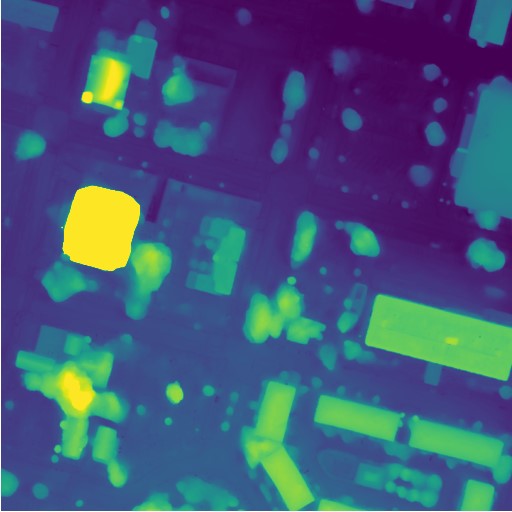}\end{subfigure} \\
    \vspace{0.5mm}

    \rotatebox{90}{\makebox[\colw]{\footnotesize OMA-212}}
    \begin{subfigure}{\colw}\myimg{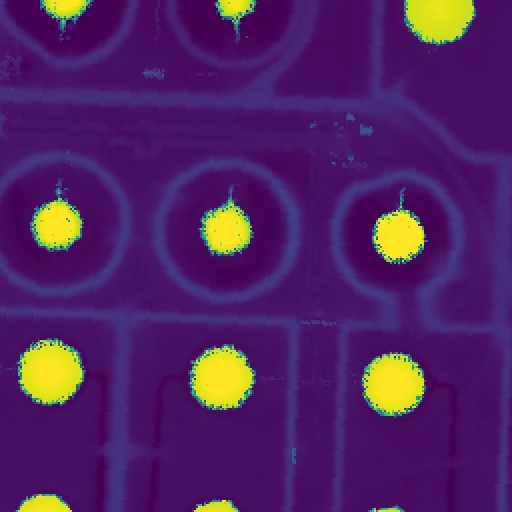}\end{subfigure}
    \begin{subfigure}{\colw}\myimg{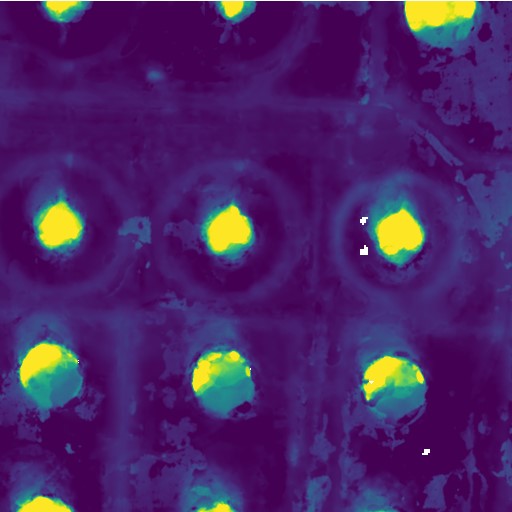}\end{subfigure}
    \begin{subfigure}{\colw}\myimg{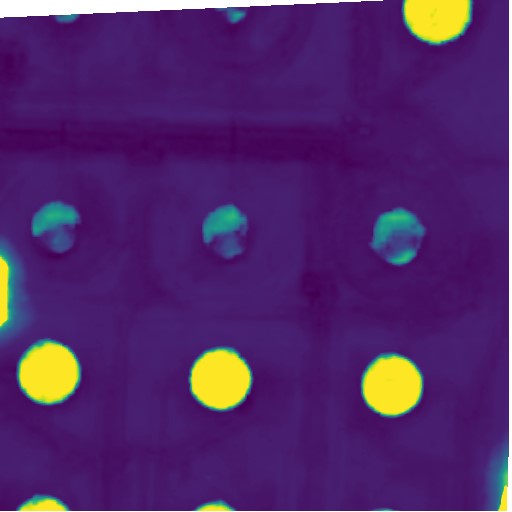}\end{subfigure}
    \begin{subfigure}{\colw}\myimg{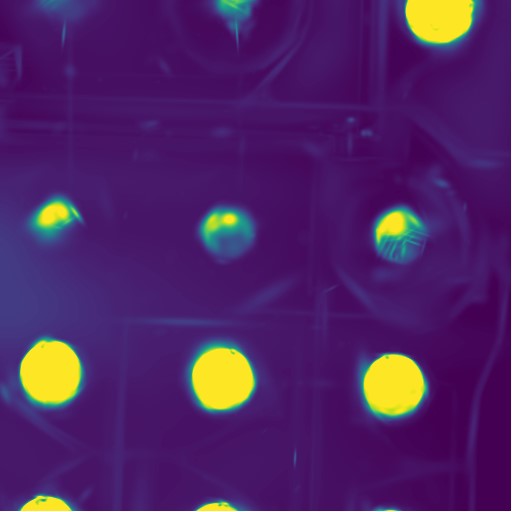}\end{subfigure}
    \begin{subfigure}{\colw}\myimg{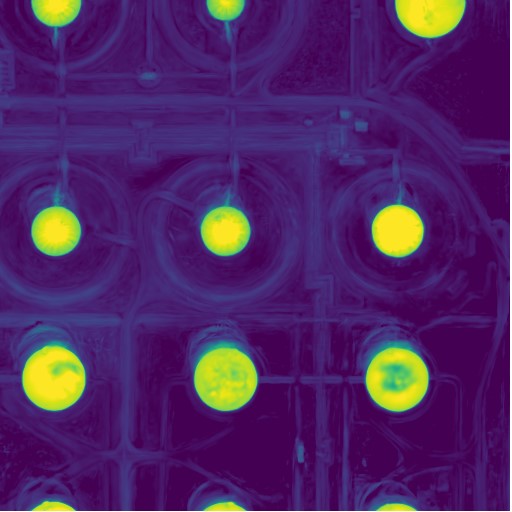}\end{subfigure}
    %\begin{subfigure}{\colw}\myimg{Figures/Figure_mae/OMA_212_GU-GS_vis.png}\end{subfigure}
    \begin{subfigure}{\colw}\myimg{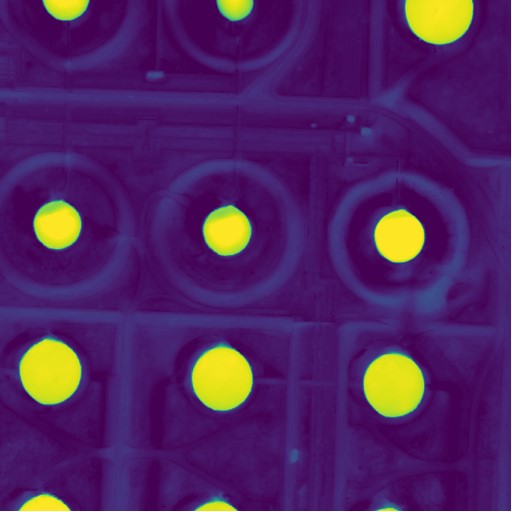}\end{subfigure} \\
    \vspace{0.5mm}

    \rotatebox{90}{\makebox[\colw]{\footnotesize OMA-315}}
    \begin{subfigure}{\colw}\myimg{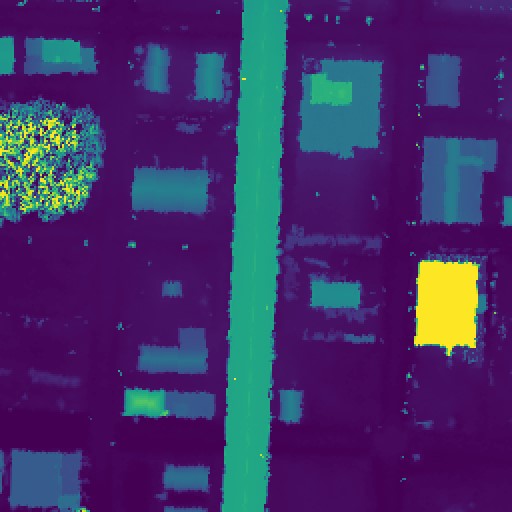}\end{subfigure}
    \begin{subfigure}{\colw}\myimg{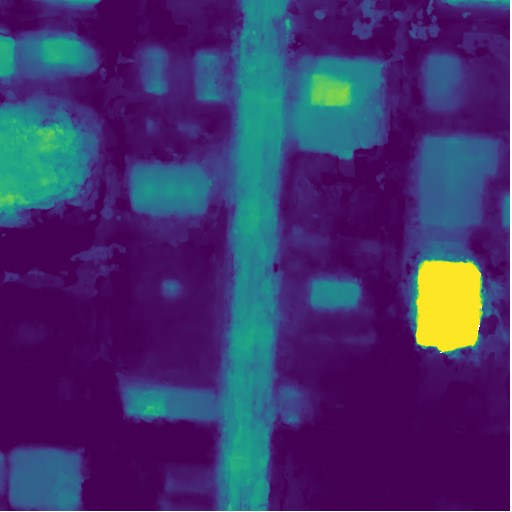}\end{subfigure}
    \begin{subfigure}{\colw}\myimg{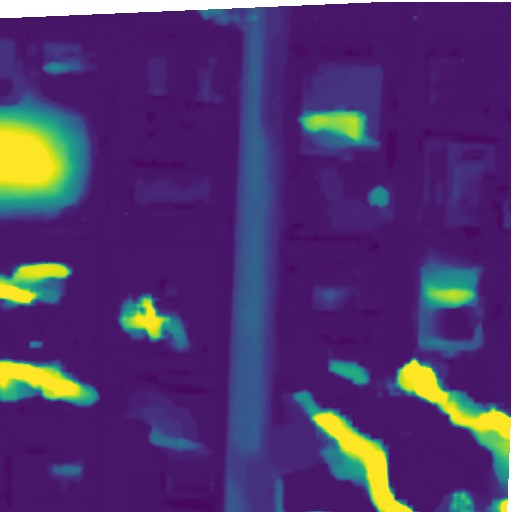}\end{subfigure}
    \begin{subfigure}{\colw}\myimg{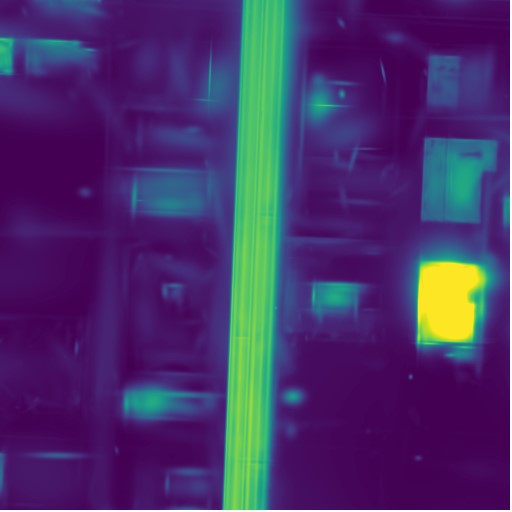}\end{subfigure}
    \begin{subfigure}{\colw}\myimg{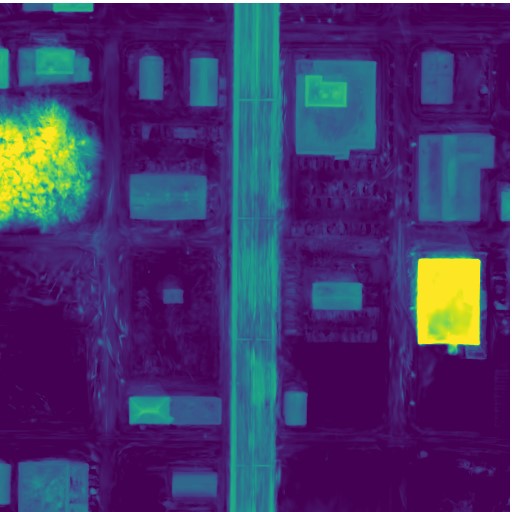}\end{subfigure}
    %\begin{subfigure}{\colw}\myimg{Figures/Figure_mae/OMA_315_GU-GS_vis.png}\end{subfigure}
    \begin{subfigure}{\colw}\myimg{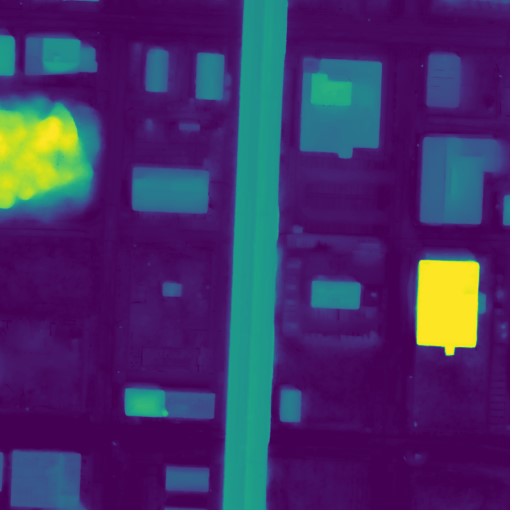}\end{subfigure} \\
    \vspace{0.5mm}

    % --- IARPA Section ---
    \rotatebox{90}{\makebox[\colw]{\footnotesize IARPA-001}}
    \begin{subfigure}{\colw}\myimg{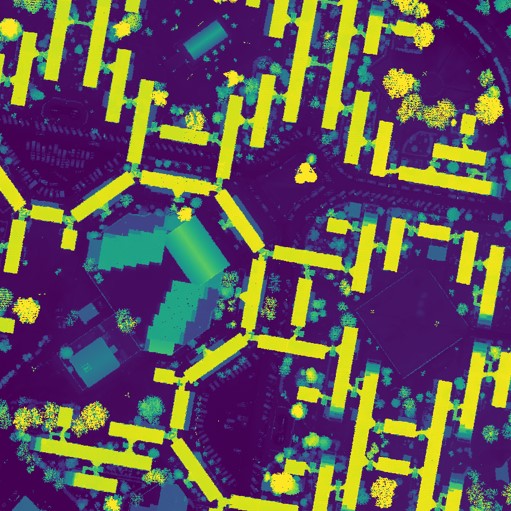}\end{subfigure}
    \begin{subfigure}{\colw}\myimg{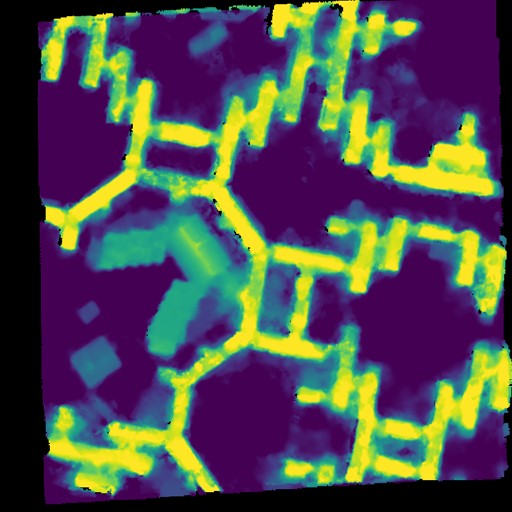}\end{subfigure}
    \begin{subfigure}{\colw}\myimg{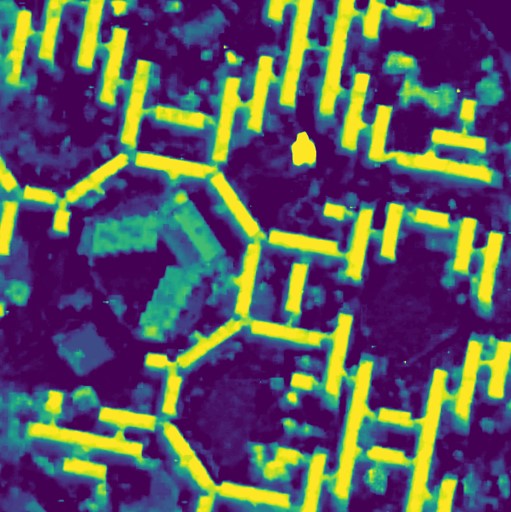}\end{subfigure}
    \begin{subfigure}{\colw}\myimg{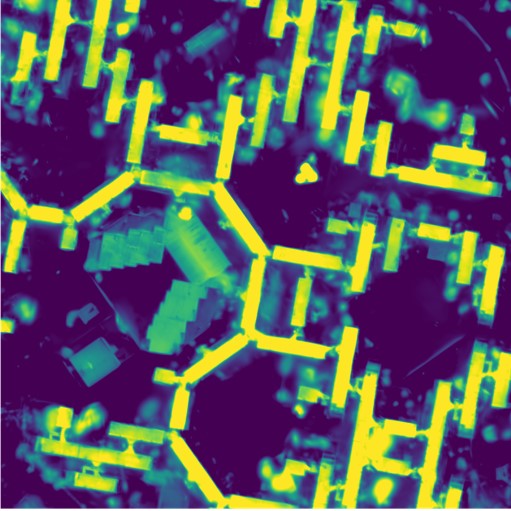}\end{subfigure}
    \begin{subfigure}{\colw}\myimg{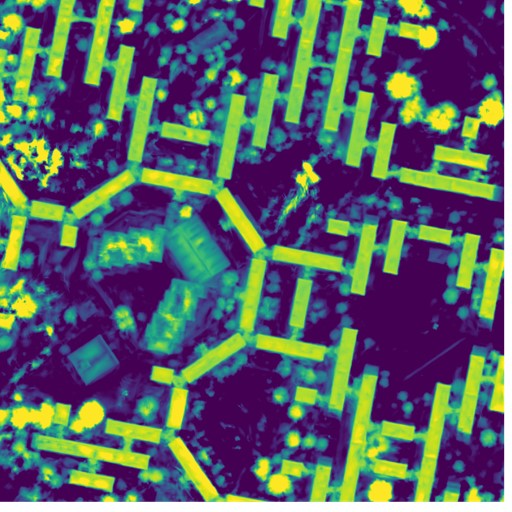}\end{subfigure}
    %\begin{subfigure}{\colw}\myimg{Figures/Figure_mae/IARPA_001_GU-GS_vis.png}\end{subfigure}
    \begin{subfigure}{\colw}\myimg{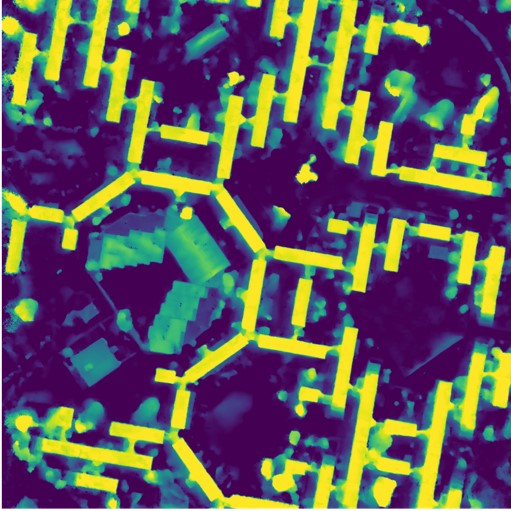}\end{subfigure} \\
    \vspace{0.5mm}

    \rotatebox{90}{\makebox[\colw]{\footnotesize IARPA-003}}
    \begin{subfigure}{\colw}\myimg{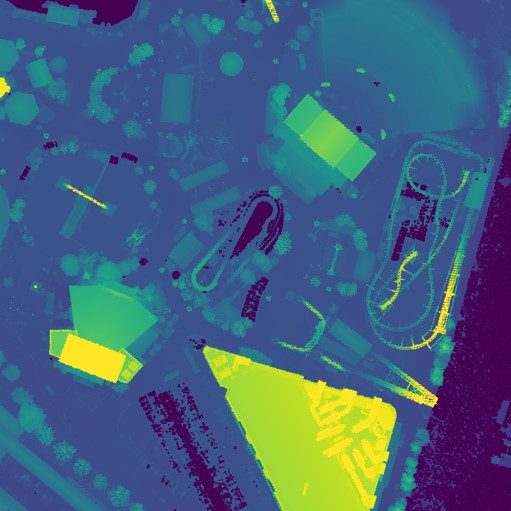}\end{subfigure}
    \begin{subfigure}{\colw}\myimg{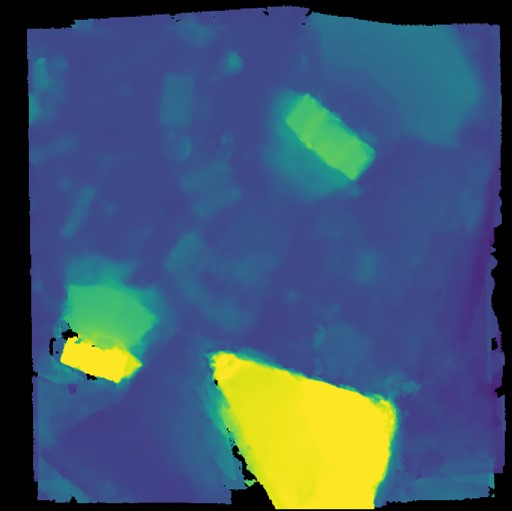}\end{subfigure}
    \begin{subfigure}{\colw}\myimg{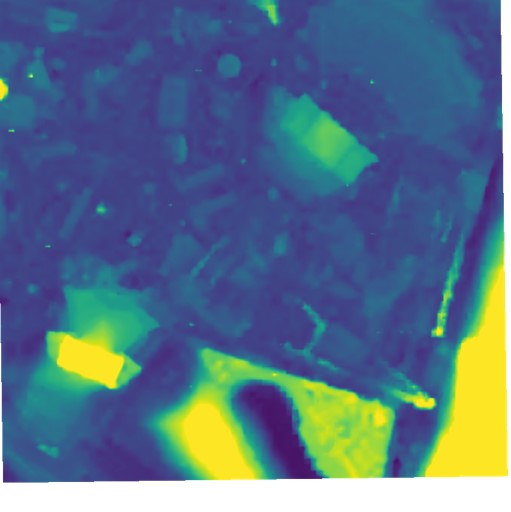}\end{subfigure}
    \begin{subfigure}{\colw}\myimg{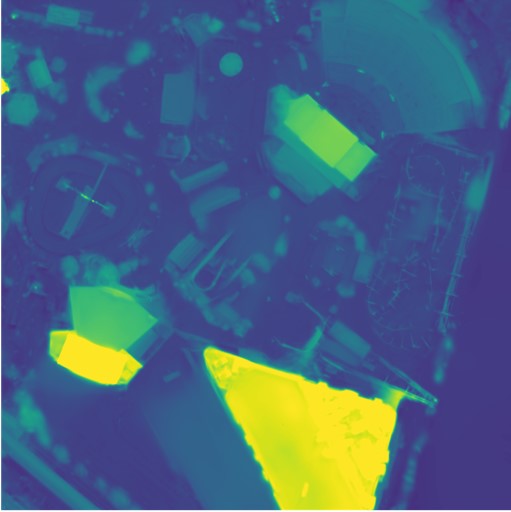}\end{subfigure}
    \begin{subfigure}{\colw}\myimg{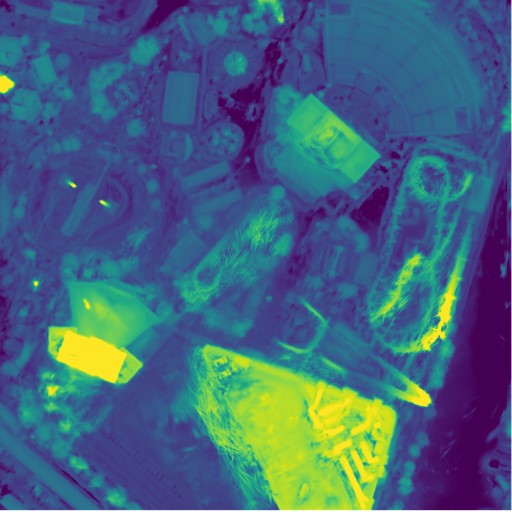}\end{subfigure}
    %\begin{subfigure}{\colw}\myimg{Figures/Figure_mae/IARPA_003_GU-GS_vis.png}\end{subfigure}
    \begin{subfigure}{\colw}\myimg{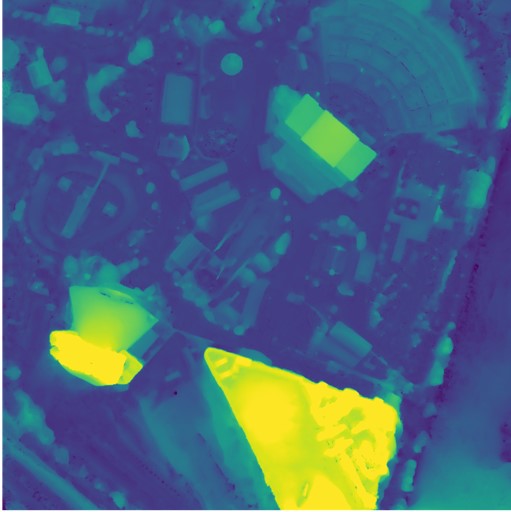}\end{subfigure} \\

    \caption{Qualitative comparison of reconstructed Digital Surface Models (DSMs) across JAX, OMA, and IARPA sites. Our method consistently generates the most sharply defined building geometries and fine structural details when compared to traditional photogrammetric (ASP) and learnable representative (Sat-NGP, EOGS, Skyfall-GS) baselines. While EOGS often produces oversmoothed surfaces that omit small structures, and Skyfall-GS suffers from transparent roofs and geometric instability, our approach maintains strict surface solidity and captures complex urban features.}
    \label{fig:comprehensive_vis_comparison}
\end{figure*}

\begin{figure*}[!h] 
    \centering
    \small
    \newcommand{\myfigcrop}[1]{%
        \includegraphics[width=\linewidth, keepaspectratio, trim=0 5 0 5, clip]{#1}%
    }
    
    \def\colwfour{0.23\linewidth} 

    \makebox[\colwfour]{LiDAR (GT)}
    \makebox[\colwfour]{EOGS}
    \makebox[\colwfour]{Skyfall-GS}
    \makebox[\colwfour]{Ours}\\
    \vspace{1mm}

    % --- JAX-068 Section ---
    \rotatebox{90}{\makebox[0.12\linewidth][c]{\footnotesize JAX-068}}
    \begin{subfigure}{\colwfour}\myfigcrop{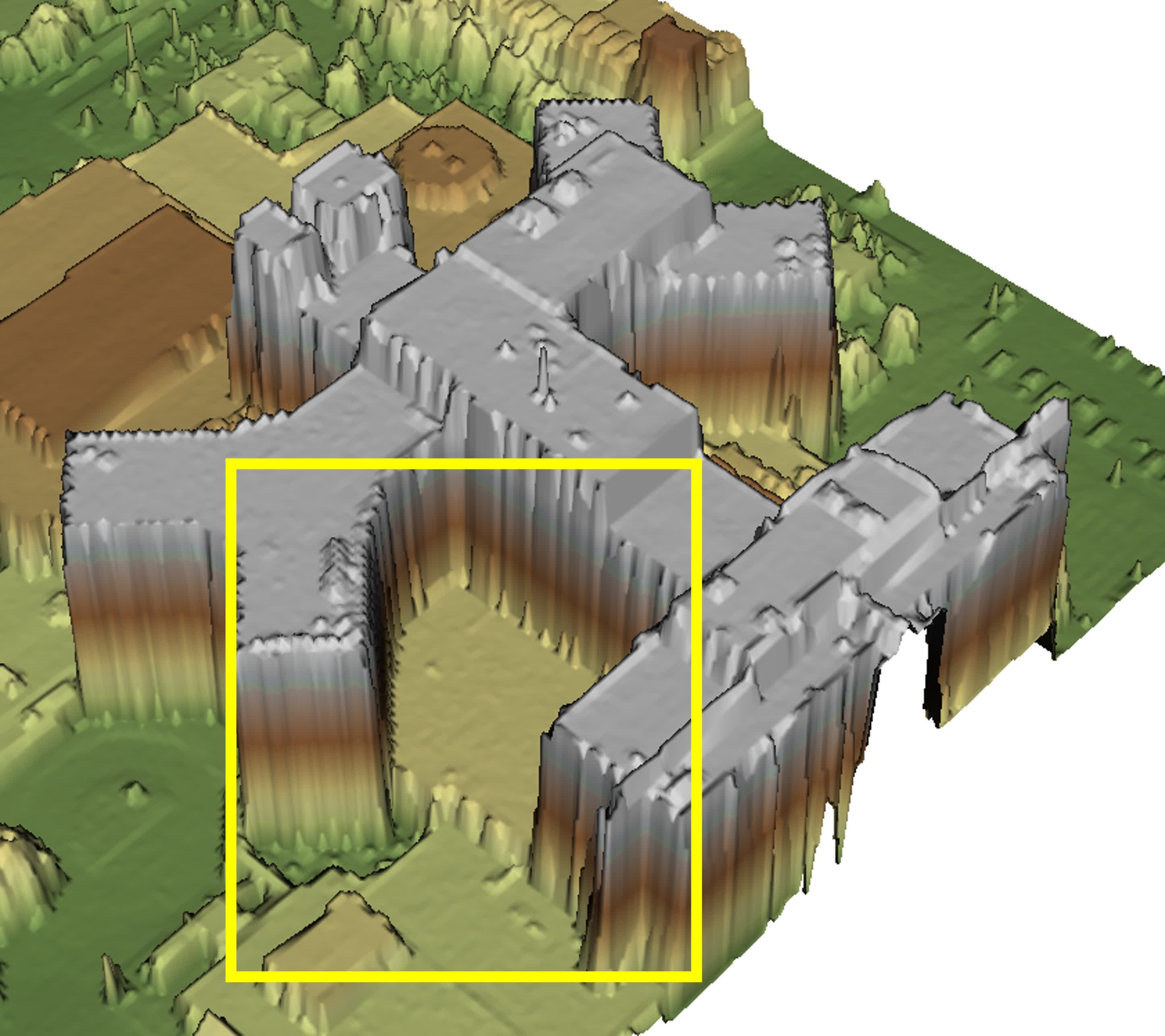}\end{subfigure}
    \begin{subfigure}{\colwfour}\myfigcrop{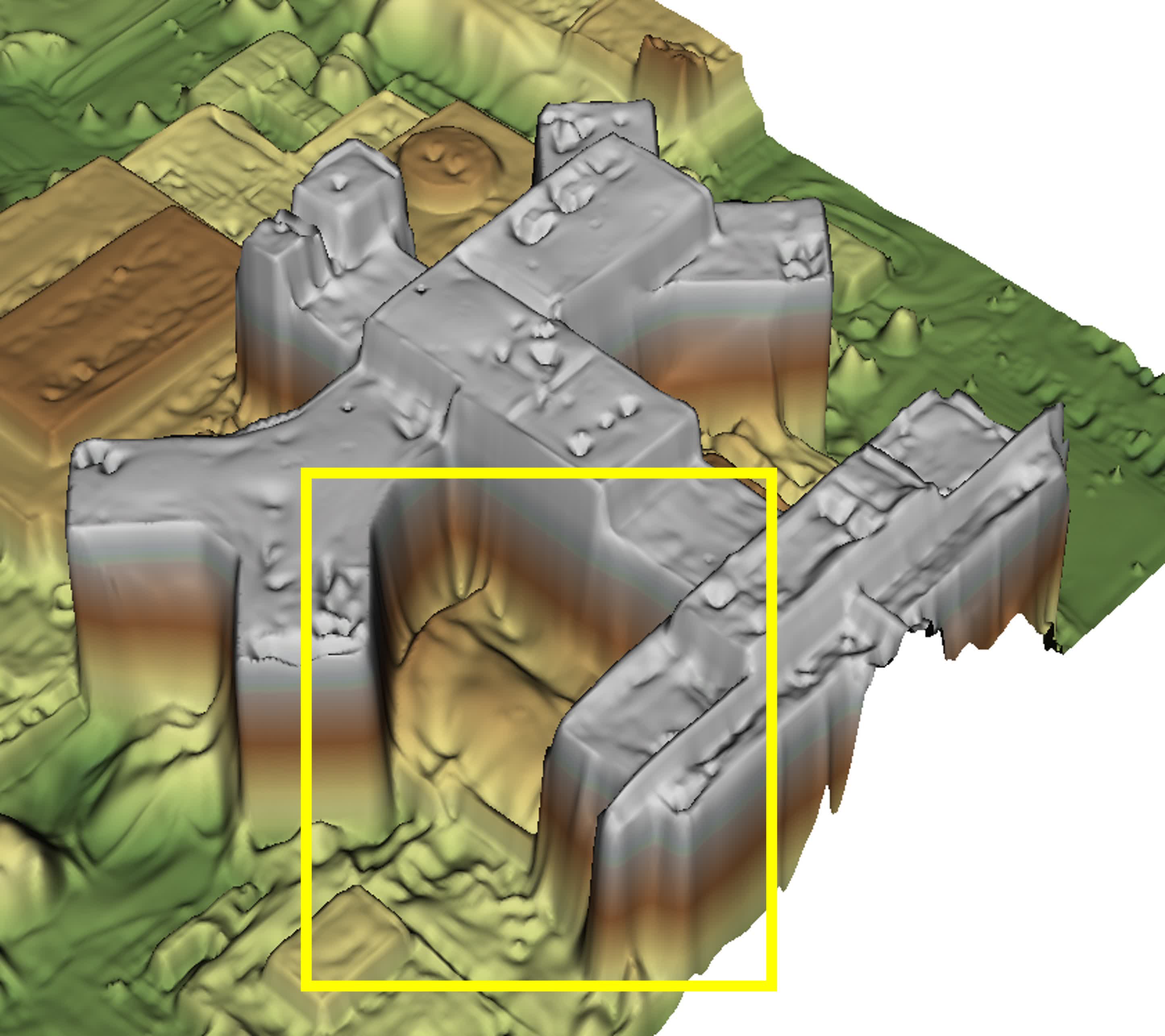}\end{subfigure}
    \begin{subfigure}{\colwfour}\myfigcrop{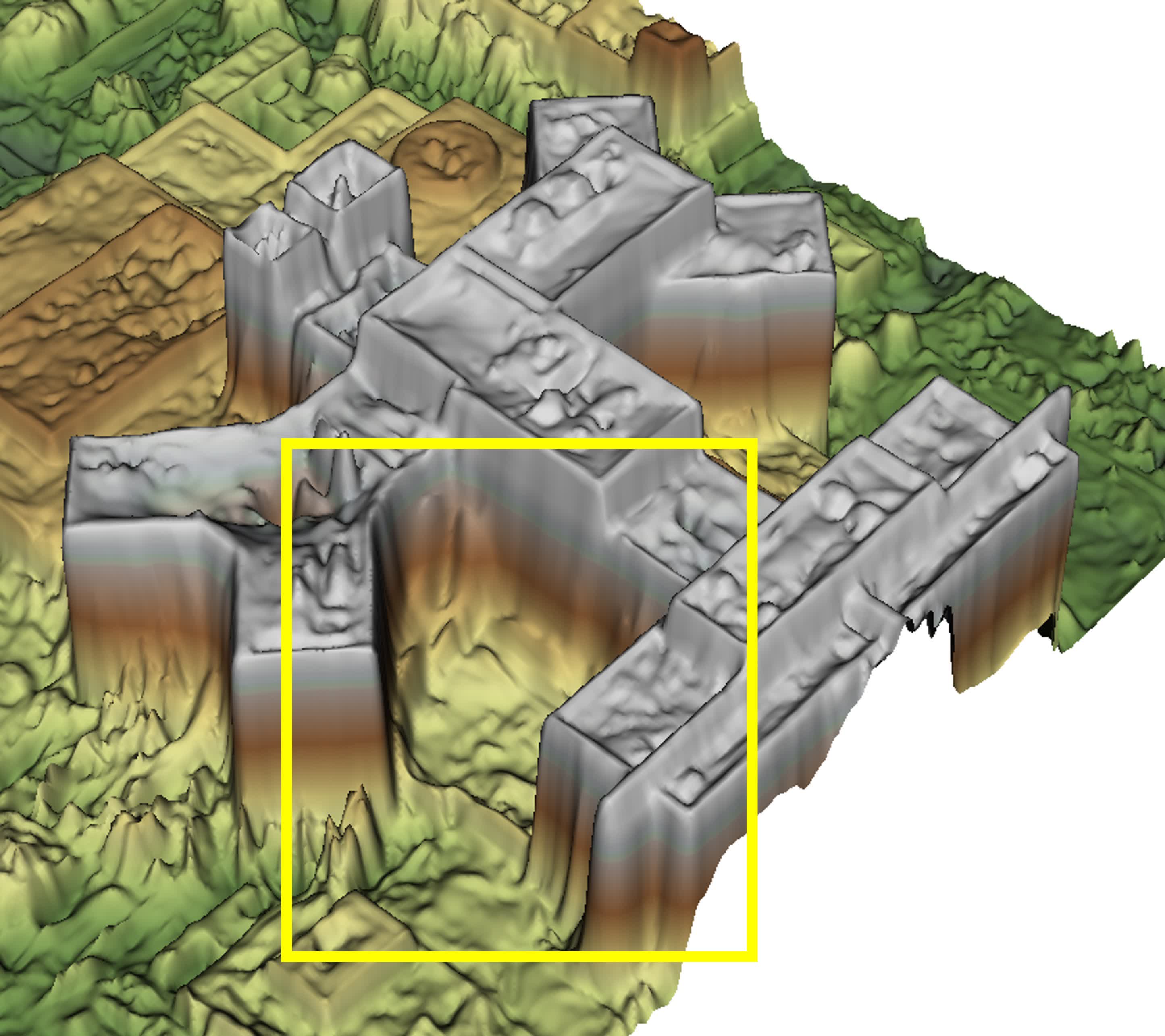}\end{subfigure}
    \begin{subfigure}{\colwfour}\myfigcrop{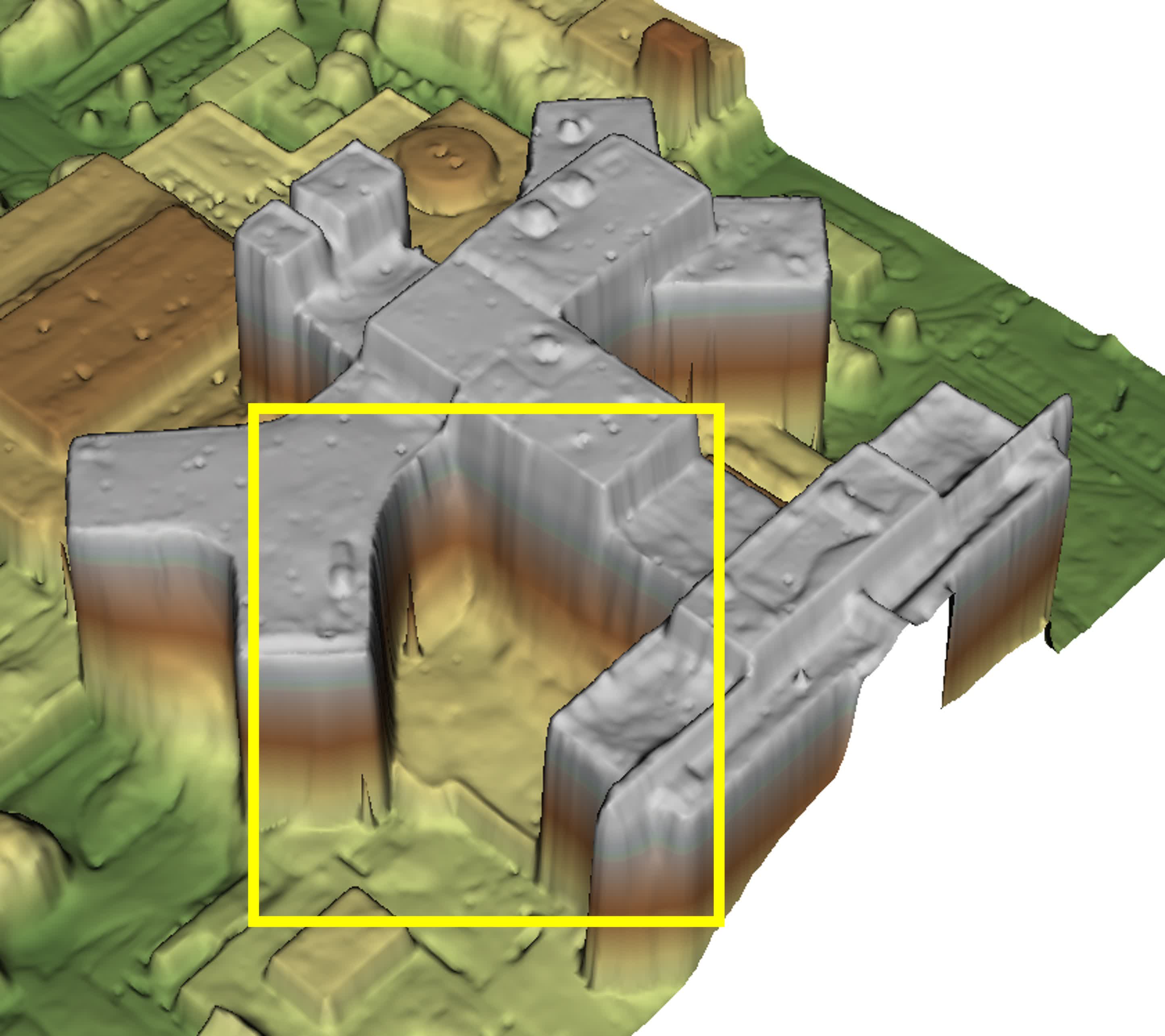}\end{subfigure} \\
    \vspace{0.5mm}

    % --- JAX-214 Section ---
    \rotatebox{90}{\makebox[0.12\linewidth][c]{\footnotesize JAX-214}}
    \begin{subfigure}{\colwfour}\myfigcrop{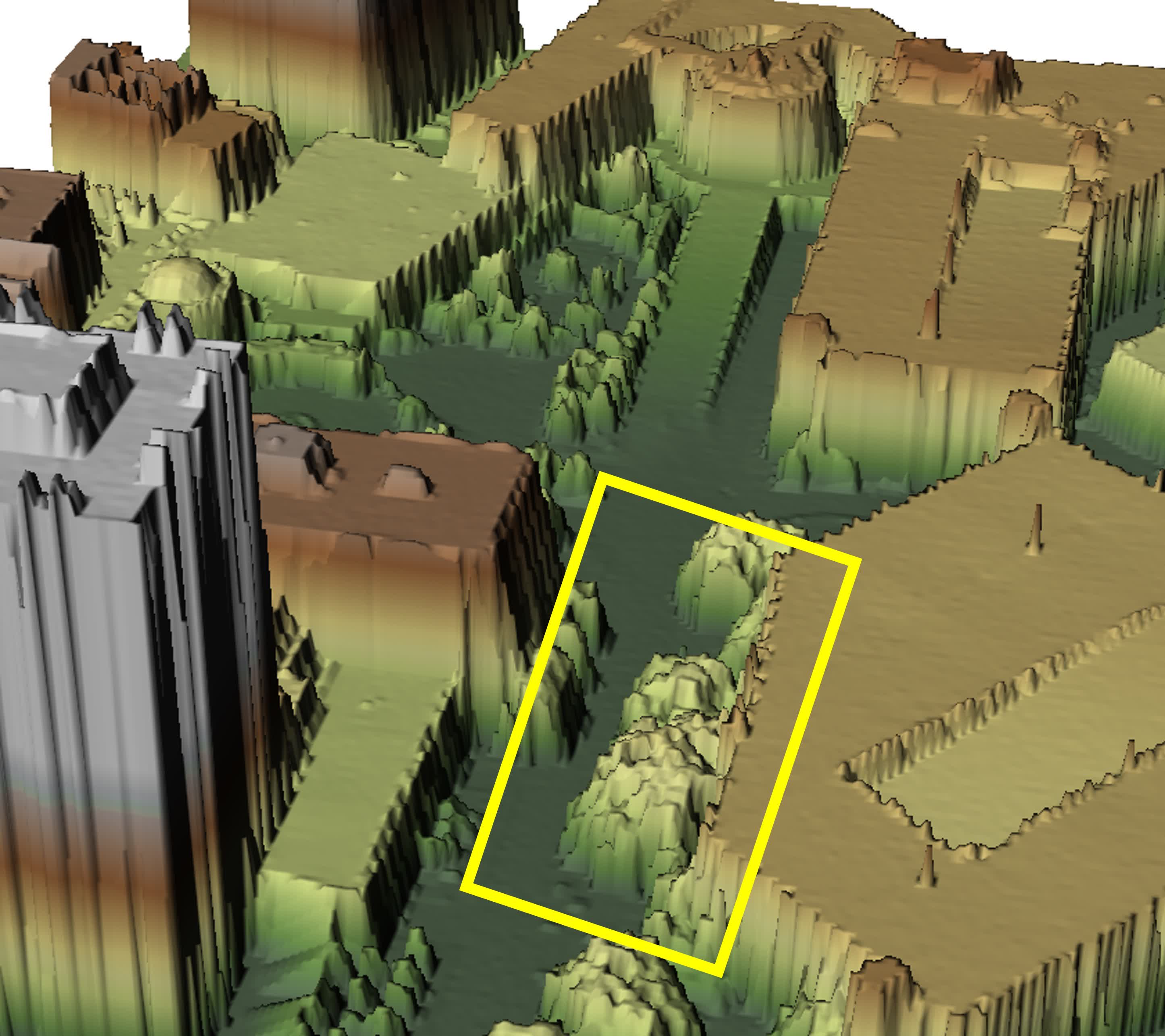}\end{subfigure}
    \begin{subfigure}{\colwfour}\myfigcrop{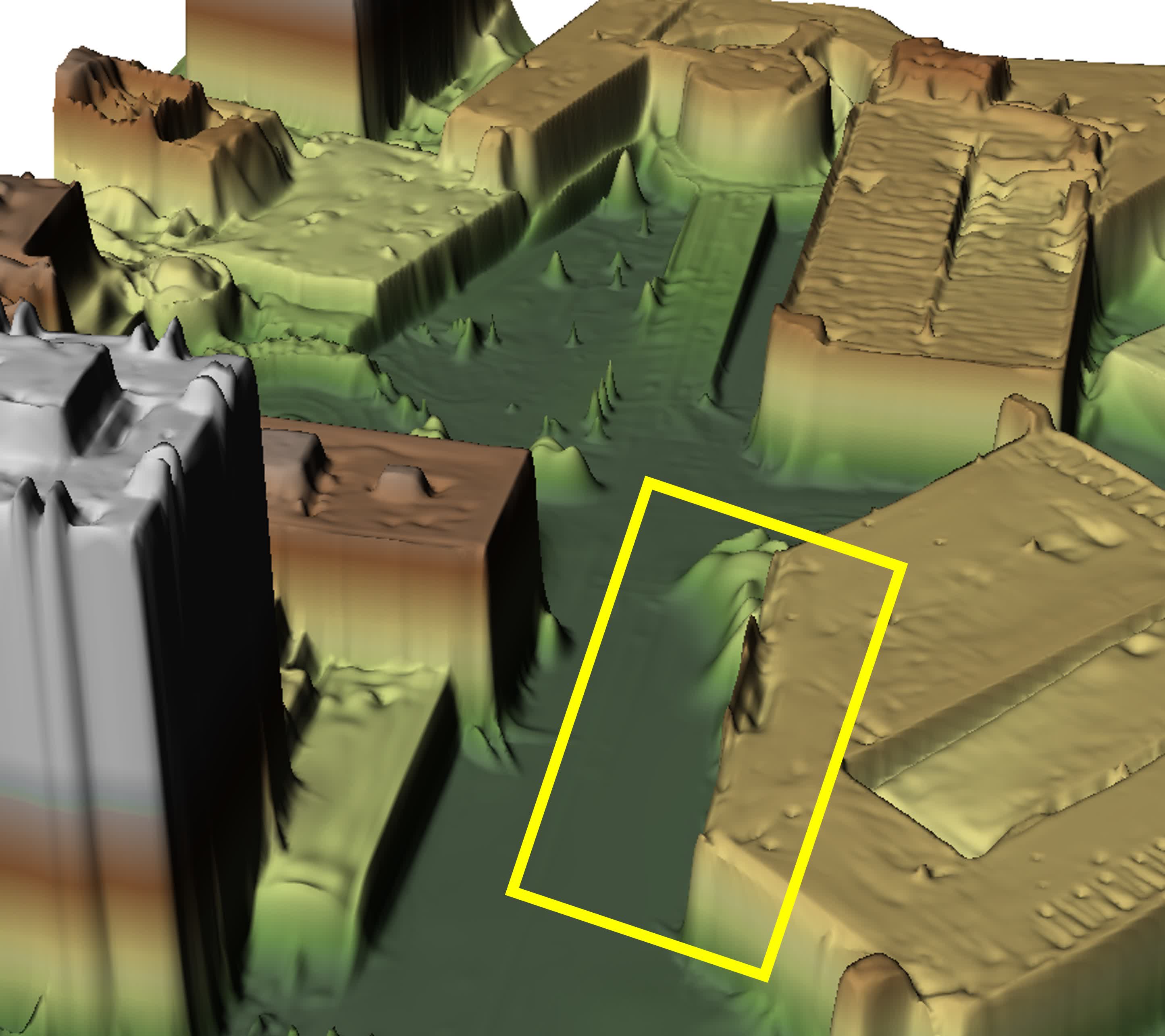}\end{subfigure}
    \begin{subfigure}{\colwfour}\myfigcrop{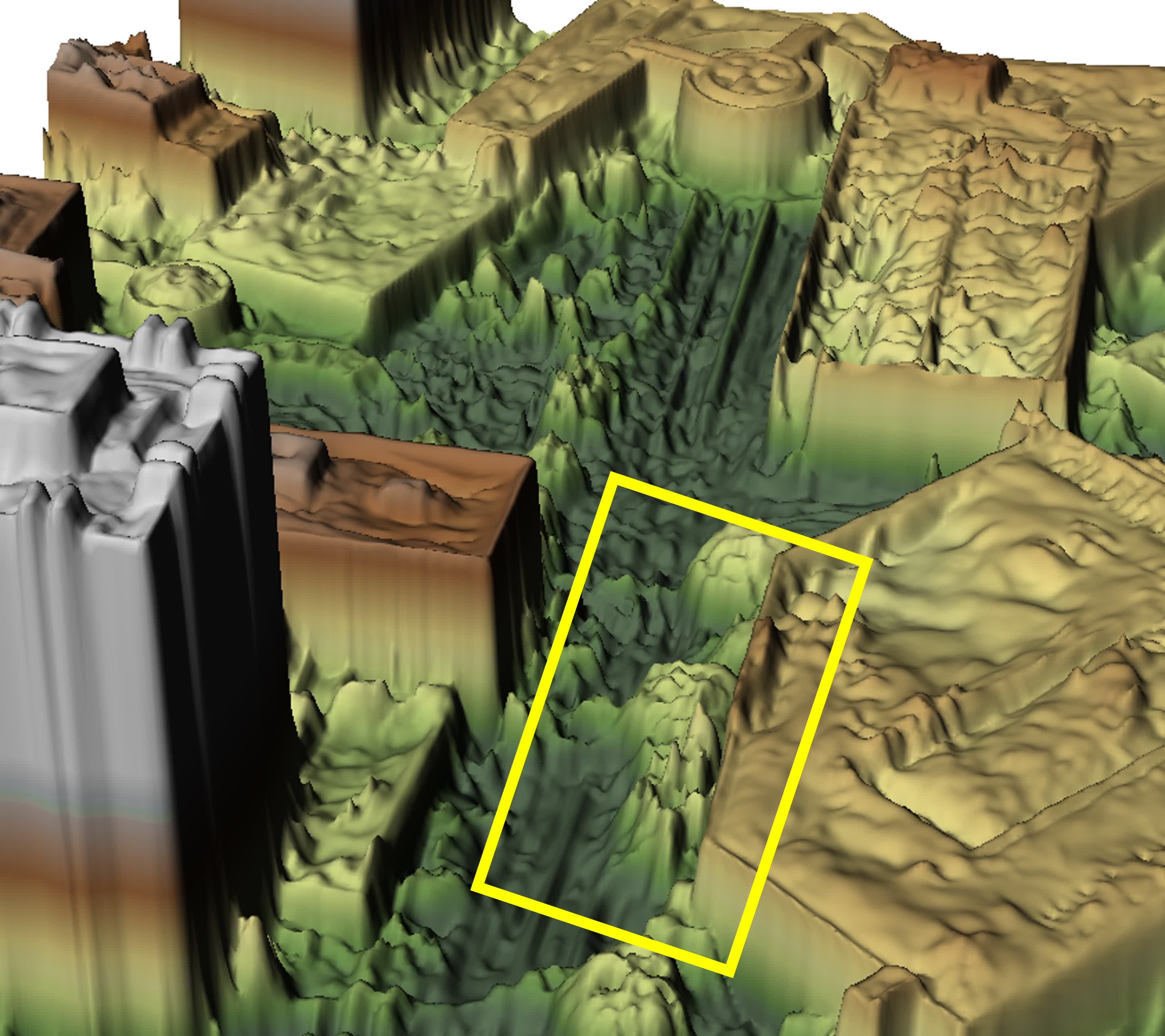}\end{subfigure}
    \begin{subfigure}{\colwfour}\myfigcrop{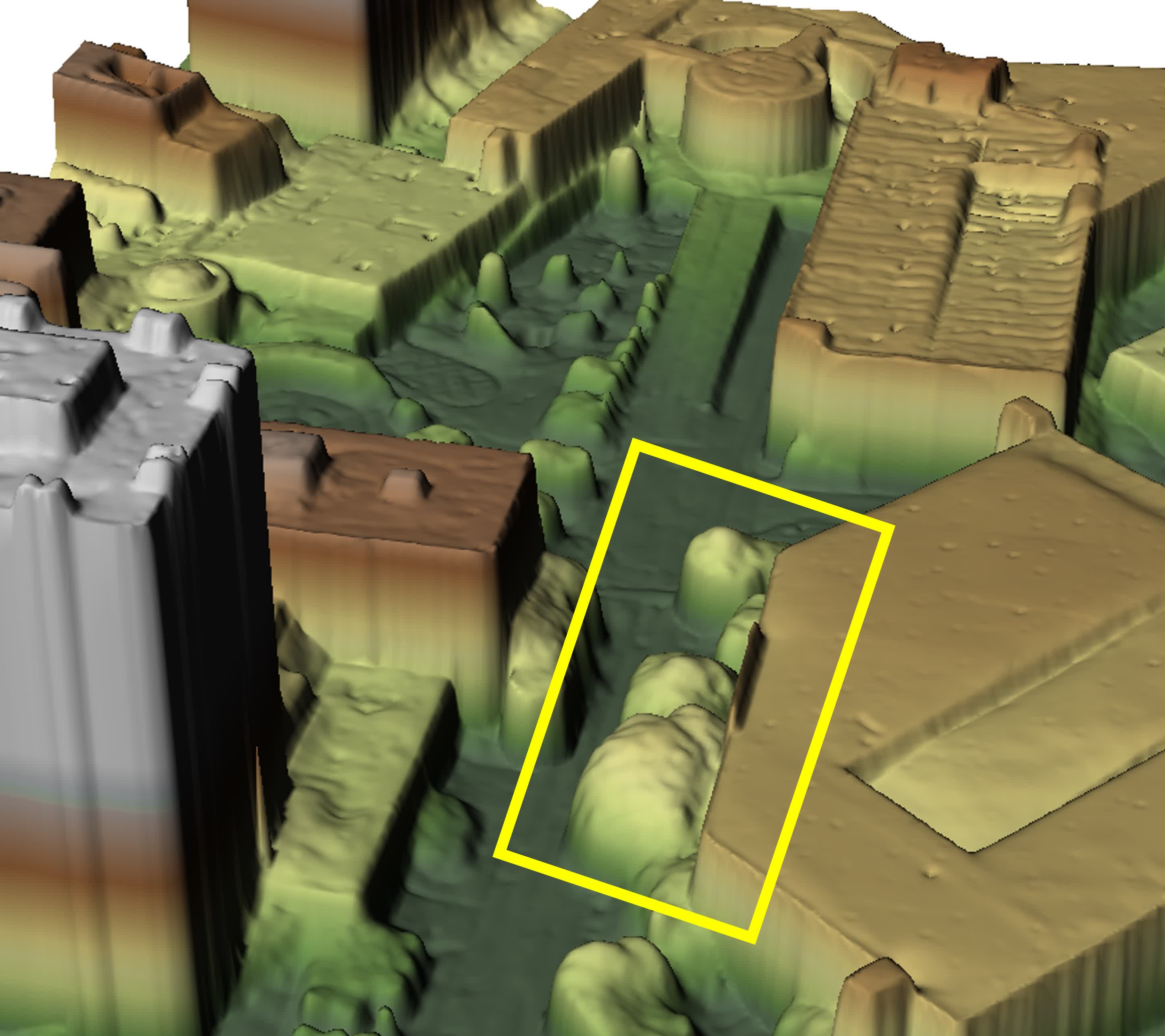}\end{subfigure} \\
    \vspace{0.5mm}

    % --- OMA-212 Section ---
    \rotatebox{90}{\makebox[0.12\linewidth][c]{\footnotesize OMA-212}}
    \begin{subfigure}{\colwfour}\myfigcrop{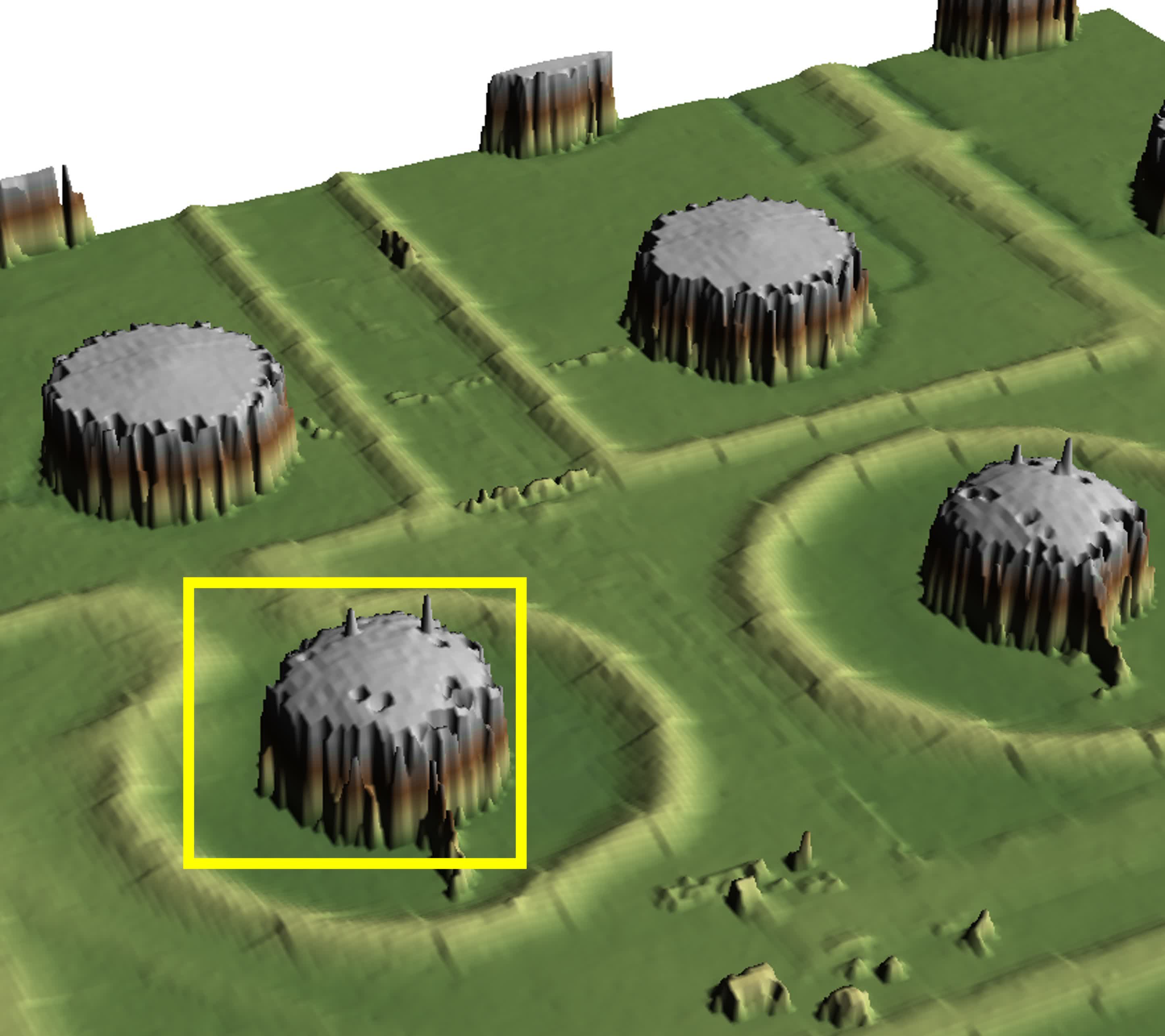}\end{subfigure}
    \begin{subfigure}{\colwfour}\myfigcrop{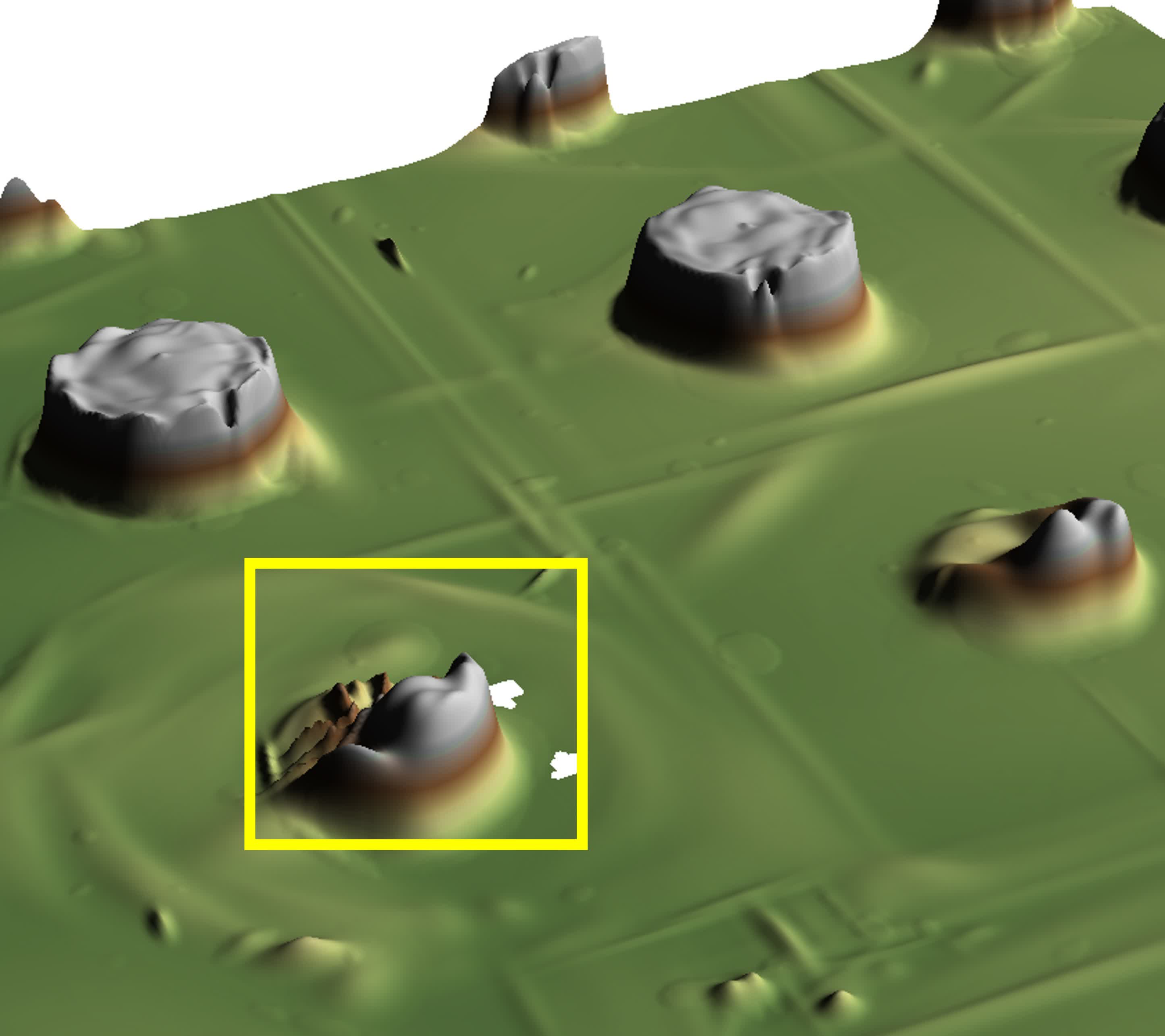}\end{subfigure}
    \begin{subfigure}{\colwfour}\myfigcrop{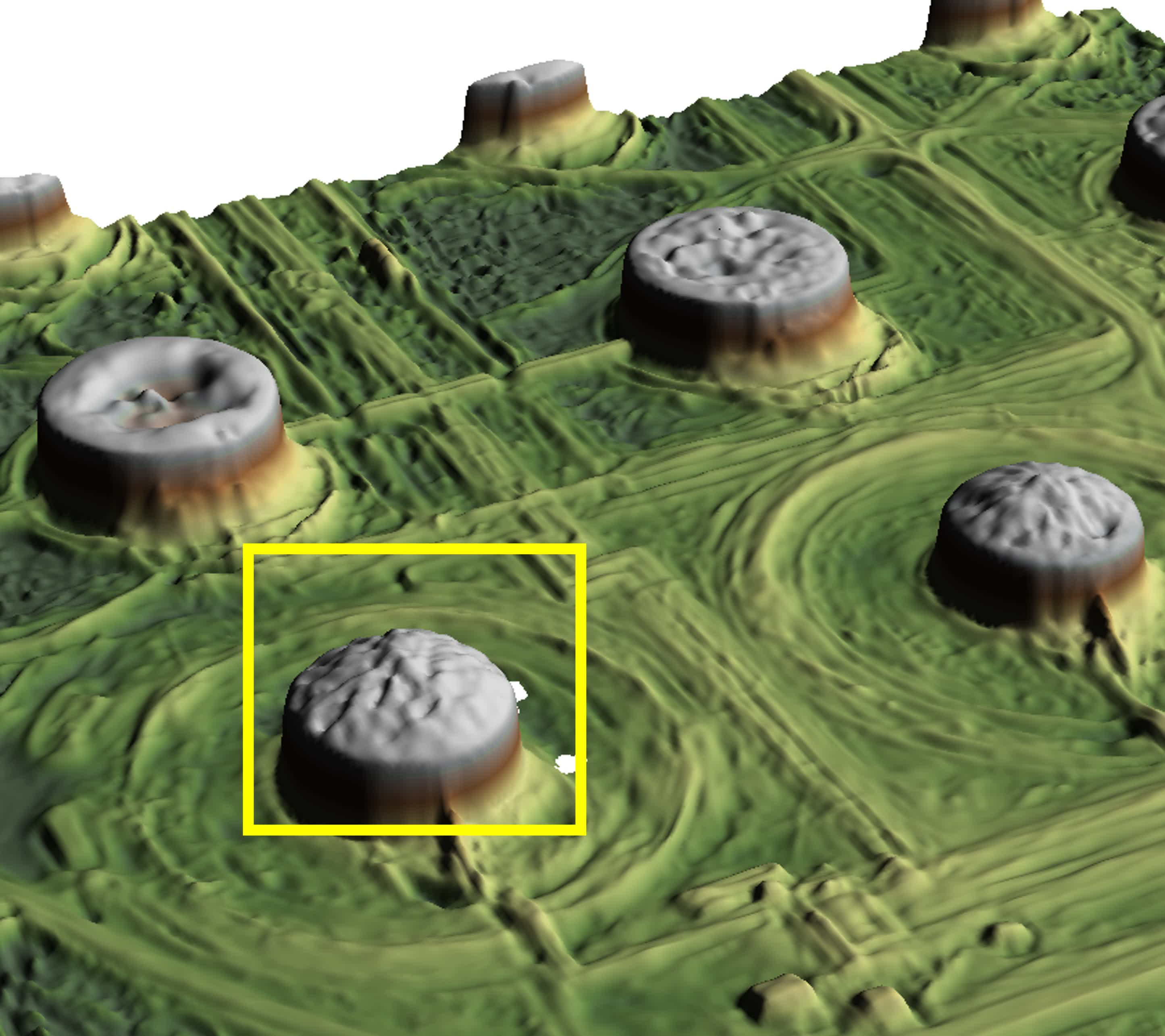}\end{subfigure}
    \begin{subfigure}{\colwfour}\myfigcrop{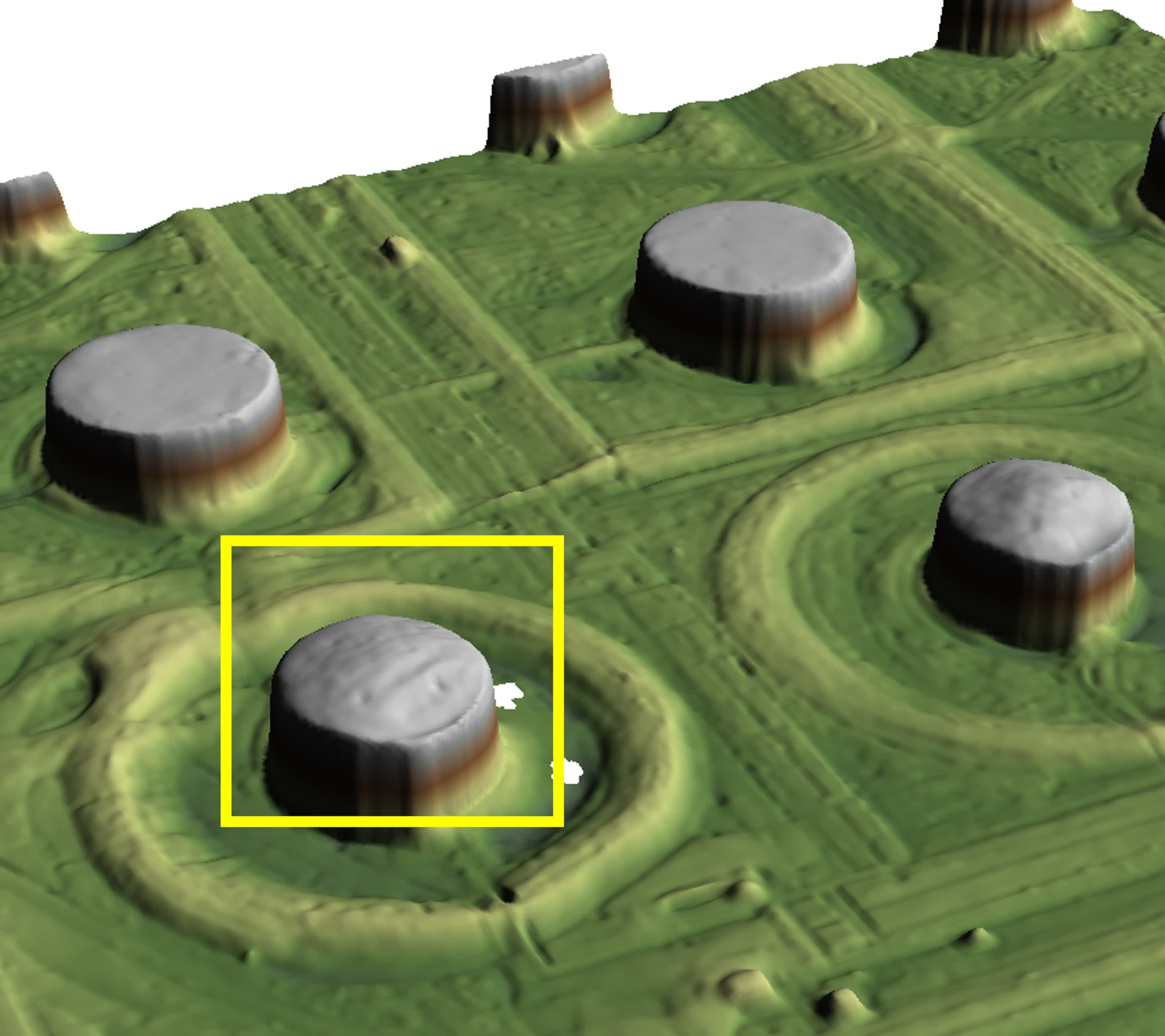}\end{subfigure} \\
    \vspace{0.5mm}

    % --- OMA-315 Section ---
    \rotatebox{90}{\makebox[0.12\linewidth][c]{\footnotesize OMA-315}}
    \begin{subfigure}{\colwfour}\myfigcrop{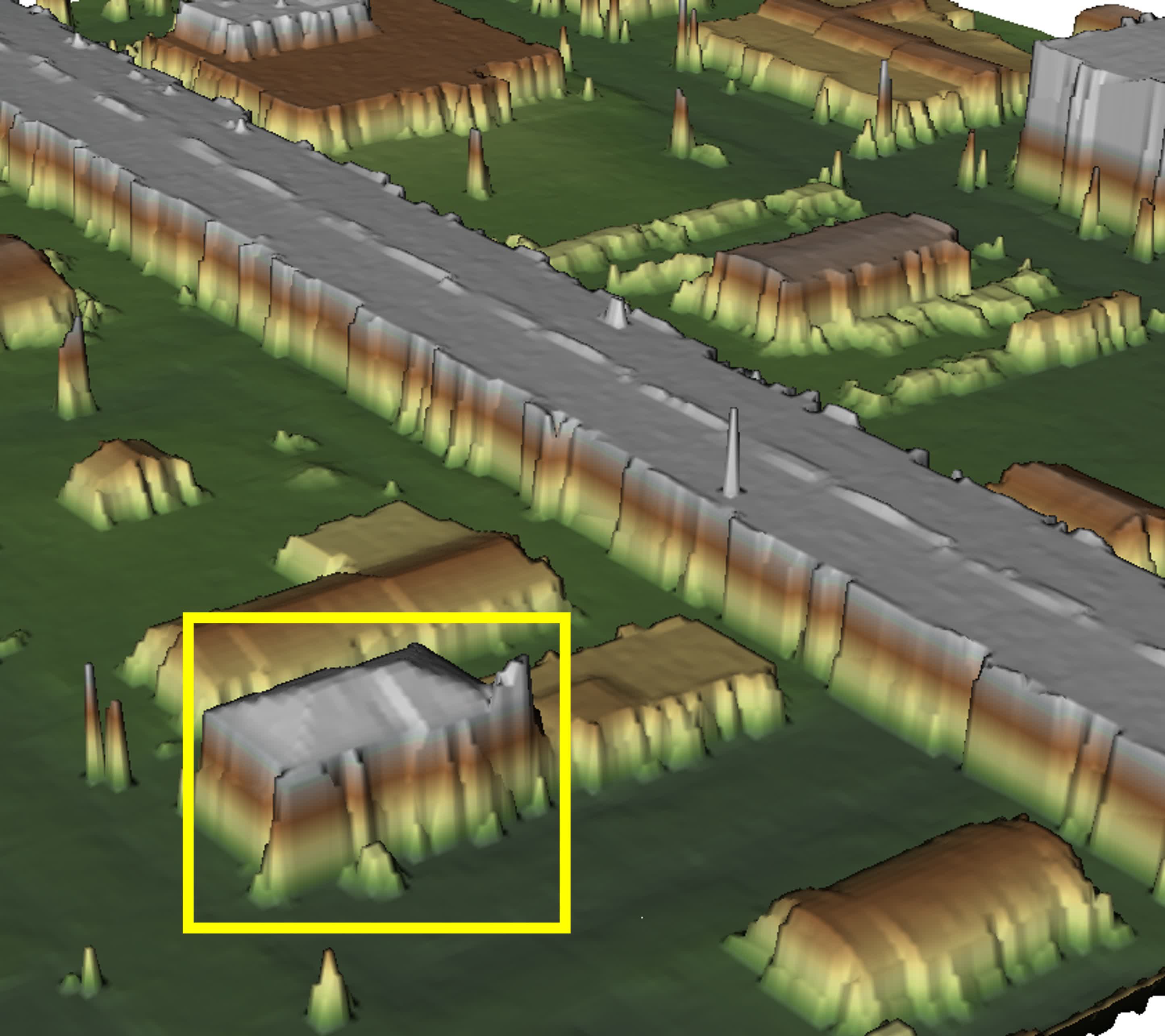}\end{subfigure}
    \begin{subfigure}{\colwfour}\myfigcrop{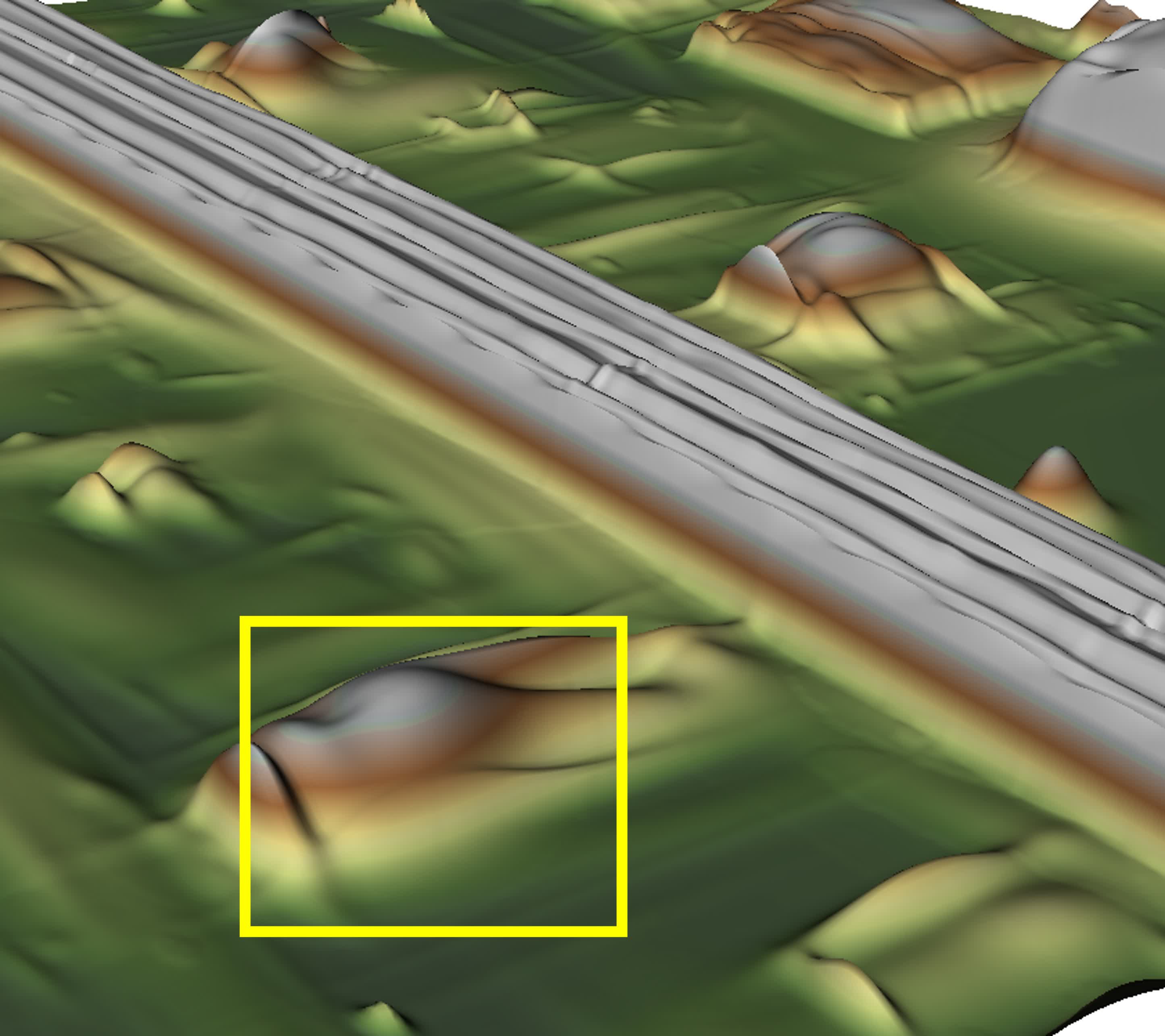}\end{subfigure}
    \begin{subfigure}{\colwfour}\myfigcrop{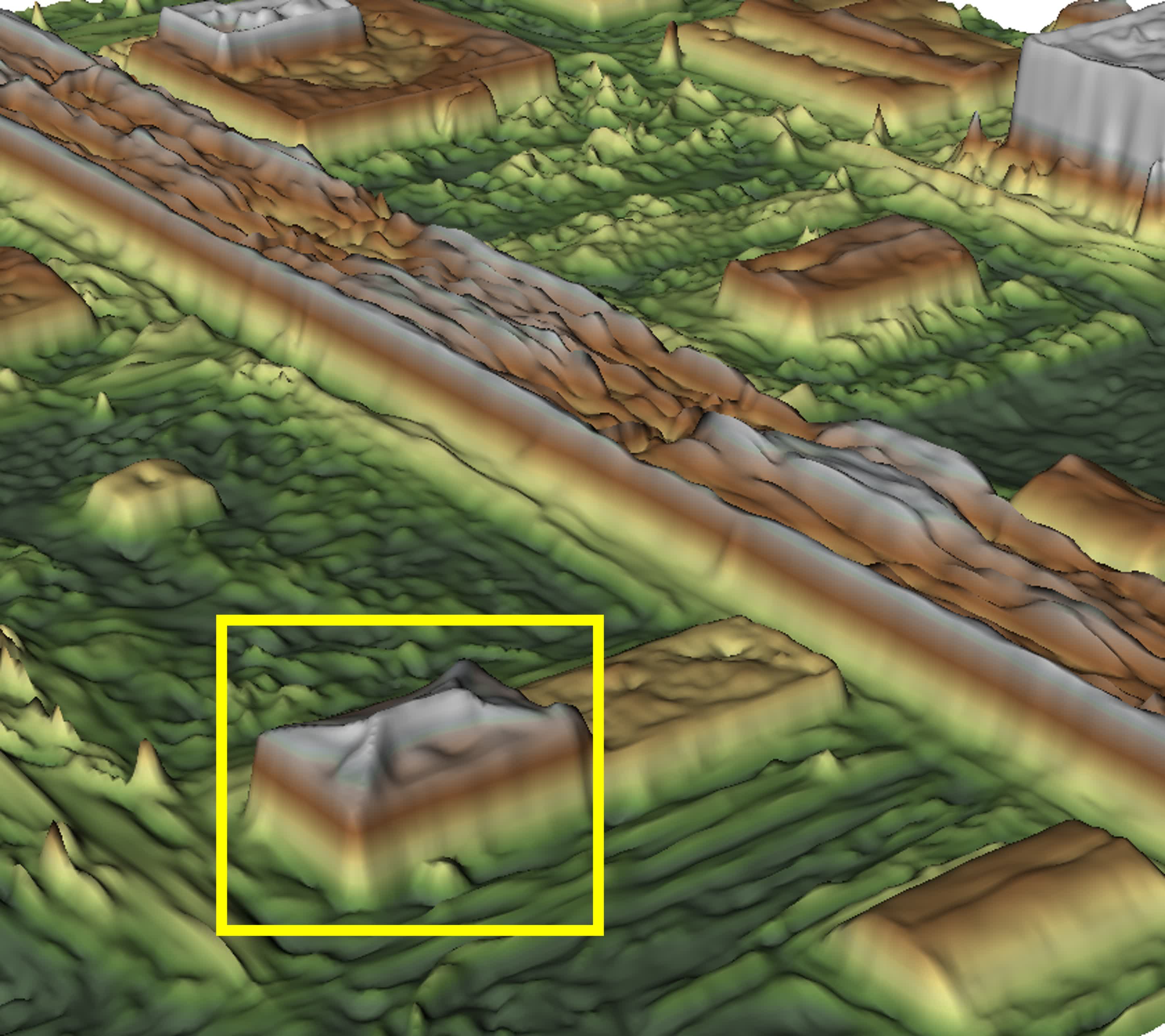}\end{subfigure}
    \begin{subfigure}{\colwfour}\myfigcrop{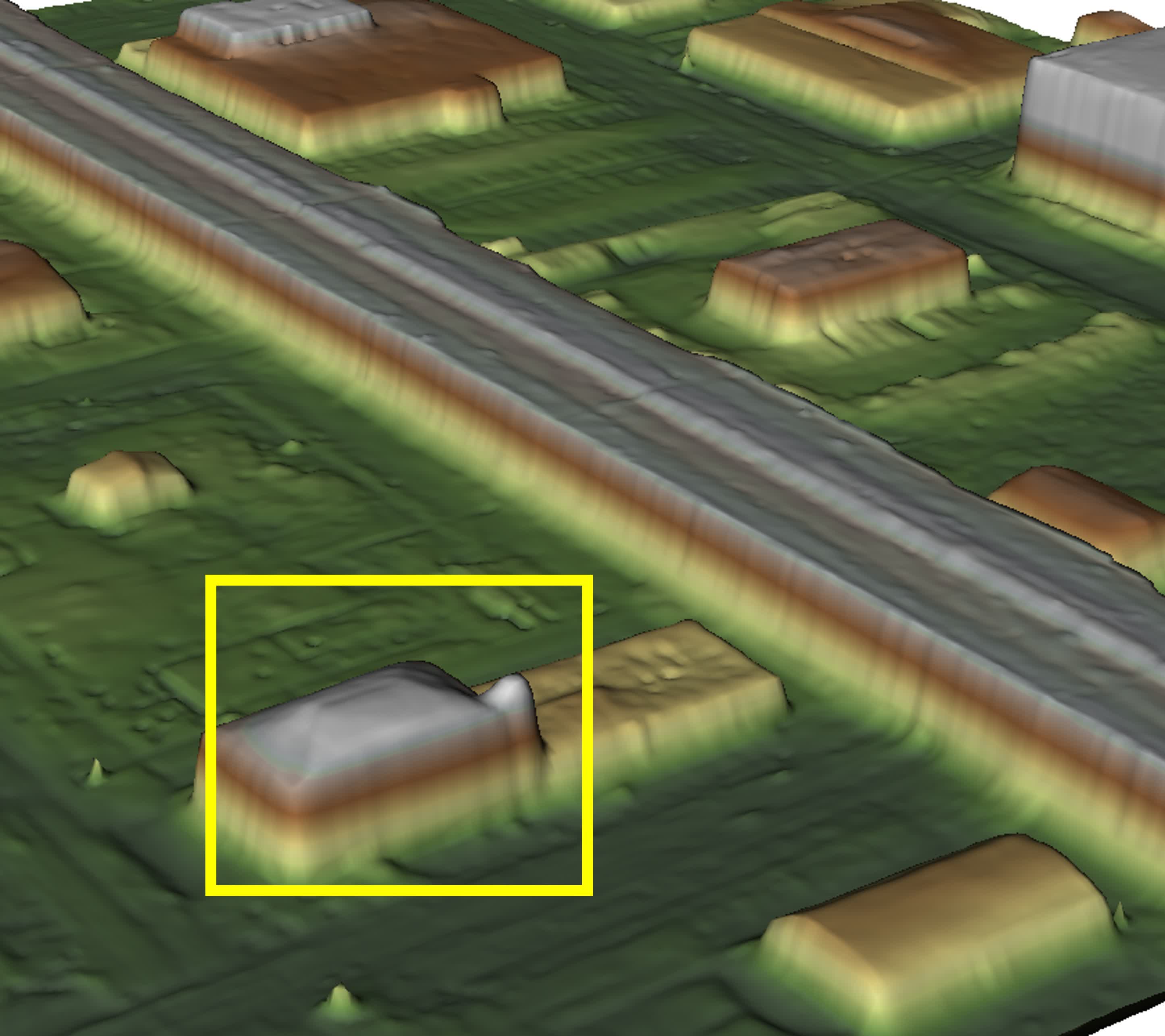}\end{subfigure} \\
    \vspace{1mm} 

    \caption{Comparative qualitative analysis of reconstructed geometry against LiDAR ground truth. Compared to existing GS-based baselines, our framework reconstructs clean surfaces with well-defined boundaries, including specular high-reflectance structures on the OMA sites.}
    \label{fig:geometry_ground}
\end{figure*}

The geometric accuracy of the proposed method is qualitatively and quantitatively evaluated with the ground truth LiDAR data, as illustrated in \Cref{fig:comprehensive_vis_comparison} and \Cref{fig:geometry_ground}. Across various pipelines for comparison, our approach consistently generates the most sharply defined and detailed DSMs.

%\RQ{how do you know quantitatively that a surface is smooth, do you refer to qualitatively?, also people understand EOGS produced GS model, have you introduced the standardized way used to derive the DSM? you need to mention that as the begining of the experimental section when you introduce how you evaluate different methods} mentioned in experimental settings

A closer qualitative inspection (\Cref{fig:geometry_ground}) reveals that EOGS tends to produce overly smoothed reconstructions, often omitting fine scene components such as vegetation and small structures, whereas Skyfall-GS captures sharper detail but struggles to maintain even, well-defined surfaces. Our method balances both characteristics, recovering structural detail while preserving surface solidity. At the JAX-214 site, for instance, EOGS and Sat-NGP fail to reconstruct vegetation entirely, and although Skyfall-GS recovers these features, it introduces transparent roof surfaces with inaccurate altitude values. In contrast, our method reconstructs both vegetation and building geometry while maintaining consistent surface solidity throughout.

%\RQ{you need to indicate where it is,  it is unclear which figure to look at to observe the effects you mentioned}. 

These clean, solid surfaces are further evidenced across both the OMA and IARPA sites (\Cref{fig:comprehensive_vis_comparison}). On the IARPA sites, EOGS produces cleaner DSMs but fails to recover certain scene components, whereas Skyfall-GS captures more detail but struggles to maintain solid surfaces. Our method addresses both limitations simultaneously, reconstructing detailed yet structurally coherent surfaces. A similar pattern emerges on the OMA-212 site, which contains high-reflectance structures that pose particular challenges for reconstruction. EOGS and Sat-NGP fail to recover these structures entirely, and while Skyfall-GS partially recovers their presence, our results remain geometrically cleaner and more consistent with the underlying surface geometry. The IARPA-003 site, however, reveals a limitation of our method: it fails to reconstruct the roller coaster structure that Skyfall-GS recovers on the right side of the DSM. We attribute this to our method's emphasis on global geometric consistency and surface smoothness, which comes at the cost of missing thin, isolated structures.
%\RQ{evidence on "what matter" across the two datasets}

%\RQ{Is this dataset opened? also have you mentioned that howm many sites in your experiment, you can take one site as an example, but the reader may also wonder if this is generalizable conclusion. }

The quantitative evaluation also supports the qualitative observation on \Cref{fig:comprehensive_vis_comparison} and \Cref{fig:geometry_ground} that our method reconstructs cleaner and solid surfaces. As shown in \Cref{tab:mae_summary} our method achieves an average $\mathrm{MAE}_{reg}$ of  1.23m on the JAX sites, which corresponds to an 18.0\% improvement over the second best method, EOGS (1.50m). While EOGS shows competitive performance, its accuracy significantly degrades in sites with more complex urban structures. In contrast, our method maintains a robust error margin below 1.4m in most cases, effectively capturing fine-grained building geometries.

\begin{table}[htbp]
\centering
\caption{
Summary of geometric accuracy $\mathrm{MAE}_{reg}$ across various satellite datasets. SatSplatDiff achieves best geometric accuracy on all sites. Comparison on only buildings demonstrates the ability of our pipeline to form solid gaussian surfaces.}
\label{tab:mae_summary}
\small
\setlength{\tabcolsep}{4pt}
\resizebox{\linewidth}{!}{
\begin{tabular}{lccccccccc}
\toprule
\multirow{3}{*}{\textbf{Method}} & \multicolumn{4}{c}{\textbf{Full Scene}} & & \multicolumn{4}{c}{\textbf{Buildings}} \\
\cmidrule(lr){2-5} \cmidrule(lr){7-10}
& \textbf{JAX} & \textbf{OMA} & \textbf{IARPA} & \textbf{Mean} & & \textbf{JAX} & \textbf{OMA} & \textbf{IARPA} & \textbf{Mean} \\
\midrule
ASP         & 2.09 & 1.01 & 2.39 & 1.83 & & 1.88 & 2.12 & 1.90 & 1.97 \\
SAT-NGP     & 3.01 & 1.82 & 2.16 & 2.33 & & 3.40 & 3.99 & 2.70 & 3.36 \\
EOGS        & 1.50 & 1.47 & 1.86 & 1.61 & & 1.25 & 2.64 & 1.37 & 1.75 \\
Skyfall-GS  & 1.89 & 1.40 & 2.60 & 1.96 & & 1.58 & 1.81 & 2.05 & 1.81 \\
GU-GS       & 1.73 & 2.31 & 1.83 & 1.96 &  & 1.69 & 2.87 & 1.35 & 1.97 \\
\midrule
\textbf{Ours} & \textbf{1.23} & \textbf{0.65} & \textbf{1.82} & \textbf{1.23} & & \textbf{0.91} & \textbf{0.81} & \textbf{1.32} & \textbf{1.01} \\
\bottomrule
\end{tabular}
}
\end{table}

The performance gap remains pronounced when including the IARPA and OMA sites. Overall, our method achieves a mean $\mathrm{MAE}_{reg}$ of 1.23m, a significant improvement over EOGS (1.61m) and other NeRF-based baselines. Notably, our approach achieves sub-meter accuracy on the Omaha sites, reducing the error to less than half that of the next-best method. When restricted to building regions, our method achieves a mean $\mathrm{MAE}_{reg}$ below one meter on both the JAX and OMA tiles. These results indicate that our method reconstructs building geometry with consistent sub-meter accuracy across both full-scene and building-specific evaluations.

\subsection{Ablations}

\subsubsection{Effect of multi-scale geometric refinement}

An ablation study was conducted to evaluate the impact of multi-scale geometric refinement (\Cref{sec:method:multiscale}) on both visual fidelity and geometric accuracy (\Cref{tab:zoom_ablation}). Multi-scale refinement introduces additional supervision from rendered images with different spatial scales, allowing Gaussians to better recover structures that are insufficiently constrained in the original satellite observations. Given that this refinement process involves an additional backpropagation step per iteration, the computational trade-off was also analyzed.

%An ablation study was conducted to evaluate the impact of multi-scale geometric refinement (\Cref{sec:method:multiscale}) on both visual fidelity and geometric accuracy. Given that this refinement process involves an additional backpropagation step per iteration, the computational trade-off was also analyzed.%\RQ{I did not recall i have seen anything related to multi-scale refinement, it was not a subsection nor i have a deep impression where you have described, when you bring it up in your first sentence, can you specify where you mentioned it?}

\begin{table}[!htbp]
\centering
\caption{Final reconstruction quality after complete optimization with and without multi-scale refinement. Complexity reports the computational cost of the geometric optimization stage.}
\label{tab:zoom_ablation}
\resizebox{\linewidth}{!}{%
\begin{tabular}{lccccc c c}
\toprule
\multirow{2}{*}{\textbf{Configuration}} 
& \multicolumn{2}{c}{\textbf{Distributional}} 
& \multicolumn{3}{c}{\textbf{Pixel-level}} 
& \multicolumn{1}{c}{\textbf{Geometric}} 
& \textbf{Complexity (Geo. Opt.)} \\
\cmidrule(lr){2-3} 
\cmidrule(lr){4-6} 
\cmidrule(lr){7-7} 
\cmidrule(lr){8-8}
& FID-CLIP $\downarrow$ & CMMD $\downarrow$ 
& PSNR $\uparrow$ & CW-SSIM $\uparrow$ & LPIPS $\downarrow$ 
& MAE$_{reg}$ $\downarrow$ / Opacity $\uparrow$ 
& \#Gaussians / Time (min) \\
\midrule
Ours (w/o multi-scale) 
& 19.56 & 1.700 
& 12.24 & 0.410 & \textbf{0.598} 
& 1.27 / 0.97 
& \textbf{2,434,791} / \textbf{42.15} \\
Ours (w/ multi-scale) 
& \textbf{19.50} & \textbf{1.681} 
& \textbf{12.26} & \textbf{0.414} & 0.611 
& \textbf{1.25} / \textbf{0.99} 
& 2,460,800 / 49.39 \\
\bottomrule
\end{tabular}%
}
\end{table}

As reported in \Cref{tab:zoom_ablation}, integrating multi-scale geometric refinement improves distribution metrics (FID-CLIP and CMMD) and pixel-level quality (PSNR and CW-SSIM). From a geometric perspective, the refinement reduces the MAE$_{reg}$ from 1.27m to 1.25m and increases the average opacity from 0.97 to 0.99, suggesting reduced ambiguity in Gaussian coverage. However, the additional multi-scale supervision increases the number of optimized Gaussians from 2,434,791 to 2,460,800 and increases the geometric optimization time from 42 to 49 minutes. These results indicate that multi-scale refinement provides measurable improvements in appearance quality and geometric consistency with a moderate increase in computational cost.

\subsubsection{Efficacy of generative refinement}

\begin{figure*}[t]
\centering
\small

\setlength{\abovecaptionskip}{2pt}
\setlength{\belowcaptionskip}{-4pt}

\def\colw{0.23\linewidth}

\newcommand{\myimg}[1]{%
    \includegraphics[width=\colw]{#1}
}

\renewcommand{\arraystretch}{0.9}
\setlength{\tabcolsep}{2pt}

\begin{tabular}{c@{\hspace{2pt}}cc|cc}

& \multicolumn{2}{c}{\textbf{Rendered Albedo}} 
& \multicolumn{2}{c}{\textbf{DSM (Height Map)}} \\

& \footnotesize w/o Diffusion
& \footnotesize w/ Diffusion
& \footnotesize w/o Diffusion
& \footnotesize w/ Diffusion \\

% JAX-068
\rotatebox{90}{\small JAX-068} &
\myimg{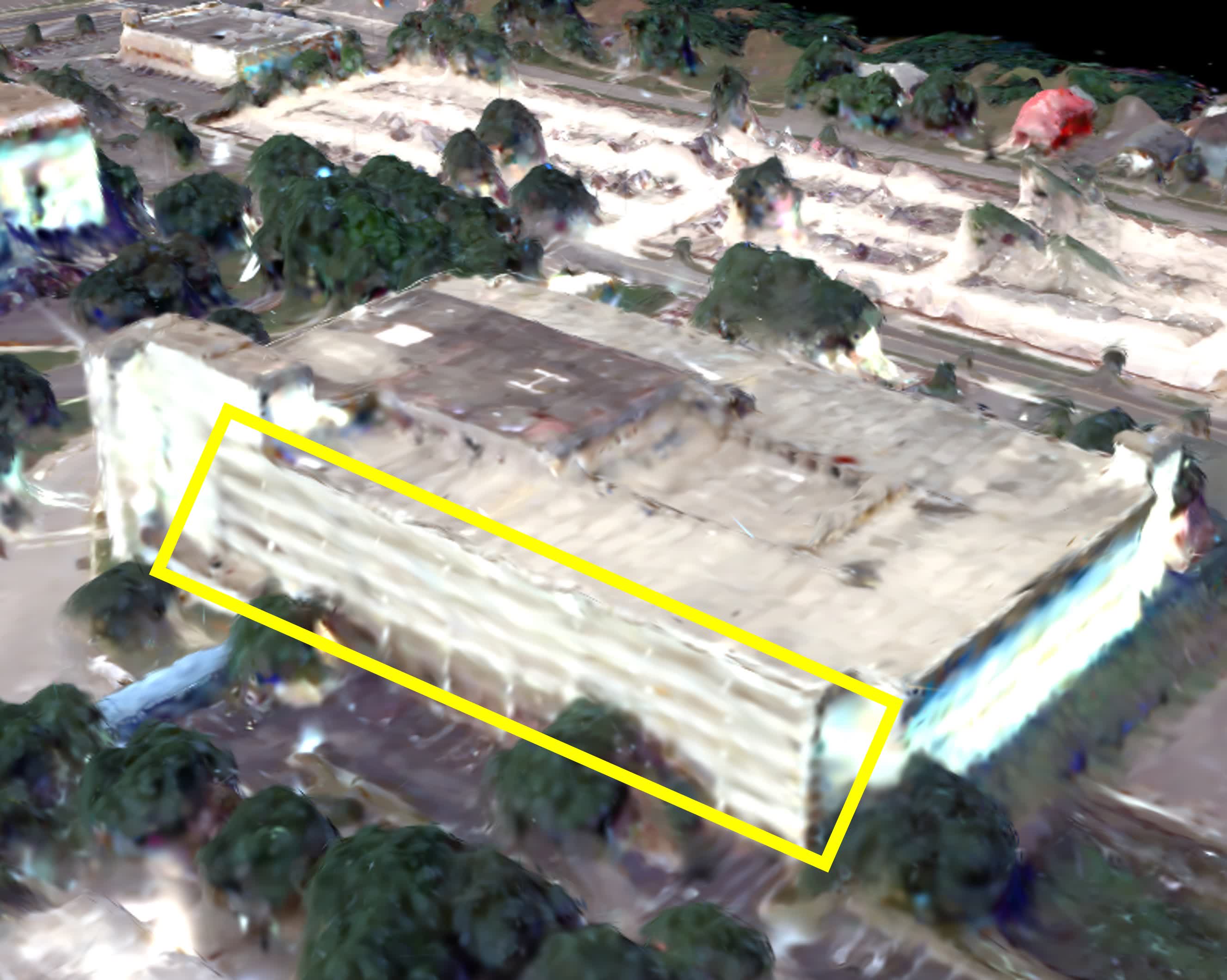} &
\myimg{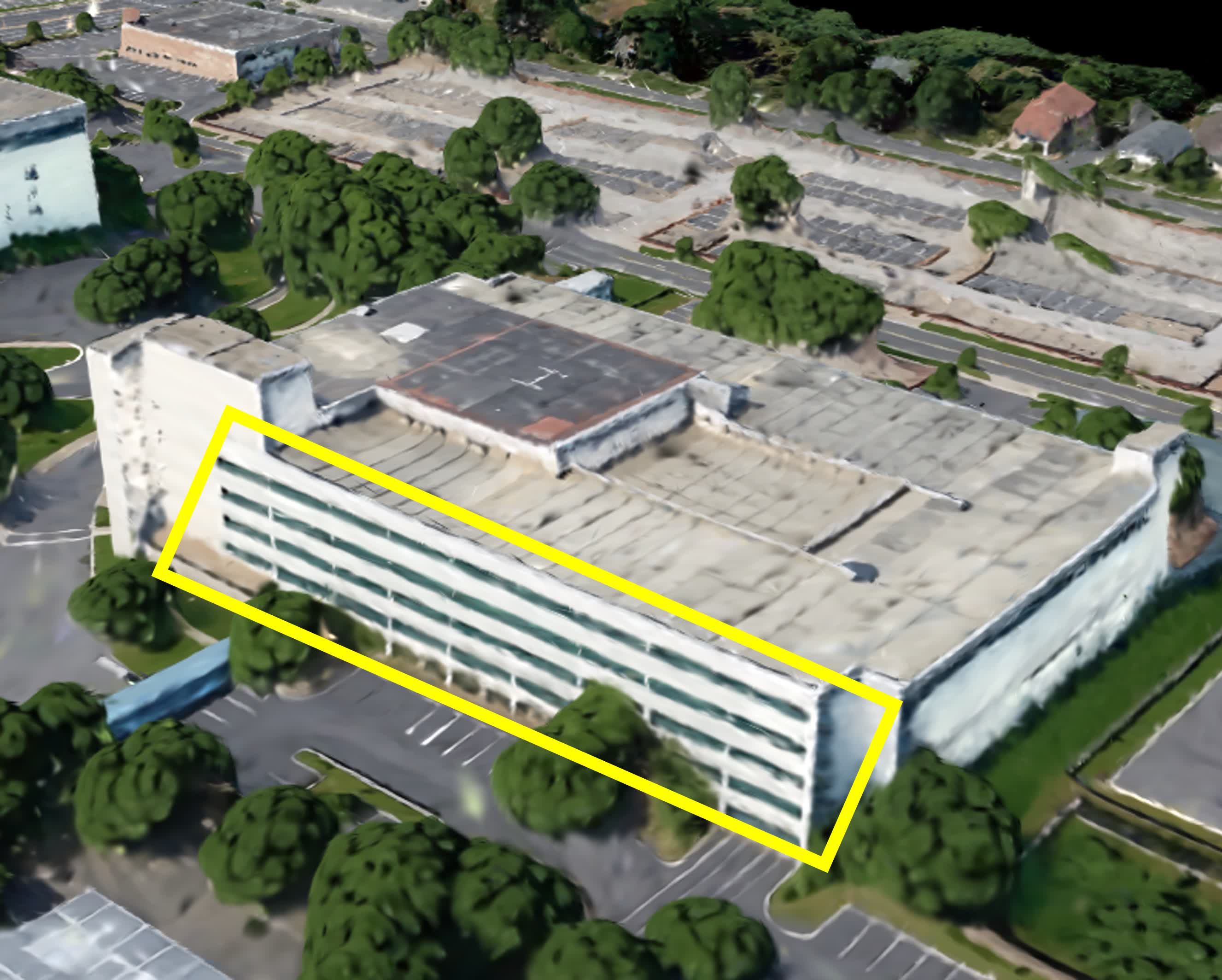} &
\myimg{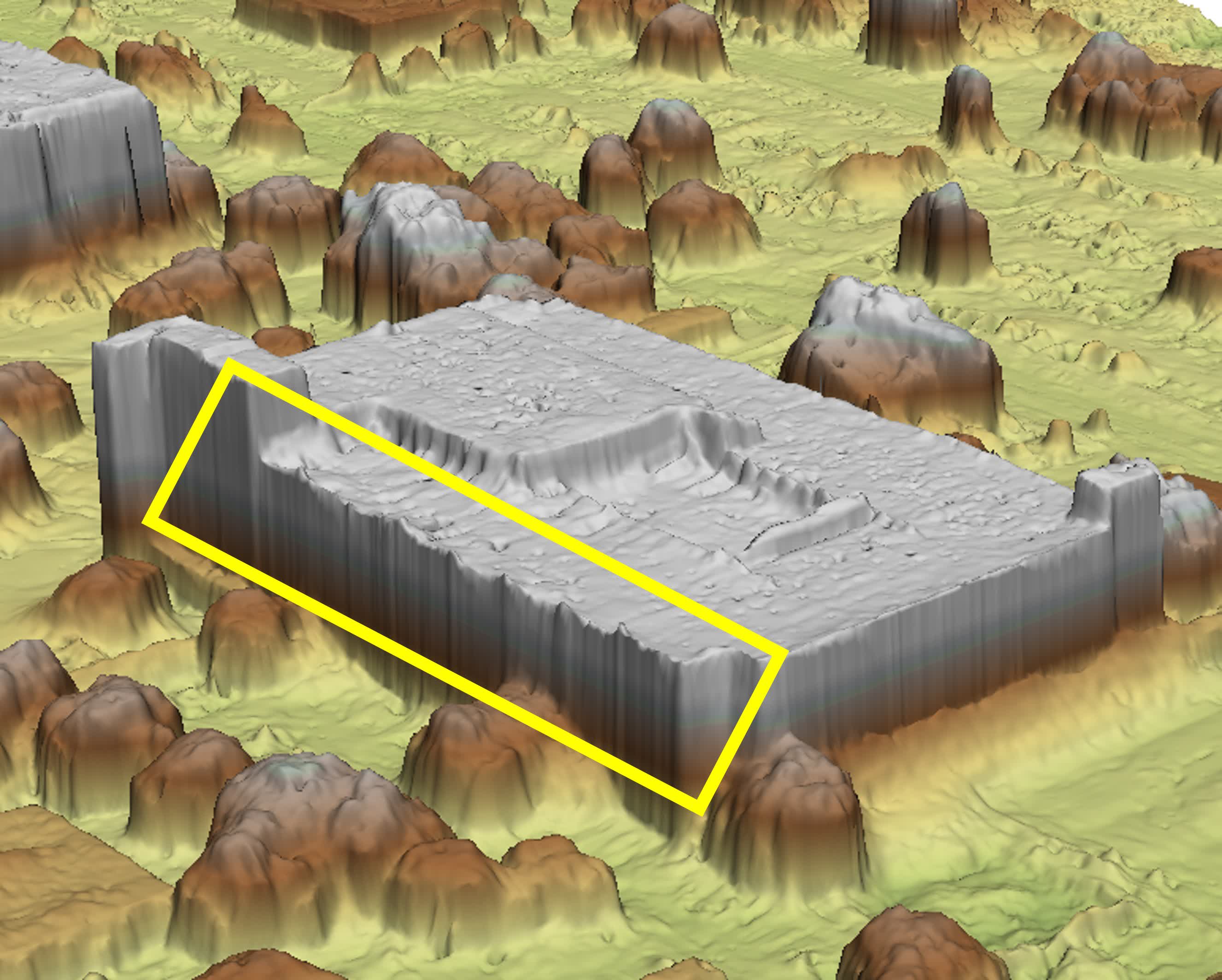} &
\myimg{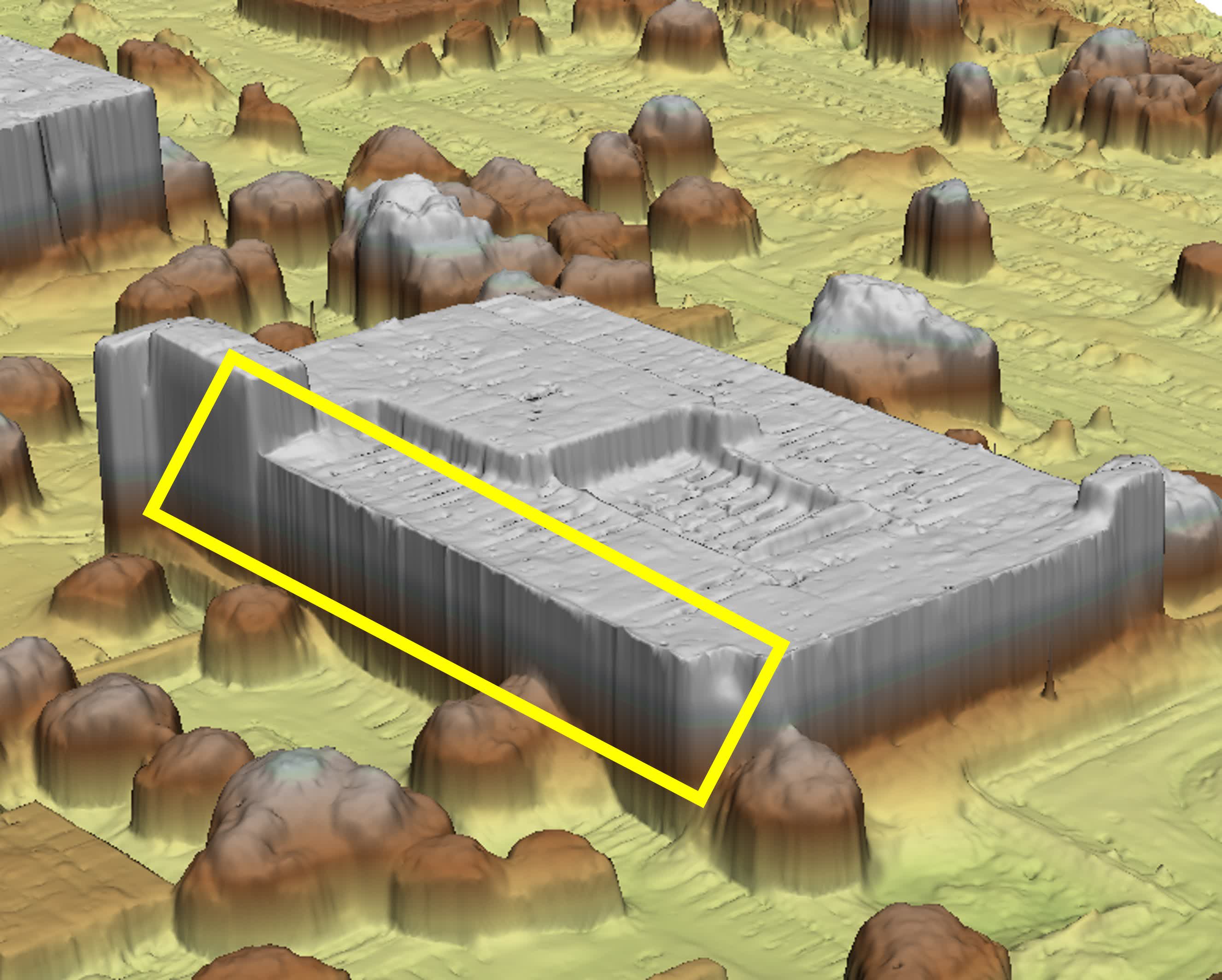} \\

% JAX-169
\rotatebox{90}{\small JAX-168} &
\myimg{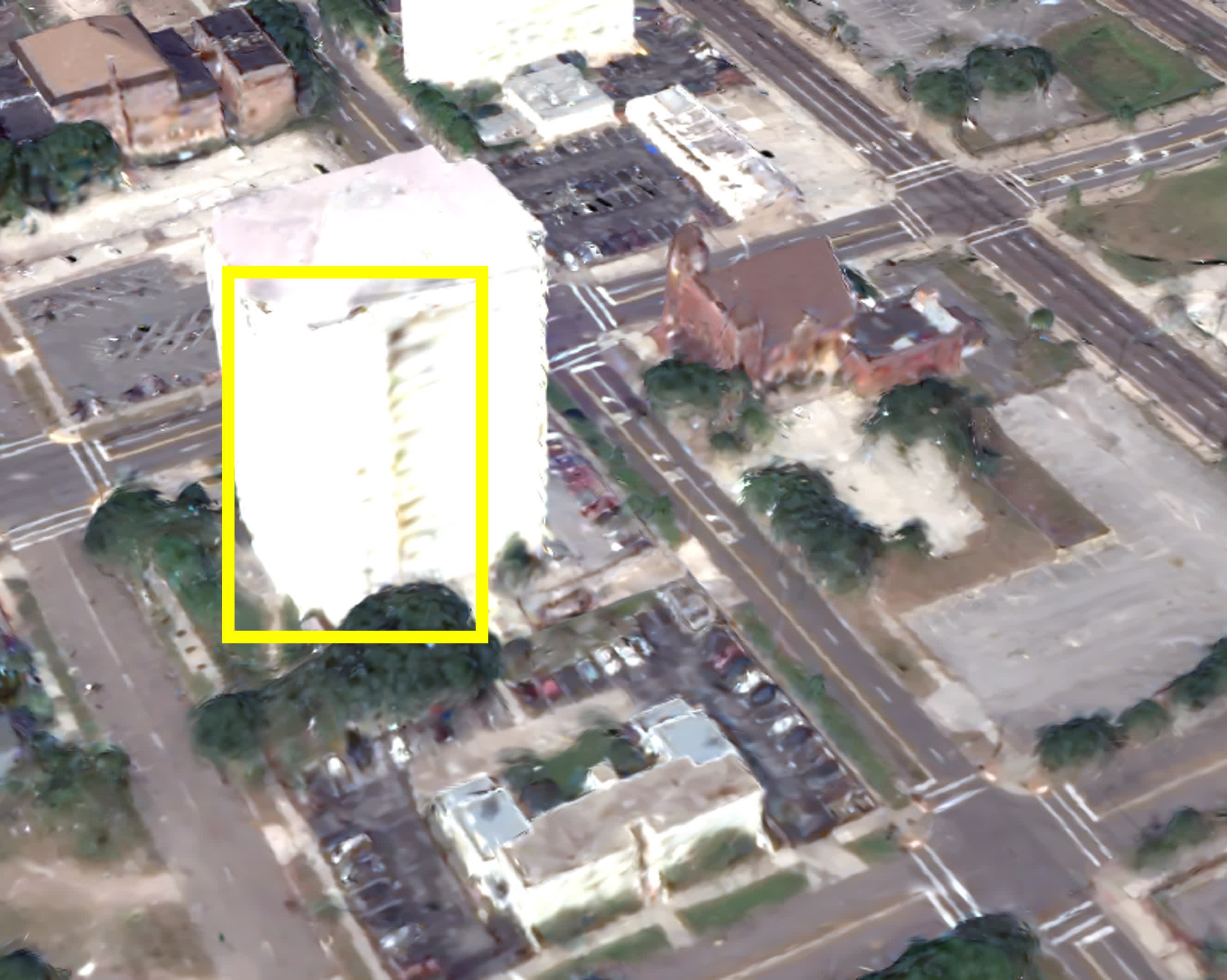} &
\myimg{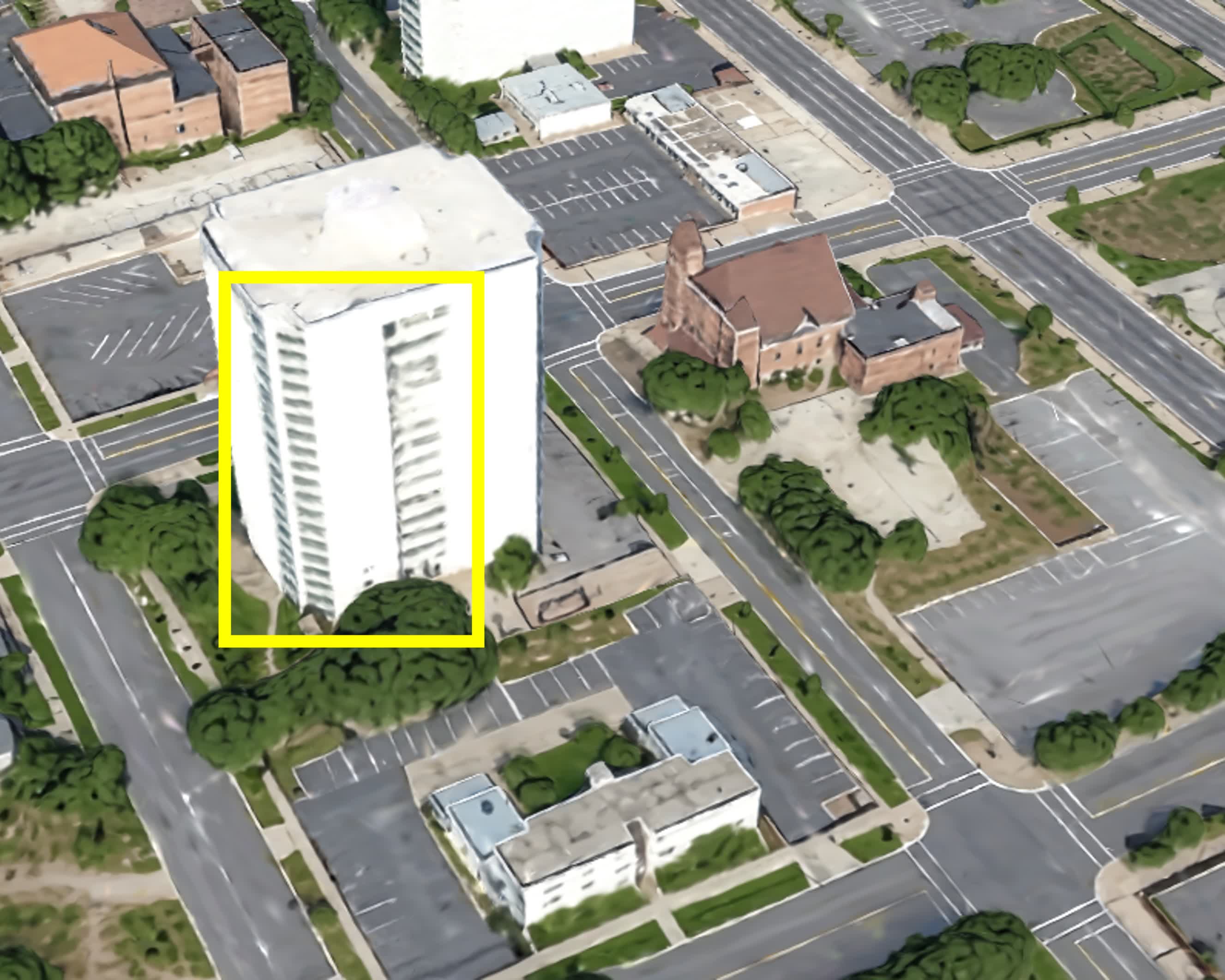} &
\myimg{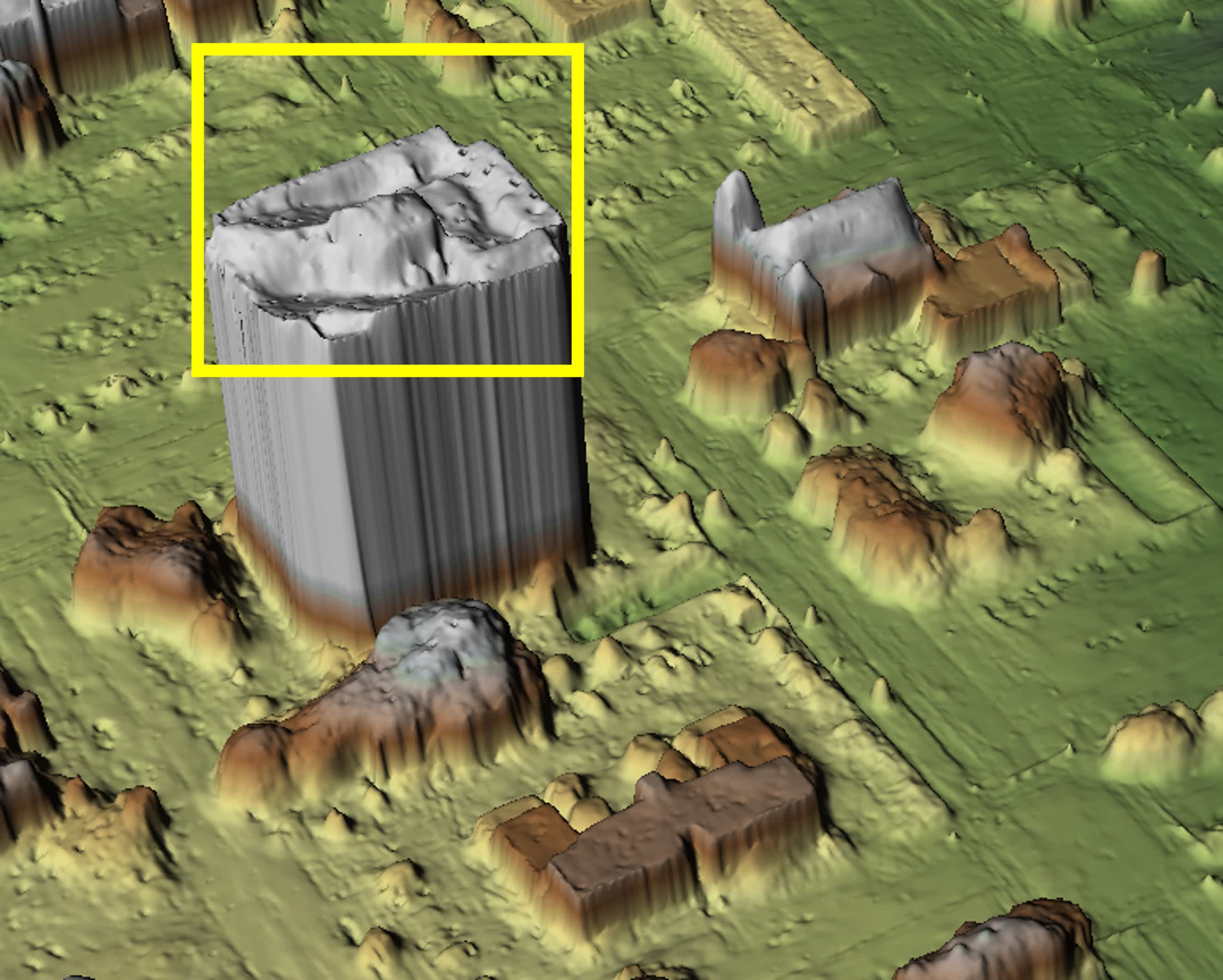} &
\myimg{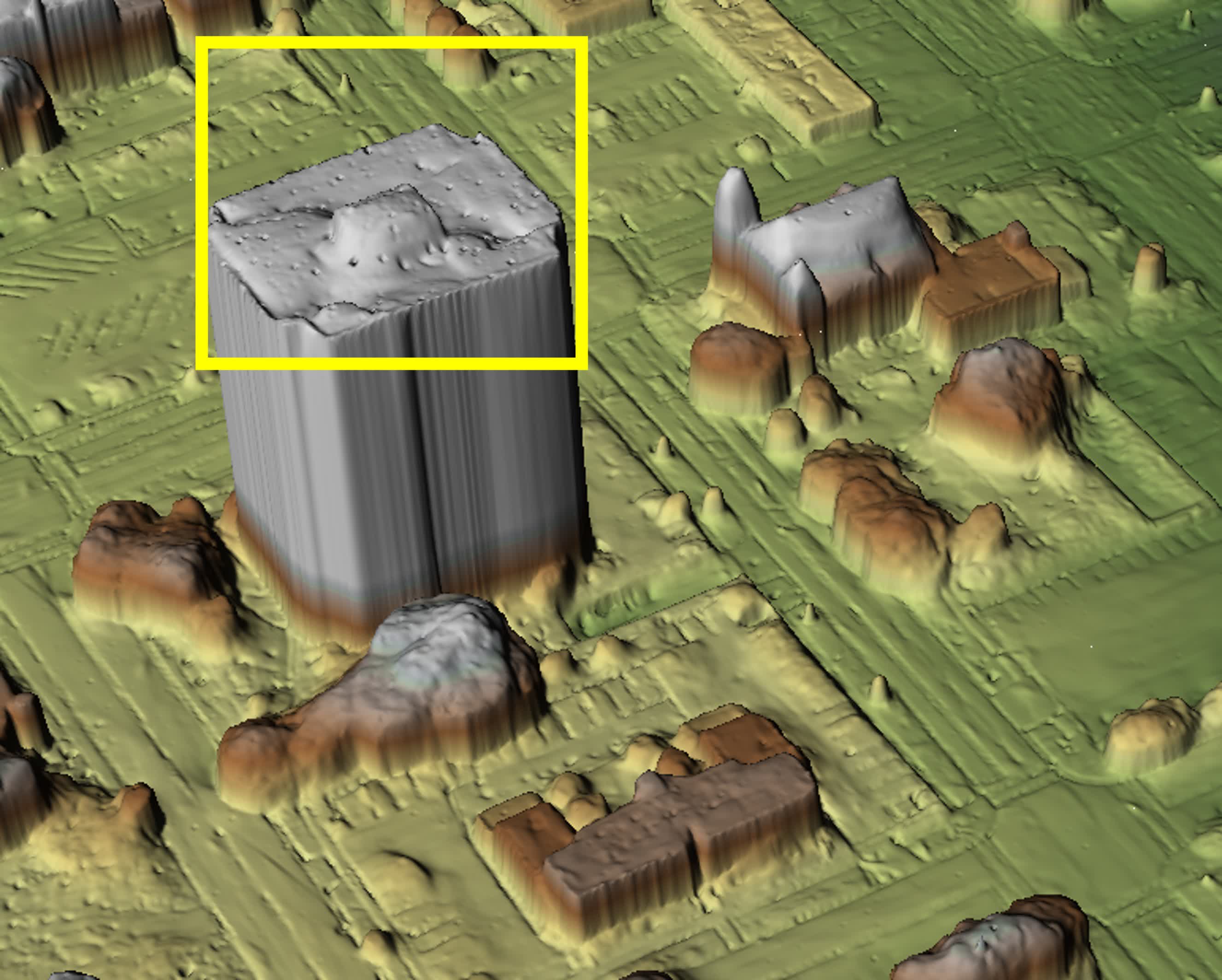} \\

% JAX-214
\rotatebox{90}{\small JAX-214} &
\myimg{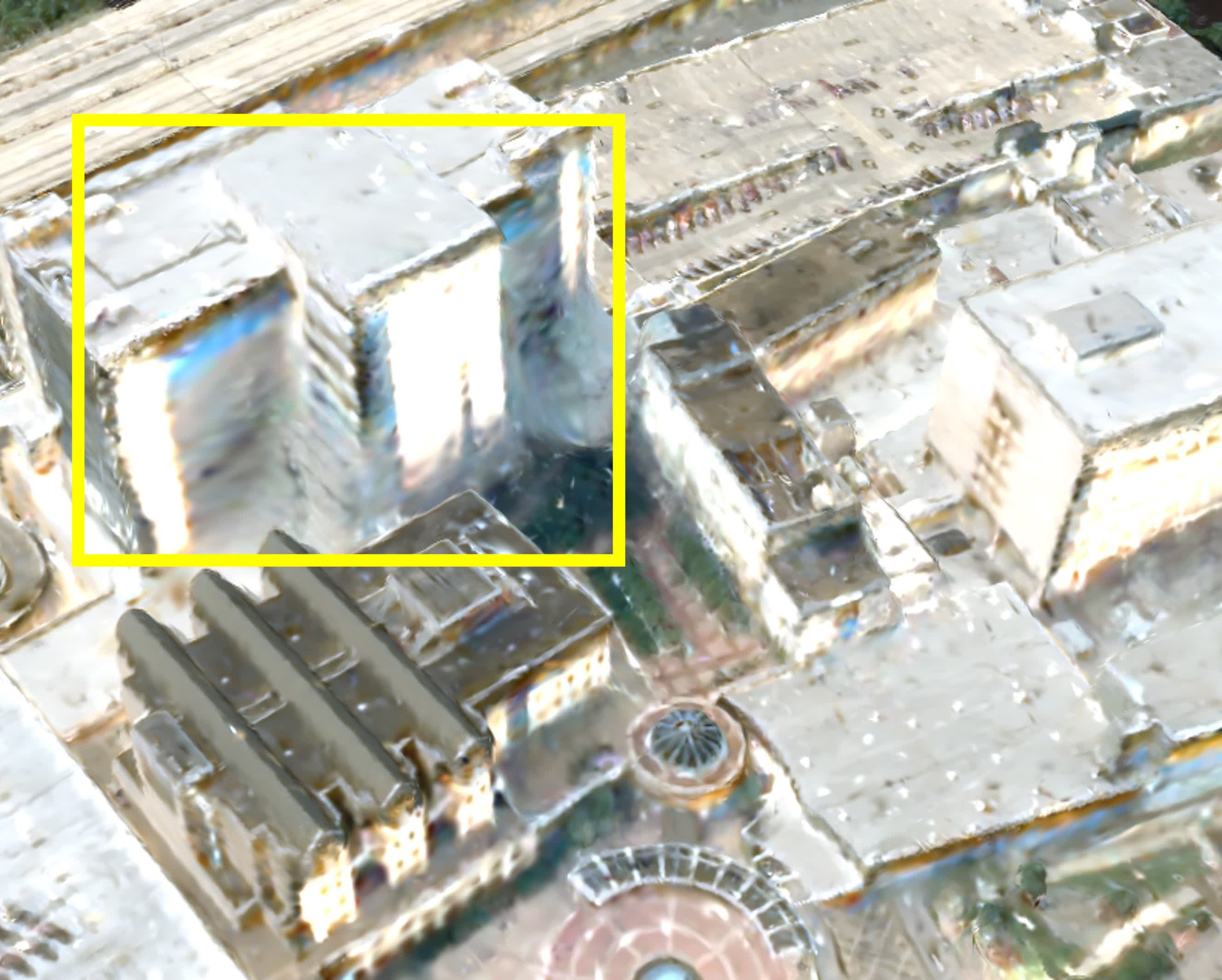} &
\myimg{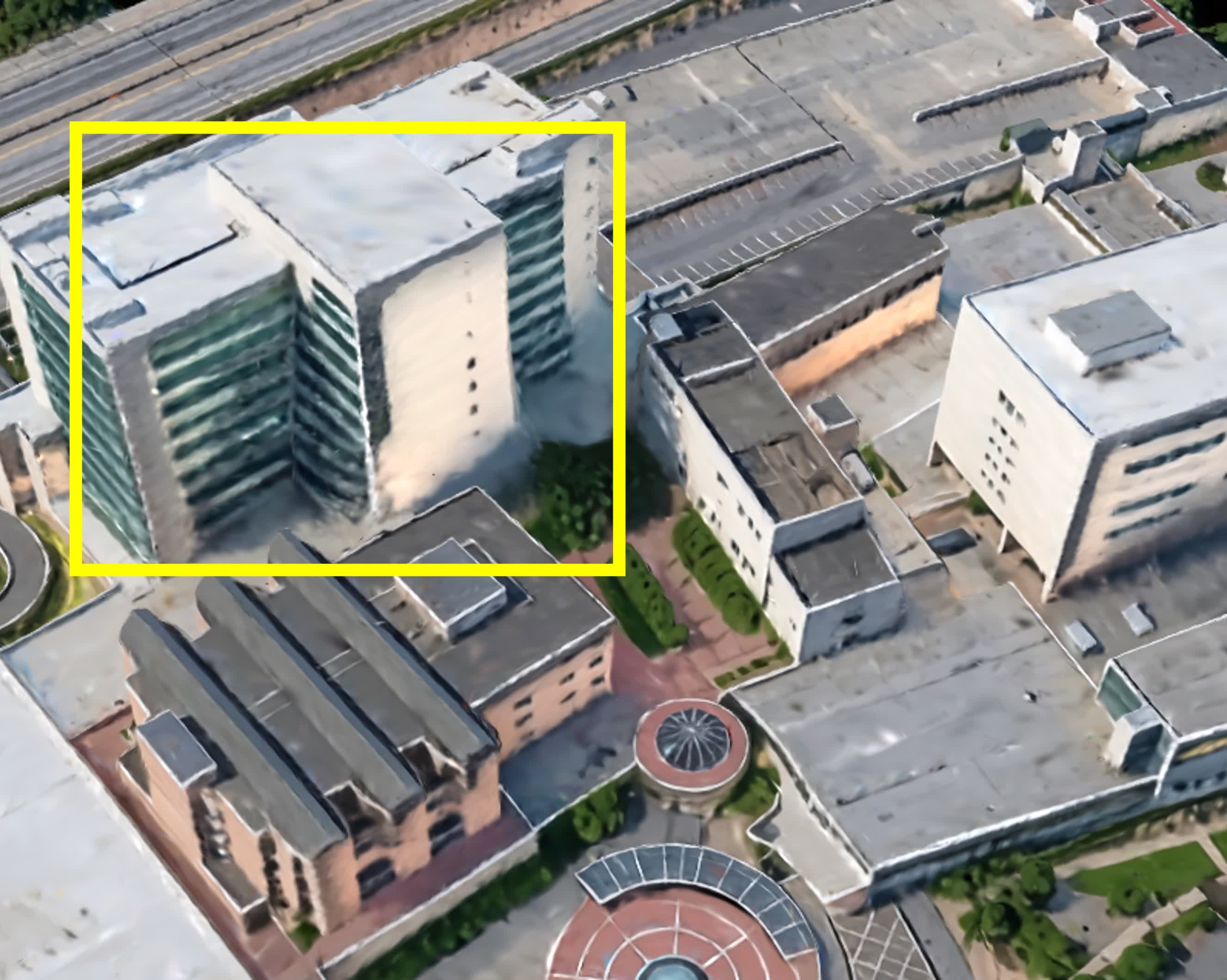} &
\myimg{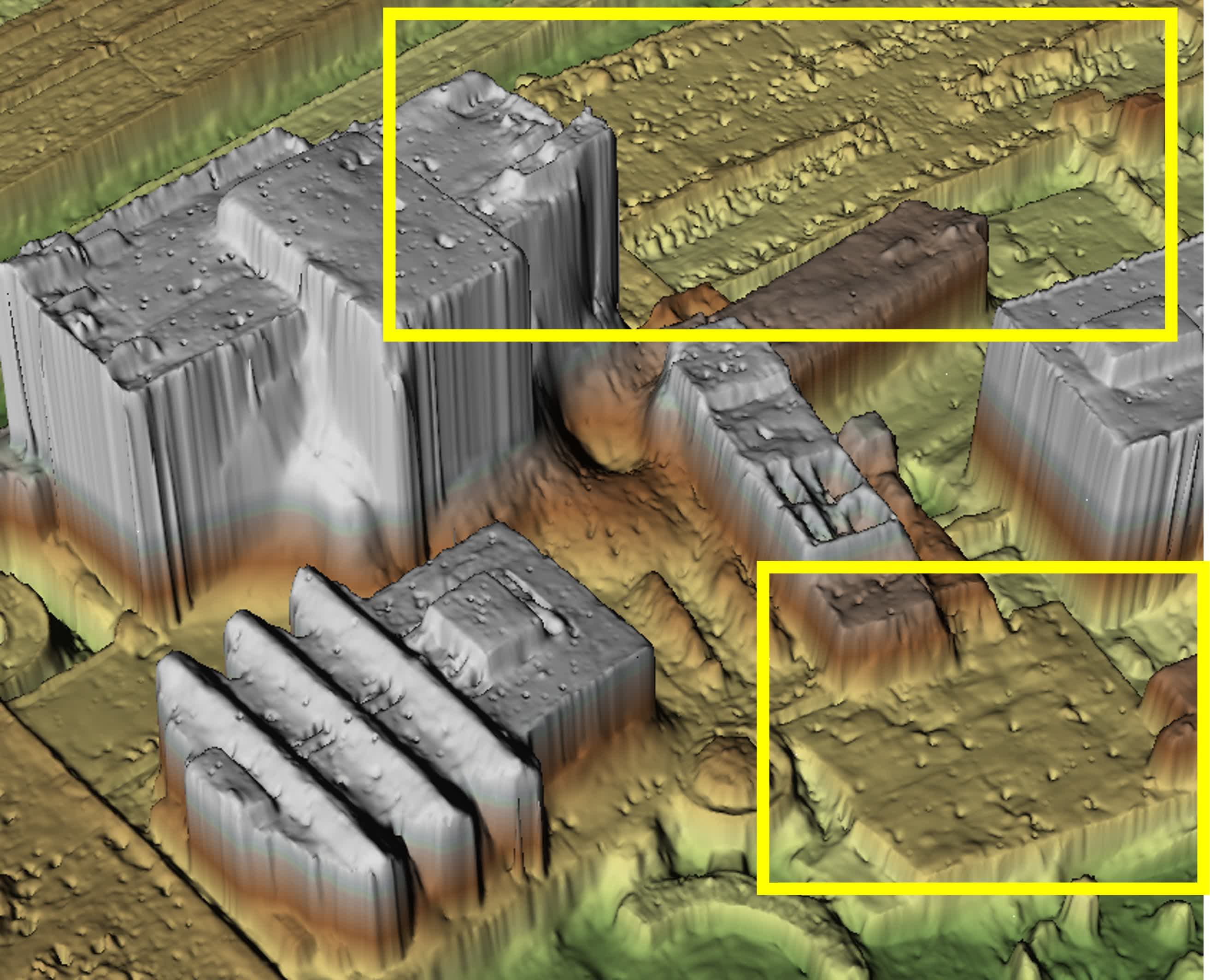} &
\myimg{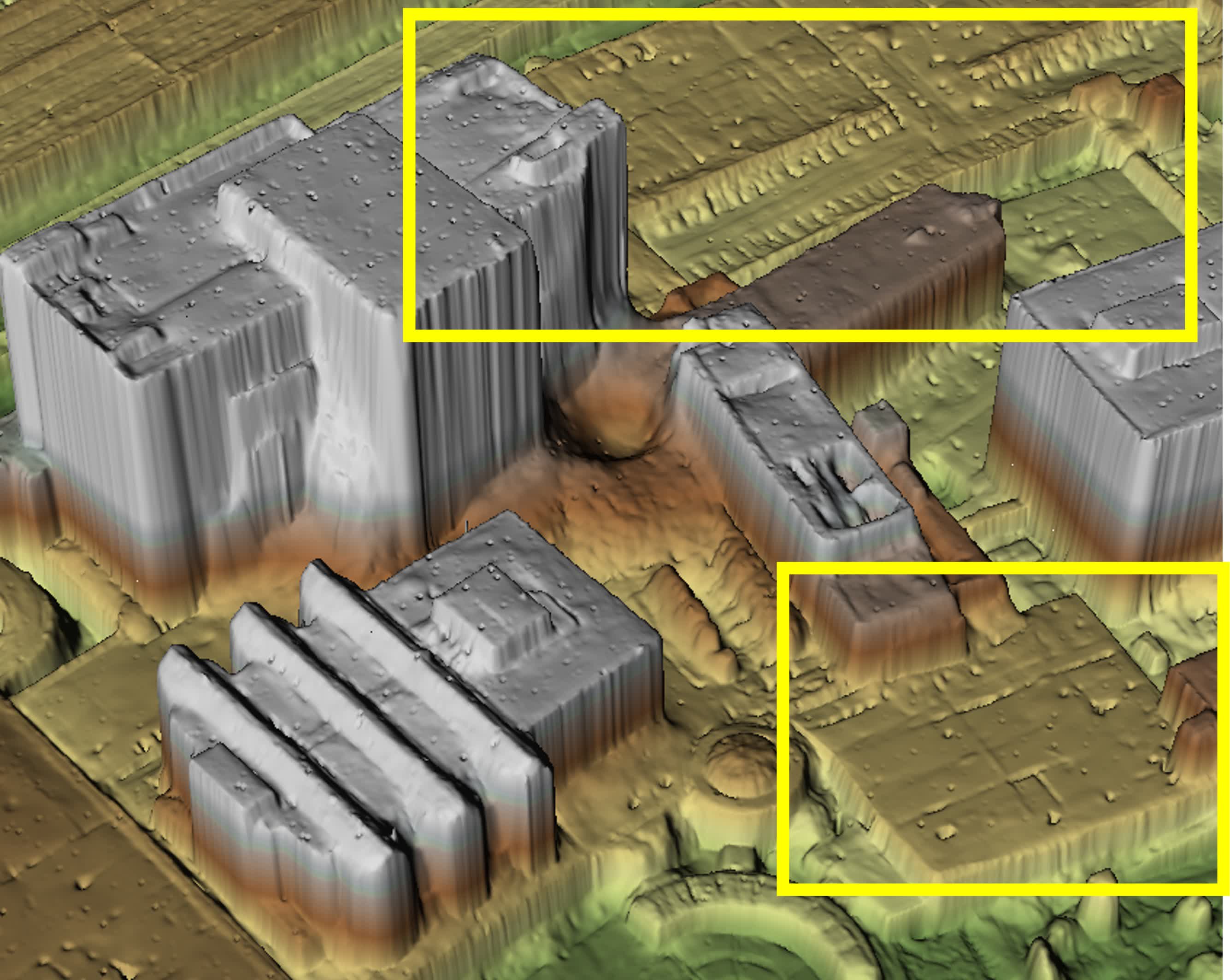} \\

% JAX-260
\rotatebox{90}{\small JAX-260} &
\myimg{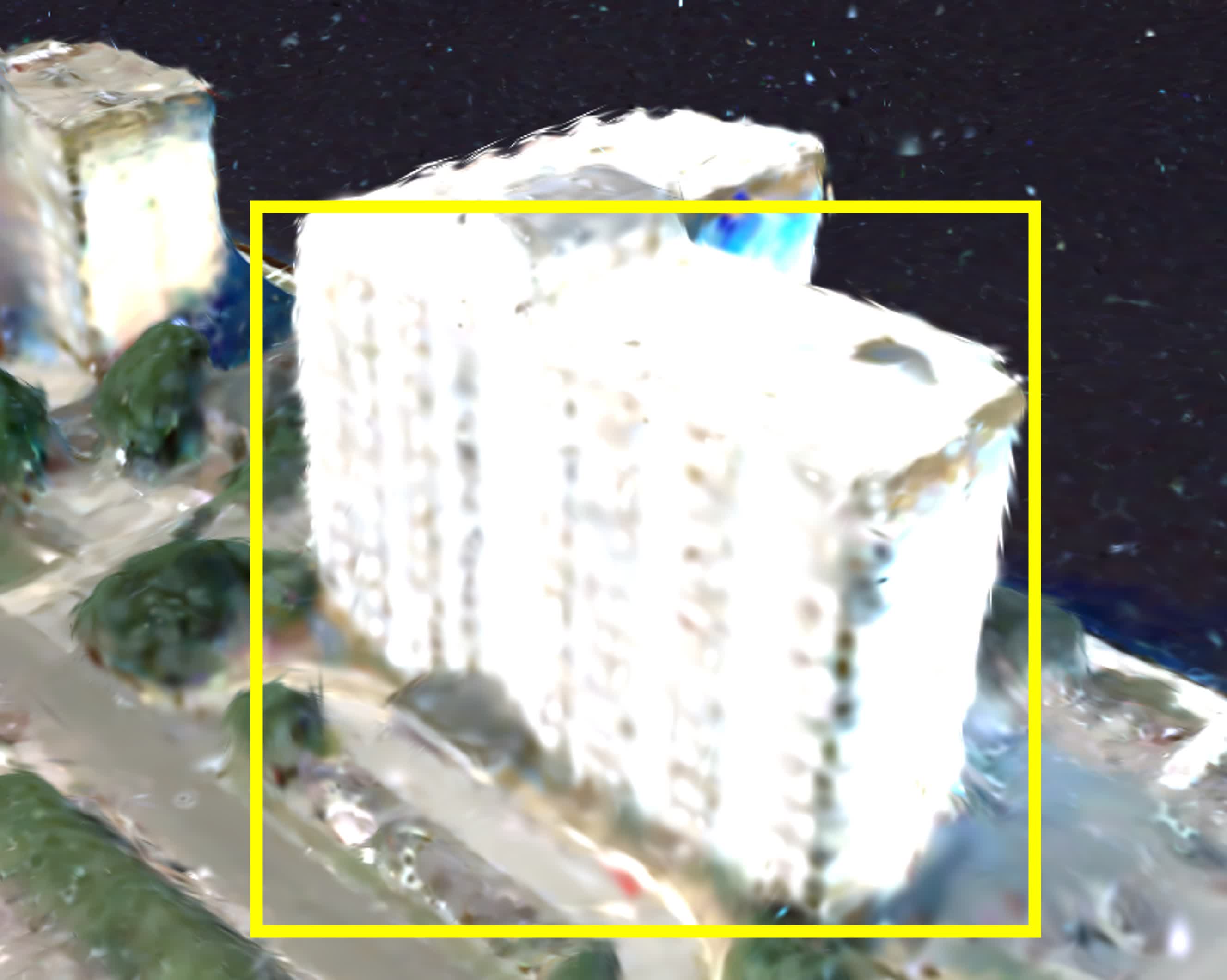} &
\myimg{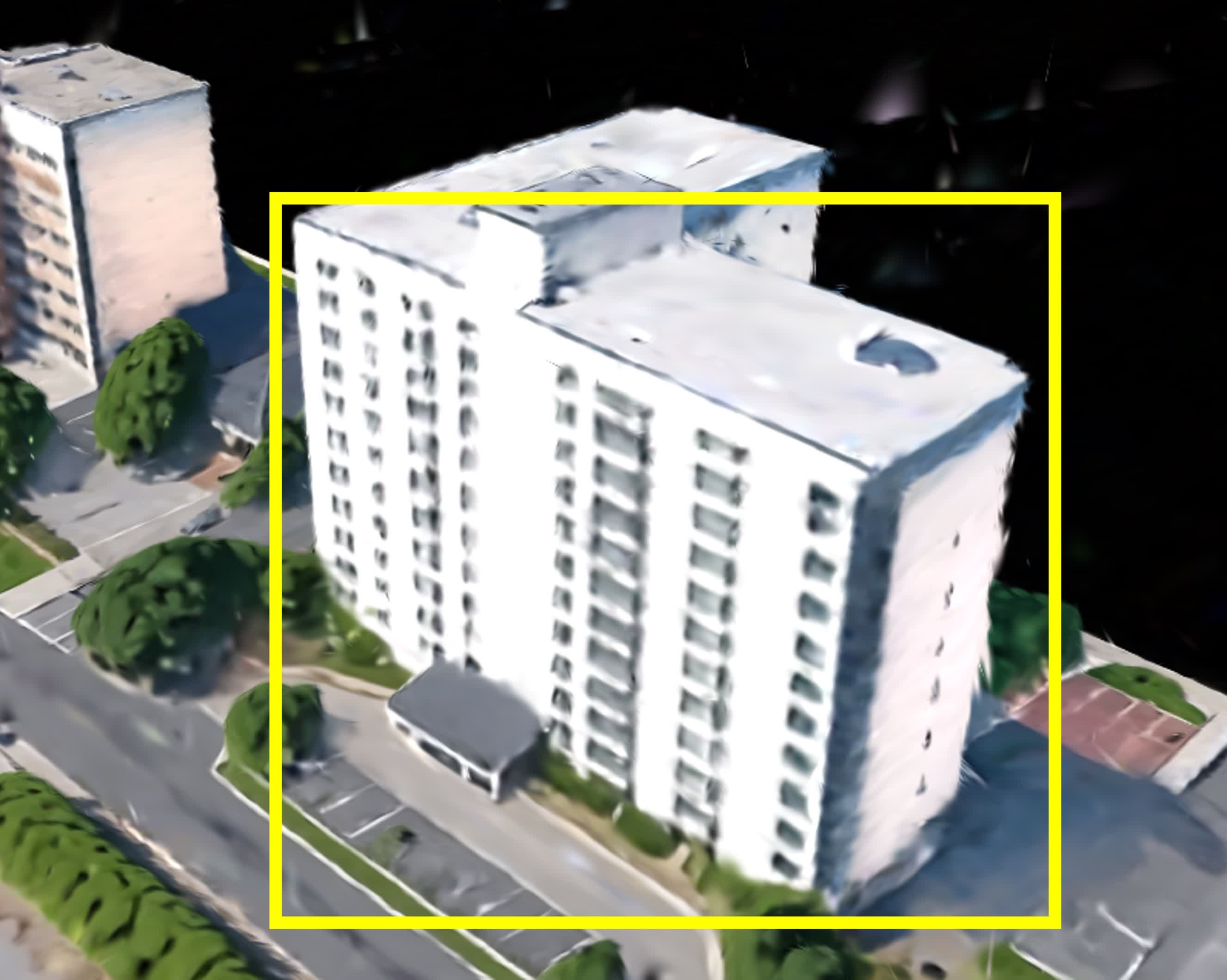} &
\myimg{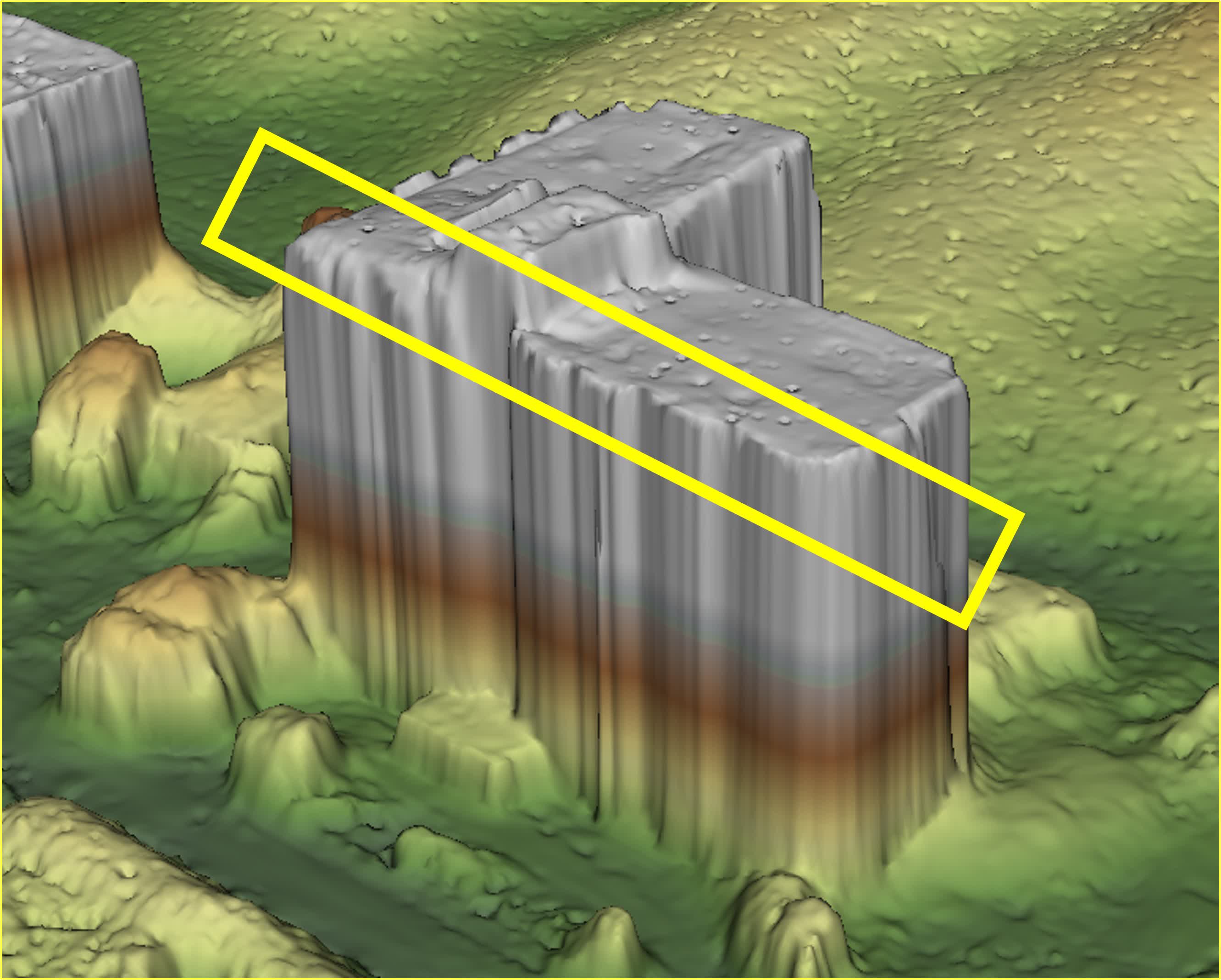} &
\myimg{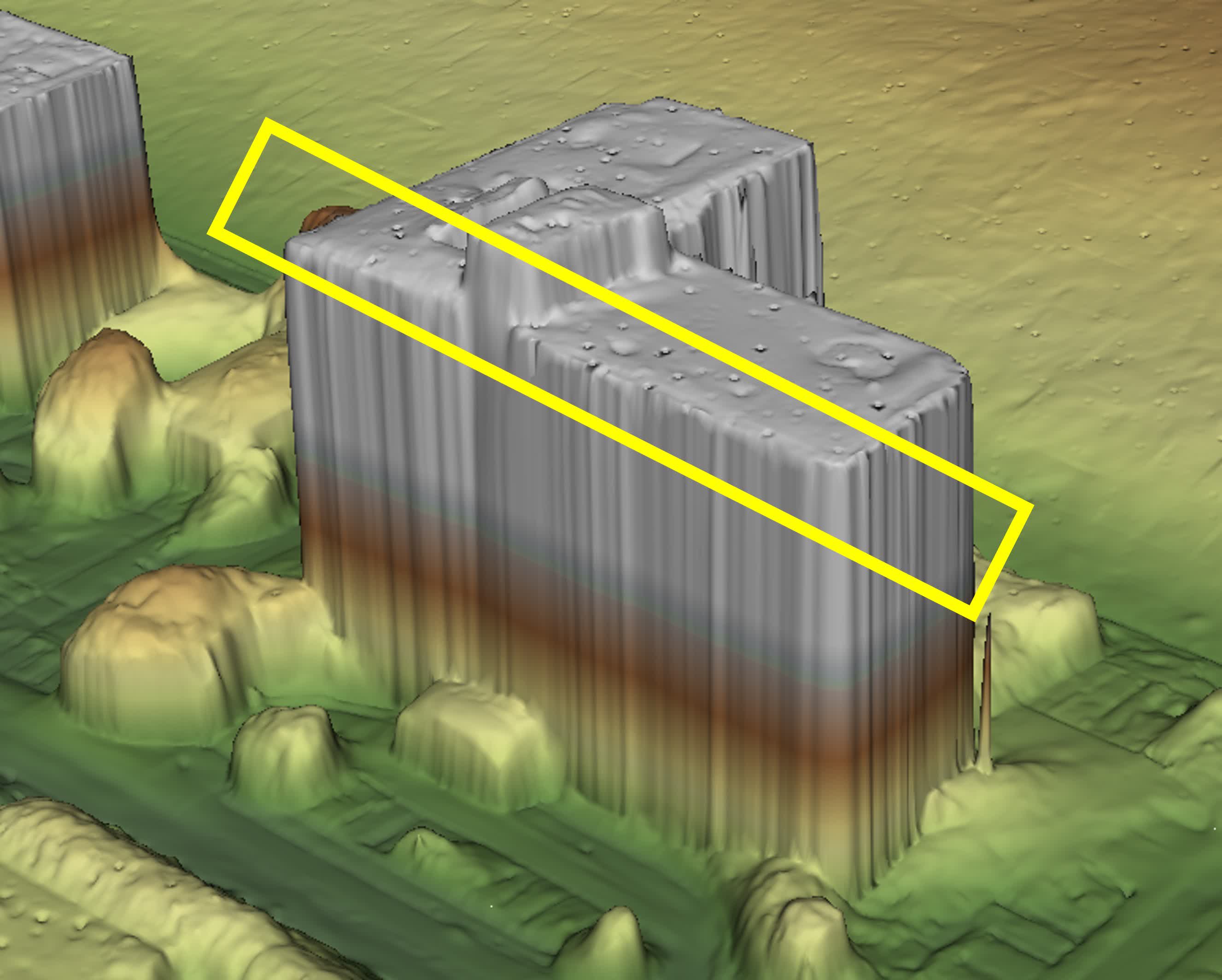} \\

\end{tabular}

\caption{Qualitative ablation of our generative refinement process. The comparison between the initial 2DGS output ("w/o Diffusion") and the refined results ("w/ Diffusion") highlights the framework's ability to recover high-frequency textures and sharpen building boundaries. }

\label{fig:diffusion_ablation_jax_final}

\end{figure*}

The effectiveness of our generative refinement pipeline is evaluated through an ablation study, as summarized in \Cref{fig:diffusion_ablation_jax_final} and \Cref{tab:ours_diffusion_ablation}. Qualitative analysis indicates improvements in building facade reconstruction. In sites JAX-068,  JAX-214 and JAX-260, which are characterized by high-rise structures, the refinement of facade textures is observed. Furthermore, comparative analysis of the reconstructed DSMs reveals that the generative refinement pipeline reduces rough textures and improve consistencies across the scene, while further regularizing structural boundaries. 

Quantitative comparison without generative refinement indicates that all evaluation criteria improve after generative refinement (\Cref{tab:ours_diffusion_ablation}). Distributional metrics show the most pronounced gains, with FID-CLIP decreasing from 48.34 to 19.50 and CMMD from 3.422 to 1.681, indicating that the refined reconstructions more closely resemble the visual structure of the Google Earth reference imagery. Pixel-level metrics improve as well, and our full configuration (FLUX.2 [klein] 4B) achieves the best PSNR, CW-SSIM, and LPIPS among all evaluated diffusion pipelines, even though the absolute gain over the unrefined baseline is comparatively modest due to the inherent pixel-level misalignment between the source satellite imagery and the Google Earth reference.

\begin{table}[htbp]
\centering
%\caption{Quantitative ablation study of the shadow-guided generative refinement stage, \RQ{evaluated using both texture similarity (Pixel-level metrics), perception metrics (Distributional metrics) and geometric accuracy ((MAE$_{reg}$). [Then, delete the rest of the texts in this caption]}This table illustrates the impact of integrating the generative refinement into the 2DGS structural foundation. The generative refinement does not merely refine textures but acts as a structural regularizer that sharpens building boundaries and aligns the DSM more accurate.\RQ{too verbiage}}
\caption{Quantitative ablation study of the shadow-guided generative refinement stage with different diffusion pipelines, evaluated using distributional metrics (FID-CLIP and CMMD), pixel-level metrics (PSNR, CW-SSIM, and LPIPS), and geometric accuracy (MAE$_{reg}$).}
\label{tab:ours_diffusion_ablation}
\resizebox{1.0\linewidth}{!}{ 
    \begin{tabular}{lcccccc}
    \toprule
    \multirow{2}{*}{\textbf{Configuration}} & \multicolumn{2}{c}{\textbf{Distributional}} & \multicolumn{3}{c}{\textbf{Pixel-level}} & \textbf{Geometric} \\
    \cmidrule(lr){2-3} \cmidrule(lr){4-6} \cmidrule(lr){7-7}
    & FID-CLIP $\downarrow$ & CMMD $\downarrow$ & PSNR $\uparrow$ & CW-SSIM $\uparrow$ & LPIPS $\downarrow$ & MAE$_{reg}$ $\downarrow$ \\
    \midrule
    No Enhancement & 48.34 & 3.422 & 11.87 & 0.368 & 0.679 & 1.28 \\
    Skyfall-GS (FlowEdit+FLUX.1 [dev]) & 27.06 & 2.125 & 11.83 & 0.386 & 0.716 & 1.89 \\
    Ours (FlowEdit+FLUX.1 [dev]) & 36.27 & 2.251 & 12.14 & 0.399 & 0.668 & 1.27 \\
    Ours (FLUX.2 [klein] 4B) & \textbf{19.50} & \textbf{1.681} & \textbf{12.26} & \textbf{0.414} & \textbf{0.611} & \textbf{1.25} \\
    \bottomrule
    \end{tabular}
}
\end{table}

\begin{figure*}[tp] 
    \centering
    \small
    
    \newcommand{\myfigcrop}[1]{%
        \includegraphics[width=\linewidth, keepaspectratio, trim=0 5 0 5, clip]{#1}%
    }
    
    \def\colwfive{0.19\linewidth} 
    {\tiny
    \makebox[\colwfive]{Google Earth}
    \makebox[\colwfive]{Skyfall-GS (FlowEdit)}
    \makebox[\colwfive]{No Enhancement}
    \makebox[\colwfive]{FlowEdit+FLUX.1 [dev]}
    \makebox[\colwfive]{FLUX.2 [klein] 4B} \\
    \vspace{1mm}
    }
    \rotatebox{90}{\makebox[0.12\linewidth][c]{\footnotesize JAX-068}}
    \begin{subfigure}{\colwfive}\myfigcrop{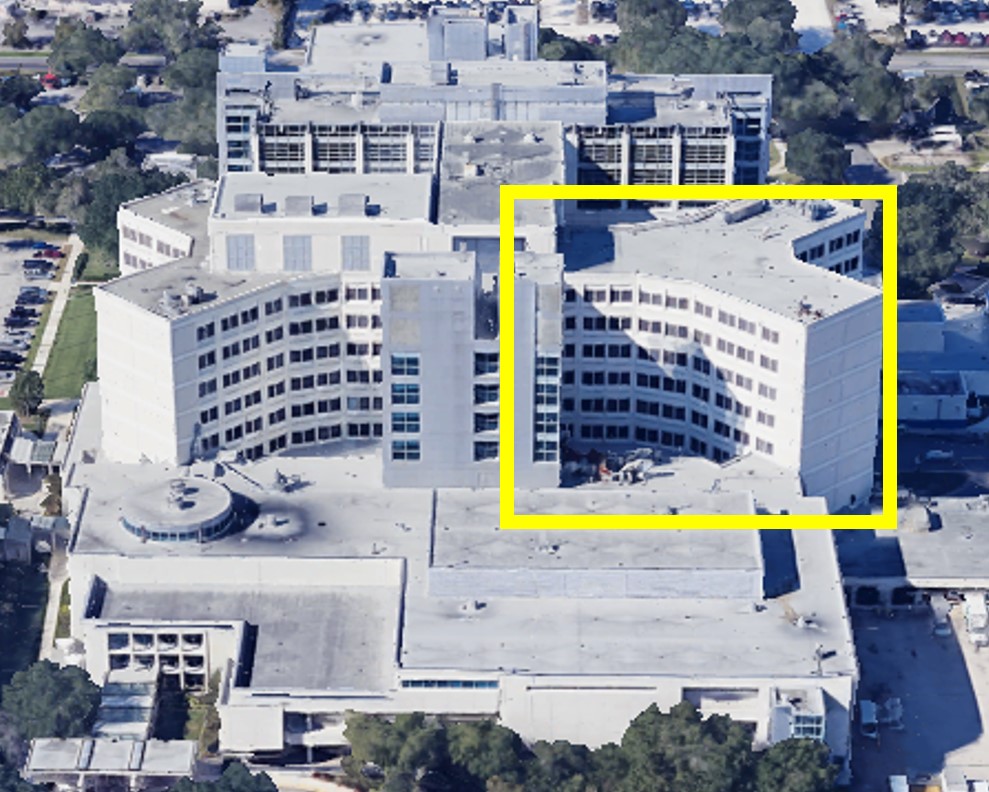}\end{subfigure}
    \begin{subfigure}{\colwfive}\myfigcrop{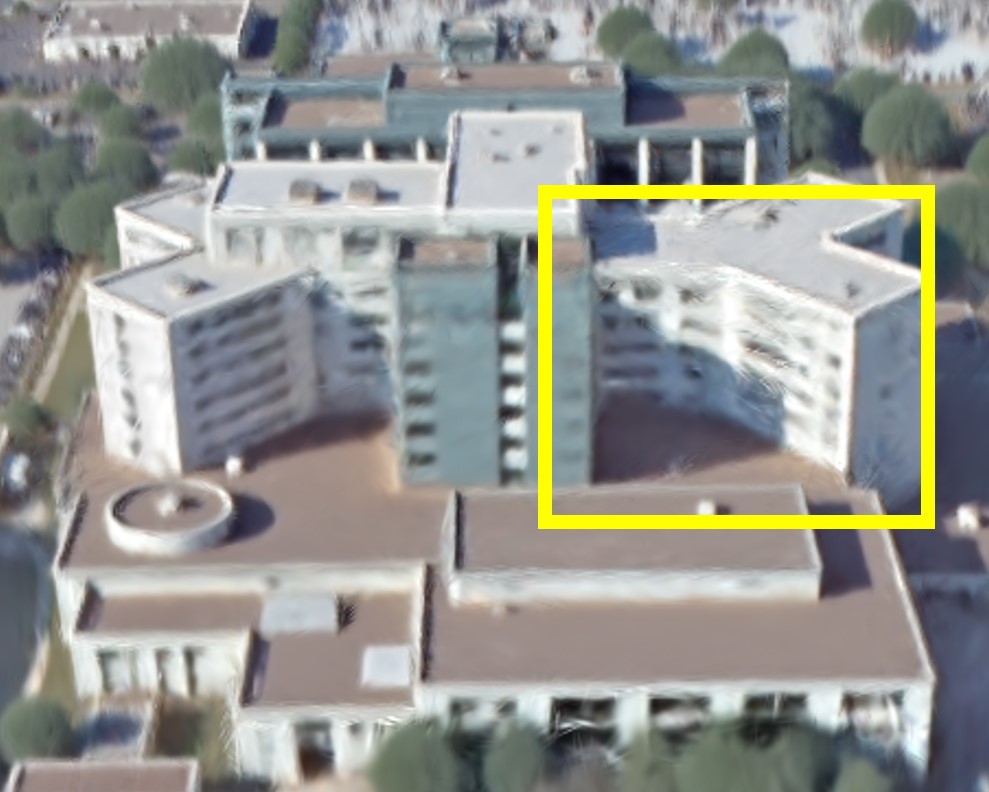}\end{subfigure}
    \begin{subfigure}{\colwfive}\myfigcrop{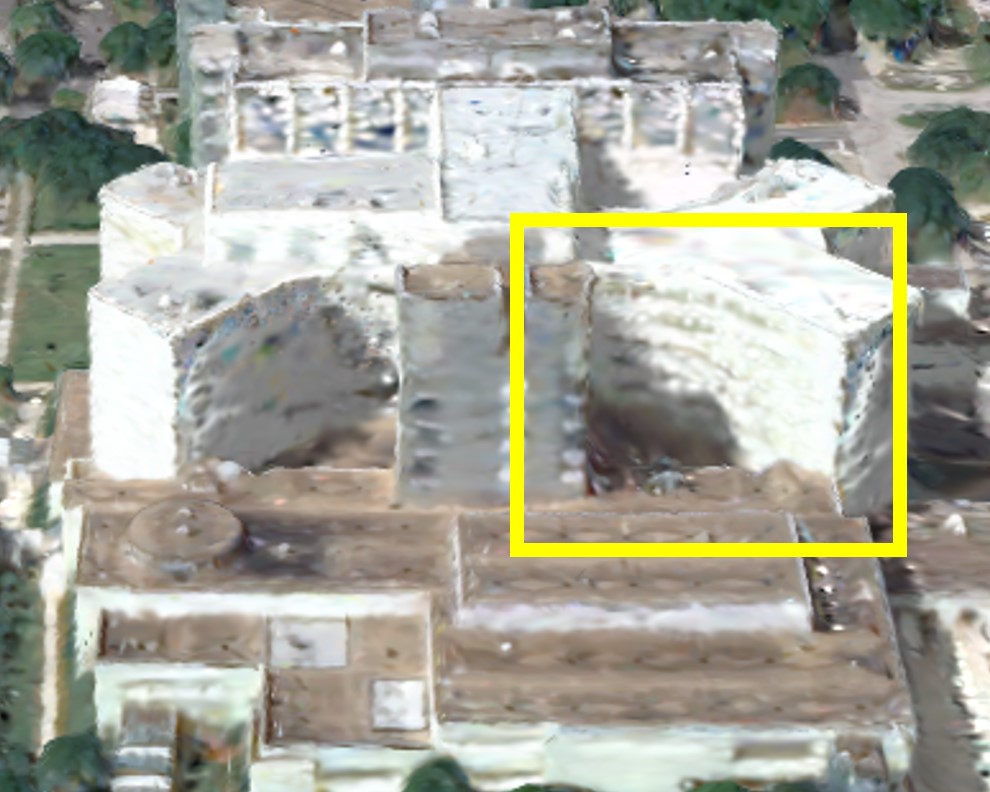}\end{subfigure}
    \begin{subfigure}{\colwfive}\myfigcrop{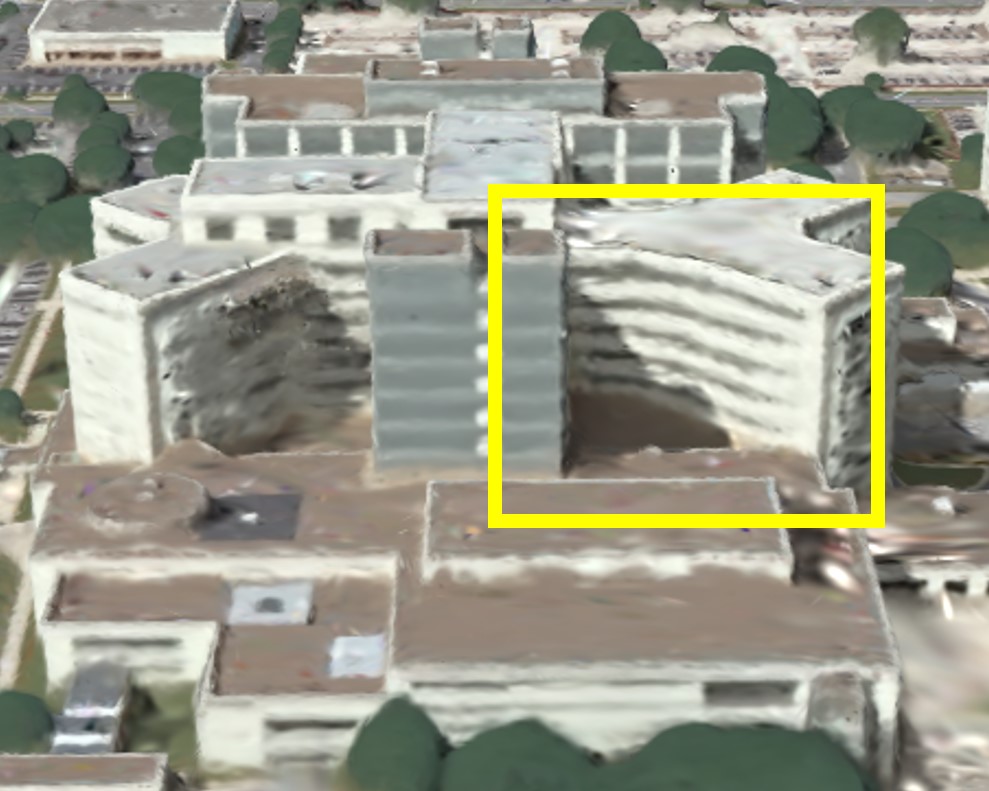}\end{subfigure}
    \begin{subfigure}{\colwfive}\myfigcrop{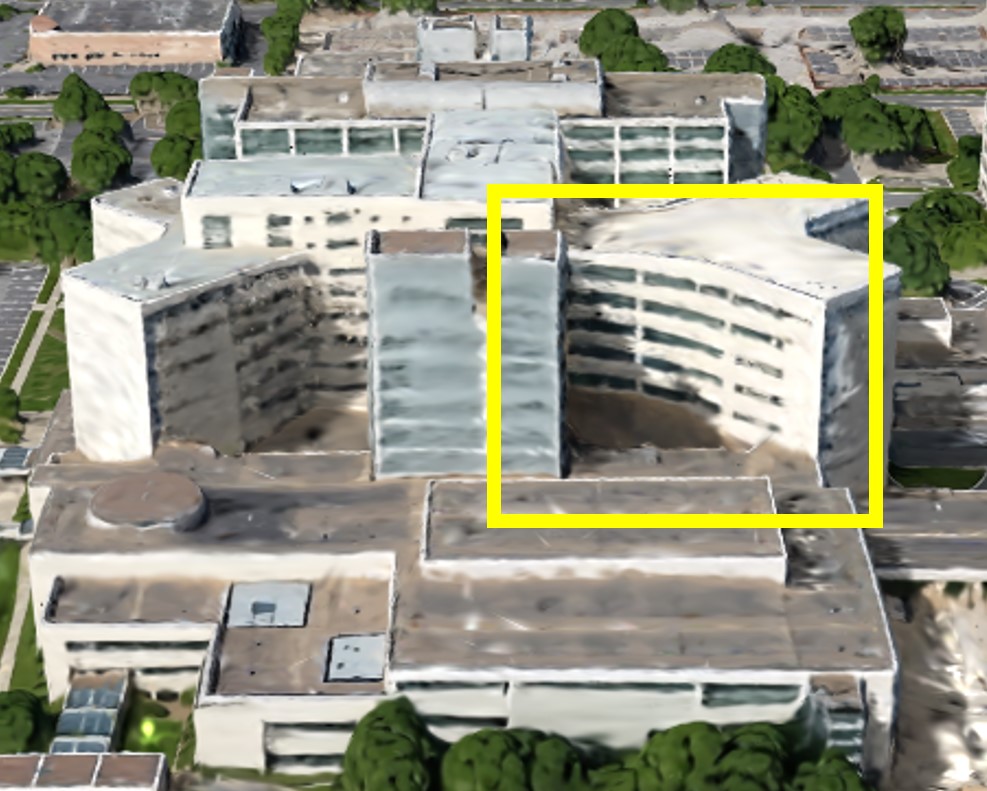}\end{subfigure} \\
    \vspace{0.5mm}

    % --- JAX-214 Section ---
    \rotatebox{90}{\makebox[0.12\linewidth][c]{\footnotesize JAX-214}}
    \begin{subfigure}{\colwfive}\myfigcrop{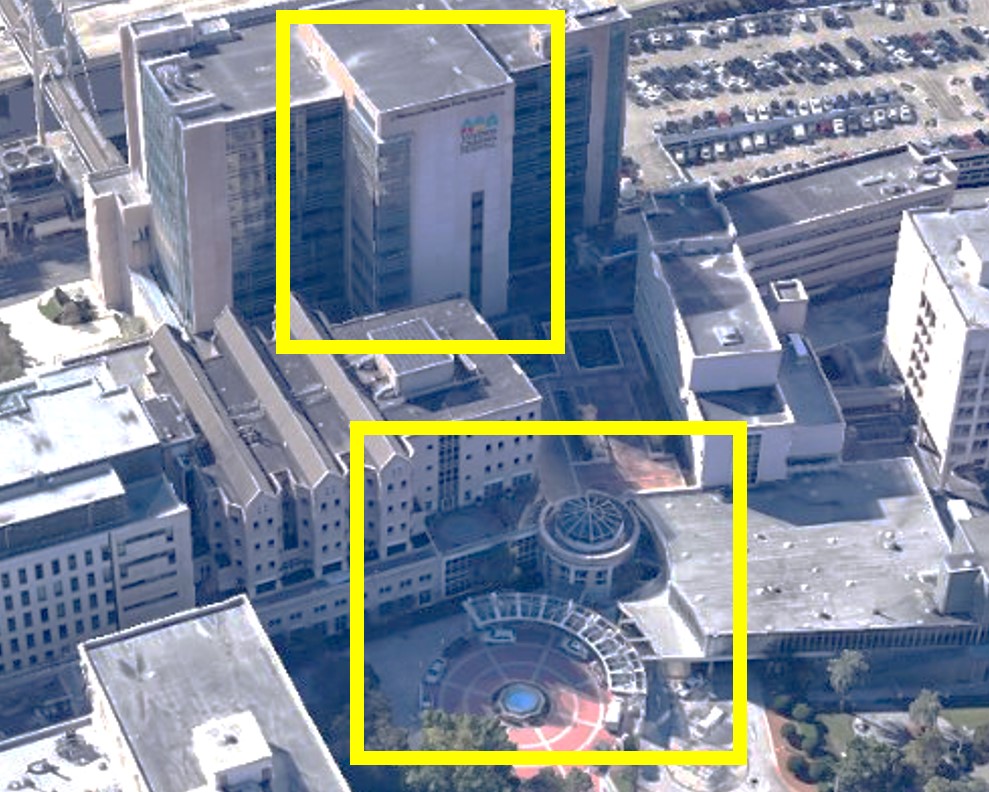}\end{subfigure}
    \begin{subfigure}{\colwfive}\myfigcrop{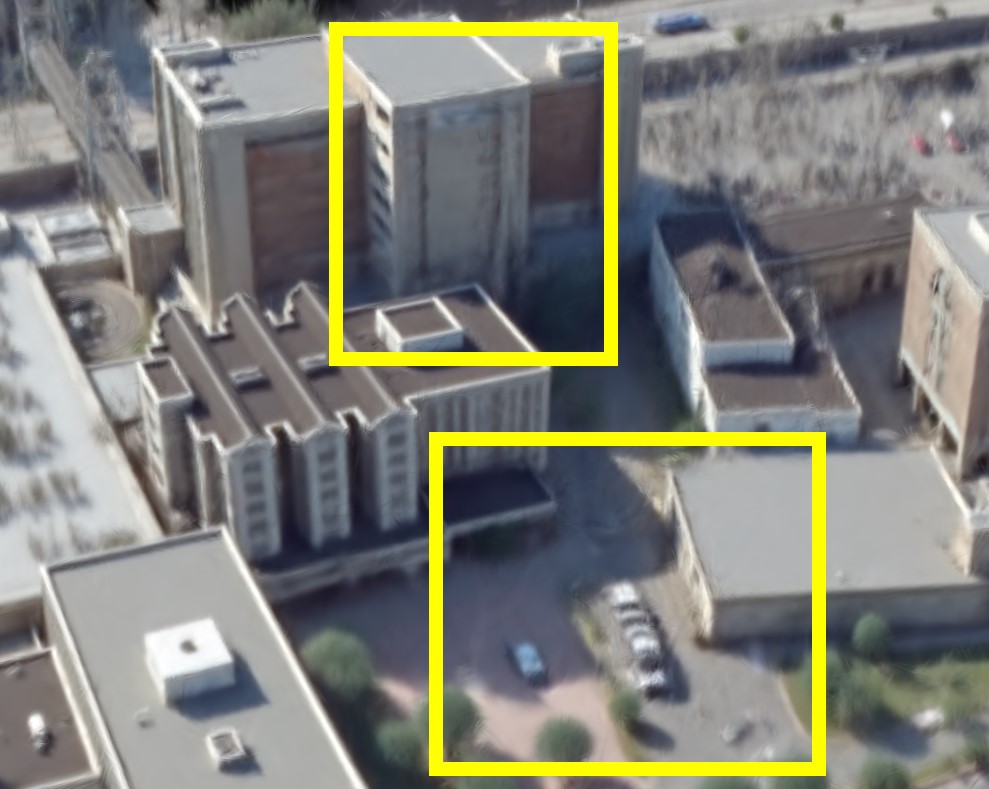}\end{subfigure}
    \begin{subfigure}{\colwfive}\myfigcrop{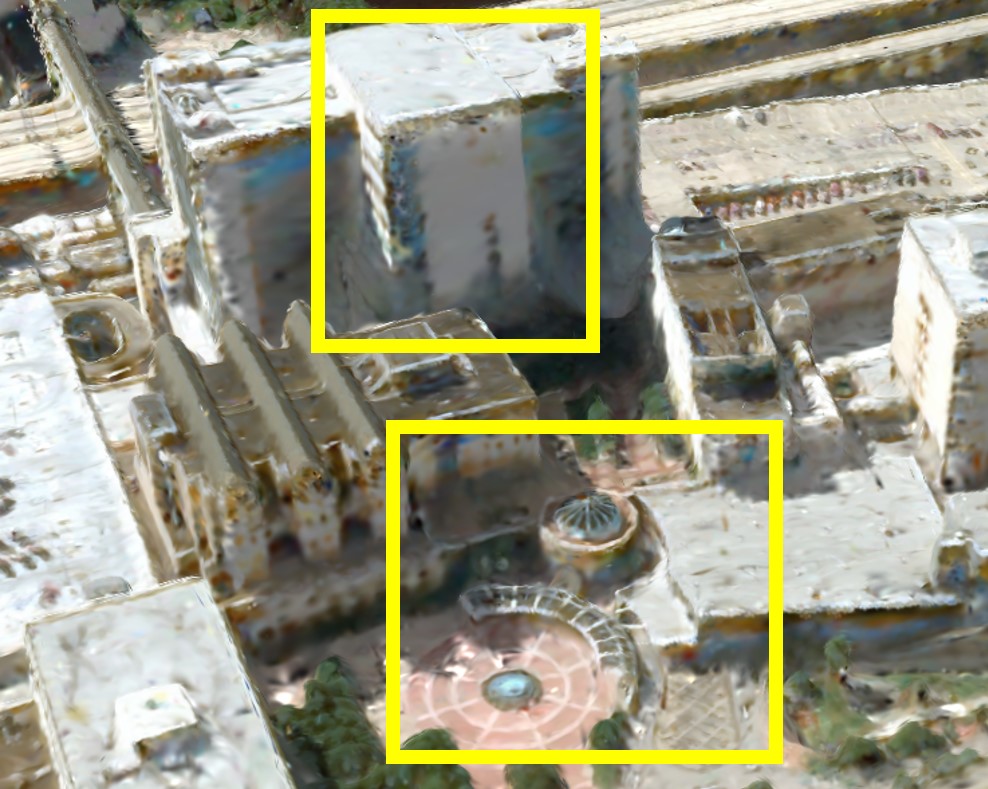}\end{subfigure}
    \begin{subfigure}{\colwfive}\myfigcrop{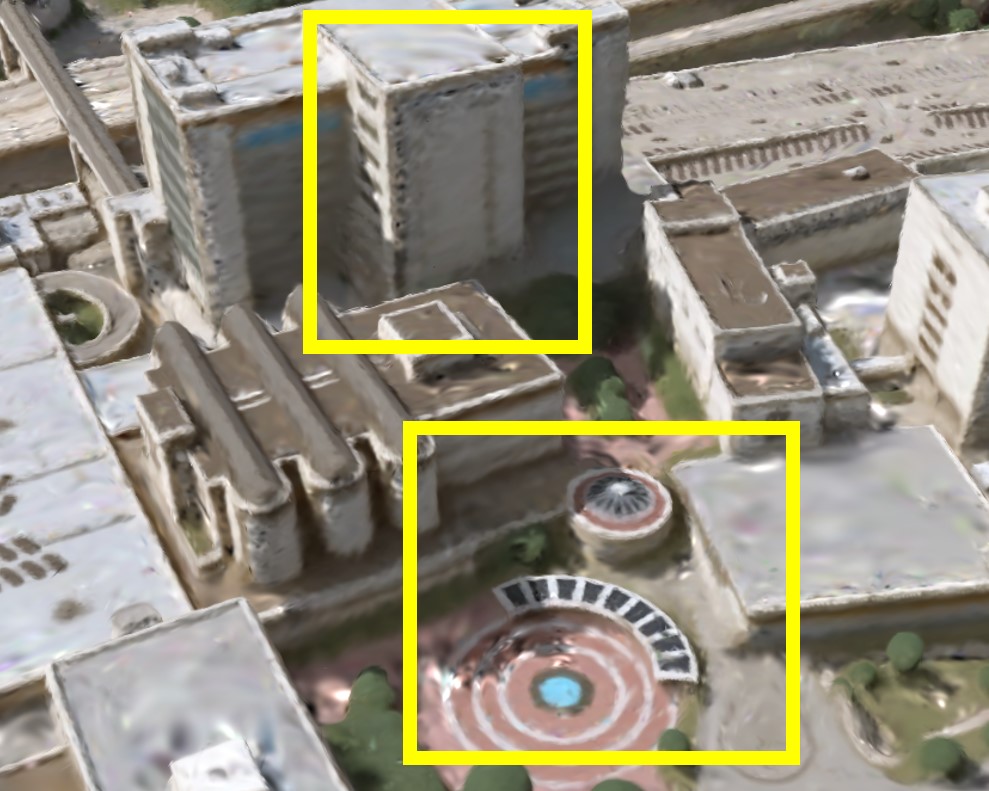}\end{subfigure}
    \begin{subfigure}{\colwfive}\myfigcrop{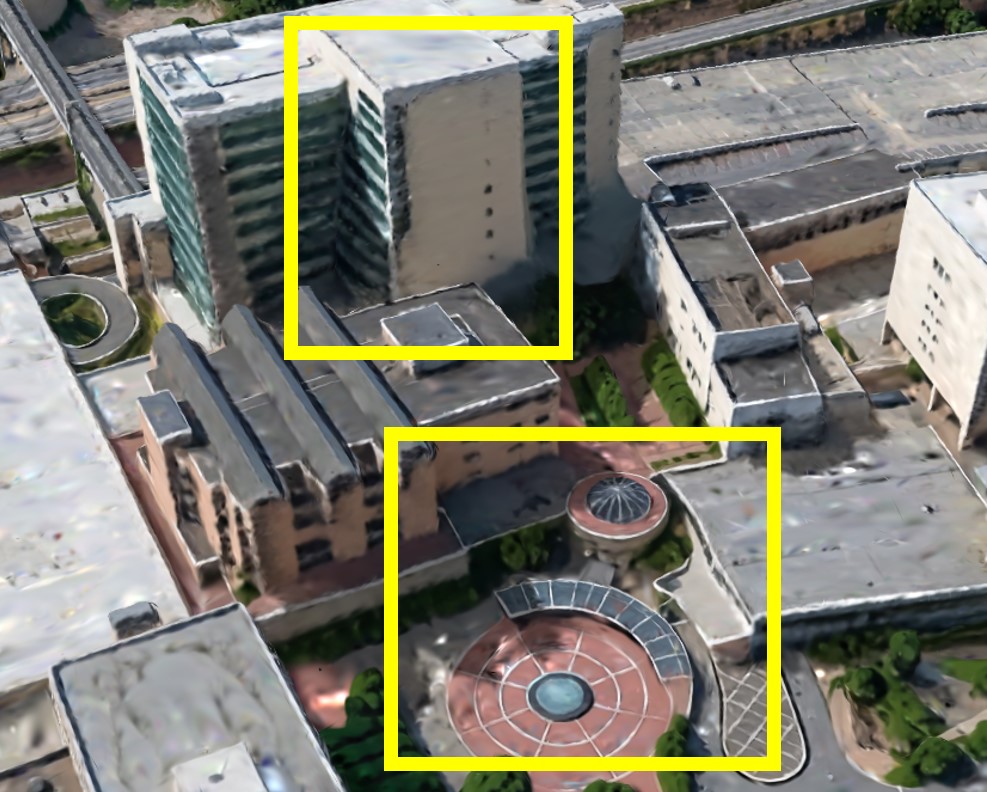}\end{subfigure} \\
    \vspace{0.5mm}

    \rotatebox{90}{\makebox[0.12\linewidth][c]{\footnotesize JAX-260}}
    \begin{subfigure}{\colwfive}\myfigcrop{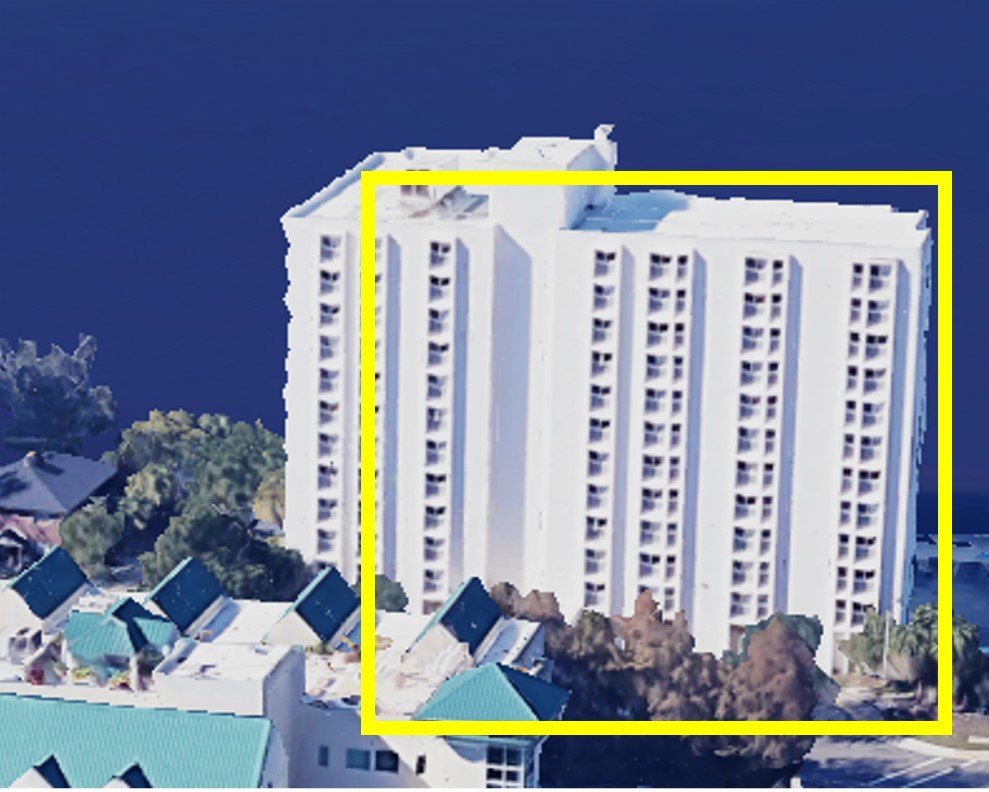}\end{subfigure}
    \begin{subfigure}{\colwfive}\myfigcrop{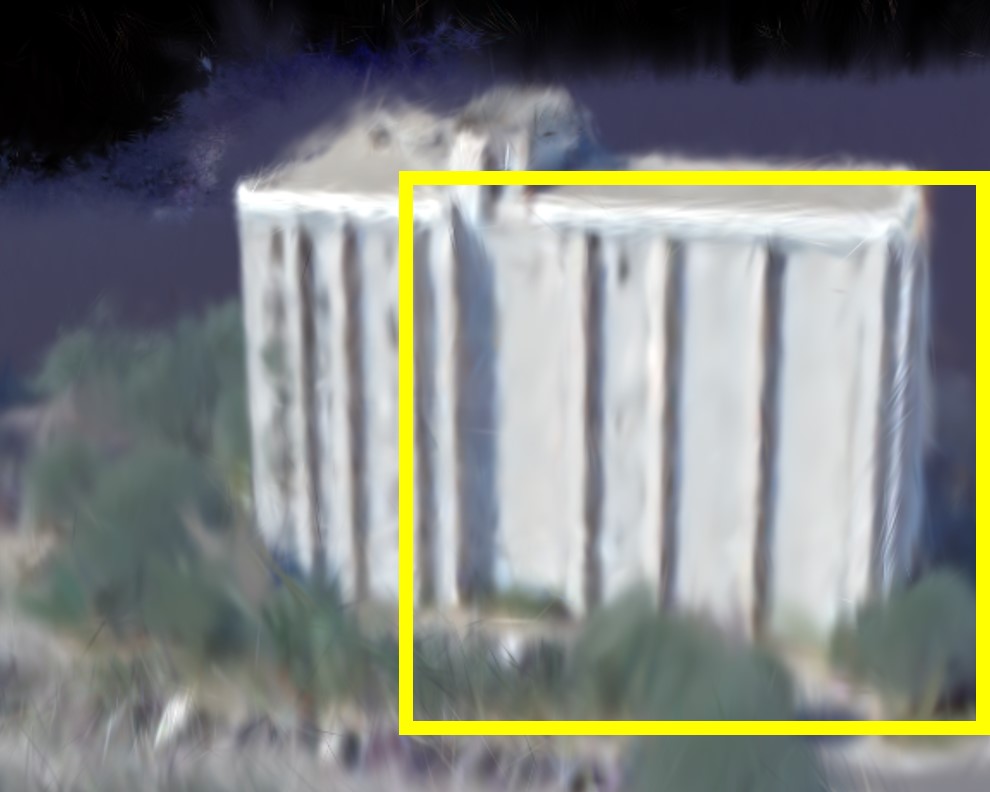}\end{subfigure}
    \begin{subfigure}{\colwfive}\myfigcrop{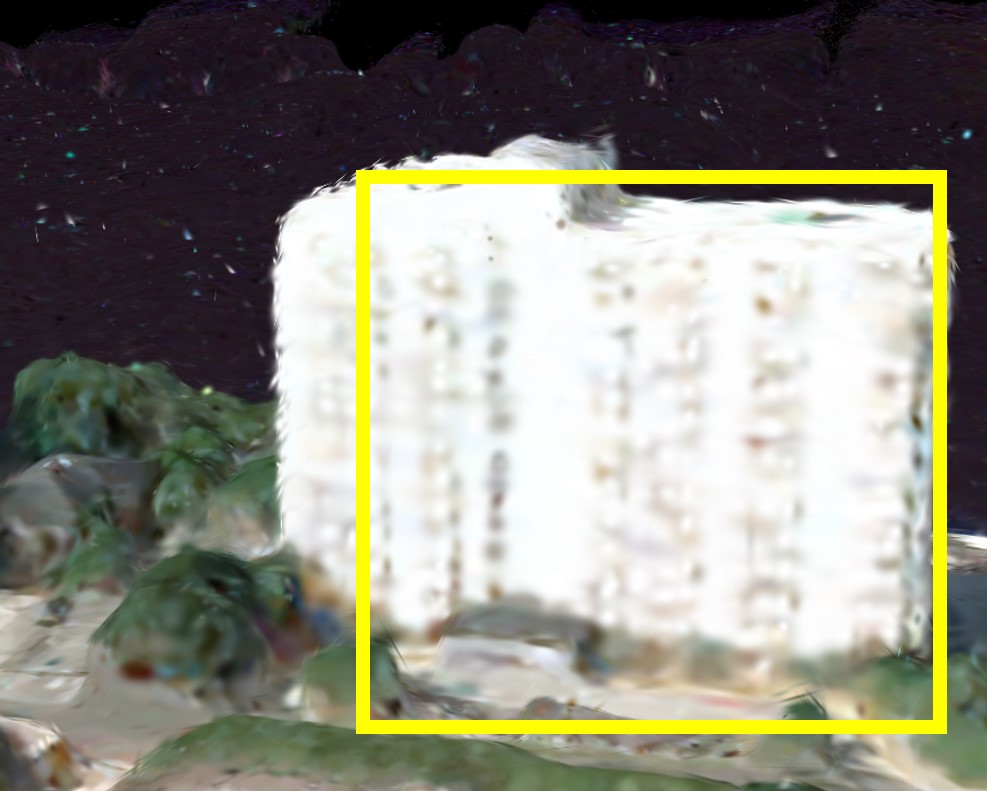}\end{subfigure}
    \begin{subfigure}{\colwfive}\myfigcrop{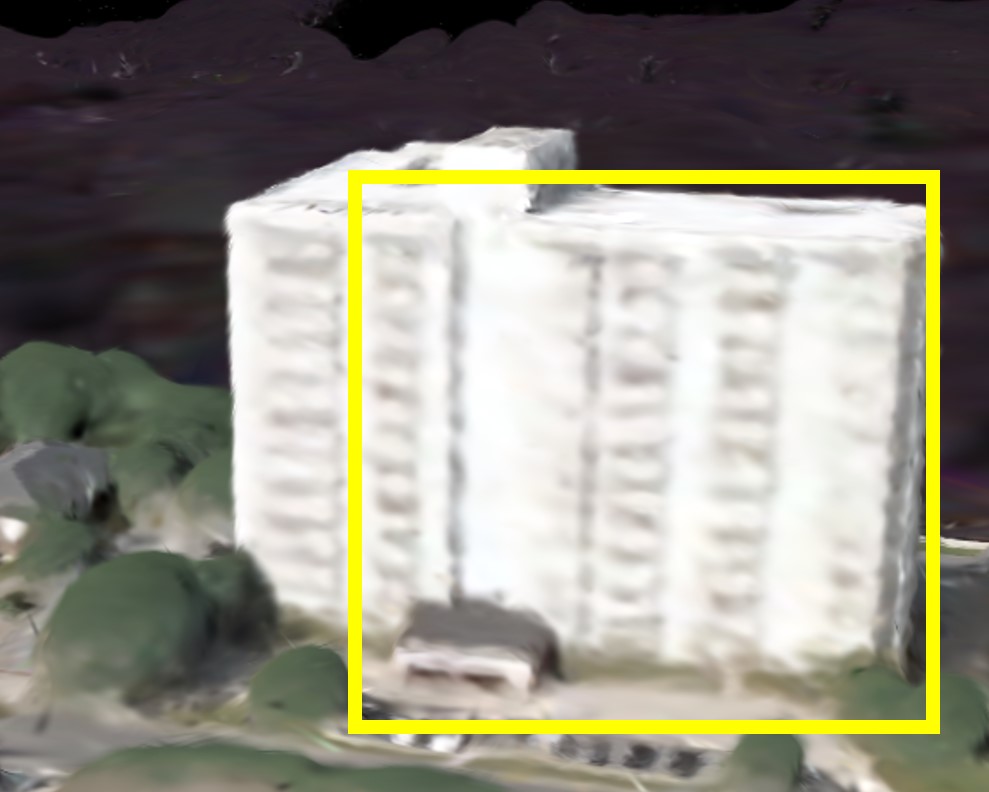}\end{subfigure}
    \begin{subfigure}{\colwfive}\myfigcrop{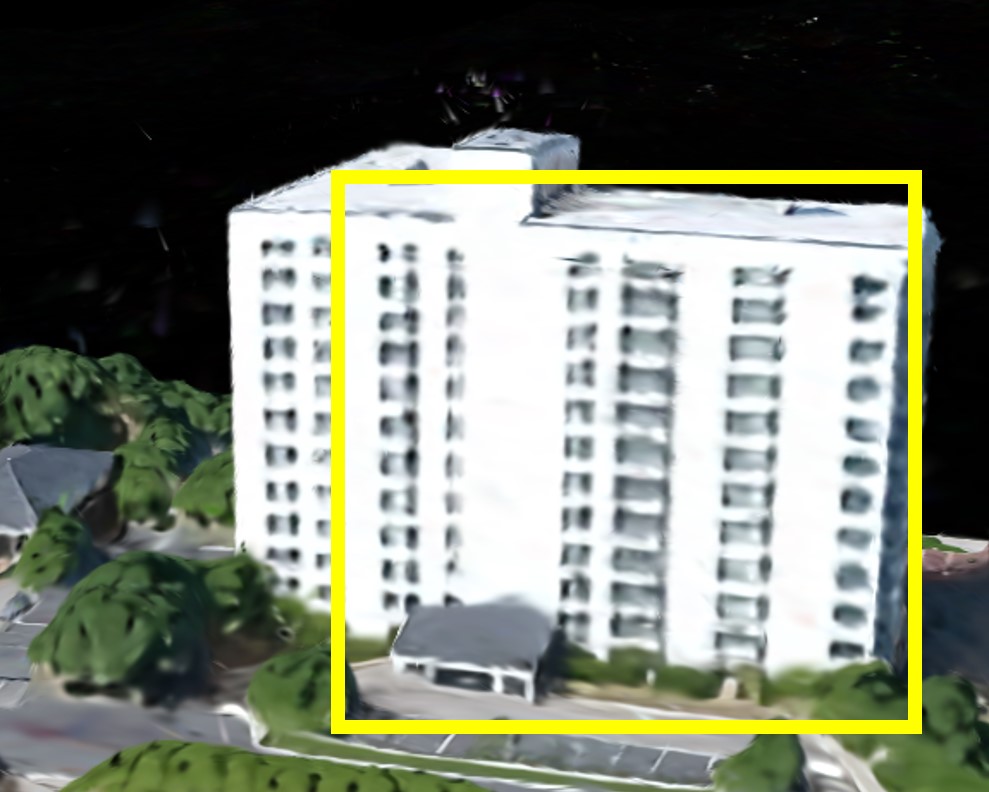}\end{subfigure} \\
    \vspace{1mm} 

    %\caption{Qualitative comparison of satellite image enhancement on different diffusion pipeline (FlowEdit). Our shadow-guided pipeline maintains geometric integrity and structural alignment regardless of the specific generative model used. This demonstrates the robustness of imposing shadow casting as geometric guidance for high-fidelity satellite scene reconstruction.\RQ{i think you do not need to over-emphasize shadow guiding since one cannot see how the shadow made the impact, you just need to state that we recover very well and clean facade}}
    \caption{Qualitative comparison of satellite image enhancement using different diffusion pipelines (FlowEdit). Our shadow-guided refinement improves visual details and recovers cleaner building facades while preserving the underlying geometric structures reconstructed from Gaussian optimization.}
    \label{fig:ablation_flowedit}
\end{figure*}

We additionally evaluate an alternative diffusion pipeline, FlowEdit with FLUX.1 [dev], to verify the robustness of our framework (\Cref{fig:ablation_flowedit}). Even with this change in diffusion backbone, performance improves consistently across all metrics relative to the unrefined baseline. Notably, when comparing our framework and Skyfall-GS under the same FlowEdit pipeline, Skyfall-GS achieves slightly better distributional metrics, while our framework achieves better pixel-level alignment and geometric accuracy ($\mathrm{MAE}_{reg}$). These results suggest that our shadow-guided refinement preserves sensor-observed geometry and improves visual fidelity regardless of the underlying diffusion backbone.

A key observation is that our diffusion pipeline improves visual fidelity while preserving the geometric accuracy established during the optimization stage. Shadow casting prevents geometric degradation that would otherwise occur during diffusion-based refinement, while the refined appearance improves visual fidelity. This demonstrates the effectiveness of our shadow-guided generative refinement in decoupling appearance enhancement from geometric change.

%\begin{table}[!htbp]
%\centering
%\caption{Quantitative evaluation on DFC2019 sites. %Comparing Step 1 and Step 2 highlights the geometric %refinement achieved through our Diffusion-based %approach.}
%\label{tab:mae_buildings}
%\small
%\setlength{\tabcolsep}{3pt}
%\begin{tabular*}{\textwidth}{@{\extracolsep{\fill}} %lcccccccccccc @{}}
%\toprule
%\multirow{2}{*}{\textbf{Method}} & \multicolumn{7}{c}{\textbf{DFC2019 JAX}} & \multicolumn{3}{c}{\textbf{DFC2019 OMA}} & \multirow{2}{*}{\textbf{Mean}} \\
%\cmidrule(lr){2-8} \cmidrule(lr){9-11}
%& 004 & 068 & 214 & 260 & 168 & 251 & 280 & 203 & 212 & 315 & \\
%\midrule
%\multicolumn{12}{l}{\textbf{Buildings-only}} \\
%\midrule
%ASP & 1.09 & 1.36 & 2.71 & 1.70 & 1.94 & 1.92 & 2.44 & 0.81 & 4.04 & 1.50 & 1.95 \\
%SAT-NGP & 1.82 & 1.81 & 4.29 & 6.33 & 3.59 & 1.65 & 4.31 & 3.08 & 4.13 & 4.75 & 3.58 \\
%EOGS & 1.53 & 0.93 & \textbf{0.97} & 1.04 & 0.98 & 0.87 & 2.43 & 1.61 & 3.68 & 2.64 & 1.67 \\
%Skyfall-GS & 1.29 & 1.32 & 1.96 & 2.01 & 1.86 & 1.55 & 1.07 & 1.47 & 2.33 & 1.62 & 1.65 \\
%\midrule
%Ours (Step 1) & 0.71 & 0.96 & 1.04 & 0.92 & \textbf{0.96} & 0.81 & 1.06 & 0.49 & 1.22 & 0.86 & 0.90 \\
%\textbf{Ours (Step 2)} & \textbf{0.69} & \textbf{0.92} & 1.04 & \textbf{0.84} & 1.01 & \textbf{0.75} & \textbf{1.02} & \textbf{0.48} & \textbf{1.09} & \textbf{0.83} & \textbf{0.87} \\
%\bottomrule
%\end{tabular*}
%\end{table}

\subsubsection{Impact of shadow casting}

\begin{figure*}[t]
    \centering
    \small
    
    \setlength{\abovecaptionskip}{2pt}
    \setlength{\belowcaptionskip}{-5pt}
    \renewcommand{\arraystretch}{0.9}
    \setlength{\tabcolsep}{2pt}
    
    \def\mycolw{0.23\linewidth}
    
    \begin{tabular}{c@{\hspace{2pt}}cc|cc}
    
    & \multicolumn{2}{c}{\textbf{Rendered Albedo}} 
    & \multicolumn{2}{c}{\textbf{DSM (Height Map)}} \\
    
    & \footnotesize w/o Shadow
    & \footnotesize w/ Shadow
    & \footnotesize w/o Shadow
    & \footnotesize w/ Shadow \\
    
    \rule{0pt}{3ex} 
    
    \rotatebox{90}{\small JAX-068} &
    \includegraphics[width=\mycolw]{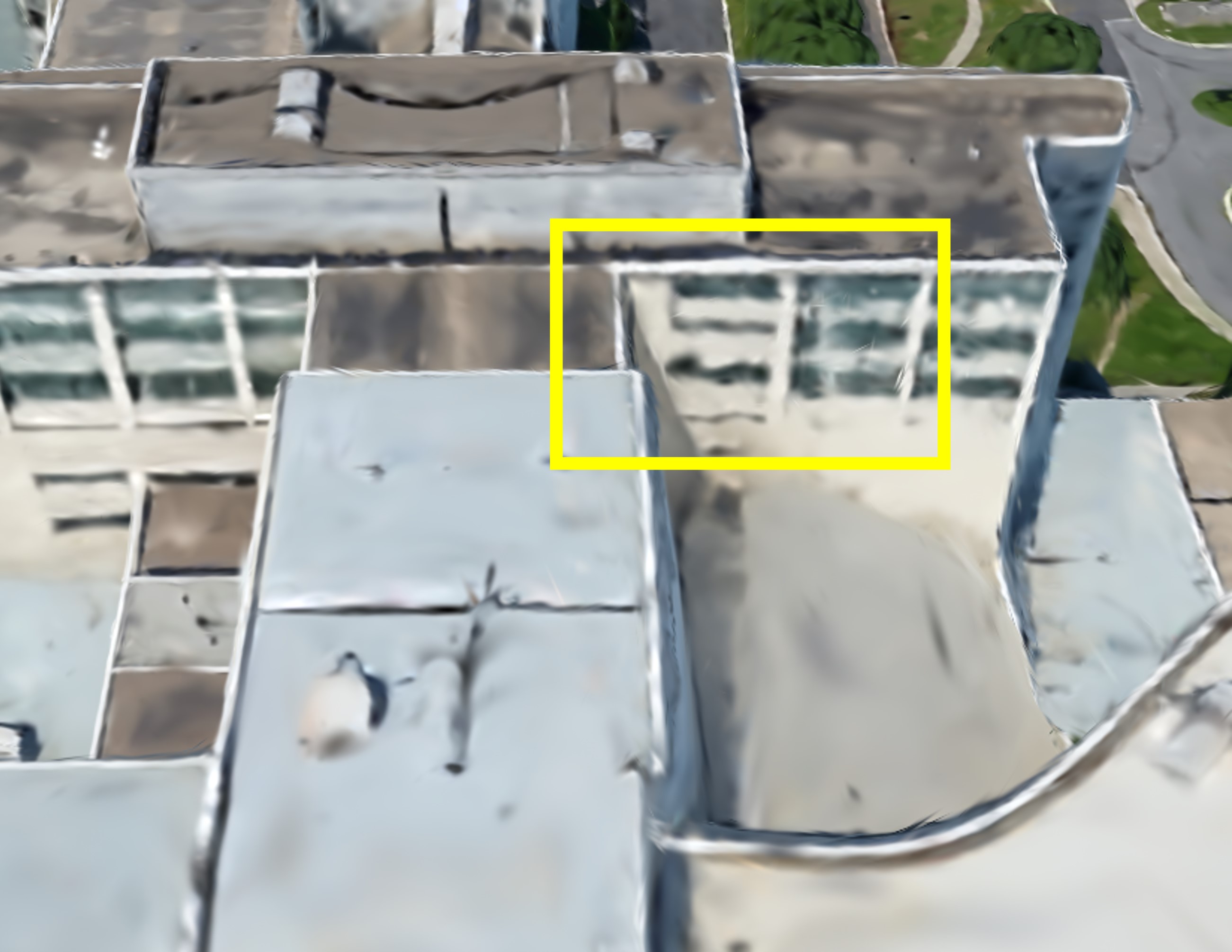} &
    \includegraphics[width=\mycolw]{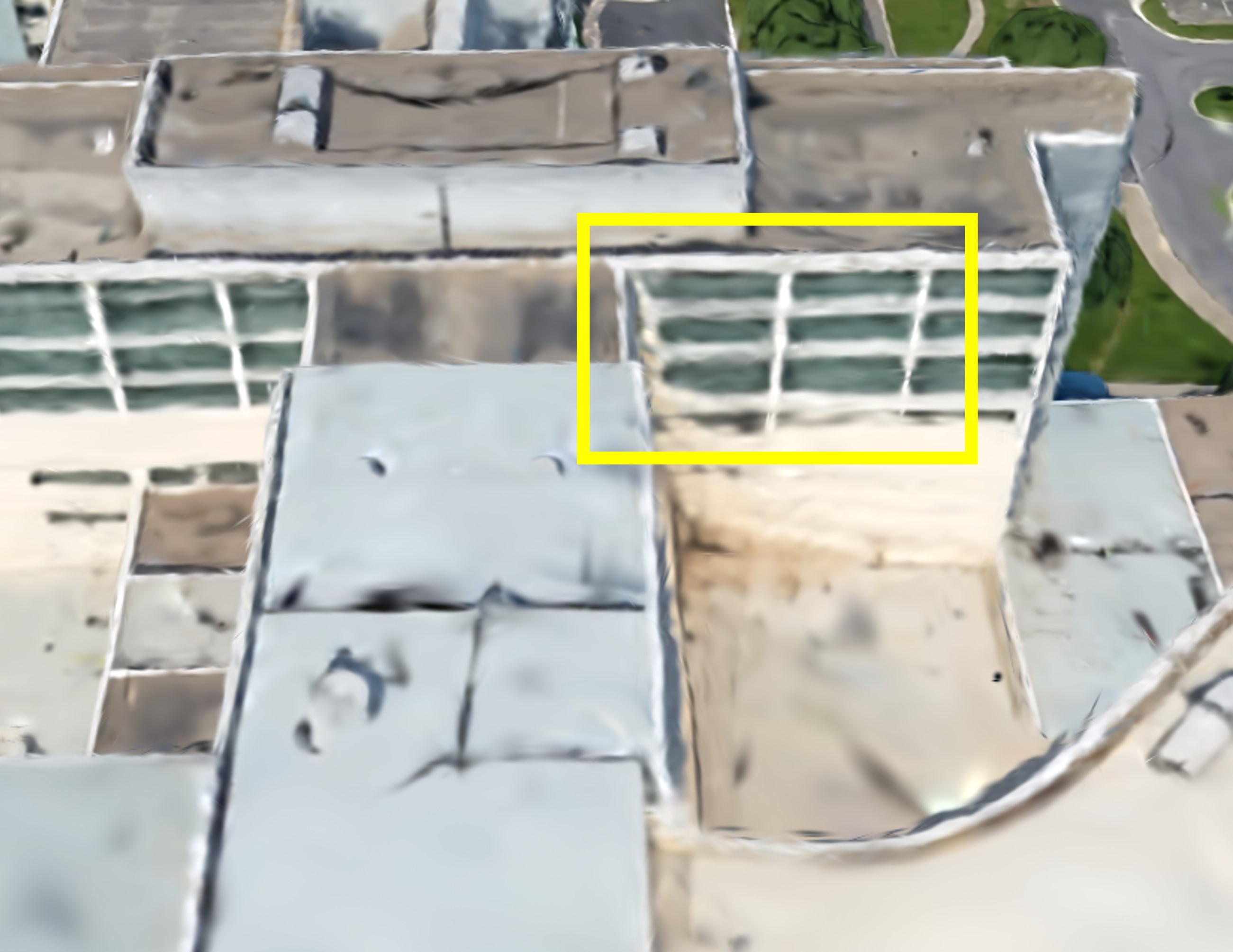} &
    \includegraphics[width=\mycolw]{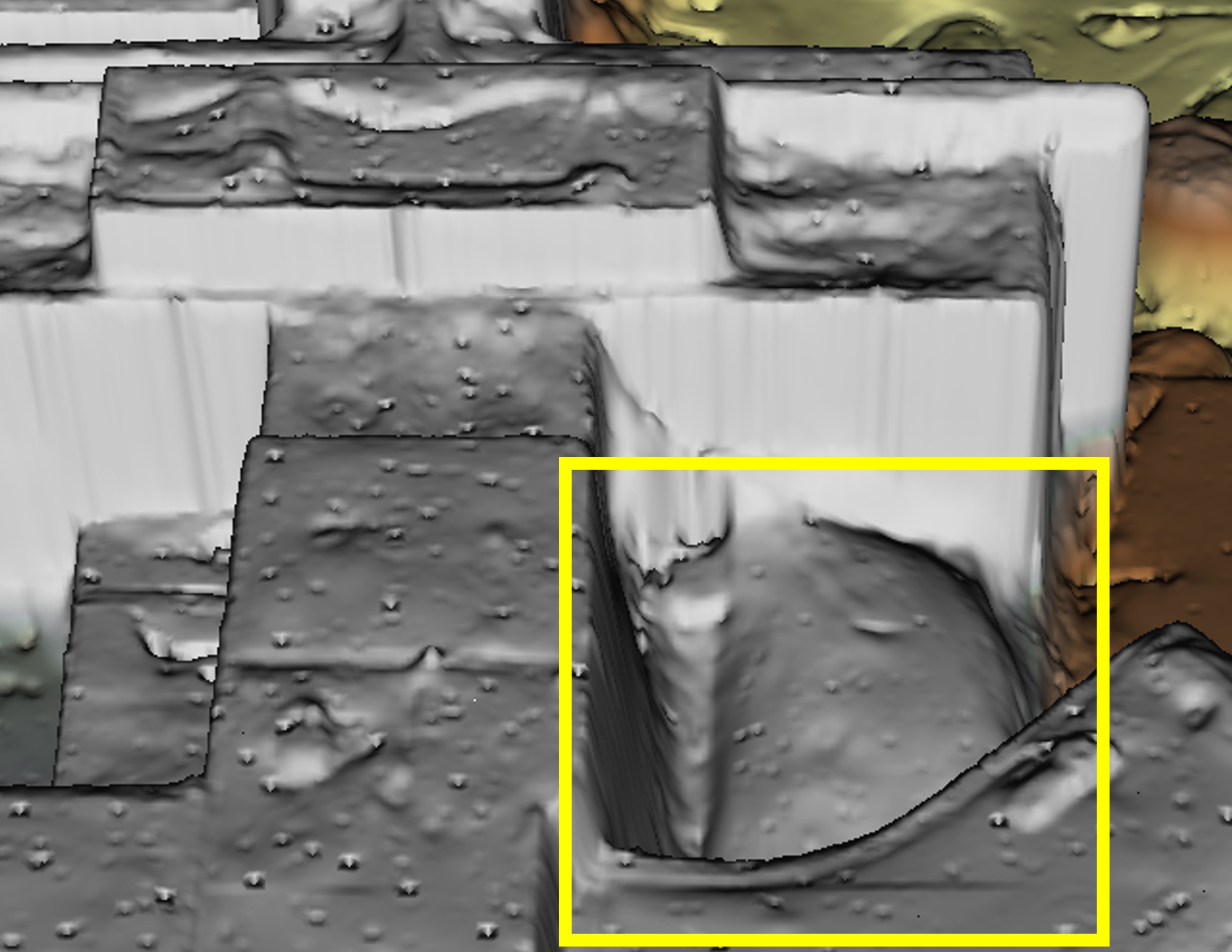} &
    \includegraphics[width=\mycolw]{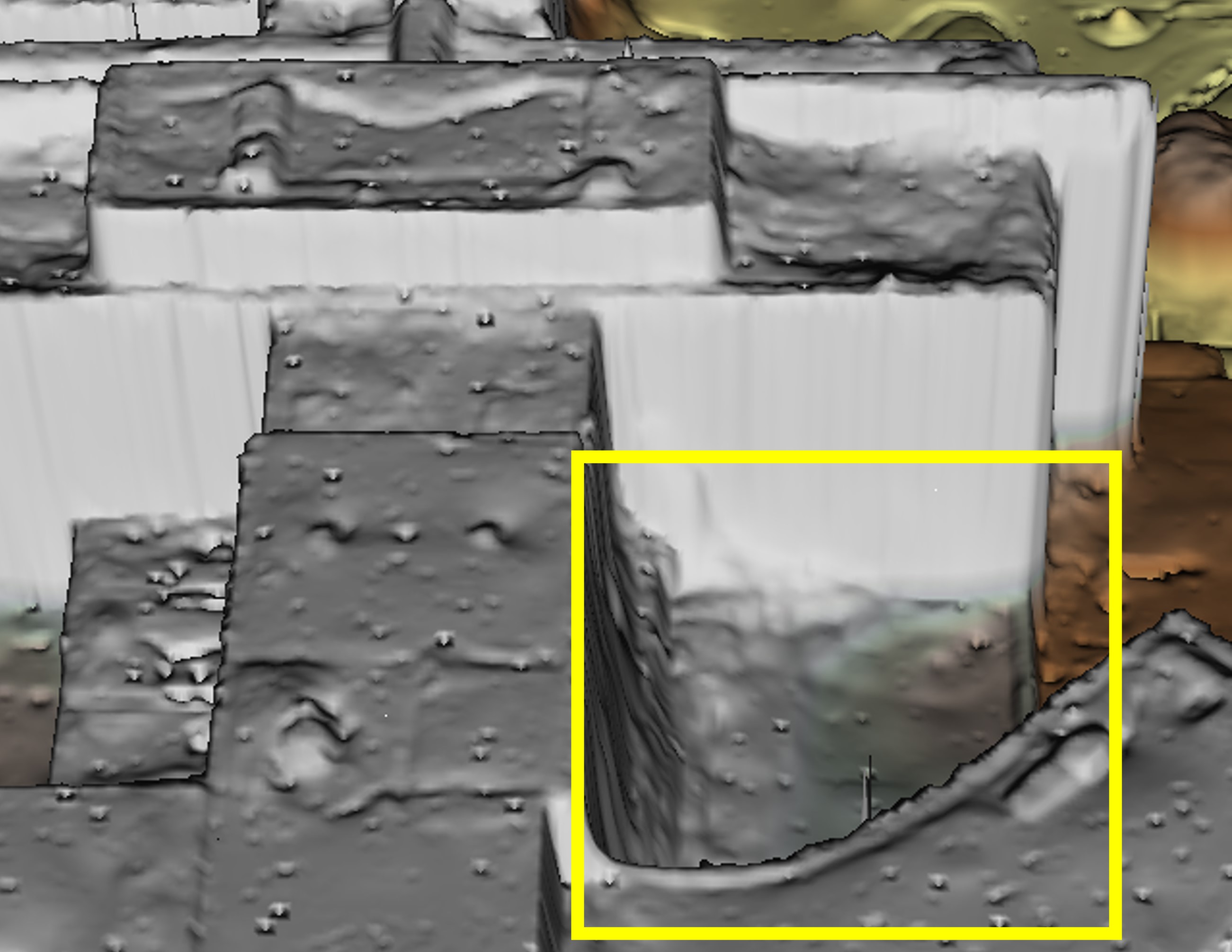} \\
    
    \rotatebox{90}{\small JAX-168} &
    \includegraphics[width=\mycolw]{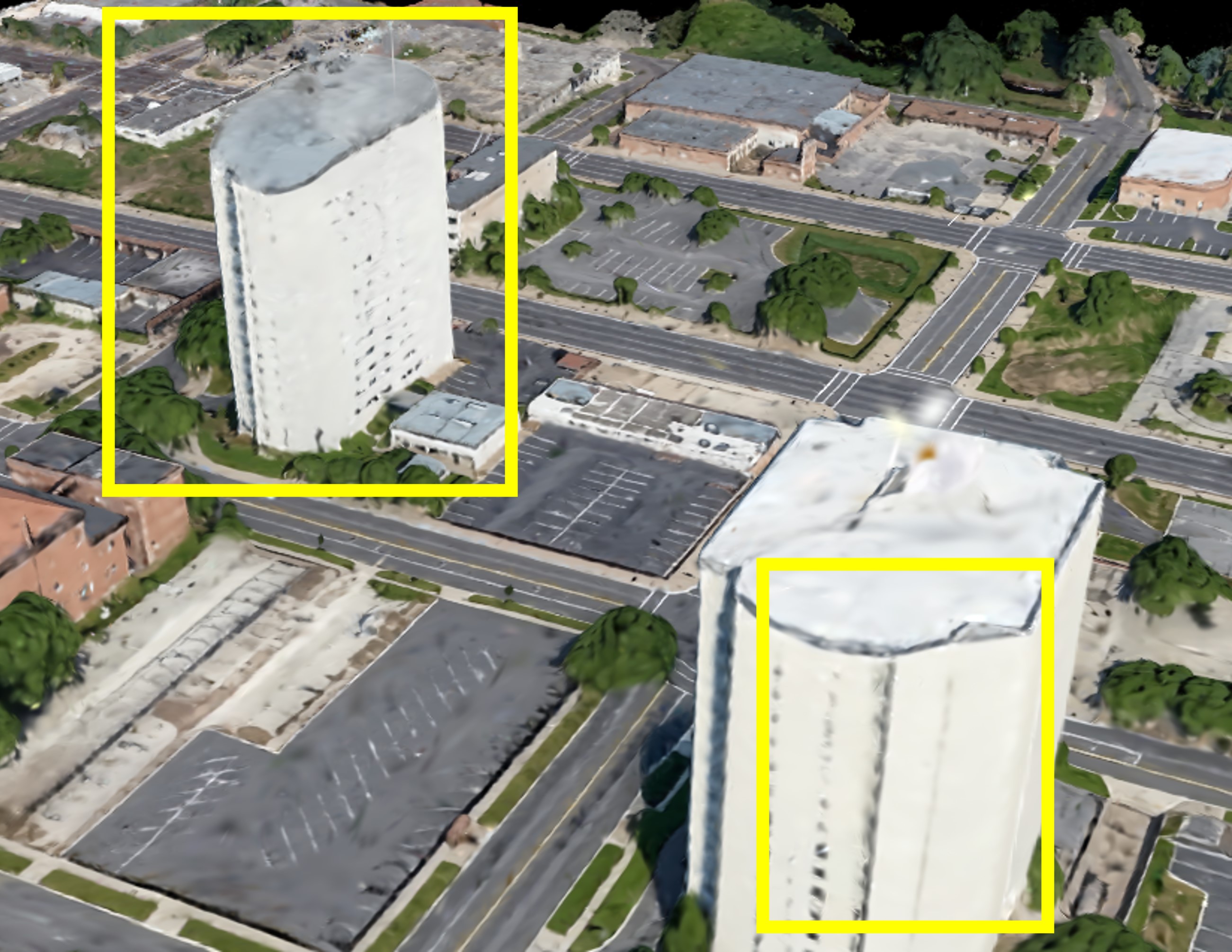} &
    \includegraphics[width=\mycolw]{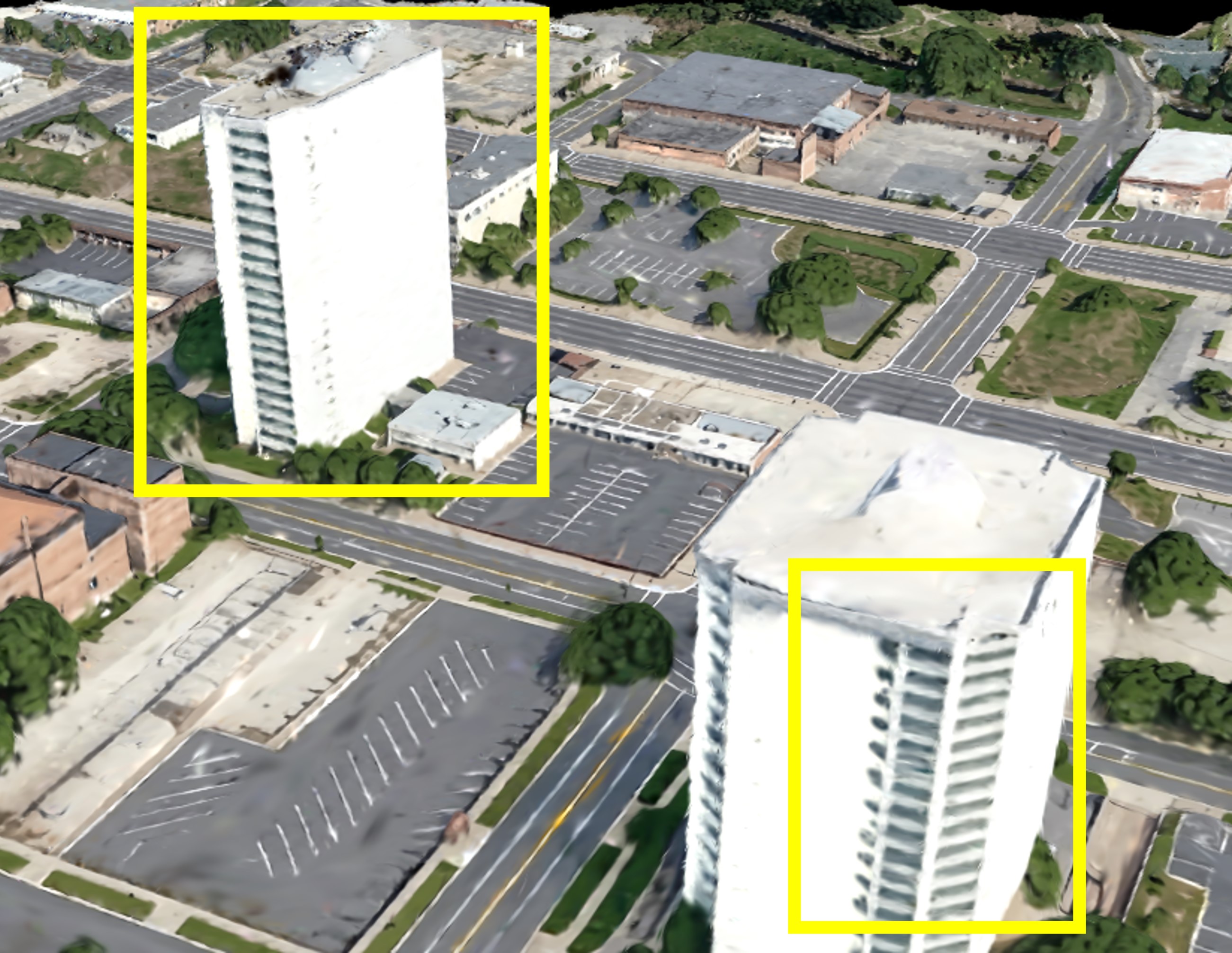} &
    \includegraphics[width=\mycolw]{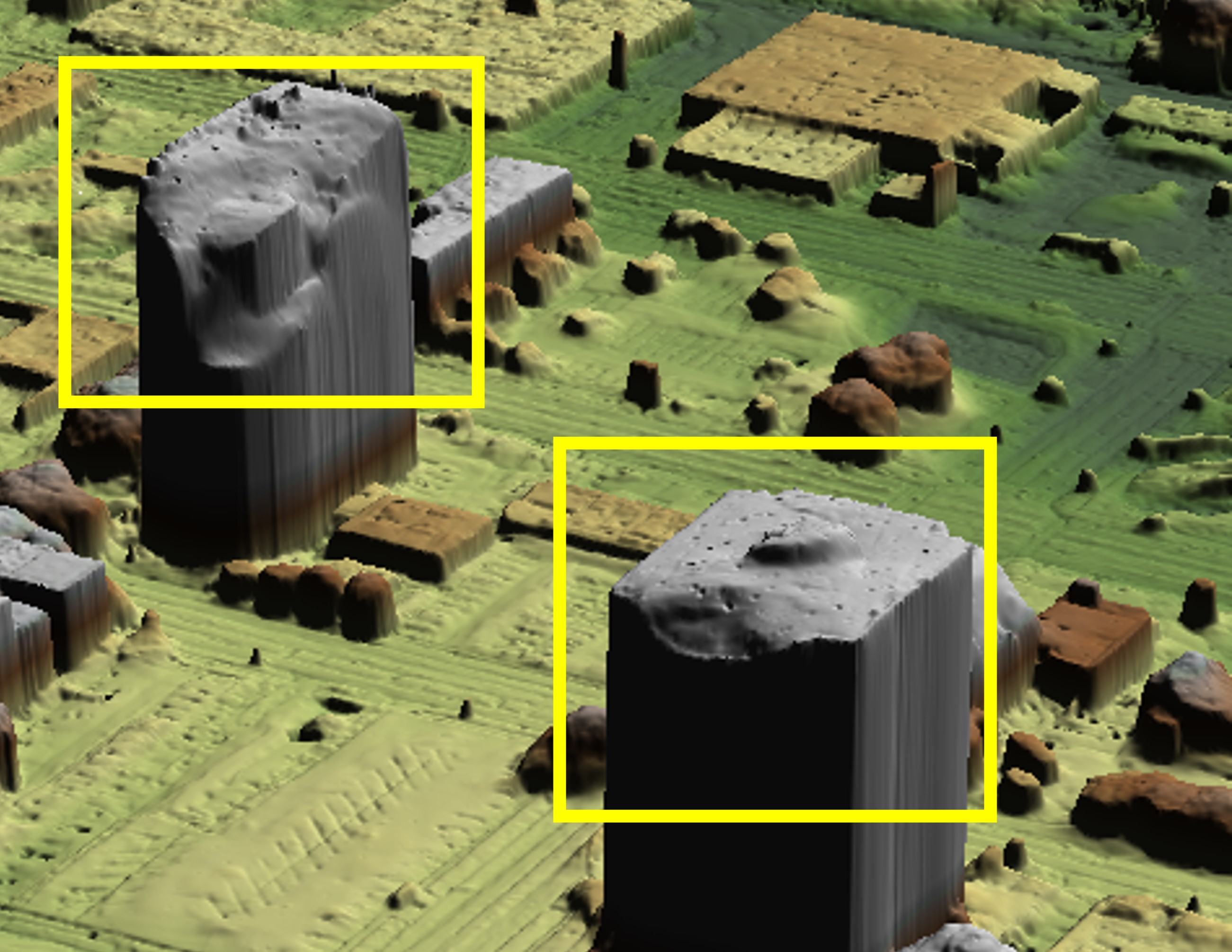} &
    \includegraphics[width=\mycolw]{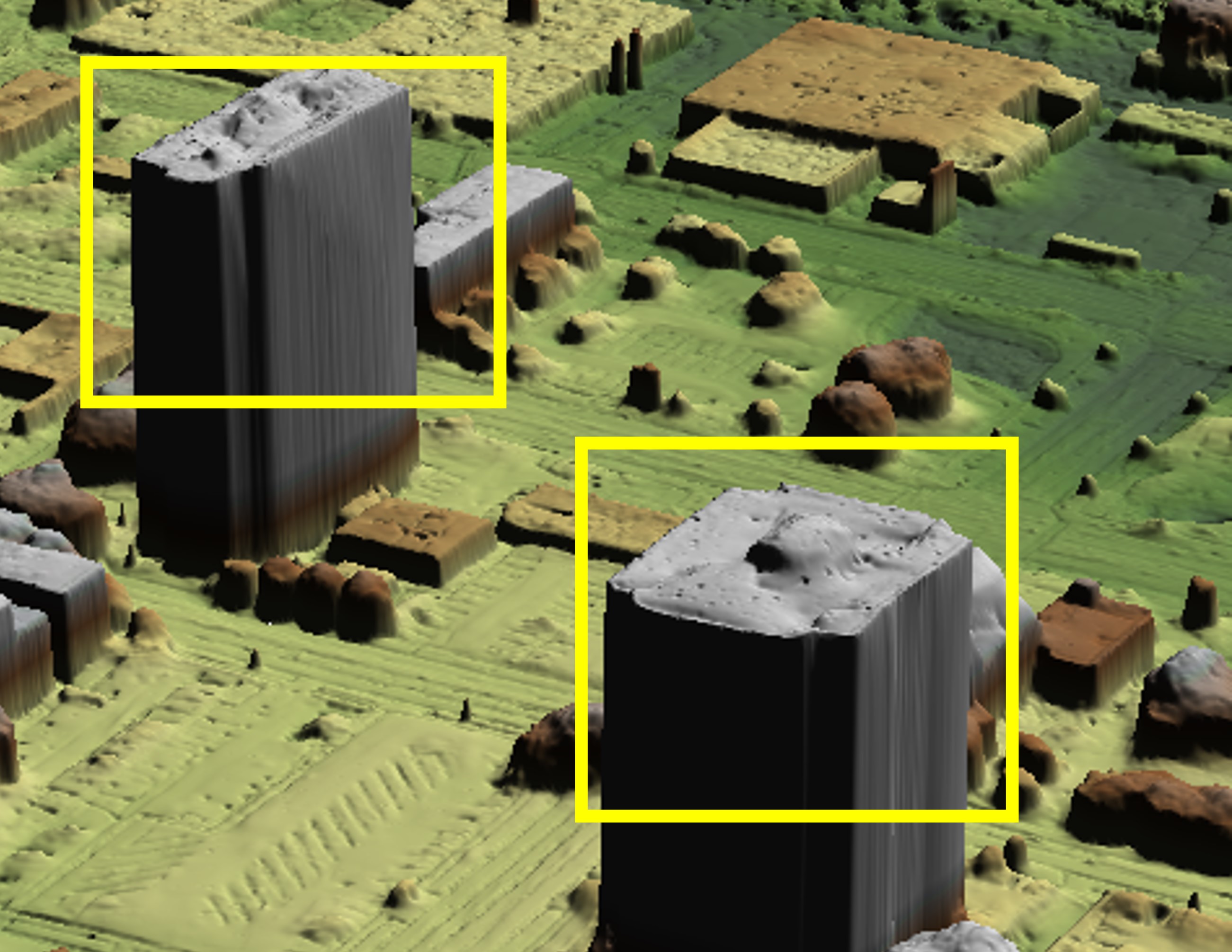} \\
    
    \rotatebox{90}{\small JAX-214} &
    \includegraphics[width=\mycolw]{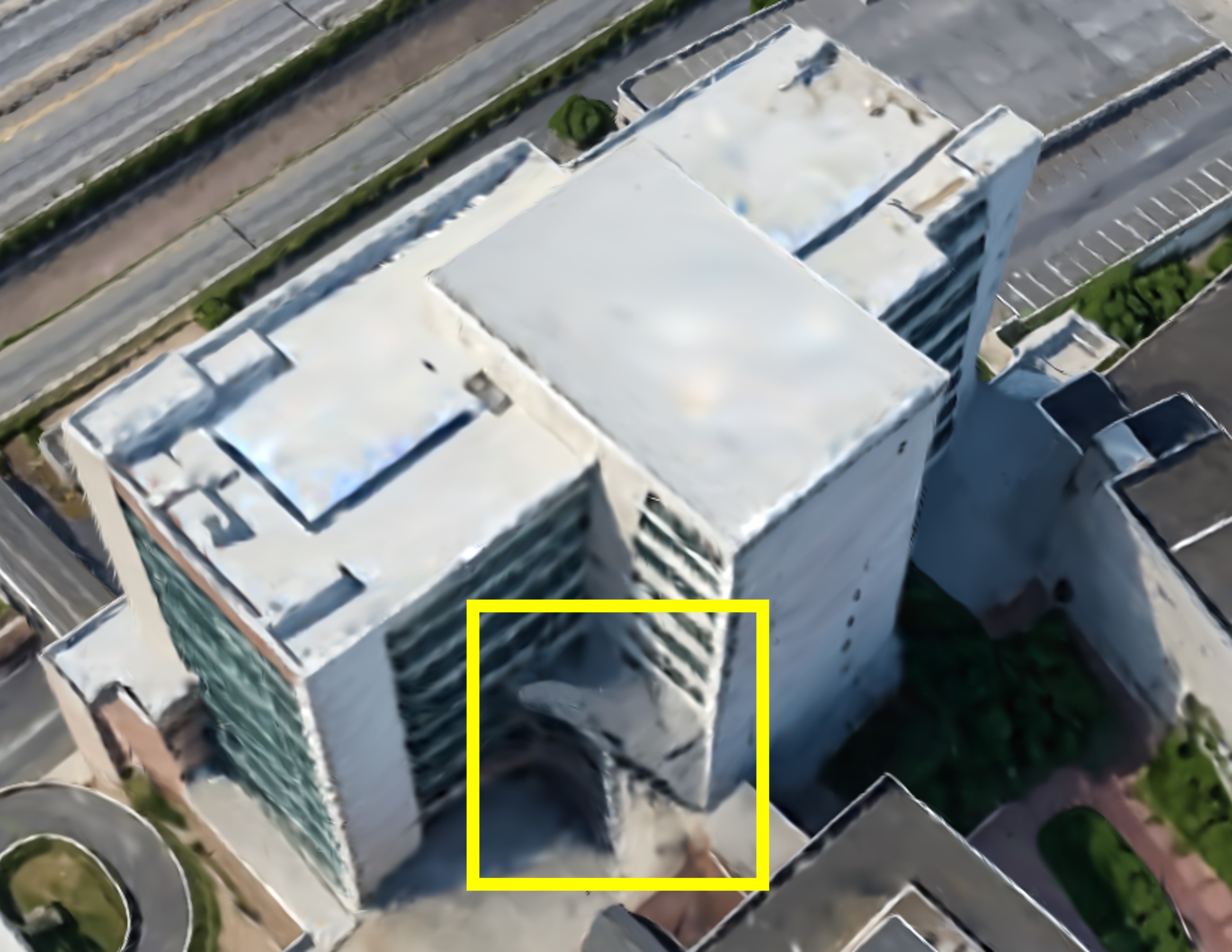} &
    \includegraphics[width=\mycolw]{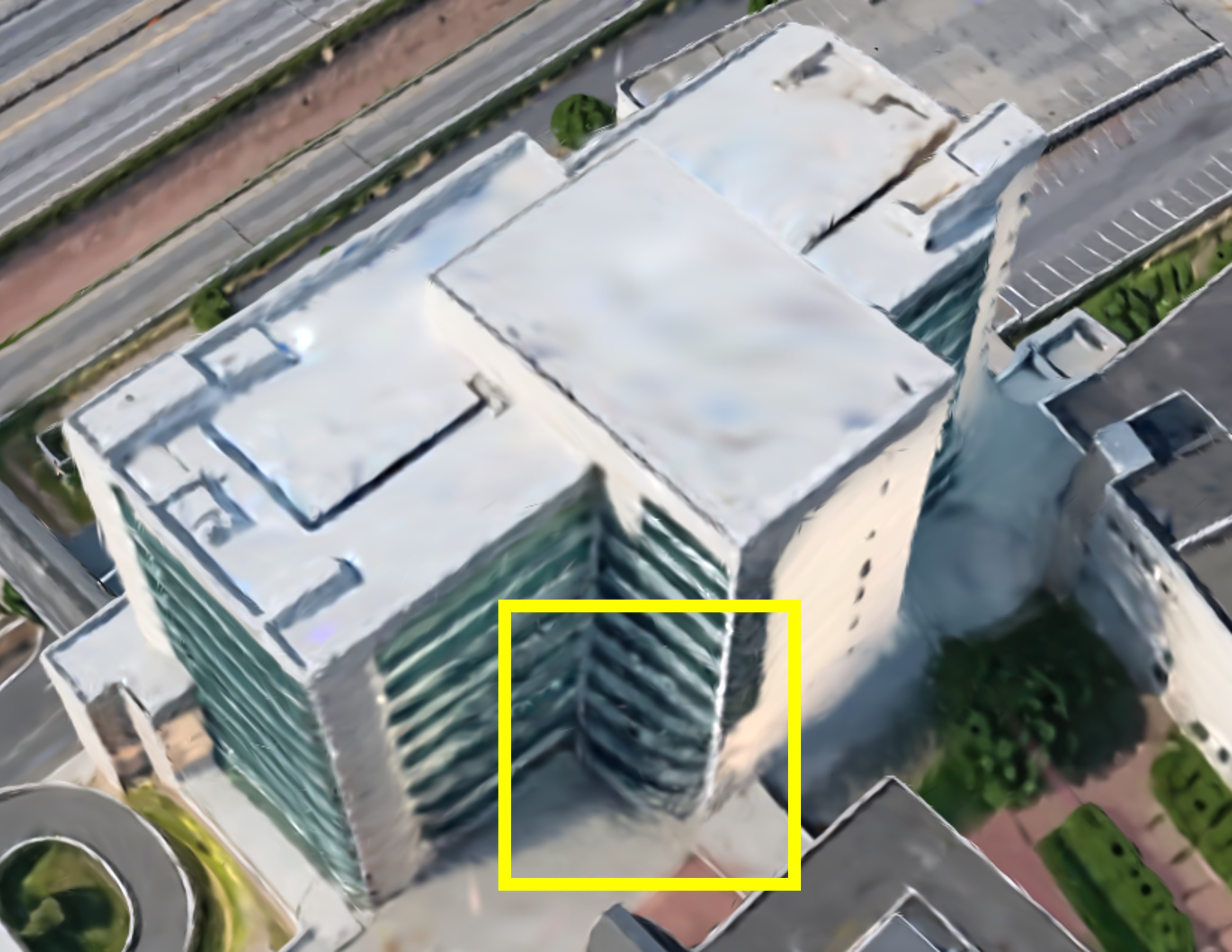} &
    \includegraphics[width=\mycolw]{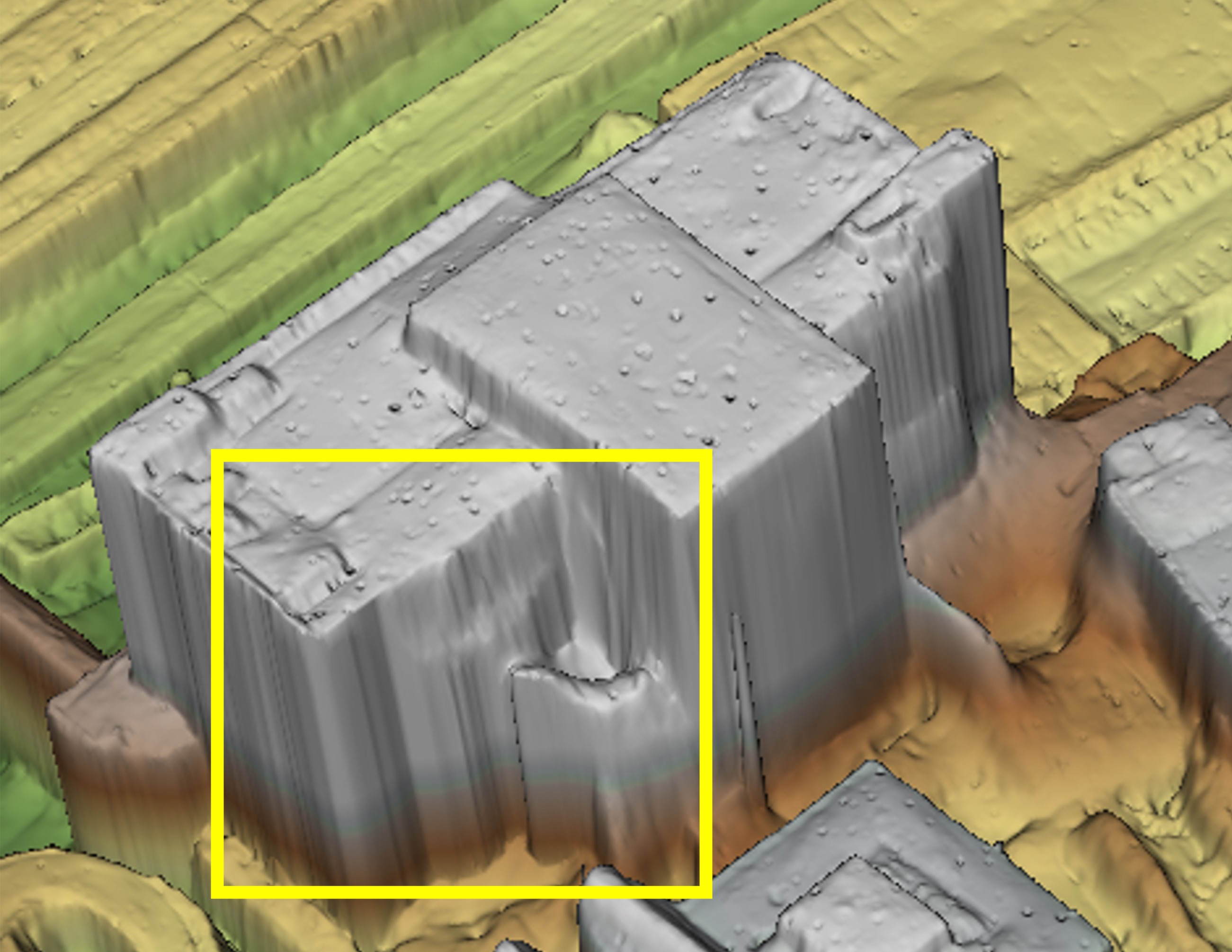} &
    \includegraphics[width=\mycolw]{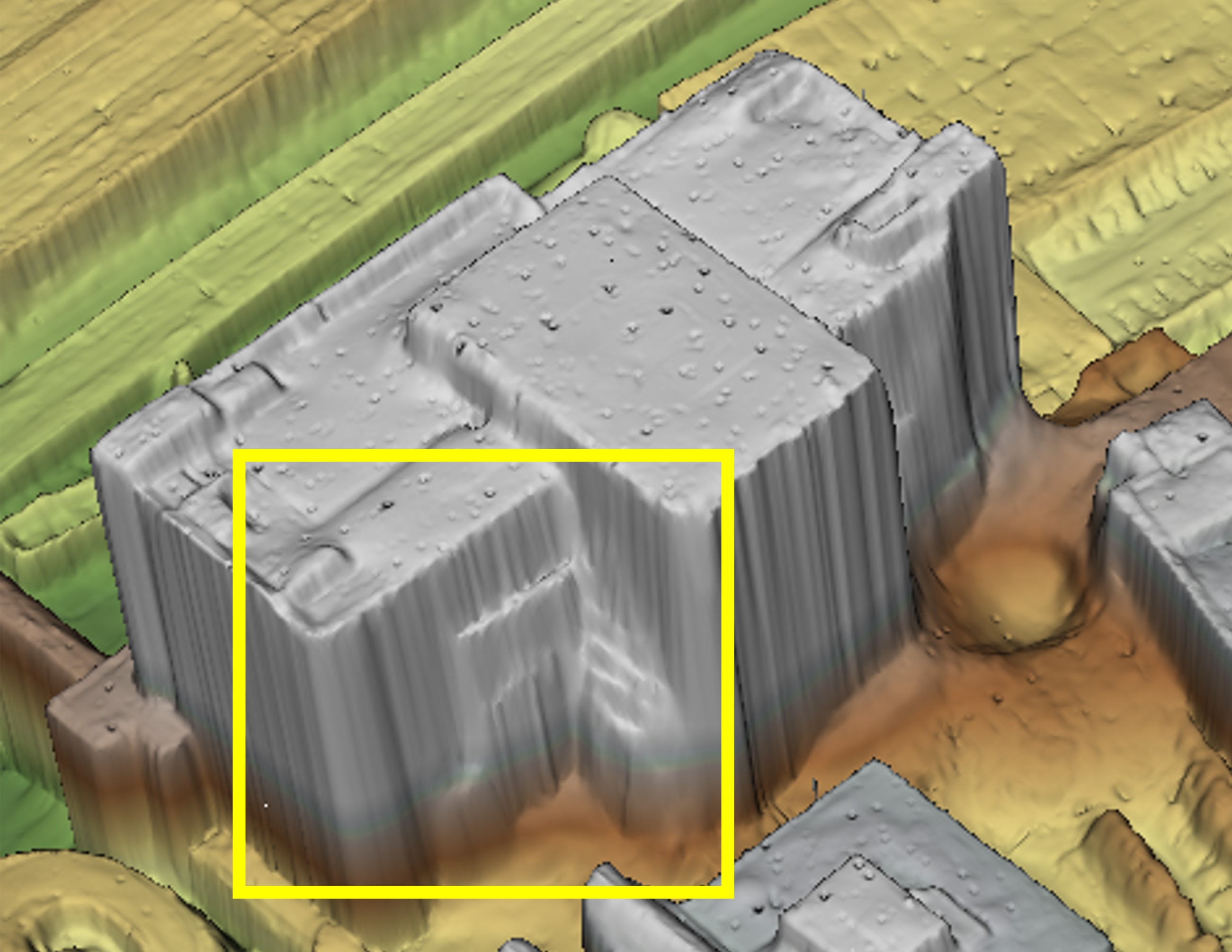} \\
    
    \rotatebox{90}{\small JAX-251} &
    \includegraphics[width=\mycolw]{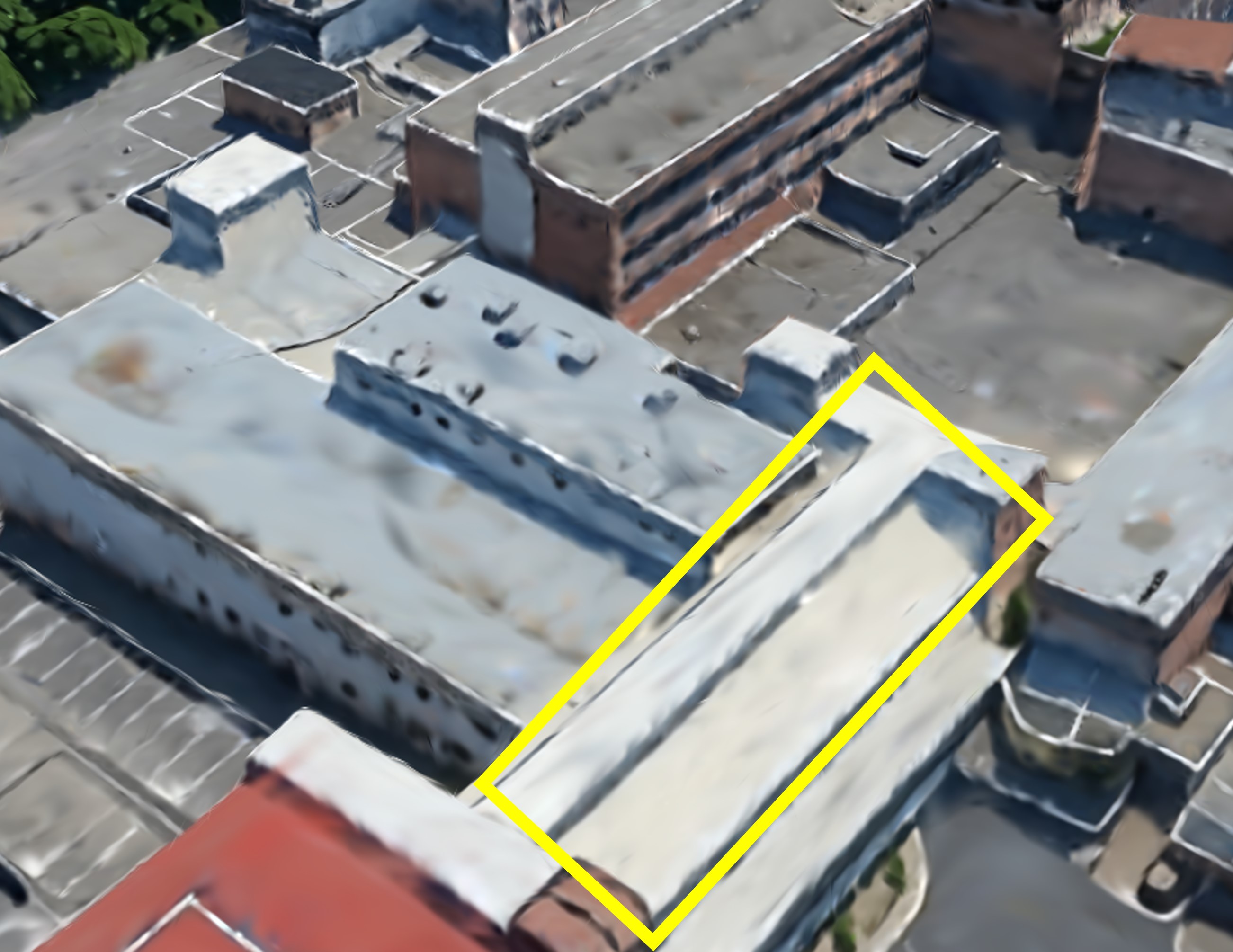} &
    \includegraphics[width=\mycolw]{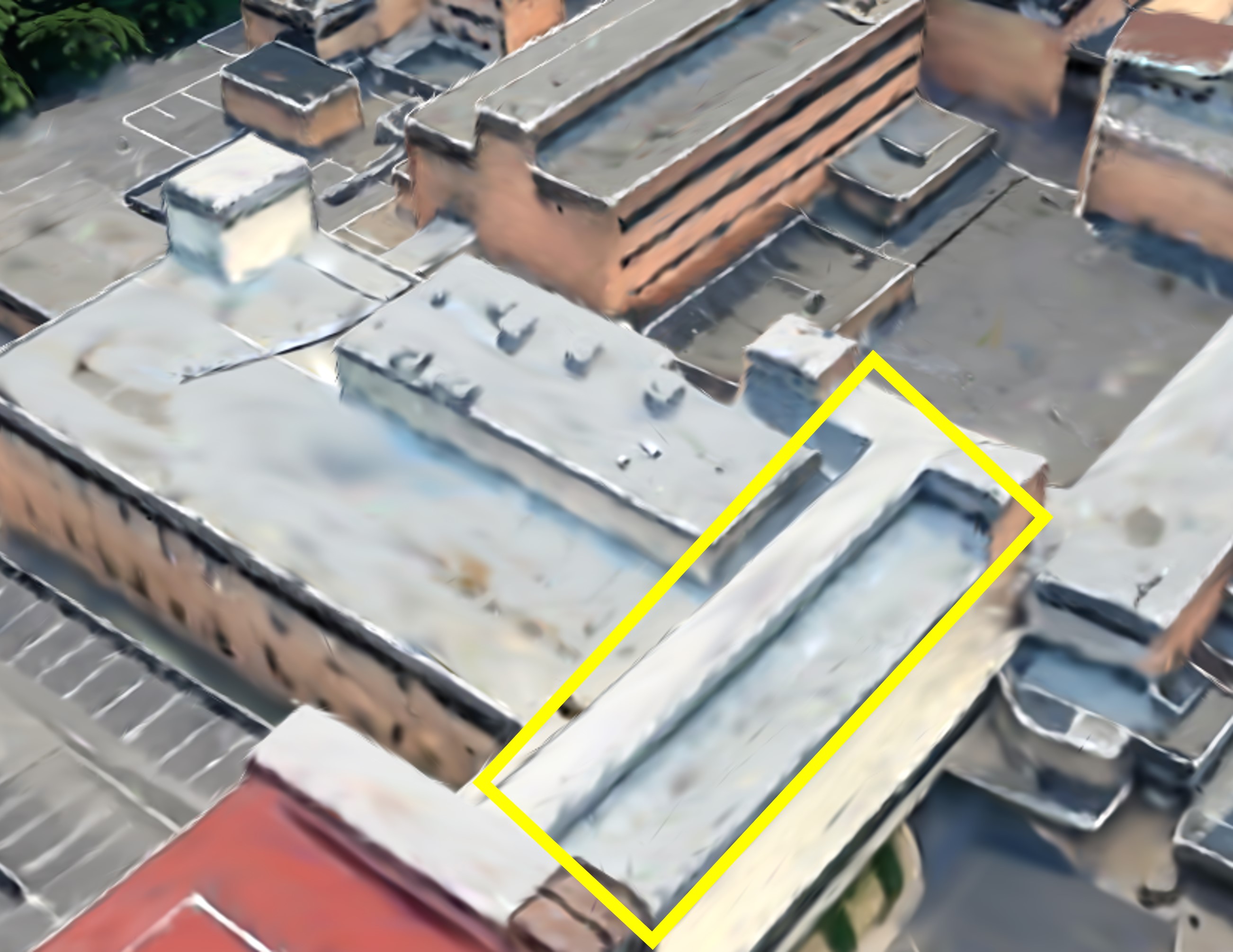} &
    \includegraphics[width=\mycolw]{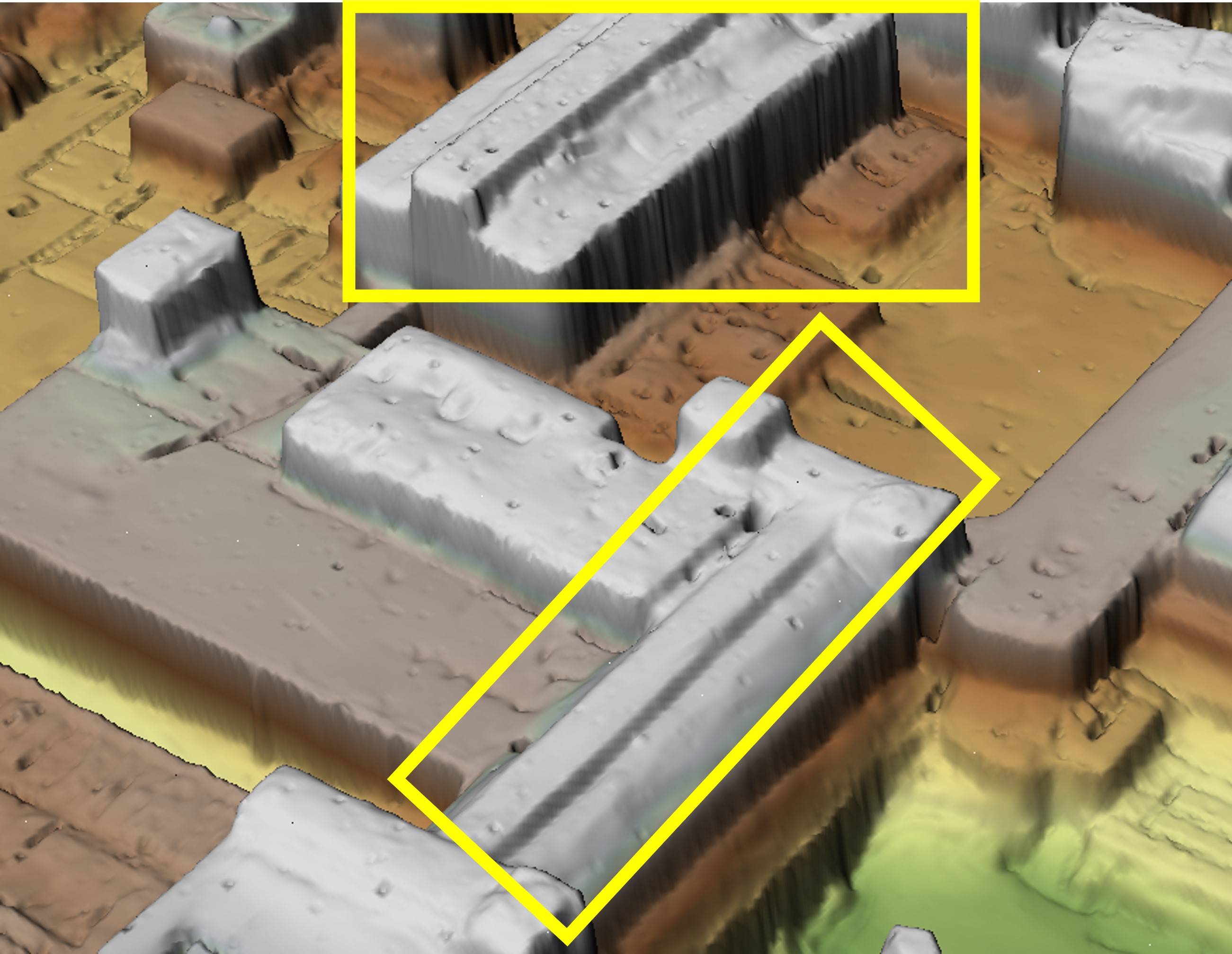} &
    \includegraphics[width=\mycolw]{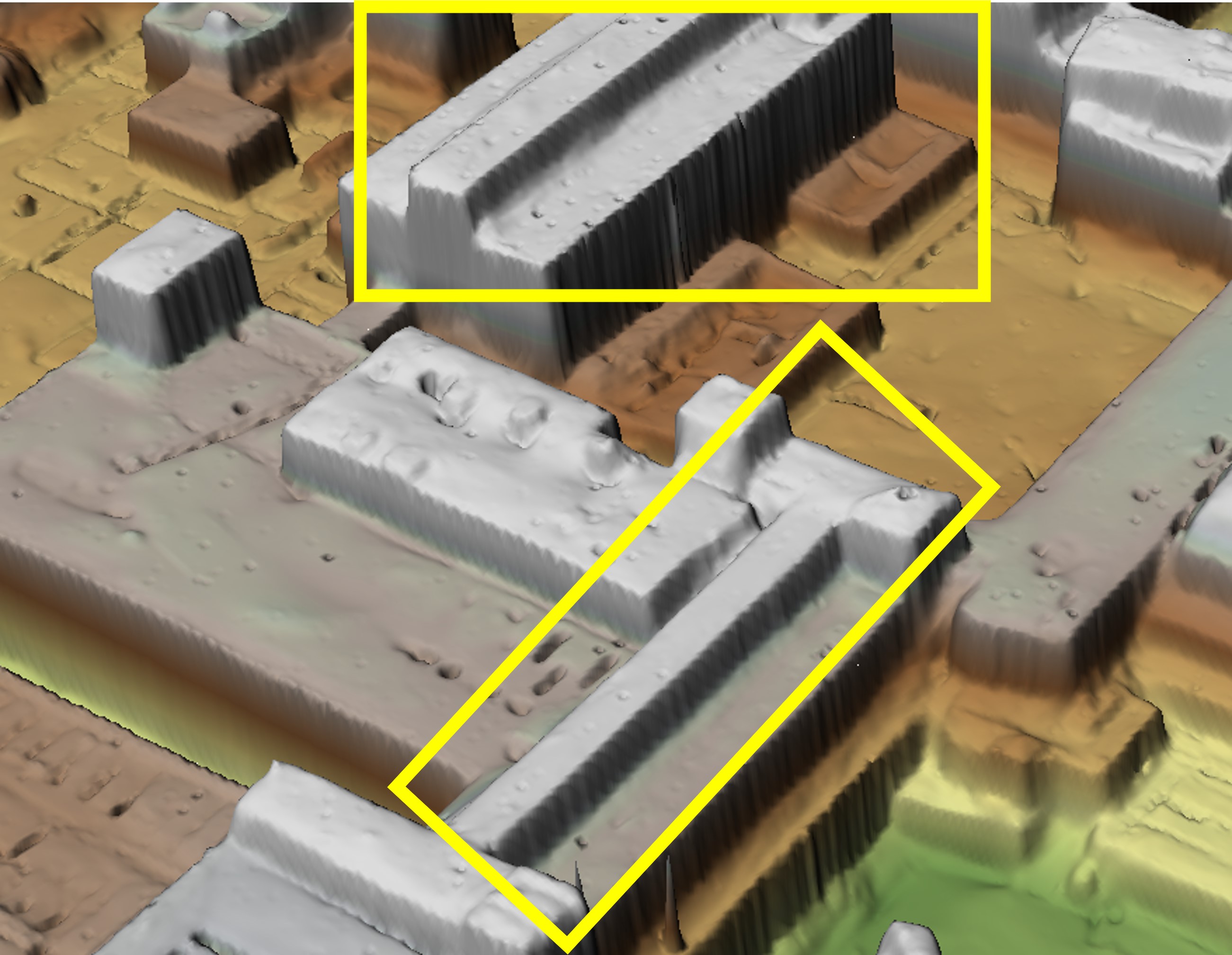} \\
    
    \end{tabular}
    
    \caption{Qualitative ablation of shadow casting across multiple JAX sites. Shadow casting imposes geometric guidance, helping Gaussians maintain geometric accuracy and visual fidelity.}
    \label{fig:shadow_ablation_jax}
\end{figure*}

We previously established that casting geometrically calculated shadows helps maintain geometric accuracy throughout the pipeline. To validate this, we conduct an ablation study on shadow casting (\Cref{fig:shadow_ablation_jax}). The results confirm that shadow casting preserves geometric accuracy relative to the generative refinement pipeline without shadow casting. In particular, the surface collapse observed when training without shadow casting is largely resolved.

A comparative analysis of geometric accuracy before and after the application of generative refinement (\Cref{tab:ablation_shadow}) further supports these findings. For our method, the mean geometric error ($\mathrm{MAE}_{reg}$) on the full scene remains nearly unchanged at 1.27m and 1.26m when the diffusion model is conditioned on albedo images ($\mathbf{R}_{albedo}$) instead of shadow-cast images. On building regions specifically, however, the degradation is more pronounced, increasing from 1.02m to 1.07m, suggesting that removing geometry-derived shadow structures particularly reduces the geometric cues available for refining facade and rooftop details. In comparison, Skyfall-GS exhibits a larger degradation on the full scene, from 1.85m to 1.93m, under the same diffusion-based refinement. When the diffusion model is instead conditioned on shadow-cast images, our method recovers a mean geometric error of 1.23m on the full scene and 1.02m on building regions, demonstrating that shadow-guided conditioning helps retain geometry-consistent appearance refinement during Gaussian optimization.

\begin{table}[htbp]
\centering
\caption{Comparative analysis of geometric accuracy during generative refinement. The results demonstrate that while generative refinement typically leads to geometric degradation, our framework with shadow casting maintains geometric accuracy.}
\label{tab:ablation_shadow}
\small
\setlength{\tabcolsep}{4pt}
\resizebox{\linewidth}{!}{
\begin{tabular}{lccccccccc}
\toprule
\multirow{2}{*}{\textbf{Method}} & \multicolumn{4}{c}{\textbf{Full Scene}} & & \multicolumn{4}{c}{\textbf{Buildings}} \\
\cmidrule(lr){2-5} \cmidrule(lr){7-10}
& \textbf{JAX} & \textbf{OMA} & \textbf{IARPA} & \textbf{Mean} & & \textbf{JAX} & \textbf{OMA} & \textbf{IARPA} & \textbf{Mean} \\
\midrule
Skyfall-GS (w/o diffusion) & 1.79 & 1.21 & 2.55 & 1.85 & & 1.35 & 1.68 & 2.14 & 1.72 \\
Skyfall-GS (w/ diffusion)  & 1.86 & 1.39 & 2.55 & 1.93 & & 1.56 & 1.80 & 2.05 & 1.80 \\
\midrule
Ours (w/o diffusion)       & 1.27 & 0.65 & 1.89 & 1.27 & & 0.92 & 0.86 & \textbf{1.29} & 1.02 \\
Ours (w/o shadow)          & 1.32 & 0.65 & 1.82 & 1.26 & & 0.98 & 0.85 & 1.37 & 1.07 \\
\textbf{Ours (w/ shadow)}  & \textbf{1.23} & \textbf{0.64} & \textbf{1.82} & \textbf{1.23} & & \textbf{0.91} & \textbf{0.81} & 1.35 & \textbf{1.02} \\
\bottomrule
\end{tabular}
}
\end{table}

In addition to geometric accuracy, shadow casting contributes to enhanced visual fidelity. By conditioning on shadow-cast images, the diffusion model generates outputs that align more closely with the radiometric distribution of real-world satellite imagery. As indicated in \Cref{tab:ours_shadow_ablation}, random shadow sampling improves visual fidelity compared to results using fixed or no shadow casting. Specifically, random shadow sampling achieves the lowest FID-CLIP (19.50) and CMMD (1.681), confirming the effectiveness of our shadow sampling strategy.

\begin{table}[htbp]
\centering
\caption{Quantitative ablation of shadow casting strategies during the shadow-guided generative refinement stage. Random shadow sampling efficiently models diverse solar configurations, providing the most robust performance across both distributional and pixel-level metrics.}
\label{tab:ours_shadow_ablation}
\resizebox{\linewidth}{!}{%
\begin{tabular}{lcccccc}
\toprule
\multirow{2}{*}{\textbf{Configuration}} & \multicolumn{2}{c}{\textbf{Distributional}} & \multicolumn{3}{c}{\textbf{Pixel-level}} & \textbf{Geometric} \\
\cmidrule(lr){2-3} \cmidrule(lr){4-6} \cmidrule(lr){7-7}
& FID-CLIP $\downarrow$ & CMMD $\downarrow$ & PSNR $\uparrow$ & CW-SSIM $\uparrow$ & LPIPS $\downarrow$ & MAE$_{reg}$ $\downarrow$ \\
\midrule
w/o shadow      & 19.57 & 1.717 & 11.74 & 0.373 & 0.621 & 1.32 \\
w/ fixed shadow & 19.61 & 1.695 & 12.23 & 0.413 & 0.614 & 1.28 \\
w/ random shadow & \textbf{19.50} & \textbf{1.681} & \textbf{12.26} & \textbf{0.414} & \textbf{0.611} & \textbf{1.25} \\
\bottomrule
\end{tabular}%
}
\end{table}
%Furthermore, a comparative analysis of geometric accuracy before and after the application of the generative refinement (\Cref{tab:ablation_shadow}) reveals a distinct performance gap between different refinement strategies. For our method, the geometric error ($\mathrm{MAE}_{reg}$) marginally increased from 1.02m to 1.07m when shadows were omitted \RQ{what do you mean by the shadow was omitted? you do not model shadow in the process? or you do not use the shaded image as the input of the diffusion model?}, whereas Skyfall-GS exhibited a more substantial degradation from 1.72m to 1.80m. However, upon incorporating the diffusion model with shadow casting, our method maintained a geometric accuracy of 1.02m. This suggests that the explicit modeling of shadows provides a structural guidance.  \RQ{your langguage lacks accuracy in general, you need to describe with clarity}

% 0.91 

%However, an efficiency trade-off was observed\RQ{what kind of trade-off is "efficient"??}. The final number of Gaussians increased slightly by approximately 1\%, and the average training time increased from 42.15 to 49.39 minutes \RQ{do you really need two digits after the decimal when talking about minutes?}. We argue that this increase in computational time (roughly 7 minutes) is marginal when weighed against the gains in overall geometric consistency.
\section{Discussion}
\label{Discussion}

%\RQ{this is a complex pipeline with many things, a single section of ablation study is not sufficient, you should also mention how the quality of different process may impact the result, for example, the quality of the depth in depth supervision, just to give you an example, there are a lot of shadwow related stuff going on and ofc with parameters, there will be a lot of puzzles, I think you should generally discuss about them and how sentitive some of the parameters and processes. we talk about significant ones, since we are backed up by the released codes, people can seek through the codes for additional unclarity as they navigate through.}
Throughout the experiments, SatSplatDiff achieves state-of-the-art geometric accuracy and visual fidelity. The geometric optimization stage produces well-regularized surfaces, and the shadow-guided generative refinement stage improves visual fidelity on occluded surfaces while preserving geometric accuracy. In the following subsections, we discuss the further analysis on key components and demonstrate scalability to large-area reconstruction.

\subsection{Initialization sensitivity}
\begin{figure}[!h]
    \centering
    \begin{subfigure}{0.46\linewidth}
        \centering
        \includegraphics[width=\linewidth]{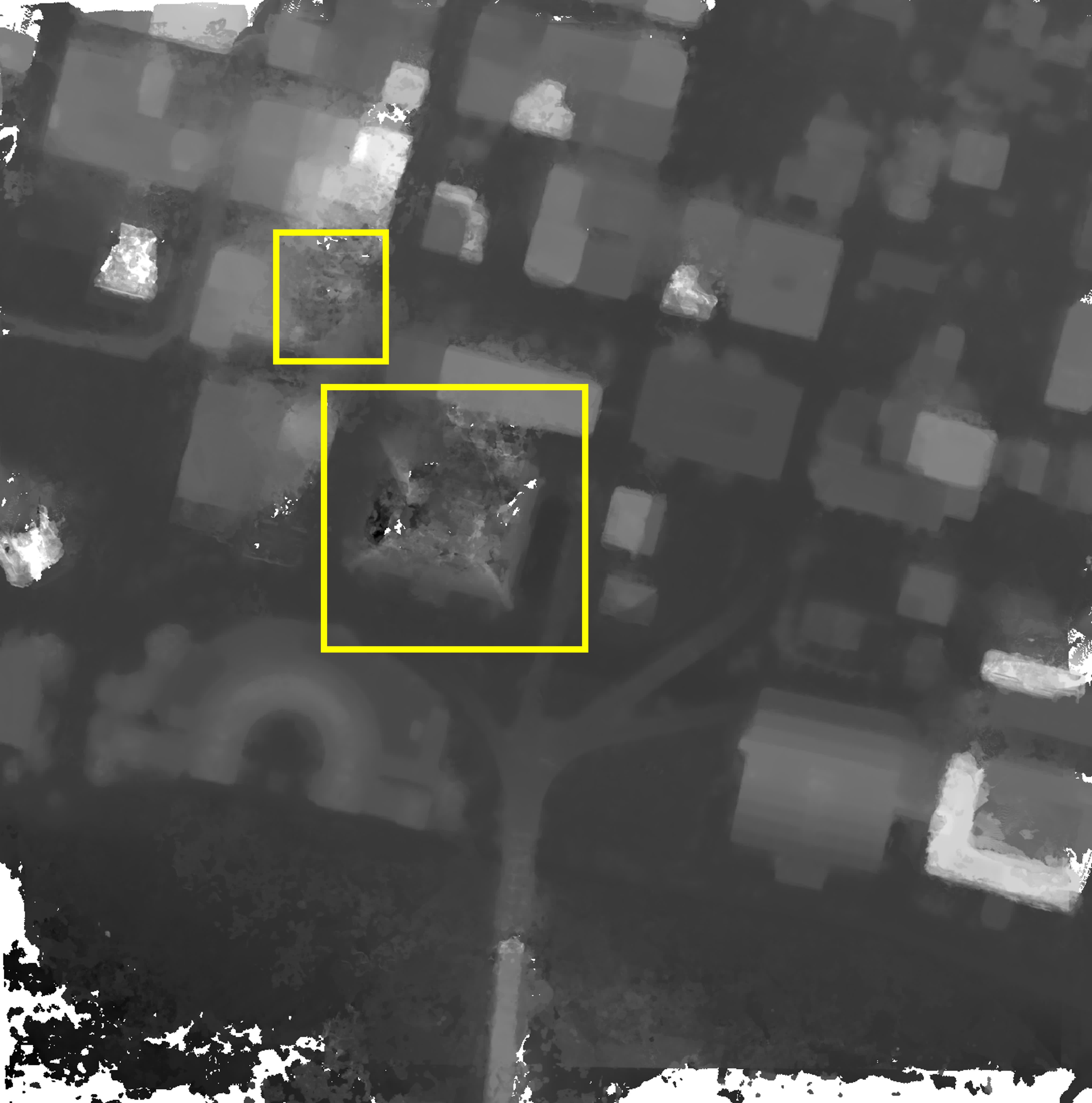} 
        \caption{Initialization DSM (failure)}
        \label{fig:init_asp}
    \end{subfigure}
    \hspace{0.02\linewidth}
    \begin{subfigure}{0.46\linewidth}
        \centering
        \includegraphics[width=\linewidth]{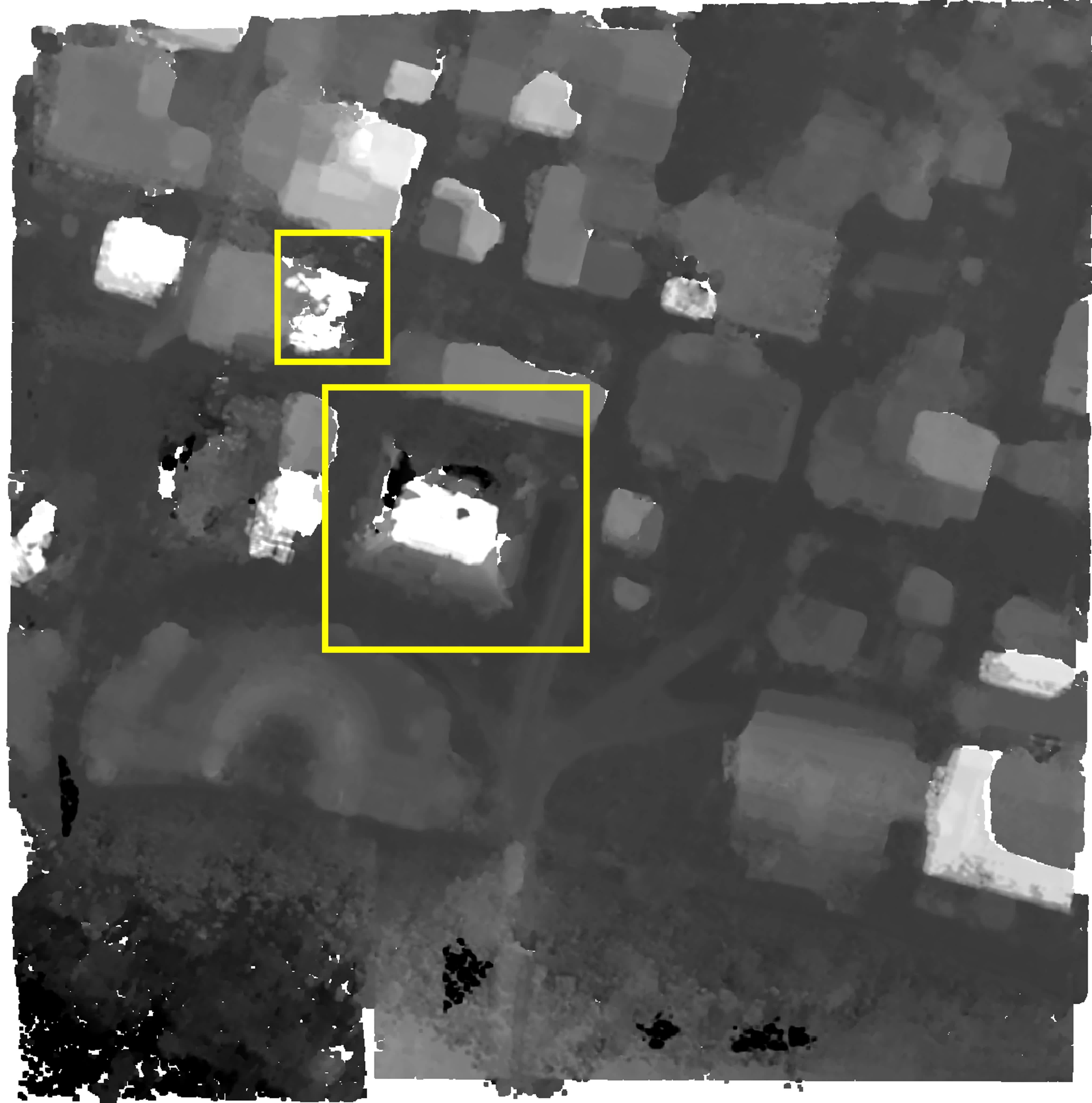} 
        \caption{Initialization DSM (partial failure)}
        \label{fig:init_s2p}
    \end{subfigure}
    
    \vspace{0.3cm}
    
    \begin{subfigure}{0.46\linewidth}
        \centering
        \includegraphics[width=\linewidth]{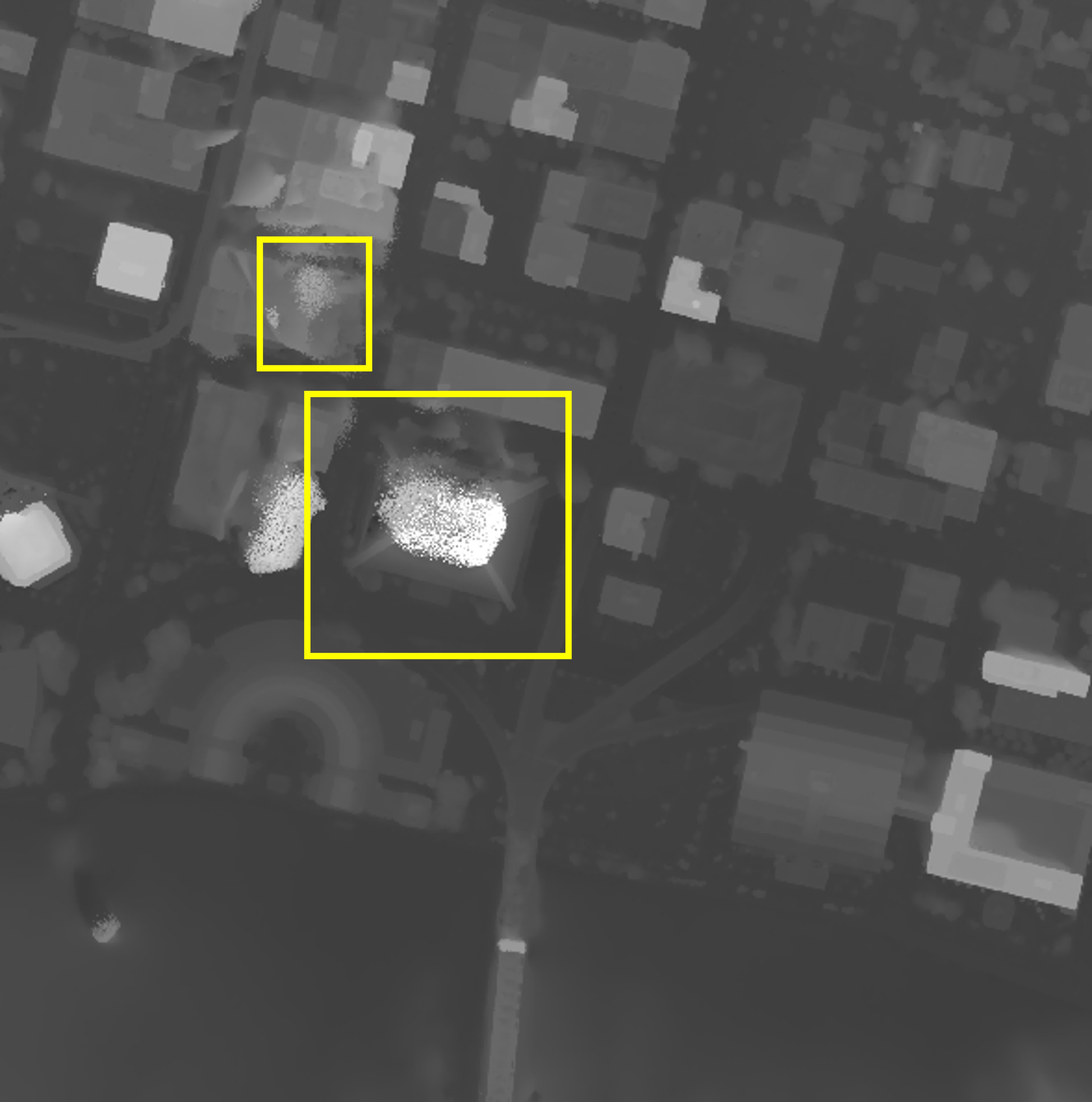} 
        \caption{Reconstructed DSM from (a)}
        \label{fig:init_asp_dsm}
    \end{subfigure}
    \hspace{0.02\linewidth}
    \begin{subfigure}{0.46\linewidth}
        \centering
        \includegraphics[width=\linewidth]{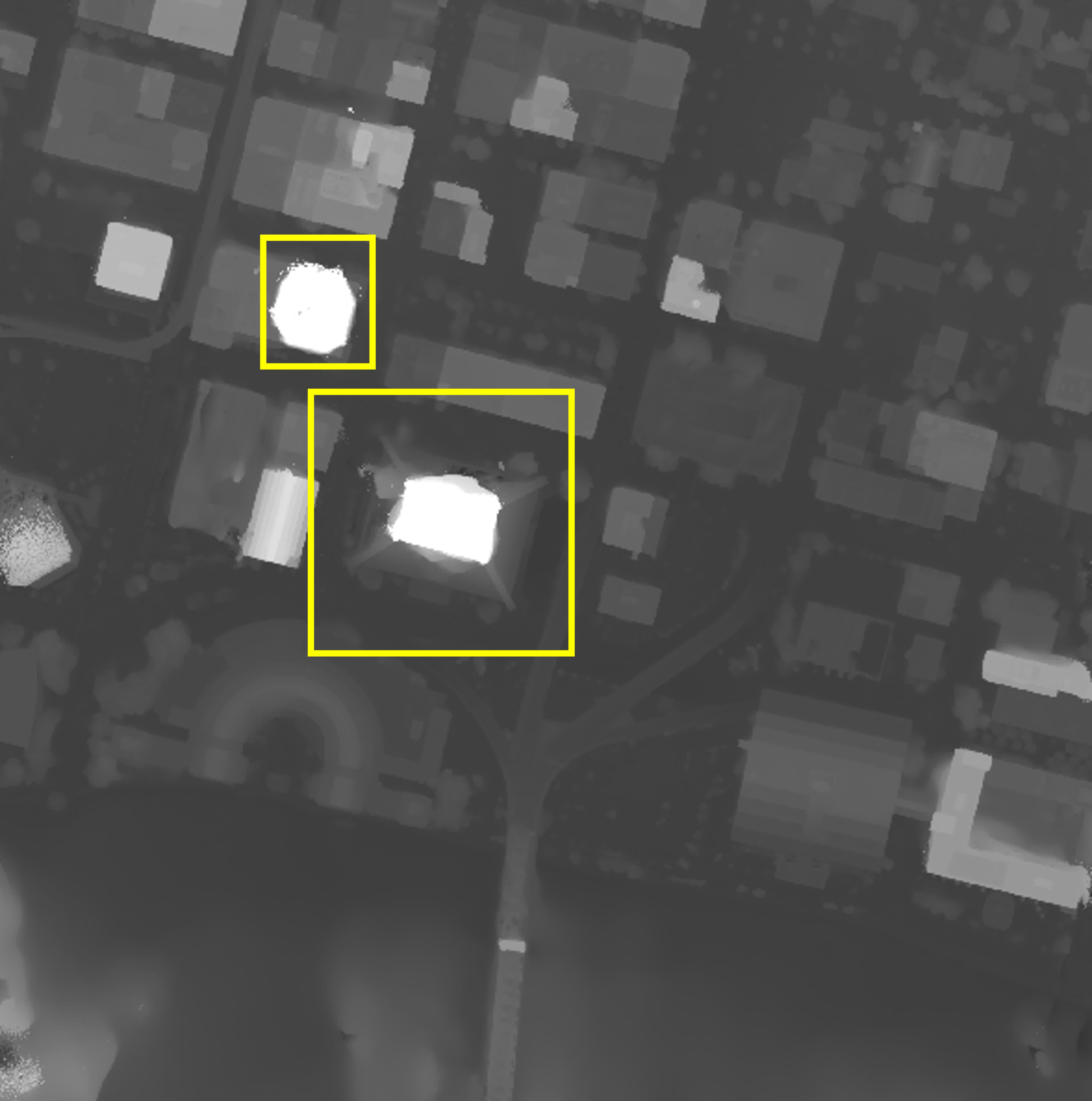} 
        \caption{Reconstructed DSM from (b)}
        \label{fig:init_s2p_dsm}
    \end{subfigure}
    
    \caption{Initialization sensitivity on JAX-167. Top row: initialization DSMs with high-rise reconstruction failure (a) and partial failure (b). Bottom row: corresponding reconstructed DSMs.}
    \label{fig:init_sensitivity}
\end{figure}

Based on our observations across multiple JAX sites, reconstruction quality relies heavily on the completeness of the photogrammetric DSM, as incomplete DSMs lead to erroneous shadow maps that reinforce incorrect geometry throughout optimization. As illustrated in \Cref{fig:init_sensitivity}, incomplete initialization leads to degraded reconstruction geometry, while a complete initialization enables stable reconstruction. In practice, ASP~\citep{beyer2018ames} offers faster runtime but is more susceptible to incomplete DSMs under limited stereo overlap, while s2p~\citep{de2014automatic} produces more complete initializations at the cost of higher computational overhead.

\subsection{Sensitivity to monocular depth supervision}

\begin{figure}[!h]
    \centering
    \includegraphics[width=\linewidth]{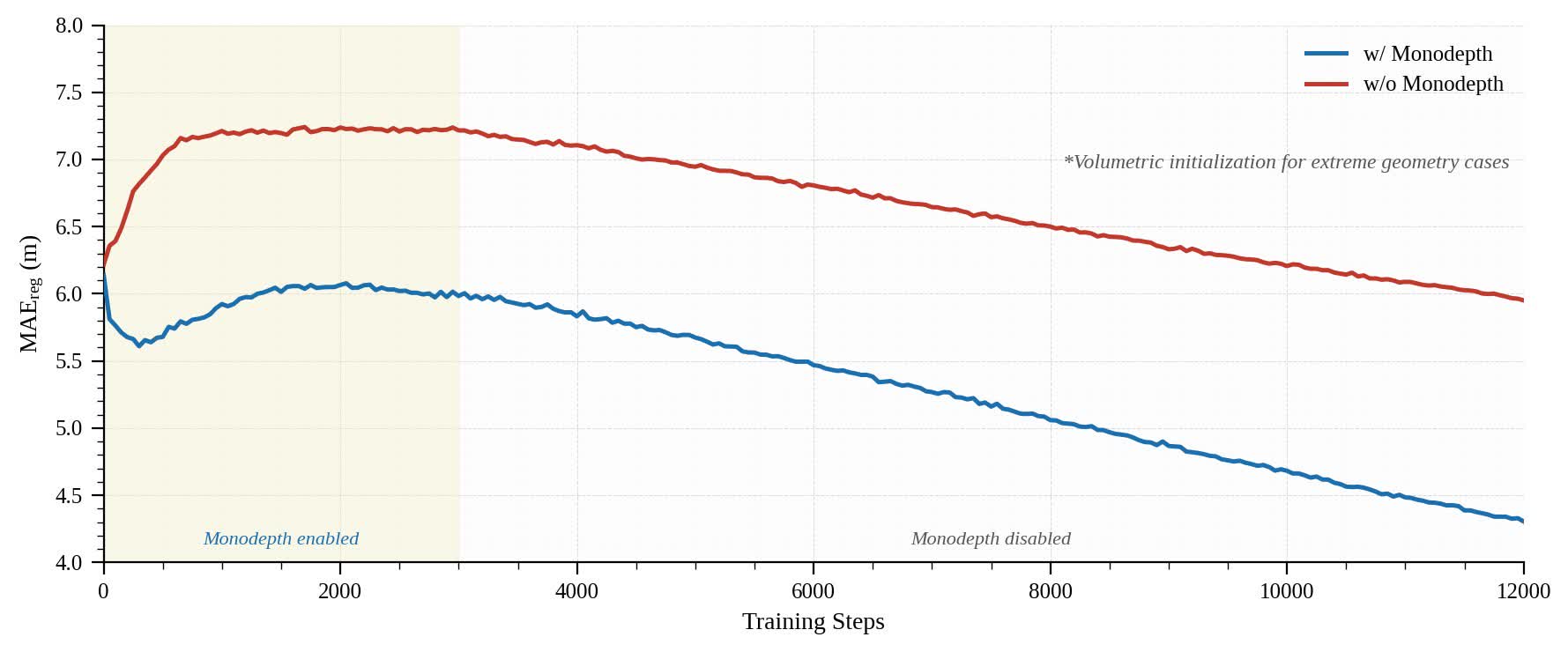}
    \caption{MAE$_{\text{reg}}$ over training steps under volumetric initialization, used as a stress test for extreme geometry cases. Monocular depth supervision accelerates early-stage convergence, while the absence of it results in slower and less stable convergence throughout training.}
    \label{fig:monodepth_graph}
\end{figure}

On our pipeline, shadow casting serves as the primary geometric supervision signal throughout optimization, while monocular depth supervision plays a supporting role during the first 3,000 iterations by accelerating early convergence. To evaluate this role under extreme conditions, we deliberately employ volumetric initialization instead of our default DSM-based initialization across four JAX sites (JAX-004, JAX-068, JAX-214, JAX-260). As shown in \Cref{fig:monodepth_graph}, the absence of monocular depth supervision results in significantly slower and less stable convergence, while its inclusion establishes a stable trajectory from early iterations, confirming its role in providing a geometric foundation for shadow casting.

\subsection{Sensitivity to number of diffusion-refined images}

%\RQ{the title is confusing, i think it was the input images. change the name of the title as Number of diffusion-refined images}

\begin{figure}[!h]
    \centering
    \includegraphics[width=\linewidth]{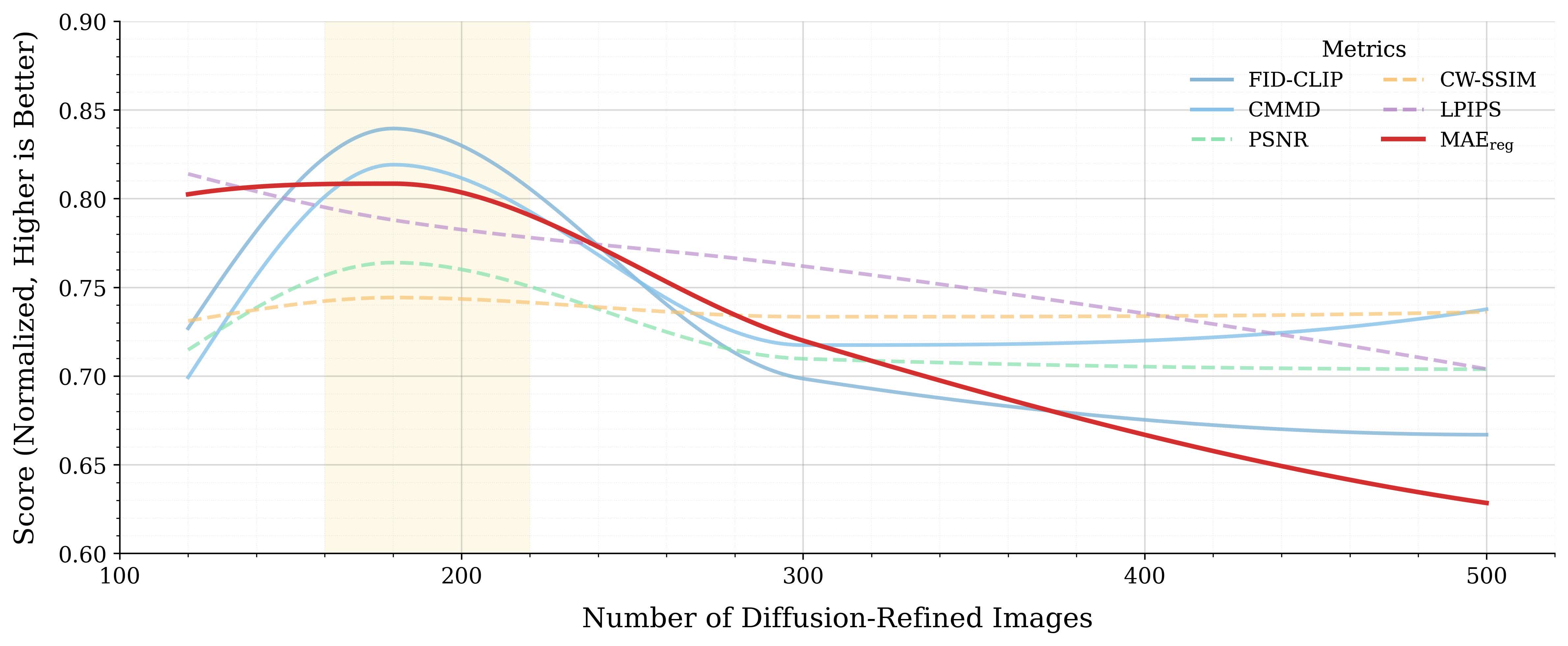}
    \caption{Sensitivity analysis on numbers of refined images on pseudo dataset. The normalized performance metrics indicate that 180 refined images provide the optimal balance. Exceeding this threshold introduces cumulative perceptual noise.}
    \label{fig:num_images}
\end{figure}

To determine the optimal balance between computational cost and reconstruction quality, we have conducted a sensitivity analysis on the number of diffusion-refined images (\Cref{fig:num_images}). We evaluate across four JAX sites (JAX-004, JAX-068, JAX-214, JAX-260) by varying the number of generated views from 120 to 500, measuring geometric accuracy ($\text{MAE}_{reg}$) alongside visual fidelity metrics (PSNR, LPIPS, FID-CLIP, CMMD, CW-SSIM). For clarity, all metrics are normalized to [0, 1] and adjusted so that higher values indicate better performance.

Our results show that 180 refined images are the optimal configuration. Increasing the count beyond this threshold does not yield further improvements but leads to slight degradation across most metrics. This suggests that while a sufficient number of views is necessary to provide multi-view constraints, an excess of diffusion-generated images may introduce cumulative perceptual noise.

\subsection{Appearance drift from diffusion refinement}

Diffusion refinement improves visual fidelity but may introduce appearance drift due to the generative prior. To analyze this behavior, we compare reconstructions with and without mixing original satellite observations during Gaussian optimization (\Cref{tab:comparison_average}). Since appearance drift is mainly associated with generative refinement, we compare against Skyfall-GS, which also employs diffusion-based enhancement, rather than conventional reconstruction methods.

\begin{table}[htbp]
\centering
\caption{Effect of incorporating original satellite images during generative refinement. Including original observations reduces distributional drift introduced by the diffusion process.}
\label{tab:comparison_average}
\small
\resizebox{0.8\linewidth}{!}{%
    \begin{tabular}{lccc}
    \toprule
    \multirow{2}{*}{\textbf{Method}} 
    & \multicolumn{3}{c}{\textbf{Pixel-level}} \\
    \cmidrule(lr){2-4}
    & PSNR $\uparrow$ & CW-SSIM $\uparrow$ & LPIPS $\downarrow$ \\
    \midrule
    Skyfall-GS
    & 14.11 & 0.273 & 0.601 \\
    \midrule
    Ours (w/o original img.)
    & 15.66 & 0.354 & 0.558 \\
    \textbf{Ours (w/ original img.)}
    & \textbf{16.83} & \textbf{0.430} & \textbf{0.529} \\
    \bottomrule
    \end{tabular}%
}
\end{table}

The proposed method achieves higher similarity to the original satellite observations compared with Skyfall-GS. While Skyfall-GS obtains a PSNR of 14.11, our method achieves 16.83. Within our framework, incorporating original satellite images during optimization further improves the similarity, increasing PSNR from 15.66 to 16.83 and reducing LPIPS from 0.558 to 0.529. These results indicate that mixing original observations with diffusion-refined images effectively constrains the refinement process and reduces deviation from the original satellite image distribution.

\subsection{Practical considerations}

Across our experiments on the DFC2019 and IARPA2016 datasets, scenes with fewer than approximately 12 images consistently exhibited unstable optimization due to insufficient multi-view coverage, leading us to identify this as the practical minimum for stable reconstruction. Additionally, accurate acquisition dates or solar metadata are required for physically accurate shadow simulation, which may not be available in all satellite products. Reconstruction quality is also sensitive to the completeness of the initialization DSM, since shadow-based supervision relies on accurate geometry for physically meaningful shadow simulation. While our pipeline consumes approximately 23 GB of GPU memory, roughly half that of Skyfall-GS, the multi-stage optimization still incurs higher computational overhead than standard Gaussian Splatting. We expect future work on accelerating diffusion-based refinement to substantially reduce this overhead.

\subsection{Scalability and large-area reconstruction}

\begin{figure*}[p]
    \centering
    \begin{subfigure}{\linewidth}
        \centering
        \includegraphics[width=\linewidth]{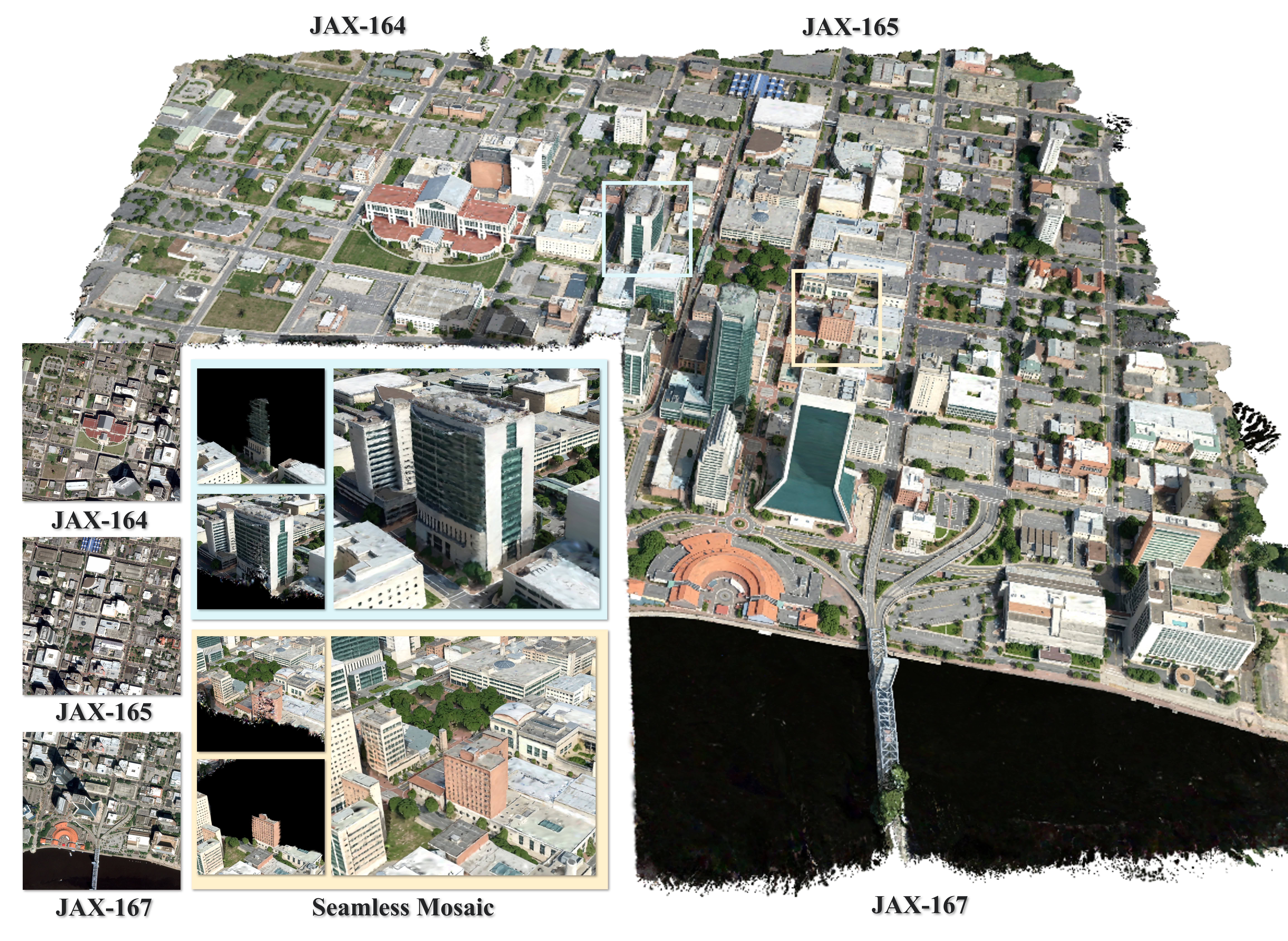} 
        \caption{Full mosaic of adjacent sites: JAX-164, JAX-165, and JAX-167.}
        \label{fig:large_scale_164_167}
    \end{subfigure}
    \vspace{0.5cm} 
    \begin{subfigure}{\linewidth}
        \centering
        \includegraphics[width=\linewidth]{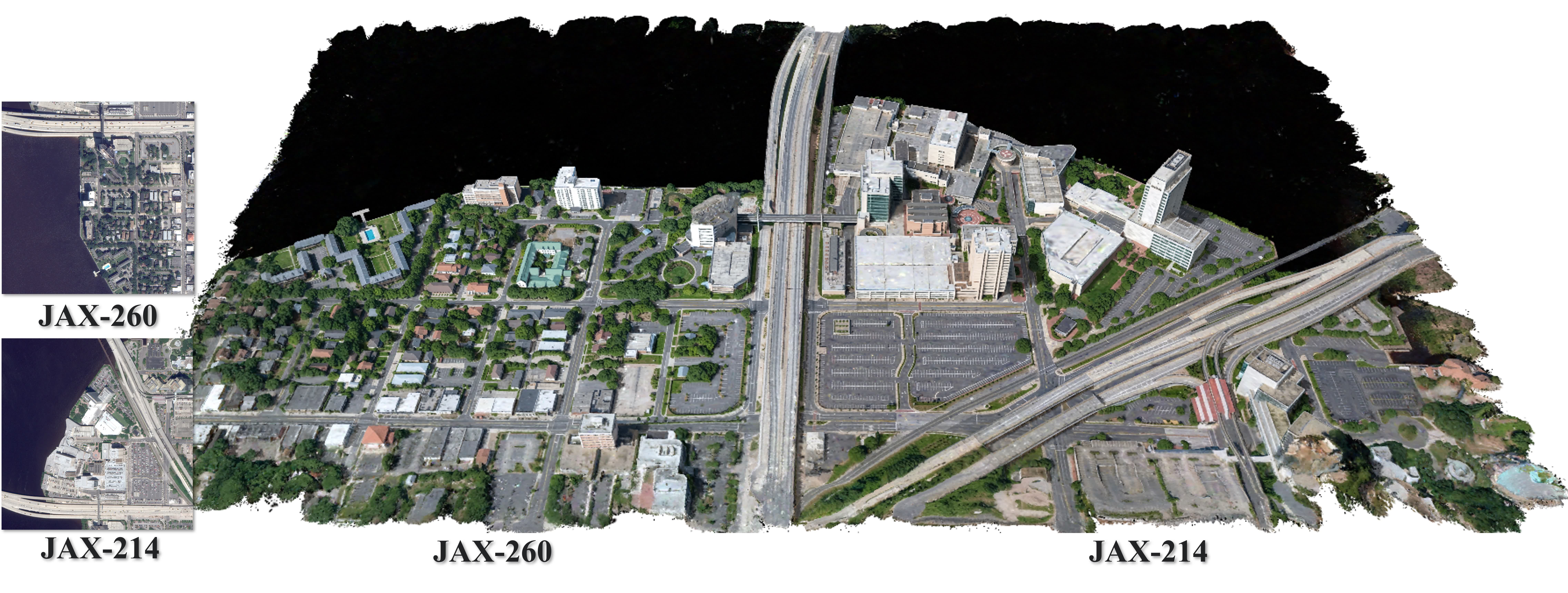} 
        \caption{Full mosaic of adjacent sites: JAX-214 and JAX-260.}
        \label{fig:large_scale_214_260}
    \end{subfigure}
    \caption{Application on larger area: seamless mosaics of adjacent JAX sites, including (a) JAX-164, JAX-165, JAX-167 and (b) JAX-214, JAX-260. Each site is reconstructed independently and merged using perpendicular bisectors along shared boundaries, demonstrating consistent visual details and scalability across different scenes.}
    \label{fig:large_scale}
\end{figure*}

Unlike full-scale reconstruction methods, our pipeline reconstructs large scenes by processing satellite images in smaller tiles. Although independent tile reconstruction may introduce inconsistencies in geometry or appearance across boundaries, our results show that adjacent reconstructed sites can be directly merged with minimal visible artifacts. As shown in \Cref{fig:large_scale}, independently reconstructed JAX-164, JAX-165, JAX-167 and JAX-214, JAX-260 produce seamless mosaics with consistent geometric structure and radiometric appearance. These results suggest that the proposed refinement process preserves sufficient consistency across independently optimized regions, enabling practical scaling to larger areas. 

%When adjacent sites from the DFC2019 dataset are reconstructed independently and merged by partitioning along perpendicular bisectors of shared boundaries, the resulting mosaics of JAX-164, JAX-165, JAX-167 and JAX-214, JAX-260 show no visible seams or radiometric discontinuities, demonstrating consistent appearance across tile boundaries.

%We attribute this scalability in part to the sensor-consistent appearance preserved throughout the pipeline. By constraining the diffusion model with geometrically calculated shadow maps, each tile maintains radiometrically coherent outputs that align naturally when merged. These results suggest that SatSplatDiff can serve as a foundation for city-scale 3D reconstruction from satellite imagery.

% geometrically accurate
% generates 'surface' rather than 'clouds', which makes it geometrically correct
% visually high fidelity
% less radiometric shift

% Memory consumption about half of Skyfall-Gs but still too much
% number of images are important
% accurate acquisition date or shadow information is required, which may not be available on many situations

\section{Conclusion}
\label{sec:conclusion}
In this paper, we presented SatSplatDiff, a unified pipeline for satellite-based 3D reconstruction that achieves state-of-the-art geometric accuracy and visual fidelity. The key idea is to extend shadow casting, previously used as a geometric cue during optimization, into the generative refinement stage. By conditioning diffusion-based refinement on shadow-cast rendered images, the resulting appearance remains consistent with the underlying geometry, mitigating the geometric degradation typically introduced by diffusion-based refinement. Together with photogrammetric DSM initialization, monocular depth supervision, multi-scale geometric refinement, and Gaussian densification, the proposed framework produces geometrically accurate and high-fidelity satellite reconstructions.

Quantitative evaluations on the IARPA2016 and DFC2019 datasets demonstrate that SatSplatDiff consistently outperforms existing methods in both geometric accuracy and visual fidelity, reducing $\mathrm{MAE}_{reg}$ by up to 18\% and improving FID-CLIP by 28--45\%, while delivering up to 5$\times$ resolution enhancement over existing baselines. Our method does exhibit limitations, including reduced fidelity in recovering extremely thin structures such as roller coasters and a practical requirement of at least 12 input images for stable reconstruction. Nonetheless, these results demonstrate the effectiveness of shadow-guided generative refinement in jointly improving geometric accuracy and visual fidelity for satellite-based 3D reconstruction. Future work will focus on extending the framework to larger and more diverse scenes and further improving robustness under sparse-view conditions.



%% References


\section*{Acknowledgements}
This work is supported by the Intelligence Advanced Research Projects Activity (IARPA) via Department of Interior/Interior Business Center (DOI/IBC) contract number 140D0423C0075 and Office of Naval Research (Award No. N000142312670). This work is also supported by the U.S. National Science Foundation (NSF) under Award No. 2331104.

\bibliographystyle{elsarticle-harv}
\bibliography{mybib}

\begin{thebibliography}{66}
\expandafter\ifx\csname natexlab\endcsname\relax\def\natexlab#1{#1}\fi
\providecommand{\url}[1]{\texttt{#1}}
\providecommand{\href}[2]{#2}
\providecommand{\path}[1]{#1}
\providecommand{\DOIprefix}{doi:}
\providecommand{\ArXivprefix}{arXiv:}
\providecommand{\URLprefix}{URL: }
\providecommand{\Pubmedprefix}{pmid:}
\providecommand{\doi}[1]{\href{http://dx.doi.org/#1}{\path{#1}}}
\providecommand{\Pubmed}[1]{\href{pmid:#1}{\path{#1}}}
\providecommand{\bibinfo}[2]{#2}
\ifx\xfnm\relax \def\xfnm[#1]{\unskip,\space#1}\fi
%Type = Inproceedings
\bibitem[{Aira et~al.(2025)Aira, Facciolo and Ehret}]{aira2025gaussian}
\bibinfo{author}{Aira, L.S.}, \bibinfo{author}{Facciolo, G.}, \bibinfo{author}{Ehret, T.}, \bibinfo{year}{2025}.
\newblock \bibinfo{title}{Gaussian splatting for efficient satellite image photogrammetry}, in: \bibinfo{booktitle}{Proceedings of the Computer Vision and Pattern Recognition Conference}, pp. \bibinfo{pages}{5959--5969}.
%Type = Article
\bibitem[{Beyer et~al.(2018)Beyer, Alexandrov and McMichael}]{beyer2018ames}
\bibinfo{author}{Beyer, R.A.}, \bibinfo{author}{Alexandrov, O.}, \bibinfo{author}{McMichael, S.}, \bibinfo{year}{2018}.
\newblock \bibinfo{title}{The ames stereo pipeline: Nasa's open source software for deriving and processing terrain data}.
\newblock \bibinfo{journal}{Earth and Space Science} \bibinfo{volume}{5}, \bibinfo{pages}{537--548}.
%Type = Inproceedings
\bibitem[{Billouard et~al.(2024)Billouard, Derksen, Sarrazin and Vallet}]{billouard2024sat}
\bibinfo{author}{Billouard, C.}, \bibinfo{author}{Derksen, D.}, \bibinfo{author}{Sarrazin, E.}, \bibinfo{author}{Vallet, B.}, \bibinfo{year}{2024}.
\newblock \bibinfo{title}{Sat-ngp: Unleashing neural graphics primitives for fast relightable transient-free 3d reconstruction from satellite imagery}, in: \bibinfo{booktitle}{IGARSS 2024-2024 IEEE International Geoscience and Remote Sensing Symposium}, \bibinfo{organization}{IEEE}. pp. \bibinfo{pages}{8749--8753}.
%Type = Inproceedings
\bibitem[{Bosch et~al.(2019)Bosch, Foster, Christie, Wang, Hager and Brown}]{bosch2019semantic}
\bibinfo{author}{Bosch, M.}, \bibinfo{author}{Foster, K.}, \bibinfo{author}{Christie, G.}, \bibinfo{author}{Wang, S.}, \bibinfo{author}{Hager, G.D.}, \bibinfo{author}{Brown, M.}, \bibinfo{year}{2019}.
\newblock \bibinfo{title}{Semantic stereo for incidental satellite images}, in: \bibinfo{booktitle}{2019 IEEE Winter Conference on Applications of Computer Vision (WACV)}, \bibinfo{organization}{IEEE}. pp. \bibinfo{pages}{1524--1532}.
%Type = Inproceedings
\bibitem[{Bosch et~al.(2016)Bosch, Kurtz, Hagstrom and Brown}]{bosch2016multiple}
\bibinfo{author}{Bosch, M.}, \bibinfo{author}{Kurtz, Z.}, \bibinfo{author}{Hagstrom, S.}, \bibinfo{author}{Brown, M.}, \bibinfo{year}{2016}.
\newblock \bibinfo{title}{A multiple view stereo benchmark for satellite imagery}, in: \bibinfo{booktitle}{2016 IEEE Applied Imagery Pattern Recognition Workshop (AIPR)}, \bibinfo{organization}{IEEE}. pp. \bibinfo{pages}{1--9}.
%Type = Article
\bibitem[{Bournez et~al.(2025)Bournez, Aira, Ehret and Facciolo}]{bournez2025eogs++}
\bibinfo{author}{Bournez, P.}, \bibinfo{author}{Aira, L.S.}, \bibinfo{author}{Ehret, T.}, \bibinfo{author}{Facciolo, G.}, \bibinfo{year}{2025}.
\newblock \bibinfo{title}{Eogs++: Earth observation gaussian splatting with internal camera refinement and direct panchromatic rendering}.
\newblock \bibinfo{journal}{arXiv preprint arXiv:2511.16542} .
%Type = Inproceedings
\bibitem[{De~Franchis et~al.(2014)De~Franchis, Meinhardt-Llopis, Michel, Morel and Facciolo}]{de2014automatic}
\bibinfo{author}{De~Franchis, C.}, \bibinfo{author}{Meinhardt-Llopis, E.}, \bibinfo{author}{Michel, J.}, \bibinfo{author}{Morel, J.M.}, \bibinfo{author}{Facciolo, G.}, \bibinfo{year}{2014}.
\newblock \bibinfo{title}{An automatic and modular stereo pipeline for pushbroom images}, in: \bibinfo{booktitle}{ISPRS Annals of the Photogrammetry, Remote Sensing and Spatial Information Sciences}.
%Type = Inproceedings
\bibitem[{Derksen and Izzo(2021)}]{derksen2021shadow}
\bibinfo{author}{Derksen, D.}, \bibinfo{author}{Izzo, D.}, \bibinfo{year}{2021}.
\newblock \bibinfo{title}{Shadow neural radiance fields for multi-view satellite photogrammetry}, in: \bibinfo{booktitle}{Proceedings of the IEEE/CVF Conference on Computer Vision and Pattern Recognition}, pp. \bibinfo{pages}{1152--1161}.
%Type = Article
\bibitem[{Ding et~al.(2026)Ding, Liu, Yin, Yang, Luo and Zhang}]{ding2026gu}
\bibinfo{author}{Ding, B.}, \bibinfo{author}{Liu, J.}, \bibinfo{author}{Yin, X.}, \bibinfo{author}{Yang, Z.}, \bibinfo{author}{Luo, Y.}, \bibinfo{author}{Zhang, D.}, \bibinfo{year}{2026}.
\newblock \bibinfo{title}{Gu-gs: Gaussian splatting-based geometry refinement and uncertainty-aware learning method for dsm generation from satellite imagery}.
\newblock \bibinfo{journal}{IEEE Transactions on Geoscience and Remote Sensing} .
%Type = Inproceedings
\bibitem[{Esser et~al.(2024)Esser, Kulal, Blattmann, Entezari, M{\"u}ller, Saini, Levi, Lorenz, Sauer, Boesel et~al.}]{esser2024scaling}
\bibinfo{author}{Esser, P.}, \bibinfo{author}{Kulal, S.}, \bibinfo{author}{Blattmann, A.}, \bibinfo{author}{Entezari, R.}, \bibinfo{author}{M{\"u}ller, J.}, \bibinfo{author}{Saini, H.}, \bibinfo{author}{Levi, Y.}, \bibinfo{author}{Lorenz, D.}, \bibinfo{author}{Sauer, A.}, \bibinfo{author}{Boesel, F.}, et~al., \bibinfo{year}{2024}.
\newblock \bibinfo{title}{Scaling rectified flow transformers for high-resolution image synthesis}, in: \bibinfo{booktitle}{Forty-first international conference on machine learning}.
%Type = Inproceedings
\bibitem[{Facciolo et~al.(2015)Facciolo, De~Franchis and Meinhardt}]{facciolo2015mgm}
\bibinfo{author}{Facciolo, G.}, \bibinfo{author}{De~Franchis, C.}, \bibinfo{author}{Meinhardt, E.}, \bibinfo{year}{2015}.
\newblock \bibinfo{title}{Mgm: A significantly more global matching for stereovision}, in: \bibinfo{booktitle}{BMVC 2015}.
%Type = Inproceedings
\bibitem[{Fischer et~al.(2025)Fischer, Bul{\`o}, Yang, Keetha, Porzi, M{\"u}ller, Schwarz, Luiten, Pollefeys and Kontschieder}]{fischer2025flowr}
\bibinfo{author}{Fischer, T.}, \bibinfo{author}{Bul{\`o}, S.R.}, \bibinfo{author}{Yang, Y.H.}, \bibinfo{author}{Keetha, N.}, \bibinfo{author}{Porzi, L.}, \bibinfo{author}{M{\"u}ller, N.}, \bibinfo{author}{Schwarz, K.}, \bibinfo{author}{Luiten, J.}, \bibinfo{author}{Pollefeys, M.}, \bibinfo{author}{Kontschieder, P.}, \bibinfo{year}{2025}.
\newblock \bibinfo{title}{Flowr: Flowing from sparse to dense 3d reconstructions}, in: \bibinfo{booktitle}{Proceedings of the IEEE/CVF International Conference on Computer Vision}, pp. \bibinfo{pages}{27702--27712}.
%Type = Article
\bibitem[{Fischler and Bolles(1981)}]{fischler1981random}
\bibinfo{author}{Fischler, M.A.}, \bibinfo{author}{Bolles, R.C.}, \bibinfo{year}{1981}.
\newblock \bibinfo{title}{Random sample consensus: a paradigm for model fitting with applications to image analysis and automated cartography}.
\newblock \bibinfo{journal}{Communications of the ACM} \bibinfo{volume}{24}, \bibinfo{pages}{381--395}.
%Type = Article
\bibitem[{Gao et~al.(2024)Gao, Holynski, Henzler, Brussee, Martin-Brualla, Srinivasan, Barron and Poole}]{gao2024cat3d}
\bibinfo{author}{Gao, R.}, \bibinfo{author}{Holynski, A.}, \bibinfo{author}{Henzler, P.}, \bibinfo{author}{Brussee, A.}, \bibinfo{author}{Martin-Brualla, R.}, \bibinfo{author}{Srinivasan, P.}, \bibinfo{author}{Barron, J.T.}, \bibinfo{author}{Poole, B.}, \bibinfo{year}{2024}.
\newblock \bibinfo{title}{Cat3d: Create anything in 3d with multi-view diffusion models}.
\newblock \bibinfo{journal}{arXiv preprint arXiv:2405.10314} .
%Type = Article
\bibitem[{He et~al.(2022)He, Li, Jiang and Jiang}]{he2022hmsm}
\bibinfo{author}{He, S.}, \bibinfo{author}{Li, S.}, \bibinfo{author}{Jiang, S.}, \bibinfo{author}{Jiang, W.}, \bibinfo{year}{2022}.
\newblock \bibinfo{title}{Hmsm-net: Hierarchical multi-scale matching network for disparity estimation of high-resolution satellite stereo images}.
\newblock \bibinfo{journal}{ISPRS Journal of Photogrammetry and Remote Sensing} \bibinfo{volume}{188}, \bibinfo{pages}{314--330}.
%Type = Article
\bibitem[{Hirschmuller(2008)}]{hirschmuller2008stereo}
\bibinfo{author}{Hirschmuller, H.}, \bibinfo{year}{2008}.
\newblock \bibinfo{title}{Stereo processing by semiglobal matching and mutual information}.
\newblock \bibinfo{journal}{IEEE Transactions on pattern analysis and machine intelligence} \bibinfo{volume}{30}, \bibinfo{pages}{328--341}.
%Type = Inproceedings
\bibitem[{Huang et~al.(2024)Huang, Yu, Chen, Geiger and Gao}]{huang20242d}
\bibinfo{author}{Huang, B.}, \bibinfo{author}{Yu, Z.}, \bibinfo{author}{Chen, A.}, \bibinfo{author}{Geiger, A.}, \bibinfo{author}{Gao, S.}, \bibinfo{year}{2024}.
\newblock \bibinfo{title}{2d gaussian splatting for geometrically accurate radiance fields}, in: \bibinfo{booktitle}{ACM SIGGRAPH 2024 conference papers}, pp. \bibinfo{pages}{1--11}.
%Type = Article
\bibitem[{Huang et~al.(2025)Huang, Liu, Wan, Zheng, Zhang, Xiong, Pei and Zhang}]{huang2025skysplat}
\bibinfo{author}{Huang, X.}, \bibinfo{author}{Liu, X.}, \bibinfo{author}{Wan, Y.}, \bibinfo{author}{Zheng, Z.}, \bibinfo{author}{Zhang, B.}, \bibinfo{author}{Xiong, M.}, \bibinfo{author}{Pei, Y.}, \bibinfo{author}{Zhang, Y.}, \bibinfo{year}{2025}.
\newblock \bibinfo{title}{Skysplat: Generalizable 3d gaussian splatting from multi-temporal sparse satellite images}.
\newblock \bibinfo{journal}{arXiv preprint arXiv:2508.09479} .
%Type = Article
\bibitem[{Huynh-Thu and Ghanbari(2008)}]{huynh2008scope}
\bibinfo{author}{Huynh-Thu, Q.}, \bibinfo{author}{Ghanbari, M.}, \bibinfo{year}{2008}.
\newblock \bibinfo{title}{Scope of validity of psnr in image/video quality assessment}.
\newblock \bibinfo{journal}{Electronics letters} \bibinfo{volume}{44}, \bibinfo{pages}{800--801}.
%Type = Inproceedings
\bibitem[{Jayasumana et~al.(2024)Jayasumana, Ramalingam, Veit, Glasner, Chakrabarti and Kumar}]{jayasumana2024rethinking}
\bibinfo{author}{Jayasumana, S.}, \bibinfo{author}{Ramalingam, S.}, \bibinfo{author}{Veit, A.}, \bibinfo{author}{Glasner, D.}, \bibinfo{author}{Chakrabarti, A.}, \bibinfo{author}{Kumar, S.}, \bibinfo{year}{2024}.
\newblock \bibinfo{title}{Rethinking fid: Towards a better evaluation metric for image generation}, in: \bibinfo{booktitle}{Proceedings of the IEEE/CVF conference on computer vision and pattern recognition}, pp. \bibinfo{pages}{9307--9315}.
%Type = Article
\bibitem[{Kerbl et~al.(2023)Kerbl, Kopanas, Leimk{\"u}hler and Drettakis}]{kerbl20233d}
\bibinfo{author}{Kerbl, B.}, \bibinfo{author}{Kopanas, G.}, \bibinfo{author}{Leimk{\"u}hler, T.}, \bibinfo{author}{Drettakis, G.}, \bibinfo{year}{2023}.
\newblock \bibinfo{title}{3d gaussian splatting for real-time radiance field rendering.}
\newblock \bibinfo{journal}{ACM Trans. Graph.} \bibinfo{volume}{42}, \bibinfo{pages}{139--1}.
%Type = Inproceedings
\bibitem[{Khanna et~al.(2024)Khanna, Liu, Zhou, Meng, Rombach, Burke, Lobell and Ermon}]{khanna2024diffusionsat}
\bibinfo{author}{Khanna, S.}, \bibinfo{author}{Liu, P.}, \bibinfo{author}{Zhou, L.}, \bibinfo{author}{Meng, C.}, \bibinfo{author}{Rombach, R.}, \bibinfo{author}{Burke, M.}, \bibinfo{author}{Lobell, D.}, \bibinfo{author}{Ermon, S.}, \bibinfo{year}{2024}.
\newblock \bibinfo{title}{Diffusionsat: A generative foundation model for satellite imagery}, in: \bibinfo{booktitle}{International Conference on Learning Representations}, pp. \bibinfo{pages}{5586--5604}.
%Type = Article
\bibitem[{Kim et~al.(2025)Kim, Cho, Chung and Kim}]{kim2025improving}
\bibinfo{author}{Kim, J.}, \bibinfo{author}{Cho, S.}, \bibinfo{author}{Chung, M.}, \bibinfo{author}{Kim, Y.}, \bibinfo{year}{2025}.
\newblock \bibinfo{title}{Improving disparity consistency with self-refined cost volumes for deep learning-based satellite stereo matching}.
\newblock \bibinfo{journal}{IEEE Journal of Selected Topics in Applied Earth Observations and Remote Sensing} .
%Type = Article
\bibitem[{Kynk{\"a}{\"a}nniemi et~al.(2022)Kynk{\"a}{\"a}nniemi, Karras, Aittala, Aila and Lehtinen}]{kynkaanniemi2022role}
\bibinfo{author}{Kynk{\"a}{\"a}nniemi, T.}, \bibinfo{author}{Karras, T.}, \bibinfo{author}{Aittala, M.}, \bibinfo{author}{Aila, T.}, \bibinfo{author}{Lehtinen, J.}, \bibinfo{year}{2022}.
\newblock \bibinfo{title}{The role of imagenet classes in fr$\backslash$'echet inception distance}.
\newblock \bibinfo{journal}{arXiv preprint arXiv:2203.06026} .
%Type = Misc
\bibitem[{Labs(2025)}]{flux-2-2025}
\bibinfo{author}{Labs, B.F.}, \bibinfo{year}{2025}.
\newblock \bibinfo{title}{{FLUX.2: Frontier Visual Intelligence}}.
\newblock \bibinfo{howpublished}{\url{https://bfl.ai/blog/flux-2}}.
%Type = Article
\bibitem[{Le~Saux et~al.(2019)Le~Saux, Yokoya, H{\"a}nsch and Brown}]{le2019data}
\bibinfo{author}{Le~Saux, B.}, \bibinfo{author}{Yokoya, N.}, \bibinfo{author}{H{\"a}nsch, R.}, \bibinfo{author}{Brown, M.}, \bibinfo{year}{2019}.
\newblock \bibinfo{title}{Data fusion contest 2019 (dfc2019); ieee dataport}.
\newblock \bibinfo{journal}{IEEE: Piscataway, NJ, USA} .
%Type = Article
\bibitem[{Lee et~al.(2025)Lee, Liu, Tsai, Chang, Wu, Chan, Zhao, Lin and Liu}]{lee2025SkyfallGS}
\bibinfo{author}{Lee, J.Y.}, \bibinfo{author}{Liu, Y.R.}, \bibinfo{author}{Tsai, S.R.}, \bibinfo{author}{Chang, W.C.}, \bibinfo{author}{Wu, C.H.}, \bibinfo{author}{Chan, J.}, \bibinfo{author}{Zhao, Z.}, \bibinfo{author}{Lin, C.H.}, \bibinfo{author}{Liu, Y.L.}, \bibinfo{year}{2025}.
\newblock \bibinfo{title}{{Skyfall-GS}: Synthesizing immersive {3D} urban scenes from satellite imagery}.
\newblock \bibinfo{journal}{arXiv preprint} \href{http://arxiv.org/abs/2510.15869}{{\tt arXiv:2510.15869}}.
%Type = Article
\bibitem[{Lee~Rodgers and Nicewander(1988)}]{lee1988thirteen}
\bibinfo{author}{Lee~Rodgers, J.}, \bibinfo{author}{Nicewander, W.A.}, \bibinfo{year}{1988}.
\newblock \bibinfo{title}{Thirteen ways to look at the correlation coefficient}.
\newblock \bibinfo{journal}{The American Statistician} \bibinfo{volume}{42}, \bibinfo{pages}{59--66}.
%Type = Inproceedings
\bibitem[{Li et~al.(2024)Li, Zhang, Bai, Zheng, Ning, Zhou and Gu}]{li2024dngaussian}
\bibinfo{author}{Li, J.}, \bibinfo{author}{Zhang, J.}, \bibinfo{author}{Bai, X.}, \bibinfo{author}{Zheng, J.}, \bibinfo{author}{Ning, X.}, \bibinfo{author}{Zhou, J.}, \bibinfo{author}{Gu, L.}, \bibinfo{year}{2024}.
\newblock \bibinfo{title}{Dngaussian: Optimizing sparse-view 3d gaussian radiance fields with global-local depth normalization}, in: \bibinfo{booktitle}{Proceedings of the IEEE/CVF conference on computer vision and pattern recognition}, pp. \bibinfo{pages}{20775--20785}.
%Type = Article
\bibitem[{Li et~al.(2023)Li, He, Jiang, Jiang and Zhang}]{li2023whu}
\bibinfo{author}{Li, S.}, \bibinfo{author}{He, S.}, \bibinfo{author}{Jiang, S.}, \bibinfo{author}{Jiang, W.}, \bibinfo{author}{Zhang, L.}, \bibinfo{year}{2023}.
\newblock \bibinfo{title}{Whu-stereo: A challenging benchmark for stereo matching of high-resolution satellite images}.
\newblock \bibinfo{journal}{IEEE Transactions on Geoscience and Remote Sensing} \bibinfo{volume}{61}, \bibinfo{pages}{1--14}.
%Type = Article
\bibitem[{Li et~al.(2025)Li, Yao, Wu, Yue, Zhao, Qin, Garc{\'\i}a-Fern{\'a}ndez, Levers, Ralph and Zhu}]{li2025ulsr}
\bibinfo{author}{Li, Z.}, \bibinfo{author}{Yao, S.}, \bibinfo{author}{Wu, T.}, \bibinfo{author}{Yue, Y.}, \bibinfo{author}{Zhao, W.}, \bibinfo{author}{Qin, R.}, \bibinfo{author}{Garc{\'\i}a-Fern{\'a}ndez, {\'A}.F.}, \bibinfo{author}{Levers, A.}, \bibinfo{author}{Ralph, J.}, \bibinfo{author}{Zhu, X.}, \bibinfo{year}{2025}.
\newblock \bibinfo{title}{Ulsr-gs: Urban large-scale surface reconstruction gaussian splatting with multi-view geometric consistency}.
\newblock \bibinfo{journal}{ISPRS Journal of Photogrammetry and Remote Sensing} \bibinfo{volume}{230}, \bibinfo{pages}{861--880}.
%Type = Inproceedings
\bibitem[{Lin et~al.(2023)Lin, Gao, Tang, Takikawa, Zeng, Huang, Kreis, Fidler, Liu and Lin}]{lin2023magic3d}
\bibinfo{author}{Lin, C.H.}, \bibinfo{author}{Gao, J.}, \bibinfo{author}{Tang, L.}, \bibinfo{author}{Takikawa, T.}, \bibinfo{author}{Zeng, X.}, \bibinfo{author}{Huang, X.}, \bibinfo{author}{Kreis, K.}, \bibinfo{author}{Fidler, S.}, \bibinfo{author}{Liu, M.Y.}, \bibinfo{author}{Lin, T.Y.}, \bibinfo{year}{2023}.
\newblock \bibinfo{title}{Magic3d: High-resolution text-to-3d content creation}, in: \bibinfo{booktitle}{Proceedings of the IEEE/CVF conference on computer vision and pattern recognition}, pp. \bibinfo{pages}{300--309}.
%Type = Inproceedings
\bibitem[{Lin et~al.(2024)Lin, Li, Tang, Liu, Liu, Liu, Lu, Wu, Xu, Yan et~al.}]{lin2024vastgaussian}
\bibinfo{author}{Lin, J.}, \bibinfo{author}{Li, Z.}, \bibinfo{author}{Tang, X.}, \bibinfo{author}{Liu, J.}, \bibinfo{author}{Liu, S.}, \bibinfo{author}{Liu, J.}, \bibinfo{author}{Lu, Y.}, \bibinfo{author}{Wu, X.}, \bibinfo{author}{Xu, S.}, \bibinfo{author}{Yan, Y.}, et~al., \bibinfo{year}{2024}.
\newblock \bibinfo{title}{Vastgaussian: Vast 3d gaussians for large scene reconstruction}, in: \bibinfo{booktitle}{Proceedings of the IEEE/CVF Conference on Computer Vision and Pattern Recognition}, pp. \bibinfo{pages}{5166--5175}.
%Type = Article
\bibitem[{Liu et~al.(2025)Liu, Zhao, Jiang and Guo}]{liu2025sat}
\bibinfo{author}{Liu, T.}, \bibinfo{author}{Zhao, S.}, \bibinfo{author}{Jiang, W.}, \bibinfo{author}{Guo, B.}, \bibinfo{year}{2025}.
\newblock \bibinfo{title}{Sat-dn: Implicit surface reconstruction from multi-view satellite images with depth and normal supervision}.
\newblock \bibinfo{journal}{IEEE Journal of Selected Topics in Applied Earth Observations and Remote Sensing} .
%Type = Inproceedings
\bibitem[{Liu et~al.(2024a)Liu, Chen, Kao, Tai and Tang}]{liu2024deceptive}
\bibinfo{author}{Liu, X.}, \bibinfo{author}{Chen, J.}, \bibinfo{author}{Kao, S.h.}, \bibinfo{author}{Tai, Y.W.}, \bibinfo{author}{Tang, C.K.}, \bibinfo{year}{2024}a.
\newblock \bibinfo{title}{Deceptive-nerf/3dgs: Diffusion-generated pseudo-observations for high-quality sparse-view reconstruction}, in: \bibinfo{booktitle}{European Conference on Computer Vision}, \bibinfo{organization}{Springer}. pp. \bibinfo{pages}{337--355}.
%Type = Inproceedings
\bibitem[{Liu et~al.(2026)Liu, Sun, Ren, Broaddus, Huang and Guigues}]{liu2026had}
\bibinfo{author}{Liu, X.}, \bibinfo{author}{Sun, W.}, \bibinfo{author}{Ren, Z.}, \bibinfo{author}{Broaddus, C.}, \bibinfo{author}{Huang, S.}, \bibinfo{author}{Guigues, L.}, \bibinfo{year}{2026}.
\newblock \bibinfo{title}{Had: Hallucination-aware diffusion priors for 3d reconstruction}, in: \bibinfo{booktitle}{Proceedings of the IEEE/CVF Conference on Computer Vision and Pattern Recognition}, pp. \bibinfo{pages}{29781--29791}.
%Type = Article
\bibitem[{Liu et~al.(2024b)Liu, Zhou and Huang}]{liu20243dgs}
\bibinfo{author}{Liu, X.}, \bibinfo{author}{Zhou, C.}, \bibinfo{author}{Huang, S.}, \bibinfo{year}{2024}b.
\newblock \bibinfo{title}{3dgs-enhancer: Enhancing unbounded 3d gaussian splatting with view-consistent 2d diffusion priors}.
\newblock \bibinfo{journal}{Advances in Neural Information Processing Systems} \bibinfo{volume}{37}, \bibinfo{pages}{133305--133327}.
%Type = Inproceedings
\bibitem[{Liu et~al.(2024c)Liu, Luo, Fan, Wang, Peng and Zhang}]{liu2024citygaussian}
\bibinfo{author}{Liu, Y.}, \bibinfo{author}{Luo, C.}, \bibinfo{author}{Fan, L.}, \bibinfo{author}{Wang, N.}, \bibinfo{author}{Peng, J.}, \bibinfo{author}{Zhang, Z.}, \bibinfo{year}{2024}c.
\newblock \bibinfo{title}{Citygaussian: Real-time high-quality large-scale scene rendering with gaussians}, in: \bibinfo{booktitle}{European Conference on Computer Vision}, \bibinfo{organization}{Springer}. pp. \bibinfo{pages}{265--282}.
%Type = Article
\bibitem[{Liu et~al.(2024d)Liu, Luo, Mao, Peng and Zhang}]{liu2024citygaussianv2}
\bibinfo{author}{Liu, Y.}, \bibinfo{author}{Luo, C.}, \bibinfo{author}{Mao, Z.}, \bibinfo{author}{Peng, J.}, \bibinfo{author}{Zhang, Z.}, \bibinfo{year}{2024}d.
\newblock \bibinfo{title}{Citygaussianv2: Efficient and geometrically accurate reconstruction for large-scale scenes}.
\newblock \bibinfo{journal}{arXiv preprint arXiv:2411.00771} .
%Type = Inproceedings
\bibitem[{Lowe(1999)}]{lowe1999object}
\bibinfo{author}{Lowe, D.G.}, \bibinfo{year}{1999}.
\newblock \bibinfo{title}{Object recognition from local scale-invariant features}, in: \bibinfo{booktitle}{Proceedings of the seventh IEEE international conference on computer vision}, \bibinfo{organization}{Ieee}. pp. \bibinfo{pages}{1150--1157}.
%Type = Article
\bibitem[{Luo et~al.(2026)Luo, Pan, Yang, Jiang, Liu and Huang}]{luo2026shadowgs}
\bibinfo{author}{Luo, F.}, \bibinfo{author}{Pan, H.}, \bibinfo{author}{Yang, X.}, \bibinfo{author}{Jiang, B.}, \bibinfo{author}{Liu, F.}, \bibinfo{author}{Huang, T.}, \bibinfo{year}{2026}.
\newblock \bibinfo{title}{Shadowgs: Shadow-aware 3d gaussian splatting for satellite imagery}.
\newblock \bibinfo{journal}{arXiv preprint arXiv:2601.00939} .
%Type = Article
\bibitem[{de~Lutio et~al.(2026)de~Lutio, Fischer, Chang, Zhang, Wu, Ren, Shen, Tothova, Gojcic and Turki}]{de2026artifixer}
\bibinfo{author}{de~Lutio, R.}, \bibinfo{author}{Fischer, T.}, \bibinfo{author}{Chang, Y.Y.}, \bibinfo{author}{Zhang, Y.}, \bibinfo{author}{Wu, J.Z.}, \bibinfo{author}{Ren, X.}, \bibinfo{author}{Shen, T.}, \bibinfo{author}{Tothova, K.}, \bibinfo{author}{Gojcic, Z.}, \bibinfo{author}{Turki, H.}, \bibinfo{year}{2026}.
\newblock \bibinfo{title}{Artifixer: Enhancing and extending 3d reconstruction with auto-regressive diffusion models}.
\newblock \bibinfo{journal}{arXiv preprint arXiv:2603.00492} .
%Type = Inproceedings
\bibitem[{Mar{\'\i} et~al.(2022)Mar{\'\i}, Facciolo and Ehret}]{mari2022sat}
\bibinfo{author}{Mar{\'\i}, R.}, \bibinfo{author}{Facciolo, G.}, \bibinfo{author}{Ehret, T.}, \bibinfo{year}{2022}.
\newblock \bibinfo{title}{Sat-nerf: Learning multi-view satellite photogrammetry with transient objects and shadow modeling using rpc cameras}, in: \bibinfo{booktitle}{Proceedings of the IEEE/CVF Conference on Computer Vision and Pattern Recognition}, pp. \bibinfo{pages}{1311--1321}.
%Type = Inproceedings
\bibitem[{Mar{\'\i} et~al.(2023)Mar{\'\i}, Facciolo and Ehret}]{mari2023multi}
\bibinfo{author}{Mar{\'\i}, R.}, \bibinfo{author}{Facciolo, G.}, \bibinfo{author}{Ehret, T.}, \bibinfo{year}{2023}.
\newblock \bibinfo{title}{Multi-date earth observation nerf: The detail is in the shadows}, in: \bibinfo{booktitle}{Proceedings of the IEEE/CVF Conference on Computer Vision and Pattern Recognition}, pp. \bibinfo{pages}{2035--2045}.
%Type = Article
\bibitem[{Marí et~al.(2021)Marí, de~Franchis, Meinhardt-Llopis, Anger and Facciolo}]{ipol.2021.352}
\bibinfo{author}{Marí, R.}, \bibinfo{author}{de~Franchis, C.}, \bibinfo{author}{Meinhardt-Llopis, E.}, \bibinfo{author}{Anger, J.}, \bibinfo{author}{Facciolo, G.}, \bibinfo{year}{2021}.
\newblock \bibinfo{title}{{A Generic Bundle Adjustment Methodology for Indirect RPC Model Refinement of Satellite Imagery}}.
\newblock \bibinfo{journal}{{Image Processing On Line}} \bibinfo{volume}{11}, \bibinfo{pages}{344--373}.
\newblock \bibinfo{note}{\url{https://doi.org/10.5201/ipol.2021.352}}.
%Type = Article
\bibitem[{Mildenhall et~al.(2021)Mildenhall, Srinivasan, Tancik, Barron, Ramamoorthi and Ng}]{mildenhall2021nerf}
\bibinfo{author}{Mildenhall, B.}, \bibinfo{author}{Srinivasan, P.P.}, \bibinfo{author}{Tancik, M.}, \bibinfo{author}{Barron, J.T.}, \bibinfo{author}{Ramamoorthi, R.}, \bibinfo{author}{Ng, R.}, \bibinfo{year}{2021}.
\newblock \bibinfo{title}{Nerf: Representing scenes as neural radiance fields for view synthesis}.
\newblock \bibinfo{journal}{Communications of the ACM} \bibinfo{volume}{65}, \bibinfo{pages}{99--106}.
%Type = Misc
\bibitem[{Oquab et~al.(2023)Oquab, Darcet, Moutakanni, Vo, Szafraniec, Khalidov, Fernandez, Haziza, Massa, El-Nouby, Howes, Huang, Xu, Sharma, Li, Galuba, Rabbat, Assran, Ballas, Synnaeve, Misra, Jegou, Mairal, Labatut, Joulin and Bojanowski}]{oquab2023dinov2}
\bibinfo{author}{Oquab, M.}, \bibinfo{author}{Darcet, T.}, \bibinfo{author}{Moutakanni, T.}, \bibinfo{author}{Vo, H.V.}, \bibinfo{author}{Szafraniec, M.}, \bibinfo{author}{Khalidov, V.}, \bibinfo{author}{Fernandez, P.}, \bibinfo{author}{Haziza, D.}, \bibinfo{author}{Massa, F.}, \bibinfo{author}{El-Nouby, A.}, \bibinfo{author}{Howes, R.}, \bibinfo{author}{Huang, P.Y.}, \bibinfo{author}{Xu, H.}, \bibinfo{author}{Sharma, V.}, \bibinfo{author}{Li, S.W.}, \bibinfo{author}{Galuba, W.}, \bibinfo{author}{Rabbat, M.}, \bibinfo{author}{Assran, M.}, \bibinfo{author}{Ballas, N.}, \bibinfo{author}{Synnaeve, G.}, \bibinfo{author}{Misra, I.}, \bibinfo{author}{Jegou, H.}, \bibinfo{author}{Mairal, J.}, \bibinfo{author}{Labatut, P.}, \bibinfo{author}{Joulin, A.}, \bibinfo{author}{Bojanowski, P.}, \bibinfo{year}{2023}.
\newblock \bibinfo{title}{Dinov2: Learning robust visual features without supervision}.
%Type = Inproceedings
\bibitem[{Peebles and Xie(2023)}]{peebles2023scalable}
\bibinfo{author}{Peebles, W.}, \bibinfo{author}{Xie, S.}, \bibinfo{year}{2023}.
\newblock \bibinfo{title}{Scalable diffusion models with transformers}, in: \bibinfo{booktitle}{Proceedings of the IEEE/CVF international conference on computer vision}, pp. \bibinfo{pages}{4195--4205}.
%Type = Article
\bibitem[{Poole et~al.(2022)Poole, Jain, Barron and Mildenhall}]{poole2022dreamfusion}
\bibinfo{author}{Poole, B.}, \bibinfo{author}{Jain, A.}, \bibinfo{author}{Barron, J.T.}, \bibinfo{author}{Mildenhall, B.}, \bibinfo{year}{2022}.
\newblock \bibinfo{title}{Dreamfusion: Text-to-3d using 2d diffusion}.
\newblock \bibinfo{journal}{arXiv preprint arXiv:2209.14988} .
%Type = Article
\bibitem[{Qin(2016)}]{qin2016rpc}
\bibinfo{author}{Qin, R.}, \bibinfo{year}{2016}.
\newblock \bibinfo{title}{Rpc stereo processor (rsp)--a software package for digital surface model and orthophoto generation from satellite stereo imagery}.
\newblock \bibinfo{journal}{ISPRS Annals of the Photogrammetry, Remote Sensing and Spatial Information Sciences} \bibinfo{volume}{3}, \bibinfo{pages}{77--82}.
%Type = Inproceedings
\bibitem[{Radford et~al.(2021)Radford, Kim, Hallacy, Ramesh, Goh, Agarwal, Sastry, Askell, Mishkin, Clark et~al.}]{radford2021learning}
\bibinfo{author}{Radford, A.}, \bibinfo{author}{Kim, J.W.}, \bibinfo{author}{Hallacy, C.}, \bibinfo{author}{Ramesh, A.}, \bibinfo{author}{Goh, G.}, \bibinfo{author}{Agarwal, S.}, \bibinfo{author}{Sastry, G.}, \bibinfo{author}{Askell, A.}, \bibinfo{author}{Mishkin, P.}, \bibinfo{author}{Clark, J.}, et~al., \bibinfo{year}{2021}.
\newblock \bibinfo{title}{Learning transferable visual models from natural language supervision}, in: \bibinfo{booktitle}{International conference on machine learning}, \bibinfo{organization}{PmLR}. pp. \bibinfo{pages}{8748--8763}.
%Type = Inproceedings
\bibitem[{Rombach et~al.(2022)Rombach, Blattmann, Lorenz, Esser and Ommer}]{rombach2022high}
\bibinfo{author}{Rombach, R.}, \bibinfo{author}{Blattmann, A.}, \bibinfo{author}{Lorenz, D.}, \bibinfo{author}{Esser, P.}, \bibinfo{author}{Ommer, B.}, \bibinfo{year}{2022}.
\newblock \bibinfo{title}{High-resolution image synthesis with latent diffusion models}, in: \bibinfo{booktitle}{Proceedings of the IEEE/CVF conference on computer vision and pattern recognition}, pp. \bibinfo{pages}{10684--10695}.
%Type = Inproceedings
\bibitem[{Rota~Bul{\`o} et~al.(2024)Rota~Bul{\`o}, Porzi and Kontschieder}]{rota2024revising}
\bibinfo{author}{Rota~Bul{\`o}, S.}, \bibinfo{author}{Porzi, L.}, \bibinfo{author}{Kontschieder, P.}, \bibinfo{year}{2024}.
\newblock \bibinfo{title}{Revising densification in gaussian splatting}, in: \bibinfo{booktitle}{European Conference on Computer Vision}, \bibinfo{organization}{Springer}. pp. \bibinfo{pages}{347--362}.
%Type = Article
\bibitem[{Saharia et~al.(2022)Saharia, Chan, Saxena, Li, Whang, Denton, Ghasemipour, Gontijo~Lopes, Karagol~Ayan, Salimans et~al.}]{saharia2022photorealistic}
\bibinfo{author}{Saharia, C.}, \bibinfo{author}{Chan, W.}, \bibinfo{author}{Saxena, S.}, \bibinfo{author}{Li, L.}, \bibinfo{author}{Whang, J.}, \bibinfo{author}{Denton, E.L.}, \bibinfo{author}{Ghasemipour, K.}, \bibinfo{author}{Gontijo~Lopes, R.}, \bibinfo{author}{Karagol~Ayan, B.}, \bibinfo{author}{Salimans, T.}, et~al., \bibinfo{year}{2022}.
\newblock \bibinfo{title}{Photorealistic text-to-image diffusion models with deep language understanding}.
\newblock \bibinfo{journal}{Advances in neural information processing systems} \bibinfo{volume}{35}, \bibinfo{pages}{36479--36494}.
%Type = Article
\bibitem[{Sampat et~al.(2009)Sampat, Wang, Gupta, Bovik and Markey}]{sampat2009complex}
\bibinfo{author}{Sampat, M.P.}, \bibinfo{author}{Wang, Z.}, \bibinfo{author}{Gupta, S.}, \bibinfo{author}{Bovik, A.C.}, \bibinfo{author}{Markey, M.K.}, \bibinfo{year}{2009}.
\newblock \bibinfo{title}{Complex wavelet structural similarity: A new image similarity index}.
\newblock \bibinfo{journal}{IEEE transactions on image processing} \bibinfo{volume}{18}, \bibinfo{pages}{2385--2401}.
%Type = Article
\bibitem[{Song et~al.(2027)Song, Kim and Qin}]{satsplat}
\bibinfo{author}{Song, S.}, \bibinfo{author}{Kim, J.}, \bibinfo{author}{Qin, R.}, \bibinfo{year}{2027}.
\newblock \bibinfo{title}{Satsplat: Geometrically-accurate gaussian splatting for satellite imagery}.
\newblock \bibinfo{journal}{Photogrammetric Engineering \& Remote Sensing} \bibinfo{note}{In press}.
%Type = Article
\bibitem[{Wang et~al.(2004)Wang, Bovik, Sheikh and Simoncelli}]{wang2004image}
\bibinfo{author}{Wang, Z.}, \bibinfo{author}{Bovik, A.C.}, \bibinfo{author}{Sheikh, H.R.}, \bibinfo{author}{Simoncelli, E.P.}, \bibinfo{year}{2004}.
\newblock \bibinfo{title}{Image quality assessment: from error visibility to structural similarity}.
\newblock \bibinfo{journal}{IEEE transactions on image processing} \bibinfo{volume}{13}, \bibinfo{pages}{600--612}.
%Type = Article
\bibitem[{Wu et~al.(2025a)Wu, Li, Zhou, Lin, Gao, Yan, Yin, Bai, Xu, Chen et~al.}]{wu2025qwen}
\bibinfo{author}{Wu, C.}, \bibinfo{author}{Li, J.}, \bibinfo{author}{Zhou, J.}, \bibinfo{author}{Lin, J.}, \bibinfo{author}{Gao, K.}, \bibinfo{author}{Yan, K.}, \bibinfo{author}{Yin, S.m.}, \bibinfo{author}{Bai, S.}, \bibinfo{author}{Xu, X.}, \bibinfo{author}{Chen, Y.}, et~al., \bibinfo{year}{2025}a.
\newblock \bibinfo{title}{Qwen-image technical report}.
\newblock \bibinfo{journal}{arXiv preprint arXiv:2508.02324} .
%Type = Inproceedings
\bibitem[{Wu et~al.(2025b)Wu, Zhang, Turki, Ren, Gao, Shou, Fidler, Gojcic and Ling}]{wu2025difix3d+}
\bibinfo{author}{Wu, J.Z.}, \bibinfo{author}{Zhang, Y.}, \bibinfo{author}{Turki, H.}, \bibinfo{author}{Ren, X.}, \bibinfo{author}{Gao, J.}, \bibinfo{author}{Shou, M.Z.}, \bibinfo{author}{Fidler, S.}, \bibinfo{author}{Gojcic, Z.}, \bibinfo{author}{Ling, H.}, \bibinfo{year}{2025}b.
\newblock \bibinfo{title}{Difix3d+: Improving 3d reconstructions with single-step diffusion models}, in: \bibinfo{booktitle}{Proceedings of the IEEE/CVF Conference on Computer Vision and Pattern Recognition}, pp. \bibinfo{pages}{26024--26035}.
%Type = Article
\bibitem[{Yang et~al.(2024)Yang, Kang, Huang, Zhao, Xu, Feng and Zhao}]{depth_anything_v2}
\bibinfo{author}{Yang, L.}, \bibinfo{author}{Kang, B.}, \bibinfo{author}{Huang, Z.}, \bibinfo{author}{Zhao, Z.}, \bibinfo{author}{Xu, X.}, \bibinfo{author}{Feng, J.}, \bibinfo{author}{Zhao, H.}, \bibinfo{year}{2024}.
\newblock \bibinfo{title}{Depth anything v2}.
\newblock \bibinfo{journal}{arXiv:2406.09414} .
%Type = Article
\bibitem[{Ye et~al.(2025)Ye, Li, Kerr, Turkulainen, Yi, Pan, Seiskari, Ye, Hu, Tancik and Kanazawa}]{ye2025gsplat}
\bibinfo{author}{Ye, V.}, \bibinfo{author}{Li, R.}, \bibinfo{author}{Kerr, J.}, \bibinfo{author}{Turkulainen, M.}, \bibinfo{author}{Yi, B.}, \bibinfo{author}{Pan, Z.}, \bibinfo{author}{Seiskari, O.}, \bibinfo{author}{Ye, J.}, \bibinfo{author}{Hu, J.}, \bibinfo{author}{Tancik, M.}, \bibinfo{author}{Kanazawa, A.}, \bibinfo{year}{2025}.
\newblock \bibinfo{title}{gsplat: An open-source library for gaussian splatting}.
\newblock \bibinfo{journal}{Journal of Machine Learning Research} \bibinfo{volume}{26}, \bibinfo{pages}{1--17}.
%Type = Article
\bibitem[{Yu et~al.(2025)Yu, Liu, Tang, Sun, Ge, Bu, Jin, Zhao, Sun, Li et~al.}]{yu2025orbit}
\bibinfo{author}{Yu, F.}, \bibinfo{author}{Liu, Y.}, \bibinfo{author}{Tang, L.}, \bibinfo{author}{Sun, M.}, \bibinfo{author}{Ge, Z.}, \bibinfo{author}{Bu, R.}, \bibinfo{author}{Jin, Y.}, \bibinfo{author}{Zhao, H.}, \bibinfo{author}{Sun, H.}, \bibinfo{author}{Li, Y.}, et~al., \bibinfo{year}{2025}.
\newblock \bibinfo{title}{From orbit to ground: Generative city photogrammetry from extreme off-nadir satellite images}.
\newblock \bibinfo{journal}{arXiv preprint arXiv:2512.07527} .
%Type = Inproceedings
\bibitem[{Yu et~al.(2024)Yu, Chen, Huang, Sattler and Geiger}]{yu2024mip}
\bibinfo{author}{Yu, Z.}, \bibinfo{author}{Chen, A.}, \bibinfo{author}{Huang, B.}, \bibinfo{author}{Sattler, T.}, \bibinfo{author}{Geiger, A.}, \bibinfo{year}{2024}.
\newblock \bibinfo{title}{Mip-splatting: Alias-free 3d gaussian splatting}, in: \bibinfo{booktitle}{Proceedings of the IEEE/CVF conference on computer vision and pattern recognition}, pp. \bibinfo{pages}{19447--19456}.
%Type = Inproceedings
\bibitem[{Zhang et~al.(2018)Zhang, Isola, Efros, Shechtman and Wang}]{zhang2018unreasonable}
\bibinfo{author}{Zhang, R.}, \bibinfo{author}{Isola, P.}, \bibinfo{author}{Efros, A.A.}, \bibinfo{author}{Shechtman, E.}, \bibinfo{author}{Wang, O.}, \bibinfo{year}{2018}.
\newblock \bibinfo{title}{The unreasonable effectiveness of deep features as a perceptual metric}, in: \bibinfo{booktitle}{Proceedings of the IEEE conference on computer vision and pattern recognition}, pp. \bibinfo{pages}{586--595}.
%Type = Article
\bibitem[{Zhao et~al.(2023)Zhao, Wang, Zhu and Song}]{zhao2023review}
\bibinfo{author}{Zhao, L.}, \bibinfo{author}{Wang, H.}, \bibinfo{author}{Zhu, Y.}, \bibinfo{author}{Song, M.}, \bibinfo{year}{2023}.
\newblock \bibinfo{title}{A review of 3d reconstruction from high-resolution urban satellite images}.
\newblock \bibinfo{journal}{International Journal of Remote Sensing} \bibinfo{volume}{44}, \bibinfo{pages}{713--748}.
%Type = Article
\bibitem[{Zhou et~al.(2024)Zhou, Wang, Lin, Cao, Li and Liu}]{zhou2024satelliterf}
\bibinfo{author}{Zhou, X.}, \bibinfo{author}{Wang, Y.}, \bibinfo{author}{Lin, D.}, \bibinfo{author}{Cao, Z.}, \bibinfo{author}{Li, B.}, \bibinfo{author}{Liu, J.}, \bibinfo{year}{2024}.
\newblock \bibinfo{title}{Satelliterf: Accelerating 3d reconstruction in multi-view satellite images with efficient neural radiance fields}.
\newblock \bibinfo{journal}{Applied Sciences} \bibinfo{volume}{14}, \bibinfo{pages}{2729}.

\end{thebibliography}






% \end{thebibliography}
%%\end{linenumbers}
\end{document}